\documentclass{pgnmreport}

\usepackage{cuted} %

\usepackage{currfile} %

\usepackage{caption} %

\usepackage{microtype}
\usepackage{multirow}
\usepackage{makecell}

\usepackage{algorithm}
\usepackage{algpseudocode}
\usepackage{tikz}
\usetikzlibrary{arrows.meta,positioning,calc}
\newlength{\vccell}
\newlength{\vclabel}
\newcommand{\vcsetup}[1]{\setlength{\tabcolsep}{0.5pt}\renewcommand{\arraystretch}{0}%
  \setlength{\vccell}{\dimexpr(\textwidth-\vclabel-\tabcolsep*#1*2-0.5pt)/#1\relax}}
\newcommand{\vcna}{{\setlength{\fboxsep}{0pt}\colorbox{black!6}{\makebox[\vccell]{%
  \rule{0pt}{\vccell}\raisebox{.47\vccell}{\tiny\color{black!55}n/a}}}}}
\newcommand{\vcrow}[1]{\makebox[\vclabel]{\rotatebox{90}{\parbox{\vccell}{\centering\scriptsize #1}}}}
\newcommand{\vchead}[1]{\parbox[b]{\vccell}{\centering\scriptsize #1\strut}}

\usepackage{placeins}

\colorlet{cvprblue}{pgnmblue}
\usepackage[breaklinks,colorlinks,allcolors=cvprblue]{hyperref}
\newcommand{\mainref}[1]{\ref{#1}}

\hypersetup{pdftitle={Parametric Gaussian Noise Mixer (PGNM): Unified Synthesis of Camera Noise and Film Grain},pdfauthor={Mingyang Song, Mart\'{\i}n Rodr\'{\i}guez Mu\~{n}oz}}
\ifdefined\pdfsuppressptexinfo \pdfsuppressptexinfo=-1 \fi

\title[PGNM: Unified synthesis of camera noise and film grain]{%
\textbf{P}arametric \textbf{G}aussian \textbf{N}oise \textbf{M}ixer (PGNM)}
\subtitle{Unified Synthesis of Camera Noise and Film Grain}
\author{Mingyang Song* \and Mart\'{\i}n Rodr\'{\i}guez Mu\~{n}oz*}
\shortauthors{Song and Rodr\'{\i}guez Mu\~{n}oz}
\affiliations{ETH Z\"urich\qquad *Equal contribution\qquad mingyang.song@inf.ethz.ch}
\abstractnotes{\sidelabel{Code}\href{https://github.com/miya9756/pgnm}{github.com/miya9756/pgnm}\par\vskip 10pt
\sidelabel{Note}This technical report will not be submitted to any conference.}
\reportdate{October 2026}

\begin{document}
\teaser{%
\centering
\includegraphics[width=\linewidth]{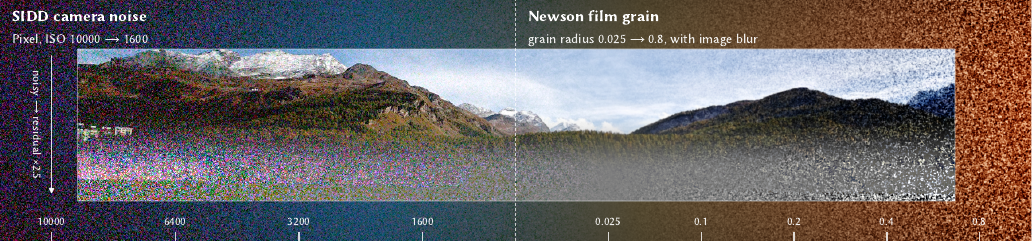}
\captionsetup{hypcap=false}
\captionof{figure}{\textbf{One model, a continuum of noise and grain.}
From left to right, PGNM interpolates between presets fitted on SIDD camera noise~\cite{abdelhamed2018sidd} and on Newson film grain~\cite{newson2017stochastic}; ticks mark the fitted presets. Film presets also apply the renderer's image blur, fitted per radius (Sec.~\ref{sec:datasets}). Within the photo, the noisy image fades into its residual from top to bottom. Please zoom in.}
\captionsetup{hypcap=true}
\label{fig:teaser}
}
\begin{abstract}
Noise synthesis is commonly approached through physical models of image formation or neural networks trained to generate noisy images. We explore a middle ground: an explicit statistical model fitted directly from paired clean and noisy images. A normalized convolution correlates a Gaussian field across pixels and color channels, and a lookup maps each Gaussian value to a residual given the clean intensity. Because the field stays standard Gaussian, the two components are fitted separately, without reproducing the physical process or training a neural generator. With one formulation for camera noise from SIDD and simulated film grain from Newson, the model has the lowest average distribution and correlation errors on SIDD among the evaluated methods and remains competitive on film grain. A denoiser trained on its noise comes closest to one trained on real noise, and a 1080p frame renders in 10\,ms on GPU.
\end{abstract}
\maketitle

\section{Introduction}
\label{sec:pgnm-introduction}

Photographic noise can be part of an image's intended appearance. Film grain contributes a texture that filmmakers may choose to preserve. For \emph{Oppenheimer}, FotoKem's~\cite{fotokem2023oppenheimer} account of the finishing process explicitly identifies grain structure as part of the distinctive photochemical texture. Reproducing this appearance requires modeling noise as well as removing it.

Noise modeling also supports efficient video delivery. Grain's spatial and temporal randomness makes it expensive to encode. AV1's~\cite{norkin2018filmgrain} film-grain synthesis framework encodes a denoised image together with estimated grain parameters and synthesizes grain at the decoder before display. This approach can reduce bandwidth while preserving the intended texture.

In visual-effects compositing, photographed and computer-generated elements must share a consistent photographic appearance. Foundry's Nuke~\cite{foundrygrain} documentation identifies grain matching as a step toward making all elements appear to have been photographed on the same film stock. A noise model therefore provides a means of reproducing the texture of live-action footage on digitally generated or modified content.

Accurate noise synthesis is also useful for denoising data augmentation. Brooks et al.~\cite{brooks2019unprocessing} show the importance of modeling sensor noise and image processing when synthesizing raw-image denoising data. Noise Flow~\cite{abdelhamed2019noiseflow} further demonstrates training a denoiser with samples from a learned noise model. These results motivate generating noise representative of the target acquisition process, even when noise removal is the eventual task.

Camera noise and film grain are mostly modeled separately. PGNM renders both by passing a convolved Gaussian field through an intensity-conditioned lookup, each fitted directly from aligned clean/noisy pairs. We find that:
\begin{itemize}
  \item \textbf{Dependence and distribution can be fitted separately.} The normalized convolution keeps each latent value standard Gaussian, so the lookup alone sets the residual distribution. The convolution is then estimated from measured correlations and the lookup from residual quantiles.
  \item \textbf{Direct fits hold up against learned generators.} The learned baselines contain every PGNM ingredient, but with many free parameters and indirect supervision, individual properties are harder to inspect and control. PGNM has the lowest average errors on SIDD, and a denoiser trained on its noise comes closest to one trained on real pairs. On film grain, it matches a physical approximation for fine grain and leads in conditional $W_1$ for coarser grain.
  \item \textbf{Intensity-only conditioning makes synthesis cheap.} The Gaussian field does not depend on the image, so it is filtered once and replayed (10\,ms per 1080p frame on GPU). On one film clip, a prototype that adds this grain after decoding suggests potential bitrate savings over coding the grain.
\end{itemize}

\section{Related Work}
\label{sec:pgnm-related-work}

\paragraph{Physical and statistical models.}
Camera models commonly combine photon-counting and readout noise; Foi et al.~\cite{foi2008noise} use a Poisson distribution for photon counts and a Gaussian distribution for readout fluctuations. Wei et al.~\cite{wei2020physical} extend this description to electronic effects in extreme low light. Unprocessing~\cite{brooks2019unprocessing} and CycleISP~\cite{zamir2020cycleisp} model the processing between raw sensor data and color images; Zou and Fu~\cite{zou2022noise} learn to estimate physical noise parameters. For film grain, Newson et al.~\cite{newson2017stochastic} render randomly placed grains through Monte Carlo simulation. Zhang et al.~\cite{zhang2023grain} derive an analytic approximation when individual grains are unresolved. Tian and Pappas~\cite{tian2026grain} extend efficient simulation to a wider range of magnification and blur. In codecs, AV1 grain synthesis~\cite{norkin2018filmgrain} combines autoregression over preceding neighbors with intensity-dependent scaling.

\paragraph{Neural camera-noise models.}
Among normalizing flows, Noise Flow~\cite{abdelhamed2019noiseflow} models raw noise, whereas sRGB-Flow~\cite{kousha2022srgb} models noise in processed images. Noise2NoiseFlow~\cite{maleky2022n2nflow} jointly learns noise modeling and denoising from noisy pairs; NAFlow~\cite{kim2024naflow} adapts sampling using a reference pair. InvDN~\cite{liu2021invdn} instead uses an invertible network to separate image content and noise.

Among generative adversarial networks (GANs), DANet~\cite{yue2020danet} jointly learns denoising and generation; C2N~\cite{jang2021c2n} uses unpaired clean and noisy images. CA-NoiseGAN~\cite{chang2020camera} conditions generation on camera information, while PNGAN~\cite{cai2021pngan} aligns generated and real images through pixel-level adversarial training. NeCA~\cite{fu2023neca} models the correlation between neighboring noise values. GMDCN~\cite{song2023camera} supplements adversarial training with a Gram-matrix style loss.

SeNM-VAE~\cite{zheng2024senm} uses a variational autoencoder with paired and unpaired data. Among diffusion models, RNSD~\cite{wu2025rnsd} uses camera conditioning, GuidNoise~\cite{kim2026guidnoise} uses one guidance pair, and PNG~\cite{ko2026png} uses learned noise prompts. YeTI~\cite{ko2026yeti} learns from two noisy observations and generates another noisy realization from a noisy input.

\paragraph{Learned film-grain models.}
Deep-FG~\cite{ameur2023deepfg} uses a GAN conditioned on grain parameters; GrainNet~\cite{lesne2025grain} uses a lightweight generator with a texture-aware loss. Style-FG~\cite{ameur2025stylefg} transfers grain style, while 3R-INN~\cite{ameur2024threeR} combines grain removal, reconstruction, and rescaling through an invertible network. Yang et al.~\cite{yang2025degradation} separate scene content from degradation to enable transfer, including film grain.

\paragraph{Shared observations.}
Read together, these models repeatedly measure or encode two observations, which PGNM adopts rather than introduces. First, residual strength depends on clean intensity. Foi et al.~\cite{foi2008noise} model it through random photon counts, Colom and Buades~\cite{colom2013analysis} estimate it as a noise curve from weakly textured regions, Noise Flow~\cite{abdelhamed2019noiseflow} builds it into an invertible transformation, and NeCA-W~\cite{fu2023neca} and AV1~\cite{norkin2018filmgrain} vary strength with intensity. In processed images, however, a strength curve captures only part of this observation: residual distributions also change in bias, asymmetry, and clipping, and deviate from common parametric forms (Fig.~\ref{fig:condpdf}). We therefore keep the full intensity-conditioned distribution as a two-dimensional lookup. Second, residuals are spatially correlated; for film grain, grain size appears as the distance over which nearby residual values remain correlated. A per-pixel transformation, such as Noise Flow's intensity-dependent one or our lookup, reshapes each residual distribution but cannot correlate values that start independent. Correlation therefore needs its own component: NeCA models neighboring correlation, NM-FlowGAN~\cite{han2023nmflowgan} adds a GAN for spatial correlation to per-pixel normalizing flows, and AV1 adds autoregression. PGNM realizes this division as a classical Gaussian copula, a correlated Gaussian field transformed pixel by pixel to target marginals~\cite{yamazaki1988digital,cario1996arta}, akin to texture synthesis~\cite{galerne2011random,heeger1995pyramid}.

\begin{figure*}[t]
    \centering
    \includegraphics[width=\linewidth]{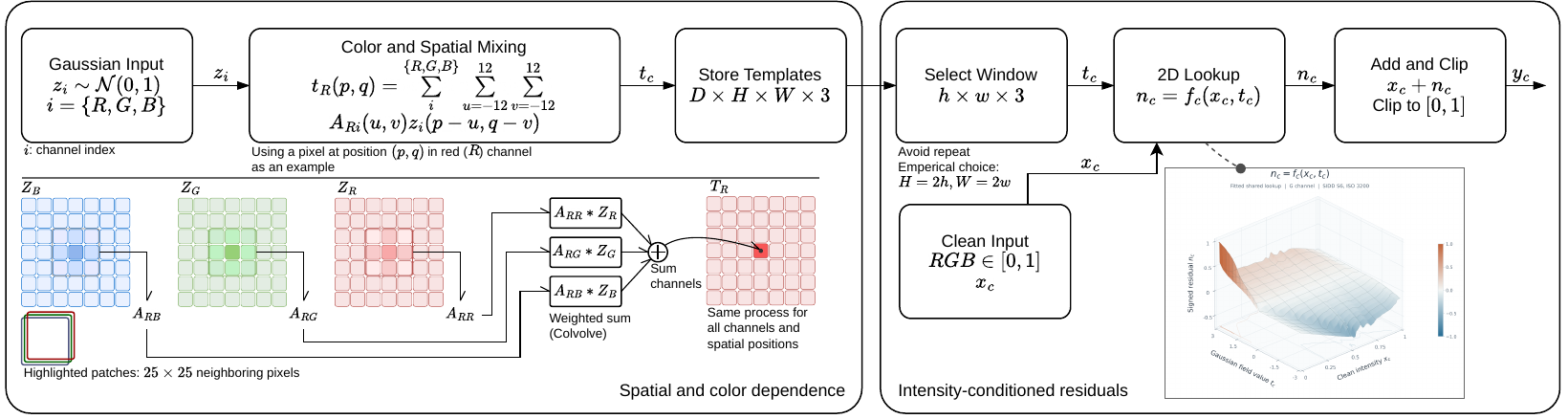}
    \caption{\textbf{Inference pipeline.} A full RGB convolution introduces spatial and color dependence into Gaussian latent templates. A shifted template window and the clean intensity index a residual lookup. We add its signed output to the clean image and clip to $[0,1]$.}
    \label{fig:infer_pipe}
\end{figure*}
\section{Method}
\label{sec:method}
PGNM separates intensity-conditioned residual distributions from spatial and color dependence. A full RGB convolution shapes a Gaussian latent field, which determines how residuals relate across pixels and channels; an empirical lookup then maps each latent value, given the clean intensity, to a signed residual (Fig.~\ref{fig:infer_pipe}). The two components are fitted separately from aligned clean/noisy RGB pairs and remain fixed during synthesis.

Let $X\in[0,1]^{h\times w\times3}$ be a clean image, $\mathbf{p}=(p,q)$ a pixel location, and $c\in\mathcal{C}=\{R,G,B\}$ a channel. For fitting pairs $\{(X^{(n)},Y^{(n)})\}_{n=1}^N$, the observed residual is
\begin{equation}\label{eq:fitterm}
    r_c^{(n)}(\mathbf{p})=
    y_c^{(n)}(\mathbf{p})-x_c^{(n)}(\mathbf{p}).
\end{equation}
We describe each component together with its fitting procedure; numerical details are provided in the supplementary material.

\subsection{Spatial and color dependence}
\label{sec:spatial_color}
\paragraph{Synthesis.}
Starting from independent Gaussian values across locations and input channels,
\begin{equation}\label{eq:iidsample}
    z_i(p,q)\sim\mathcal{N}(0,1),\qquad i\in\mathcal{C},
\end{equation}
we construct the latent field with a full $25\times25\times3\times3$ convolution tensor:
\begin{equation}\label{eq:filter}
    t_c(p,q)=\sum_{i\in\mathcal{C}}
    \sum_{u=-12}^{12}\sum_{v=-12}^{12}
    A_{ci}(u,v)z_i(p-u,q-v).
\end{equation}
Its nine filters $A_{ci}$ couple input channel $i$ to output channel $c$, jointly introducing spatial and color dependence. The per-output energy constraint
\begin{equation}\label{eq:normalize}
    \sum_{i\in\mathcal{C}}\sum_{u=-12}^{12}\sum_{v=-12}^{12}
    A_{ci}(u,v)^2=1\qquad\text{for each }c
\end{equation}
makes each latent value standard Gaussian whatever the filter shapes; for film grain, the spatial extent of the filters reflects grain size. For efficiency, we cache four independently constructed $2h\times2w\times3$ latent templates and reuse them by selecting random cyclic windows (details in the supplementary material).

\paragraph{Correlation fitting.}
We fit the filters so that the latent field reproduces residual correlations, measured where image content and clipping affect residuals least. Following noise estimation from weakly textured regions~\cite{liu2013single,colom2013analysis}, we define a clean-image flat mask $\mathcal{M}_{\mathrm{flat}}^{(n)}$ by requiring the maximum RGB local standard deviation in a reflected $7\times7$ window to be at most a threshold. Guarding against clipping and normalizing residuals both need a strength estimate, so we fit a pilot strength curve $\sigma_c^{\mathrm{pilot}}(x)$ on these flat pixels by matching observed second moments to a zero-mean Gaussian residual after image clipping. It serves only the correlation fit; the lookup (Sec.~\ref{sec:lookup}) is fitted separately on all pixels.

Correlation estimation additionally excludes clean intensities near the clipping boundaries. At each pixel, write $s_c=\sigma_c^{\mathrm{pilot}}(x_c)$, omitting the image index and location; one shared RGB mask is
\begin{equation}\label{eq:correlation_mask}
\begin{aligned}
    \mathcal{M}_{\mathrm{clip}}^{(n)}(\mathbf{p})
    &=\prod_{c\in\mathcal{C}}
      \mathbf{1}\{2.5s_c<x_c<1-2.5s_c\},\\
    \mathcal{M}_{\mathrm{corr}}^{(n)}(\mathbf{p})
    &=\mathcal{M}_{\mathrm{flat}}^{(n)}(\mathbf{p})
      \mathcal{M}_{\mathrm{clip}}^{(n)}(\mathbf{p}).
\end{aligned}
\end{equation}
This clean-based guard reduces clipping contamination without selecting on the observed noisy value. If the combined mask is empty across the entire fitting set, which occurs only for the coarsest simulated film grain, we instead set $\mathcal{M}_{\mathrm{corr}}^{(n)}=\mathcal{M}_{\mathrm{flat}}^{(n)}$ for all images. Dividing by the pilot strength places residuals on the unit scale of the latent field, so their correlations estimate those of the field; we also center within each image:
\begin{equation}\label{eq:normalize_before_spatial}
    u_c^{(n)}(\mathbf{p})=
    \frac{r_c^{(n)}(\mathbf{p})}{s_c^{(n)}(\mathbf{p})}
    -\mu_c^{(n)},
\end{equation}
where $\mu_c^{(n)}$ is the mean of the normalized residuals on $\mathcal{M}_{\mathrm{corr}}^{(n)}$.

For displacement $\mathbf{d}$, let $\mathcal{P}_{\mathbf{d}}$ contain all $(n,\mathbf{p})$ whose two endpoints remain within image $n$ and pass its correlation mask. Pooling valid products and counts across images gives all nine channel-pair covariances and correlations:
\begin{align}
    \widehat{C}_{ab}(\mathbf{d})
    &=\frac{1}{|\mathcal{P}_{\mathbf{d}}|}
      \sum_{(n,\mathbf{p})\in\mathcal{P}_{\mathbf{d}}}
      u_a^{(n)}(\mathbf{p})
      u_b^{(n)}(\mathbf{p}+\mathbf{d}),
      \label{eq:spatial_covariance}\\
    \widehat{R}_{ab}(\mathbf{d})
    &=\frac{\widehat{C}_{ab}(\mathbf{d})}
    {\sqrt{
        \widehat{C}_{aa}(\mathbf{0})
        \widehat{C}_{bb}(\mathbf{0})
    }}.
    \label{eq:makesure}
\end{align}
We use lags in $[-16,16]^2$, covering both same-pixel and shifted color dependence. By Eqs.~\ref{eq:iidsample} and~\ref{eq:filter}, the latent field's correlation is that of its filters, with $\mathbf{k}=(u,v)$:
\begin{equation}\label{eq:model_correlation}
    R_{ab}(\mathbf{d})=\sum_{i\in\mathcal{C}}\sum_{\mathbf{k}}
    A_{ai}(\mathbf{k})A_{bi}(\mathbf{k}+\mathbf{d}),
\end{equation}
so Eq.~\ref{eq:normalize} is exactly $R_{cc}(\mathbf{0})=1$, and fitting means choosing filters with $R_{ab}=\widehat{R}_{ab}$. Eq.~\ref{eq:model_correlation} is quadratic in the filters and couples all displacements, but the Fourier transform turns it into one $3\times3$ product per frequency: the filter spectrum times its conjugate transpose. We therefore factor the spectrum of the symmetrized and tapered $\widehat{R}_{ab}$: at each frequency, we project it onto positive semidefinite matrices and take its matrix square root (\emph{spectral factorization}; supplementary material). We convert the spectral factor to spatial kernels, crop to $25\times25$, and normalize by Eq.~\ref{eq:normalize}. Projection and finite support make the fit approximate; the subsequent nonlinear lookup can also alter residual correlations.

\subsection{Intensity-conditioned residuals}
\label{sec:lookup}
\begin{figure}[t]
    \centering
    \includegraphics[width=\linewidth]{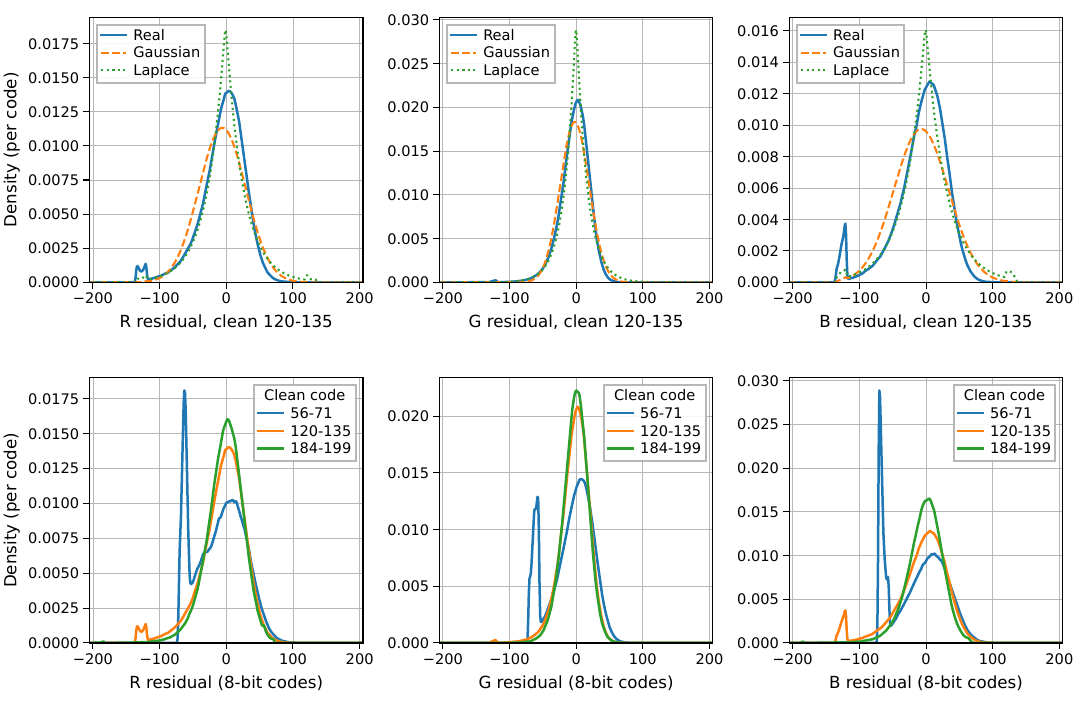}
    \caption{\textbf{Intensity-conditioned residual distributions} from SIDD~\cite{abdelhamed2018sidd}. Top: empirical densities versus common parametric distributions. Bottom: variation with clean intensity.}
    \label{fig:condpdf}
\end{figure}

We represent the intensity-conditioned residual distributions (Fig.~\ref{fig:condpdf}) with a two-dimensional lookup $f_c(x,t)$ indexed by clean intensity $x$ and latent Gaussian value $t$.

\begin{figure}[t]
    \centering
    \includegraphics[width=\linewidth]{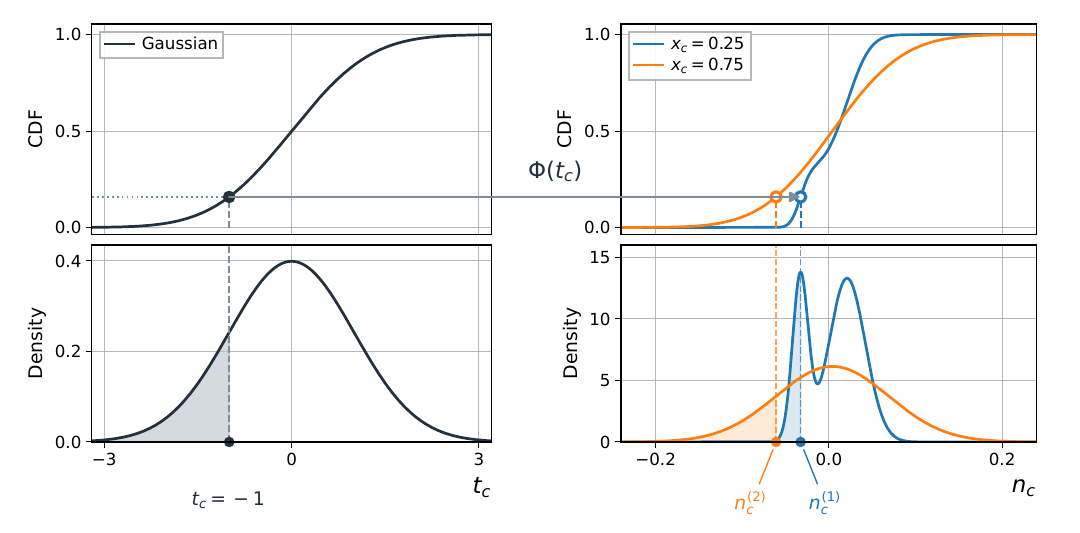}
    \caption{\textbf{Quantile mapping.} The Gaussian value $t_c$ selects a percentile and clean intensity $x_c$ selects a residual distribution. The lookup returns the corresponding signed residual $n_c$.}
    \label{fig:qmap}
\end{figure}

\paragraph{Fitting.}
We construct residual histograms by channel and clean intensity using \emph{all training pixels}, including textured and clipped observations. Pooling nearby intensity bins stabilizes the empirical cumulative distributions $\widehat{F}_{c,x}$. At grid points $(x_\ell,t_m)$, we store their quantiles:
\begin{equation}\label{eq:fit_lookup}
    f_c(x_\ell,t_m)=
    \widehat{F}_{c,x_\ell}^{-1}\!\left(\Phi(t_m)\right),
\end{equation}
where $\Phi$ is the standard Gaussian cumulative distribution function. In this \emph{conditional quantile mapping}, the Gaussian value selects a percentile and the clean intensity a residual distribution (Fig.~\ref{fig:qmap}). Because each latent value is standard Gaussian (Sec.~\ref{sec:spatial_color}), $\Phi(t_c)$ is uniform on $[0,1]$, so the lookup reproduces $\widehat{F}_{c,x}$ up to interpolation and clipping (Eq.~\ref{eq:noise_output}), whatever dependence the convolution introduces.

\paragraph{Synthesis.}
We evaluate $n_c(\mathbf{p})=f_c(x_c(\mathbf{p}),t_c(\mathbf{p}))$ by bilinear interpolation, with linear extrapolation along the Gaussian axis beyond its stored range (Fig.~\ref{fig:qmap}). The final image is
\begin{equation}\label{eq:noise_output}
    \widehat{y}_c(\mathbf{p})=\operatorname{clip}
    \bigl(x_c(\mathbf{p})+n_c(\mathbf{p}),0,1\bigr).
\end{equation}

\section{Evaluation protocol}
\label{sec:metrics}

\subsection{Noise synthesis fidelity}
\label{sec:noise_fidelity}
We evaluate residual-value distributions and their dependence using four complementary errors; lower values are better for all four. Real and generated images undergo the same clipping and round-half-up conversion to 8 bits before evaluation. Let $r_c=y_c-x_c$ and $\widehat{r}_c=\widehat{y}_c-x_c$ denote their signed residuals in normalized image units.

\paragraph{KL divergence \(D_{\mathrm{KL}}\).}
We follow the residual-histogram estimator released with sRGB-Flow~\cite{kousha2022srgb}. Within each $32\times32$ tile, we pool all RGB residuals into 130 bins of width four code values over $[-260,260]$, adding one count per bin before normalization. For real and generated bin probabilities $p_j$ and $q_j$, the divergence is
\begin{equation}\label{eq:metric_kl}
    D_{\mathrm{KL}}(p\|q)
    =\sum_{j=1}^{130}p_j\log\frac{p_j}{q_j}.
\end{equation}

\paragraph{Conditional $W_1$.}
We compare residual distributions at similar clean intensities with the 1-Wasserstein (earth mover's) distance, which in one dimension averages the absolute difference between matching percentiles. Let $F_{cb}$ and $G_{cb}$ be the empirical cumulative distributions of real and generated residuals in channel $c$ and clean-intensity bin $b$. We measure
\begin{equation}\label{eq:metric_conditional_w1}
    E_{W_1}
    =255\sum_{c,b}w_{cb}
      \int_0^1
      \left|F_{cb}^{-1}(a)-G_{cb}^{-1}(a)\right|\,da,
\end{equation}
where $w_{cb}$ weights bins by their pixel counts and channels equally, with $\sum_{c,b}w_{cb}=1$, and 255 gives code units. It responds to conditional bias, strength, and distribution shape.

\paragraph{Spatial and color correlation.}
We measure \emph{spatial correlation error} within each channel across nearby locations and \emph{color correlation error} between channels at neighboring locations. Both compare normalized correlations of real and generated residuals using root mean square error. Normalization, lag supports, and estimator details are provided in the supplementary material.

\subsection{Datasets}
\label{sec:datasets}
\paragraph{Smartphone Image Denoising Dataset (SIDD).}
We evaluate noise synthesis on the official sRGB validation split of the SIDD~\cite{abdelhamed2018sidd}, released for the NTIRE 2019 real image denoising challenge~\cite{abdelhamed2019ntire}. The release contains 1,280 paired $256\times256$ blocks from 40 source images, with 32 blocks per image. PGNM fits one preset per camera--ISO class on SIDD-Medium sRGB pairs; we exclude the iPhone~7 ISO~1250 source because its class is absent from SIDD-Medium. The resulting evaluation set contains 1,248 blocks from 39 source images, spanning five cameras, 16 camera--ISO classes, and eight physical scenes. Source IDs, acquisition-condition coverage, and the fitting crops are listed in the supplementary material.

\paragraph{Simulated film grain.}
We use the Boolean film-grain renderer of Newson et al.~\cite{newson2017stochastic}, which places random grains of one radius over the image and blurs the result optically, so grain size is an explicit parameter. We render eight radii from 0.025 to 0.8, covering the fine grain evaluated by GrainNet~\cite{lesne2025grain} and coarser grain as a stress case. Fitting uses the grayscale training pairs released with GrainNet, about 100 per radius. Because the renderer blurs the image as well as adding grain, the clean image is not the expected grainy image. For each radius we therefore fit a $7\times7$ linear filter $B$ on the training pairs and measure residuals against $\mu=B\ast x$. PGNM and Zhang23~\cite{zhang2023grain} add their grain to $\mu$, while GrainNet and an independent Newson render keep their native pipelines. Evaluation uses GrainNet's 60 natural test images, each rendered by Newson with five independent seeds. On this grayscale data, we report conditional $W_1$ and spatial correlation error, and add DISTS~\cite{ding2022dists} and LPIPS~\cite{zhang2018lpips} to compare the overall appearance of the delivered images. Results cover all radii and, separately, radii up to 0.2, where Zhang23's small-grain approximation applies.

\subsection{Denoising}
\label{sec:denoising}
Denoising gives an independent view of realism: a denoiser trained on synthetic pairs works on real images only as far as the synthesized noise resembles real noise. We therefore train the same denoiser on real pairs and on pairs from each generator, and compare each generator with this real-data reference. Training follows the protocol released with NM-FlowGAN~\cite{han2023nmflowgan}: a DnCNN~\cite{zhang2017beyond} trained for 30 epochs on $96\times96$ patches from 246 SIDD-Medium images, which cover exactly its 15 supported camera--ISO classes. PGNM is fitted per class on the same images; neural generators use the checkpoints of Tab.~\ref{tab:sidd} at native strength. Noise is drawn anew for every patch and epoch and quantized to 8 bits, and all noise sources share initialization, patch order, and augmentation. NAFlow-NAS denotes NAFlow's published setting, which samples noise guided by each patch's real noisy observation. We train one seed per noise source and report PSNR and SSIM at the final epoch on all 1,280 SIDD validation blocks (supplementary material).

\section{Results and analysis}
\label{sec:analysis}

\paragraph{Camera noise.}
PGNM has the lowest mean error on all four measures over all 39 SIDD sources (Tab.~\ref{tab:sidd}) and in most of the 16 classes (supplementary), although NM-FlowGAN's spatial error is slightly lower on the 19 sources it covers. The learned baselines already contain every ingredient of PGNM, most with far more free parameters fitted by gradient descent. Each network, however, fits its own training objective, and how its modules interact under that objective is hard to inspect; capacity in the wrong place, or supervision that constrains a property only indirectly, can leave that property off target. Their deviations indeed differ in kind: one baseline makes neighboring pixels too similar, another too independent, and the closest one, GMDCN~\cite{song2023camera}, matches noise strength better than PGNM but the distribution shape and the dependence between channels at neighboring pixels less well (supplementary). PGNM instead gives each property a form just flexible enough for what the observations show and estimates it directly from them. Visually, it is on par with the baselines (Fig.~\ref{fig:main_sidd_vis}).

\begin{table}[t]
    \centering
    \small
    \setlength{\tabcolsep}{7.5pt}
    \begin{tabular}{@{}lcccc@{}}
        \toprule
        & & & \multicolumn{2}{c}{Correlation$\downarrow$} \\
        \cmidrule(l){4-5}
        Method & \(D_{\mathrm{KL}}\)$\downarrow$ & $W_1\downarrow$ & Spatial & Color \\
        \midrule
        \multicolumn{5}{@{}l}{\textit{All 39 sources (16 camera--ISO classes)}} \\
        PGNM (ours) & \textbf{0.022} & \textbf{1.319} & \textbf{0.009} & \textbf{0.014} \\
        NAFlow~\cite{kim2024naflow} & 0.035 & 2.105 & 0.019 & 0.047 \\
        GMDCN$^{*}$~\cite{song2023camera} & 0.031 & 1.553 & 0.013 & 0.032 \\
        DANet~\cite{yue2020danet} & 0.143 & 2.589 & 0.016 & 0.045 \\
        \midrule
        \multicolumn{5}{@{}l}{\textit{19 sources in sRGB-Flow and NM-FlowGAN}} \\
        PGNM (ours) & \textbf{0.022} & \textbf{1.187} & 0.011 & \textbf{0.014} \\
        NAFlow~\cite{kim2024naflow} & 0.033 & 1.746 & 0.018 & 0.046 \\
        GMDCN$^{*}$~\cite{song2023camera} & 0.032 & 1.434 & 0.013 & 0.033 \\
        DANet~\cite{yue2020danet} & 0.150 & 2.334 & 0.017 & 0.042 \\
        sRGB-Flow~\cite{kousha2022srgb} & 0.051 & 4.197 & 0.023 & 0.045 \\
        NM-FlowGAN~\cite{han2023nmflowgan} & 0.037 & 1.998 & \textbf{0.010} & 0.036 \\
        \bottomrule
    \end{tabular}
    \caption{\textbf{Camera-noise synthesis on SIDD}~\cite{abdelhamed2018sidd}. Mean over five generator runs, cameras weighted equally (supplementary); the best value in each block is in bold. $^{*}$Our implementation of the published model.}
    \label{tab:sidd}
\end{table}

\begin{figure}[t]
    \centering
    \tiny
    \setlength{\tabcolsep}{0.35pt}
    \renewcommand{\arraystretch}{0.5}
    
    \begin{tabular}{cccccccc}
        Clean & Real & \textbf{PGNM} & GMDCN & NM-FlowG. & NAFlow & sRGB-Flow & DANet \\
        \includegraphics[width=0.122\linewidth]{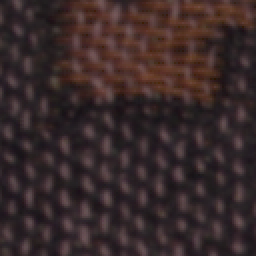} &
        \includegraphics[width=0.122\linewidth]{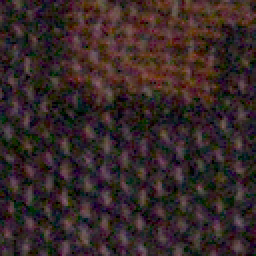} &
        \includegraphics[width=0.122\linewidth]{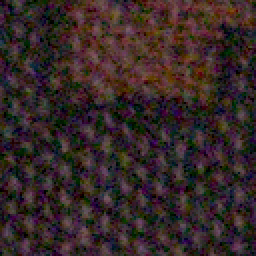} &
        \includegraphics[width=0.122\linewidth]{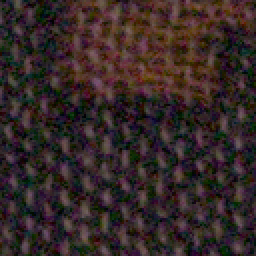} &
        \includegraphics[width=0.122\linewidth]{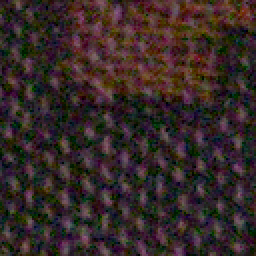} &
        \includegraphics[width=0.122\linewidth]{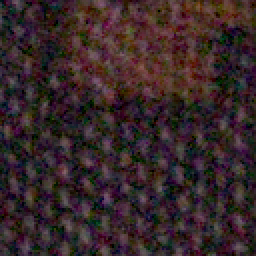} &
        \includegraphics[width=0.122\linewidth]{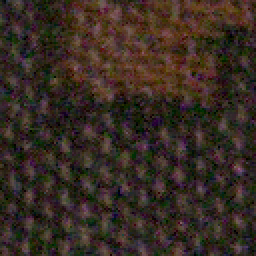} &
        \includegraphics[width=0.122\linewidth]{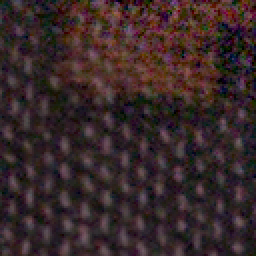} \\
        \raisebox{11pt}{S6 400} &
        \includegraphics[width=0.122\linewidth]{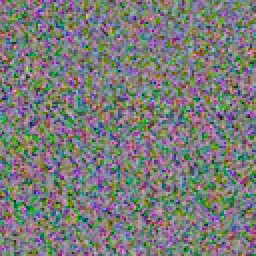} &
        \includegraphics[width=0.122\linewidth]{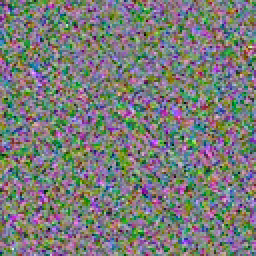} &
        \includegraphics[width=0.122\linewidth]{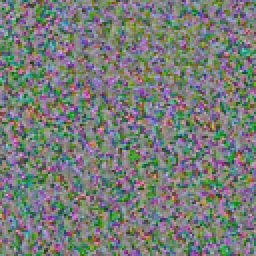} &
        \includegraphics[width=0.122\linewidth]{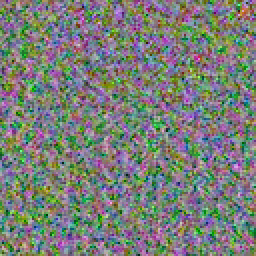} &
        \includegraphics[width=0.122\linewidth]{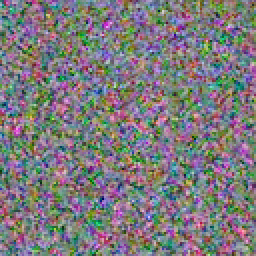} &
        \includegraphics[width=0.122\linewidth]{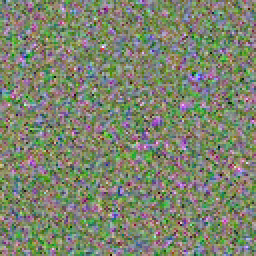} &
        \includegraphics[width=0.122\linewidth]{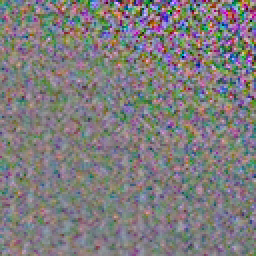} \\

        \includegraphics[width=0.122\linewidth]{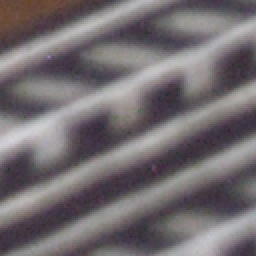} &
        \includegraphics[width=0.122\linewidth]{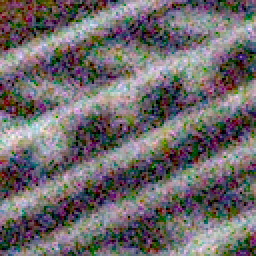} &
        \includegraphics[width=0.122\linewidth]{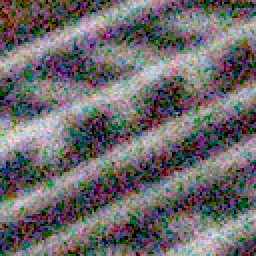} &
        \includegraphics[width=0.122\linewidth]{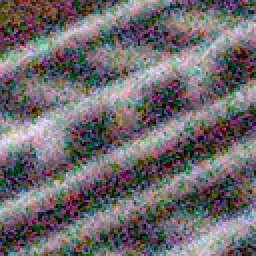} &
        \includegraphics[width=0.122\linewidth]{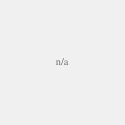} &
        \includegraphics[width=0.122\linewidth]{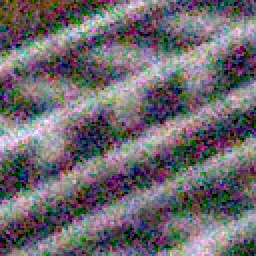} &
        \includegraphics[width=0.122\linewidth]{figures/na.pdf} &
        \includegraphics[width=0.122\linewidth]{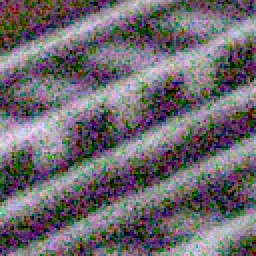} \\
        \raisebox{11pt}{GP 6400} &
        \includegraphics[width=0.122\linewidth]{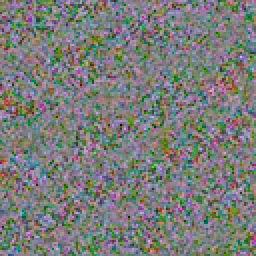} &
        \includegraphics[width=0.122\linewidth]{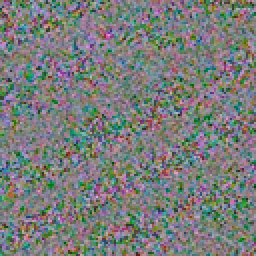} &
        \includegraphics[width=0.122\linewidth]{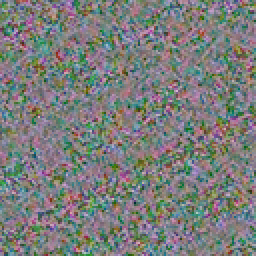} &
        \includegraphics[width=0.122\linewidth]{figures/na.pdf} &
        \includegraphics[width=0.122\linewidth]{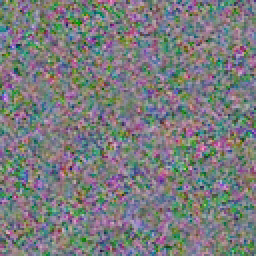} &
        \includegraphics[width=0.122\linewidth]{figures/na.pdf} &
        \includegraphics[width=0.122\linewidth]{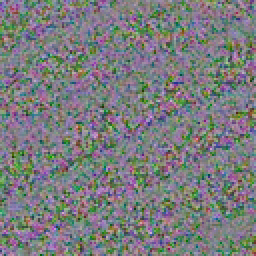} \\
    \end{tabular}
    \caption{\textbf{Visual comparisons on SIDD}~\cite{abdelhamed2018sidd}: random (S6 400) and most textured (GP 6400) windows. PGNM is visually on par with the baselines and shows fewer repetitive patterns in the residual (S6 400). Second row of each pair: residual relative to the clean image, amplified by one gain per row (gray is zero). n/a: class not covered by the checkpoint. More in the supplementary material.}
    \label{fig:main_sidd_vis}
\end{figure}

PGNM is content-aware through each pixel's clean intensity, which its lookup measures directly, but not tied to the surrounding structure, so on a new image its noise changes only as the intensities do. We applied all SIDD fits and checkpoints, unchanged, to a textured Kodak image (supplementary). NAFlow's~\cite{kim2024naflow} noise grows well above its SIDD level, and DANet~\cite{yue2020danet}, which has no camera or ISO input, gives about the same level for every class; the others, PGNM included, change less. GMDCN~\cite{song2023camera} and NM-FlowGAN~\cite{han2023nmflowgan} keep the edge-following part of the residual, which PGNM does not model (Sec.~\ref{sec:ending}).

\paragraph{Denoising.}
As in Tab.~\ref{tab:sidd}, PGNM ranks first among synthetic sources: a DnCNN trained on its noise is 0.72\,dB below one trained on real pairs (Tab.~\ref{tab:denoise_sidd}). On all 40 images, the 95\% scene-bootstrap interval of its lead over each baseline excludes zero; on the 21 images outside the training classes, its lead over NAFlow-NAS is a tie (supplementary).

\begin{table}[t]
\centering
\setlength{\tabcolsep}{10pt}
\begin{tabular}{@{}lcc@{}}
\toprule
Training noise & PSNR$\uparrow$ & SSIM$\uparrow$ \\
\midrule
Real noise (reference) & 37.94 & 0.896 \\
\midrule
GMDCN & 36.15 & 0.872 \\
NM-FlowGAN & 35.15 & 0.875 \\
NAFlow & 36.72 & 0.878 \\
NAFlow-NAS & 36.88 & 0.883 \\
PGNM (ours) & \textbf{37.22} & \textbf{0.889} \\
\bottomrule
\end{tabular}
\caption{\textbf{Real-noise denoising on SIDD Validation} (40 images, 1{,}280 blocks). One DnCNN per training noise (Sec.~\ref{sec:denoising}), training seed 1. NAFlow-NAS: noise-aware sampling, which also reads the real noisy training patch.}
\label{tab:denoise_sidd}
\end{table}

\begin{table}[t]
    \centering
    \small
    \setlength{\tabcolsep}{7pt}
    \begin{tabular}{@{}lcccc@{}}
        \toprule
        & \multicolumn{2}{c}{Residual$\downarrow$} & \multicolumn{2}{c}{Appearance$\downarrow$} \\
        \cmidrule(lr){2-3} \cmidrule(l){4-5}
        Method & $W_1$ & Spatial & DISTS & LPIPS \\
        \midrule
        \multicolumn{5}{@{}l}{\textit{All eight radii (0.025--0.8)}} \\
        PGNM (ours)                          & \textbf{0.626} & \textbf{0.0082} & 0.0731 & 0.0809 \\
        Zhang23$^{*}$~\cite{zhang2023grain}  & 1.402 & 0.0203 & 0.1014 & 0.1397 \\
        GrainNet~\cite{lesne2025grain}       & 3.072 & 0.0202 & \textbf{0.0690} & \textbf{0.0799} \\
        \cmidrule{1-5}
        Ind. Newson render            & 0.424 & 0.0068 & 0.0653 & 0.0785 \\
        \midrule
        \multicolumn{5}{@{}l}{\textit{Radii up to 0.2}} \\
        PGNM (ours)                          & \textbf{0.208} & 0.0079 & \textbf{0.0442} & \textbf{0.0338} \\
        Zhang23$^{*}$~\cite{zhang2023grain}  & 0.232 & \textbf{0.0073} & \textbf{0.0442} & \textbf{0.0338} \\
        GrainNet~\cite{lesne2025grain}       & 3.313 & 0.0275 & 0.0473 & 0.0357 \\
        \cmidrule{1-5}
        Ind. Newson render            & 0.143 & 0.0066 & 0.0430 & 0.0333 \\
        \bottomrule
    \end{tabular}
    \caption{\textbf{Simulated film grain}~\cite{newson2017stochastic}. Mean over the 25 output/reference pairs of each image and radius, then over the 60 test images and the radii of each block; the best fitted method is in bold. $W_1$: conditional $W_1$ in 8-bit code units; Spatial: spatial correlation error. The independent Newson render compares two realizations of the simulator and is not bolded. $^{*}$Our implementation of the published model, with the shared blur $B$ applied to the clean image.}
    \label{tab:film}
\end{table}

\begin{figure}[t]
    \centering
    \tiny
    \setlength{\tabcolsep}{0.35pt}
    \renewcommand{\arraystretch}{0.5}
    \begin{tabular}{ccccc}
        Clean & Newson & \textbf{PGNM}+\textit{B} & Zhang23~\cite{zhang2023grain}+\textit{B} & GrainNet \\

        \includegraphics[width=0.195\linewidth]{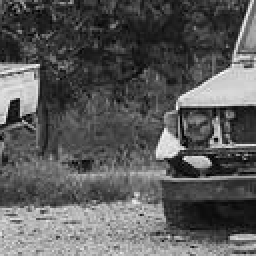} & 
        \includegraphics[width=0.195\linewidth]{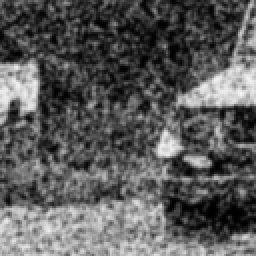} & 
        \includegraphics[width=0.195\linewidth]{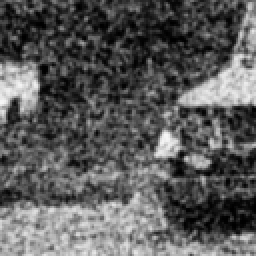} & 
        \includegraphics[width=0.195\linewidth]{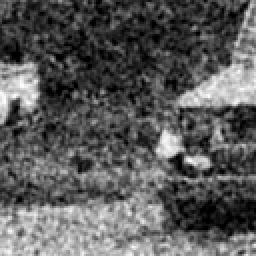} & 
        \includegraphics[width=0.195\linewidth]{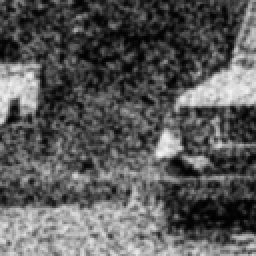} \\
        \raisebox{19pt}{\(r=0.4\)} & 
        \includegraphics[width=0.195\linewidth]{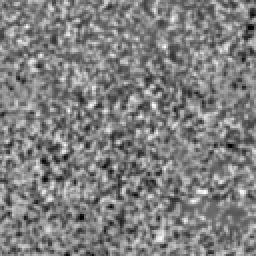} & 
        \includegraphics[width=0.195\linewidth]{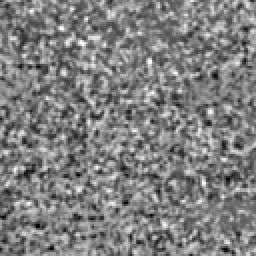} & 
        \includegraphics[width=0.195\linewidth]{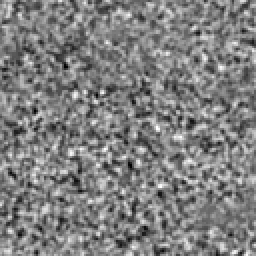} & 
        \includegraphics[width=0.195\linewidth]{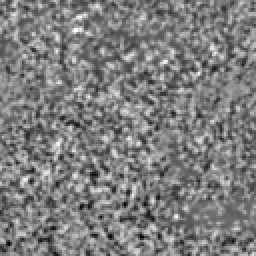} \\

        \includegraphics[width=0.195\linewidth]{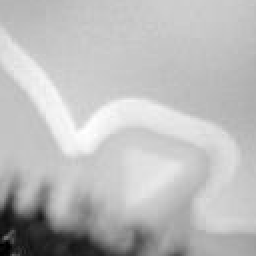} & 
        \includegraphics[width=0.195\linewidth]{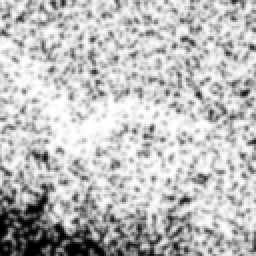} & 
        \includegraphics[width=0.195\linewidth]{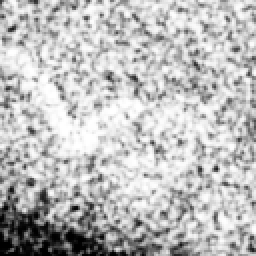} & 
        \includegraphics[width=0.195\linewidth]{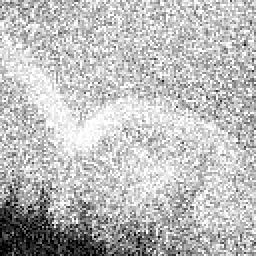} & 
        \includegraphics[width=0.195\linewidth]{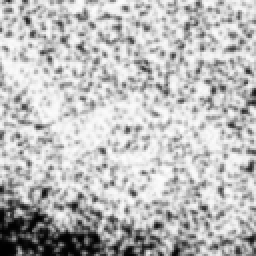} \\
        \raisebox{19pt}{\(r=0.8\)} & 
        \includegraphics[width=0.195\linewidth]{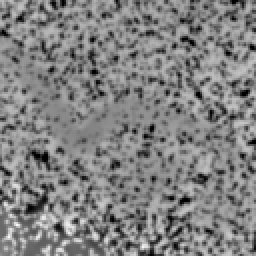} & 
        \includegraphics[width=0.195\linewidth]{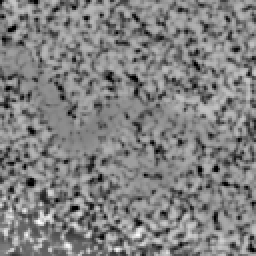} & 
        \includegraphics[width=0.195\linewidth]{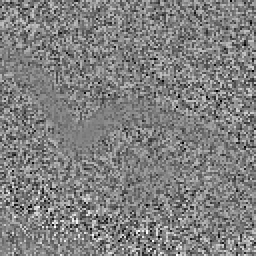} & 
        \includegraphics[width=0.195\linewidth]{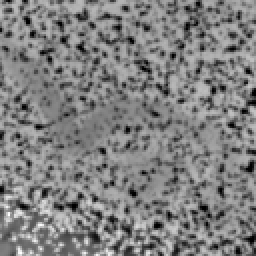} \\
    \end{tabular}
    \caption{\textbf{Uncurated film-grain comparison.} PGNM and Zhang23~\cite{zhang2023grain} add grain to the shared training-fitted blur \(\mu=B\ast x\). Second row of each pair: residual relative to \(\mu\), amplified for visibility.}
    \label{fig:main_newson_vis}
\end{figure}

\paragraph{Simulated film grain.}
Because the Boolean renderer is a physical model, we can compare PGNM with Zhang23's analytic approximation of it (Tab.~\ref{tab:film}). For small grain, Zhang23 is on par with PGNM, and up to $r=0.6$ it is slightly closer in spatial correlation, as its isotropic correlation model follows the renderer's isotropic grain; as grains become resolved, the approximation degrades, and at the coarsest radius it no longer reproduces the renderer (Fig.~\ref{fig:main_newson_vis}). Without a physical model, PGNM has the lowest conditional $W_1$ from a radius of 0.075 upward. GrainNet matches the overall appearance best at coarse radii, where its spatial correlation error is also slightly lower than PGNM's (supplementary). Its output, however, is offset from the expected image by up to 7 codes at every radius, and its finest grain is too strong. The residual errors register this offset, while the appearance distances largely tolerate it.

\paragraph{Model size and speed.}
A PGNM model stores the $25\times25\times3\times3$ convolution tensor (5,625
values) and a $3\times256\times25$ residual lookup (19,200 values;
Tab.~\ref{tab:cost}); on grayscale film, only 7,025 of its values per
radius are distinct. One model is 5.7--1{,}250$\times$ smaller than GMDCN,
NAFlow, DANet, and NM-FlowGAN, but larger than sRGB-Flow and GrainNet, and
because PGNM fits one model per condition, its total over the 16 SIDD classes
(397,200) exceeds GMDCN. Drawing and filtering a new noise field for every frame, as the
baselines do, PGNM is 5--19$\times$ faster than the neural baselines on GPU and
3.6--57$\times$ faster on CPU; on CPU, Zhang23 remains
cheaper. Because the filter does not depend on the image, all reported results read a
random window from four templates filtered once per model and resolution
(Sec.~\ref{sec:spatial_color}), as codec grain synthesis reuses a grain
template~\cite{norkin2018filmgrain}. Counting the time to build it, this bank is
faster than per-frame filtering after about 30 frames on GPU and 24 on CPU.

\begin{table}[t]
\centering
\small
\setlength{\tabcolsep}{4.5pt}
\begin{tabular}{@{}lrrrr@{}}
\toprule
& \multicolumn{2}{c}{Parameters} & \multicolumn{2}{c}{Time per frame} \\
\cmidrule(lr){2-3}\cmidrule(l){4-5}
Method & Per model & Models & GPU ms & CPU s \\
\midrule
\multicolumn{5}{@{}l}{\emph{SIDD (16 camera--ISO classes)}} \\
sRGB-Flow$^\dagger$~\cite{kousha2022srgb} & 6,160 & 1 & 112.1 & 3.79 \\
GMDCN~\cite{song2023camera} & 140,688 & 1 & 395.1 & 35.39 \\
NAFlow~\cite{kim2024naflow} & 1,078,091 & 1 & 105.0 & 10.97 \\
DANet~\cite{yue2020danet} & 9,153,859 & 1 & 163.7 & 16.04 \\
NM-FlowGAN$^\dagger$~\cite{han2023nmflowgan} & 31,057,359 & 1 & 220.0 & 59.47 \\
PGNM (ours) & 24,825 & 16 & & \\
\quad new field per frame & & & 20.4 & 1.04 \\
\quad bank playback & & & 10.2 & 0.051 \\
\quad bank construction & & & 307 & 23.45 \\
\midrule
\multicolumn{5}{@{}l}{\emph{Boolean film grain (8 radii)}} \\
Zhang23~\cite{zhang2023grain} & 51 & 8 & -- & 0.59 \\
GrainNet~\cite{lesne2025grain} & 9,857 & 1 & 118.2 & 8.62 \\
PGNM (ours) & 24,874 & 8 & & \\
\bottomrule
\end{tabular}
\caption{\textbf{Model size and speed.} Parameters: learned values read at
inference (neural baselines: the noise generator; PGNM's fixed knot grid
not counted; film rows include the shared $7\times7$ blur). Models: separately
fitted models. Time: one $1920\times1080$ RGB frame, 8-bit clean to noisy image
in host memory (median, batch 1, fp32), on one RTX A6000 (GPU) or two
i9-14900K efficiency-core threads (CPU); setup details in the supplementary
material. $^\dagger$Trained on NM-FlowGAN's 15 camera--ISO classes, 7 of the 16 evaluated. --: no GPU implementation.}
\label{tab:cost}
\end{table}
\paragraph{Grain synthesis for video delivery.}
Netflix's \emph{Nocturne}~\cite{netflixnocturne} has a clean master and a
film-grain version. In a prototype outside any codec, we code 100 frames of the
grain version, or code the master and add PGNM grain after decoding
(Tab.~\ref{tab:bandwidth}). Coding the grain costs 9--14$\times$ the bits at equal
quantizer (Q\,15--31). Even at their highest lossy rates, both encoders stay
farther from the grain version in DISTS than PGNM at 0.94\,Mbps. They have lower
spatial error, and VVC slightly lower LPIPS. Matched DISTS saves at least 90\%
(85\% with nlmeans denoising). At the same rate, PGNM is also closer than AV1's
grain synthesis in DISTS, LPIPS and $W_1$, with nearly equal spatial error. Sent
once, its model costs less than AV1's parameters beyond about 20\,s.

\begin{table}[t]
\centering
\footnotesize
\setlength{\tabcolsep}{3.2pt}
\begin{tabular}{@{}llrcccc@{}}
\toprule
Encoder & Grain & Mbps & DISTS$\downarrow$ & LPIPS$\downarrow$ & $W_1\downarrow$ & Spatial$\downarrow$ \\
\midrule
AV1, Q\,15 & coded & 73.0 & 0.089 & 0.102 & 1.30 & 0.056 \\
VVC, QP\,17 & coded & 94.5 & 0.091 & 0.085 & 1.21 & 0.046 \\
AV1, Q\,31 & AV1~\cite{norkin2018filmgrain} & 0.92 & 0.098 & 0.118 & 1.79 & 0.064 \\
VVC, 2-pass & VVC~\cite{menon2024vvc} & 0.99 & 0.201 & 0.250 & 1.09 & 0.076 \\
AV1, Q\,31 & \textbf{PGNM} & 1.63 & 0.055 & 0.080 & 0.73 & 0.037 \\
AV1, Q\,55 & \textbf{PGNM} & 0.94 & 0.069 & 0.086 & 0.94 & 0.065 \\
\bottomrule
\end{tabular}
\caption{\textbf{Video delivery} (\emph{Nocturne}, 100 frames, 1080p60; VVC:
VVenC~\cite{wieckowski2021vvenc}). Coded: grain encoded, temporal filter on or off,
the closer one. Otherwise synthesized after decoding (VVC: estimated from
the grain version); PGNM's rate includes its 161\,KB model.}
\label{tab:bandwidth}
\end{table}

\begin{figure*}[t]
\centering
\setlength{\tabcolsep}{0pt}
\begin{tabular*}{\linewidth}{@{\extracolsep{\fill}}cccccc@{}}
\begin{tabular}[b]{@{}c@{}}{\footnotesize Kernel scores}\\[1pt]\includegraphics[height=0.92in]{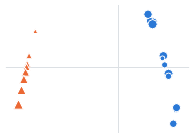}\end{tabular} & \begin{tabular}[b]{@{}c@{}}{\footnotesize Kernel PC1 (96\%)}\\[1pt]\includegraphics[height=0.92in]{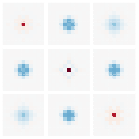}\end{tabular} & \begin{tabular}[b]{@{}c@{}}{\footnotesize Kernel PC2 (2\%)}\\[1pt]\includegraphics[height=0.92in]{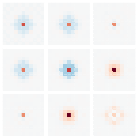}\end{tabular} &
\begin{tabular}[b]{@{}c@{}}{\footnotesize Lookup scores}\\[1pt]\includegraphics[height=0.92in]{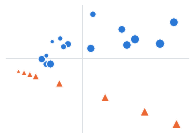}\end{tabular} & \begin{tabular}[b]{@{}c@{}}{\footnotesize Lookup PC1 (82\%)}\\[1pt]\includegraphics[height=0.92in]{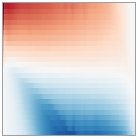}\end{tabular} & \begin{tabular}[b]{@{}c@{}}{\footnotesize Lookup PC2 (11\%)}\\[1pt]\includegraphics[height=0.92in]{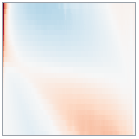}\end{tabular} \\
\end{tabular*}
\caption{\textbf{Joint PCA of the 24 fitted presets}: convolution tensors (left) and residual lookups (right). Scores in PC1 (horizontal) and PC2: SIDD classes as \textcolor[HTML]{2A78D6}{blue circles} and Newson radii as \textcolor[HTML]{EB6834}{orange triangles}, larger for higher ISO or radius. Kernel PC1 and lookup PC2 separate grain from camera noise; kernel PC2 groups SIDD by camera, and lookup PC1 orders presets by strength. Components: central $11\times11$ taps (rows output, columns input) and the green lookup over clean intensity (horizontal) and latent value ($-3$ to 3); red positive, blue negative.}
\label{fig:kernel_pca}
\end{figure*}

\section{Ablation}
\label{sec:ablation}
We ablate one fitted component at a time on SIDD, and the kernel support also on film grain (Tab.~\ref{tab:ablation}). Each variant is refitted on the same pairs and rendered from the same random streams, so differences come from that component alone; the full variant reproduces the main-table PGNM scores. The ablation was specified after the main evaluation, before its outputs existed, and did not change the method.

\paragraph{Fitting masks.}
The lookup does not use the fitting masks, so removing them changes only the correlation fit. Without the flat mask, textured pixels enter that fit and weaken the fitted cross-channel dependence between neighboring pixels, which raises color correlation error and, slightly, KL. The flat mask alone recovers full PGNM; the clipping guard alone does not (supplementary material).

\paragraph{Kernel support.}
Spatial support is essential, but the fitted dependence is compact. On SIDD, the errors level off by $5\times5$. On Newson film grain, $1\times1$ and $3\times3$ supports lose the grain's spatial structure and appearance, increasingly so for coarser grain, whereas $5\times5$ and $11\times11$ match $25\times25$ up to the coarsest radius, with slightly lower spatial correlation error, as on SIDD. Distribution measures improve slightly as the support shrinks (KL on SIDD, conditional $W_1$ on film grain): spatially independent noise makes each image's empirical distribution fluctuate less, and these measures do not see dependence. We nevertheless fixed a generous $25\times25$ support in advance for every dataset, leaving room for longer-range dependence (correlations are estimated up to 16 pixels) instead of tuning the support on evaluation data. The margin costs little: all timings in Tab.~\ref{tab:cost} use $25\times25$, and with template playback the support affects only the one-time filtering. In deployment, a smaller support such as $11\times11$, with five times fewer kernel values, is a free option. The fitted presets are also highly structured: across all 24 presets, two principal components explain 98\% of the kernel and 93\% of the lookup variation (Fig.~\ref{fig:kernel_pca}).

\paragraph{Conditional distribution.}
Replacing the lookup by a Gaussian with the same mean and variance at each clean intensity more than doubles KL and conditional $W_1$, while the correlation errors stay tied: a Gaussian cannot reproduce the skewed, heavy-tailed shape of SIDD residuals at a fixed intensity (Fig.~\ref{fig:condpdf}), which the lookup preserves.

\providecommand{\abltie}[1]{\textcolor{black!45}{#1}}
\providecommand{\ablworse}[1]{\textcolor[HTML]{B2182B}{#1}}
\providecommand{\ablbetter}[1]{\textcolor[HTML]{1B7837}{#1}}
\begin{table}[t]
\centering
\small
\setlength{\tabcolsep}{3.9pt}
\begin{tabular}{@{}lrrrr@{}}
\toprule
\emph{SIDD} & $D_{\mathrm{KL}}\downarrow$ & $W_1\downarrow$ & Spatial$\downarrow$ & Color$\downarrow$ \\
\midrule
Full PGNM & 0.0219 & 1.32 & 0.0094 & 0.0144 \\
No fitting masks & \ablworse{$+$1.2\%} & \abltie{$+$0.0\%} & \abltie{$-$1.7\%} & \ablworse{$+$46\%} \\
Kernel $1\times1$ & \ablbetter{$-$5.4\%} & \abltie{$+$0.0\%} & \ablworse{$+$586\%} & \ablworse{$+$770\%} \\
Kernel $5\times5$ & \ablbetter{$-$0.8\%} & \abltie{$+$0.2\%} & \abltie{$-$6.5\%} & \abltie{$+$0.1\%} \\
Kernel $11\times11$ & \ablbetter{$-$0.9\%} & \abltie{$+$0.1\%} & \ablbetter{$-$8.5\%} & \abltie{$+$1.2\%} \\
Gaussian conditional & \ablworse{$+$138\%} & \ablworse{$+$108\%} & \abltie{$+$4.5\%} & \abltie{$+$2.0\%} \\
\midrule
\emph{Film grain} & $W_1\downarrow$ & Spatial$\downarrow$ & DISTS$\downarrow$ & LPIPS$\downarrow$ \\
\midrule
Full PGNM & 0.626 & 0.0082 & 0.0731 & 0.0809 \\
Kernel $1\times1$ & \ablbetter{$-$11\%} & \ablworse{$+$1164\%} & \ablworse{$+$193\%} & \ablworse{$+$230\%} \\
Kernel $5\times5$ & \abltie{$-$0.2\%} & \ablbetter{$-$8.6\%} & \abltie{$+$0.2\%} & \abltie{$-$0.2\%} \\
Kernel $11\times11$ & \abltie{$-$0.2\%} & \ablbetter{$-$5.9\%} & \abltie{$+$0.2\%} & \abltie{$+$0.0\%} \\
\bottomrule
\end{tabular}
\caption{\textbf{Ablation.} Full PGNM's scores, then each variant's change (lower is better): \ablworse{worse}, \ablbetter{better}, or \abltie{tie}. SIDD (39 sources; $D_{\mathrm{KL}}$ on $32\times32$ tiles): tie when the 95\% scene-bootstrap interval includes zero. Film grain (mean over eight Newson radii): colored when most of the 60 images change in that direction at every radius. Further variants and intervals are in the supplementary material.}
\label{tab:ablation}
\end{table}

\section{Limitations and Conclusion}
\label{sec:ending}

\paragraph{Limitations.}
PGNM represents only what its two components express: a stationary latent filter and a lookup that sees the clean image through each pixel's intensity alone. Higher-capacity generators can learn residual structure that follows image content, varies between captures, or has dependence far from Gaussian; PGNM cannot. On SIDD, part of the residual follows image edges: if this stems from references averaged over many captures and thus softer than the noisy images, PGNM avoids fitting a dataset artifact; if it is a real noise pattern, PGNM misses it. A gap to real noise remains: training a denoiser on PGNM noise falls short of training on real pairs (Tab.~\ref{tab:denoise_sidd}), and the independent Newson render is closer to the reference on every measure (Tab.~\ref{tab:film}). Each condition needs its own fit from aligned clean/noisy pairs; one-shot fitting is not yet supported.

\paragraph{Conclusion.}
The idea behind PGNM is to model noise only as far as it can be measured. Each component is fitted to an observable statistic, so a model can be checked component by component, and camera noise and film grain become points in one parameter space (Fig.~\ref{fig:kernel_pca}). Two such measurements already account for much of the realism of far larger generators; what they miss marks where learned capacity is needed.

\section*{AI use disclosure}
This work used Claude (Anthropic) through Claude Code as an assistant for software
development (the PGNM implementation, the experiment, evaluation and figure scripts,
and the supplementary video) and for drafting and editing the text. The authors
directed its use, reviewed its output, and take full responsibility for the content.

{
    \small
    \bibliographystyle{ieeenat_fullname}
    \bibliography{references}

\begin{thebibliography}{51}
\providecommand{\natexlab}[1]{#1}
\providecommand{\url}[1]{\texttt{#1}}
\expandafter\ifx\csname urlstyle\endcsname\relax
  \providecommand{\doi}[1]{doi: #1}\else
  \providecommand{\doi}{doi: \begingroup \urlstyle{rm}\Url}\fi

\bibitem[Abdelhamed et~al.(2018)Abdelhamed, Lin, and Brown]{abdelhamed2018sidd}
Abdelrahman Abdelhamed, Stephen Lin, and Michael~S. Brown.
\newblock A high-quality denoising dataset for smartphone cameras.
\newblock In \emph{2018 IEEE/CVF Conference on Computer Vision and Pattern
  Recognition}, pages 1692--1700. IEEE, 2018.

\bibitem[Abdelhamed et~al.(2019{\natexlab{a}})Abdelhamed, Brubaker, and
  Brown]{abdelhamed2019noiseflow}
Abdelrahman Abdelhamed, Marcus~A. Brubaker, and Michael~S. Brown.
\newblock {Noise Flow}: Noise modeling with conditional normalizing flows.
\newblock In \emph{2019 IEEE/CVF International Conference on Computer Vision
  (ICCV)}, pages 3165--3173. IEEE, 2019{\natexlab{a}}.

\bibitem[Abdelhamed et~al.(2019{\natexlab{b}})Abdelhamed, Timofte, Brown,
  et~al.]{abdelhamed2019ntire}
Abdelrahman Abdelhamed, Radu Timofte, Michael~S. Brown, et~al.
\newblock {NTIRE} 2019 challenge on real image denoising: Methods and results.
\newblock In \emph{IEEE/CVF Conference on Computer Vision and Pattern
  Recognition Workshops (CVPRW)}, 2019{\natexlab{b}}.

\bibitem[Ameur et~al.(2023)Ameur, Hamidouche, Fran{\c{c}}ois,
  Radosavljevi{\'c}, Menard, and Demarty]{ameur2023deepfg}
Zoubida Ameur, Wassim Hamidouche, Edouard Fran{\c{c}}ois, Milo{\v{s}}
  Radosavljevi{\'c}, Daniel Menard, and Claire-H{\'e}l{\`e}ne Demarty.
\newblock {Deep-Based Film Grain Removal and Synthesis}.
\newblock \emph{IEEE Transactions on Image Processing}, 32:\penalty0
  5046--5059, 2023.

\bibitem[Ameur et~al.(2024)Ameur, Demarty, Menard, and {Le
  Meur}]{ameur2024threeR}
Zoubida Ameur, Claire-H{\'e}l{\`e}ne Demarty, Daniel Menard, and Olivier {Le
  Meur}.
\newblock {3R-INN: How to be climate friendly while consuming/delivering
  videos?}
\newblock In \emph{European Conference on Computer Vision (ECCV)}, 2024.

\bibitem[Ameur et~al.(2025)Ameur, Demarty, Le~Meur, and
  M{\'e}nard]{ameur2025stylefg}
Zoubida Ameur, Claire-H{\'e}l{\`e}ne Demarty, Olivier Le~Meur, and Daniel
  M{\'e}nard.
\newblock {Style-FG: A Style-based Framework for Film Grain Analysis and
  Synthesis}.
\newblock \emph{ACM Transactions on Multimedia Computing, Communications, and
  Applications}, 21\penalty0 (9):\penalty0 1--24, 2025.

\bibitem[Brooks et~al.(2019)Brooks, Mildenhall, Xue, Chen, Sharlet, and
  Barron]{brooks2019unprocessing}
Tim Brooks, Ben Mildenhall, Tianfan Xue, Jiawen Chen, Dillon Sharlet, and
  Jonathan~T. Barron.
\newblock Unprocessing images for learned raw denoising.
\newblock In \emph{2019 IEEE/CVF Conference on Computer Vision and Pattern
  Recognition (CVPR)}, pages 11028--11037. IEEE, 2019.

\bibitem[Cai et~al.(2021)Cai, Hu, Wang, Zhang, Pfister, and Wei]{cai2021pngan}
Yuanhao Cai, Xiaowan Hu, Haoqian Wang, Yulun Zhang, Hanspeter Pfister, and
  Donglai Wei.
\newblock {Learning to Generate Realistic Noisy Images via Pixel-level
  Noise-aware Adversarial Training}.
\newblock In \emph{Advances in Neural Information Processing Systems}, 2021.

\bibitem[Cario and Nelson(1996)]{cario1996arta}
Marne~C. Cario and Barry~L. Nelson.
\newblock {Autoregressive to Anything: Time-Series Input Processes for
  Simulation}.
\newblock \emph{Operations Research Letters}, 19\penalty0 (2):\penalty0 51--58,
  1996.

\bibitem[Chang et~al.(2020)Chang, Wang, Lin, Liu, Chen, Chang, and
  Chen]{chang2020camera}
Ke-Chi Chang, Ren Wang, Hung-Jin Lin, Yu-Lun Liu, Chia-Ping Chen, Yu-Lin Chang,
  and Hwann-Tzong Chen.
\newblock {Learning Camera-Aware Noise Models}.
\newblock In \emph{European Conference on Computer Vision (ECCV)}, 2020.

\bibitem[Colom and Buades(2013)]{colom2013analysis}
Miguel Colom and Antoni Buades.
\newblock Analysis and extension of the {Ponomarenko} et al. method, estimating
  a noise curve from a single image.
\newblock \emph{Image Processing On Line}, 3:\penalty0 173--197, 2013.

\bibitem[Ding et~al.(2022)Ding, Ma, Wang, and Simoncelli]{ding2022dists}
Keyan Ding, Kede Ma, Shiqi Wang, and Eero~P. Simoncelli.
\newblock Image quality assessment: Unifying structure and texture similarity.
\newblock \emph{IEEE Transactions on Pattern Analysis and Machine
  Intelligence}, 44\penalty0 (5):\penalty0 2567--2581, 2022.

\bibitem[Foi et~al.(2008)Foi, Trimeche, Katkovnik, and
  Egiazarian]{foi2008noise}
Alessandro Foi, Mejdi Trimeche, Vladimir Katkovnik, and Karen Egiazarian.
\newblock {Practical Poissonian-Gaussian Noise Modeling and Fitting for
  Single-Image Raw-Data}.
\newblock \emph{IEEE Transactions on Image Processing}, 17\penalty0
  (10):\penalty0 1737--1754, 2008.

\bibitem[{FotoKem}(2023)]{fotokem2023oppenheimer}
{FotoKem}.
\newblock {DP Van Hoytema and FotoKem talk 65mm and Oppenheimer}, 2023.
\newblock Published November 22, 2023. Accessed October 13, 2026.

\bibitem[{Foundry}()]{foundrygrain}
{Foundry}.
\newblock \emph{Applying Grain}.
\newblock Nuke 16.0 documentation. Accessed October 13, 2026.

\bibitem[Fu et~al.(2023)Fu, Guo, and Wen]{fu2023neca}
Zixuan Fu, Lanqing Guo, and Bihan Wen.
\newblock {sRGB Real Noise Synthesizing with Neighboring Correlation-Aware
  Noise Model}.
\newblock In \emph{IEEE/CVF Conference on Computer Vision and Pattern
  Recognition (CVPR)}, pages 1683--1691, 2023.

\bibitem[Galerne et~al.(2011)Galerne, Gousseau, and Morel]{galerne2011random}
Bruno Galerne, Yann Gousseau, and Jean-Michel Morel.
\newblock {Random Phase Textures: Theory and Synthesis}.
\newblock \emph{IEEE Transactions on Image Processing}, 20\penalty0
  (1):\penalty0 257--267, 2011.

\bibitem[Han and Yu(2023)]{han2023nmflowgan}
Young~Joo Han and Ha-Jin Yu.
\newblock {NM-FlowGAN: Modeling sRGB Noise without Paired Images using a Hybrid
  Approach of Normalizing Flows and GAN}, 2023.
\newblock arXiv:2312.10112.

\bibitem[Heeger and Bergen(1995)]{heeger1995pyramid}
David~J. Heeger and James~R. Bergen.
\newblock {Pyramid-Based Texture Analysis/Synthesis}.
\newblock In \emph{Proceedings of the 22nd Annual Conference on Computer
  Graphics and Interactive Techniques (SIGGRAPH)}, pages 229--238, 1995.

\bibitem[Jang et~al.(2021)Jang, Lee, Son, and Lee]{jang2021c2n}
Geonwoon Jang, Wooseok Lee, Sanghyun Son, and Kyoung~Mu Lee.
\newblock {C2N: Practical Generative Noise Modeling for Real-World Denoising}.
\newblock In \emph{IEEE/CVF International Conference on Computer Vision
  (ICCV)}, pages 2350--2359, 2021.

\bibitem[Khintchine(1934)]{khintchine1934korrelationstheorie}
A. Khintchine.
\newblock {Korrelationstheorie der station\"aren stochastischen Prozesse}.
\newblock \emph{Mathematische Annalen}, 109\penalty0 (1):\penalty0 604--615,
  1934.

\bibitem[Kim et~al.(2026)Kim, Lee, and Yoo]{kim2026guidnoise}
Changjin Kim, HyeokJun Lee, and YoungJoon Yoo.
\newblock {GuidNoise: Single-Pair Guided Diffusion for Generalized Noise
  Synthesis}.
\newblock \emph{Proceedings of the AAAI Conference on Artificial Intelligence},
  40\penalty0 (7):\penalty0 5639--5647, 2026.

\bibitem[Kim et~al.(2024)Kim, Jung, Baik, and Kim]{kim2024naflow}
Dongjin Kim, Donggoo Jung, Sungyong Baik, and Tae~Hyun Kim.
\newblock {sRGB Real Noise Modeling via Noise-Aware Sampling with Normalizing
  Flows}.
\newblock In \emph{International Conference on Learning Representations
  (ICLR)}, 2024.

\bibitem[Ko et~al.(2026{\natexlab{a}})Ko, Kim, Lee, Wang, and Kim]{ko2026png}
Jaekyun Ko, Dongjin Kim, Soomin Lee, Guanghui Wang, and Tae~Hyun Kim.
\newblock {Diffusion-Based sRGB Real Noise Generation via Prompt-Driven Noise
  Representation Learning}.
\newblock In \emph{IEEE/CVF Conference on Computer Vision and Pattern
  Recognition (CVPR)}, pages 35956--35966, 2026{\natexlab{a}}.

\bibitem[Ko et~al.(2026{\natexlab{b}})Ko, Lim, Lee, Kim, and Kim]{ko2026yeti}
Jaekyun Ko, Byung~Wan Lim, Soomin Lee, Dongjin Kim, and Tae~Hyun Kim.
\newblock {YeTI: You Only Need Two Noisy Images for Real-World sRGB Noise
  Generation}.
\newblock In \emph{European Conference on Computer Vision (ECCV)},
  2026{\natexlab{b}}.

\bibitem[Kousha et~al.(2022)Kousha, Maleky, Brown, and
  Brubaker]{kousha2022srgb}
Shayan Kousha, Ali Maleky, Michael~S. Brown, and Marcus~A. Brubaker.
\newblock {Modeling sRGB Camera Noise with Normalizing Flows}.
\newblock In \emph{IEEE/CVF Conference on Computer Vision and Pattern
  Recognition (CVPR)}, pages 17463--17471, 2022.

\bibitem[Lesn{\'e} et~al.(2025)Lesn{\'e}, Gousseau, Ladjal, and
  Newson]{lesne2025grain}
Gwilherm Lesn{\'e}, Yann Gousseau, Sa{\"i}d Ladjal, and Alasdair Newson.
\newblock Neural film grain rendering.
\newblock \emph{Computer Graphics Forum}, 44\penalty0 (2):\penalty0 e70076,
  2025.

\bibitem[Liu et~al.(2013)Liu, Tanaka, and Okutomi]{liu2013single}
Xinhao Liu, Masayuki Tanaka, and Masatoshi Okutomi.
\newblock Single-image noise level estimation for blind denoising.
\newblock \emph{IEEE Transactions on Image Processing}, 22\penalty0
  (12):\penalty0 5226--5237, 2013.

\bibitem[Liu et~al.(2021)Liu, Qin, Anwar, Ji, Kim, Caldwell, and
  Gedeon]{liu2021invdn}
Yang Liu, Zhenyue Qin, Saeed Anwar, Pan Ji, Dongwoo Kim, Sabrina Caldwell, and
  Tom Gedeon.
\newblock {Invertible Denoising Network: A Light Solution for Real Noise
  Removal}.
\newblock In \emph{IEEE/CVF Conference on Computer Vision and Pattern
  Recognition (CVPR)}, pages 13365--13374, 2021.

\bibitem[Maleky et~al.(2022)Maleky, Kousha, Brown, and
  Brubaker]{maleky2022n2nflow}
Ali Maleky, Shayan Kousha, Michael~S. Brown, and Marcus~A. Brubaker.
\newblock {Noise2NoiseFlow: Realistic Camera Noise Modeling without Clean
  Images}.
\newblock In \emph{IEEE/CVF Conference on Computer Vision and Pattern
  Recognition (CVPR)}, 2022.

\bibitem[Menon et~al.(2024)Menon, Wieckowski, Stoffers, Brandenburg, Lehmann,
  Bross, Schierl, and Marpe]{menon2024vvc}
Vignesh~V Menon, Adam Wieckowski, Christian Stoffers, Jens Brandenburg,
  Christian Lehmann, Benjamin Bross, Thomas Schierl, and Detlev Marpe.
\newblock {Enhancing Film Grain Coding in VVC: Improving Encoding Quality and
  Efficiency}, 2024.
\newblock arXiv:2407.12465; accepted at IBC 2024.

\bibitem[{Netflix}(2018)]{netflixnocturne}
{Netflix}.
\newblock {Nocturne}.
\newblock Netflix Open Content, Creative Commons Attribution 4.0, 2018.
\newblock Clean master and film-grain version. Accessed October 21, 2026.

\bibitem[Newson et~al.(2017)Newson, Delon, and Galerne]{newson2017stochastic}
Alasdair Newson, Julie Delon, and Bruno Galerne.
\newblock {A Stochastic Film Grain Model for Resolution-Independent Rendering}.
\newblock \emph{Computer Graphics Forum}, 36\penalty0 (8):\penalty0 684--699,
  2017.

\bibitem[Norkin and Birkbeck(2018)]{norkin2018filmgrain}
Andrey Norkin and Neil Birkbeck.
\newblock Film grain synthesis for {AV1} video codec.
\newblock In \emph{2018 Data Compression Conference}, pages 3--12. IEEE, 2018.

\bibitem[Papoulis and Pillai(2002)]{papoulis2002probability}
Athanasios Papoulis and S.~Unnikrishna Pillai.
\newblock \emph{{Probability, Random Variables, and Stochastic Processes}}.
\newblock McGraw-Hill, 4th edition, 2002.

\bibitem[Shinozuka(1971)]{shinozuka1971simulation}
Masanobu Shinozuka.
\newblock {Simulation of Multivariate and Multidimensional Random Processes}.
\newblock \emph{The Journal of the Acoustical Society of America}, 49\penalty0
  (1B):\penalty0 357--368, 1971.

\bibitem[Song et~al.(2023)Song, Zhang, Ayd{\i}n, Mansour, and
  Schroers]{song2023camera}
Mingyang Song, Yang Zhang, Tun{\c{c}}~O. Ayd{\i}n, Elham~Amin Mansour, and
  Christopher Schroers.
\newblock {A Generative Model for Digital Camera Noise Synthesis}, 2023.
\newblock arXiv:2303.09199.

\bibitem[Tian and Pappas(2026)]{tian2026grain}
Daizong Tian and Thrasyvoulos~N. Pappas.
\newblock {Efficient Digital Film Grain Simulation}.
\newblock \emph{IEEE Transactions on Image Processing}, 35:\penalty0
  8851--8862, 2026.

\bibitem[Wei et~al.(2020)Wei, Fu, Yang, and Huang]{wei2020physical}
Kaixuan Wei, Ying Fu, Jiaolong Yang, and Hua Huang.
\newblock {A Physics-based Noise Formation Model for Extreme Low-light Raw
  Denoising}.
\newblock In \emph{IEEE/CVF Conference on Computer Vision and Pattern
  Recognition (CVPR)}, 2020.

\bibitem[Wieckowski et~al.(2021)Wieckowski, Brandenburg, Hinz, Bartnik, George,
  Hege, Helmrich, Henkel, Lehmann, Stoffers, Zupancic, Bross, and
  Marpe]{wieckowski2021vvenc}
Adam Wieckowski, Jens Brandenburg, Tobias Hinz, Christian Bartnik, Valeri
  George, Gabriel Hege, Christian Helmrich, Anastasia Henkel, Christian
  Lehmann, Christian Stoffers, Ivan Zupancic, Benjamin Bross, and Detlev Marpe.
\newblock {VVenC}: An open and optimized {VVC} encoder implementation.
\newblock In \emph{2021 IEEE International Conference on Multimedia \& Expo
  Workshops (ICMEW)}, pages 1--2. IEEE, 2021.

\bibitem[Wiener(1930)]{wiener1930generalized}
Norbert Wiener.
\newblock {Generalized Harmonic Analysis}.
\newblock \emph{Acta Mathematica}, 55:\penalty0 117--258, 1930.

\bibitem[Wu et~al.(2025)Wu, Han, Jiang, Jiang, Luo, Jiang, Fan, and
  Liu]{wu2025rnsd}
Qi Wu, Mingyan Han, Ting Jiang, Chengzhi Jiang, Jinting Luo, Man Jiang,
  Haoqiang Fan, and Shuaicheng Liu.
\newblock {Realistic Noise Synthesis with Diffusion Models}.
\newblock \emph{Proceedings of the AAAI Conference on Artificial Intelligence},
  39\penalty0 (8):\penalty0 8432--8440, 2025.

\bibitem[Yamazaki and Shinozuka(1988)]{yamazaki1988digital}
Fumio Yamazaki and Masanobu Shinozuka.
\newblock {Digital Generation of Non-Gaussian Stochastic Fields}.
\newblock \emph{Journal of Engineering Mechanics}, 114\penalty0 (7):\penalty0
  1183--1197, 1988.

\bibitem[Yang et~al.(2025)Yang, Wang, and Wang]{yang2025degradation}
Wenbo Yang, Zhongling Wang, and Zhou Wang.
\newblock {Towards a Universal Image Degradation Model via Content-Degradation
  Disentanglement}.
\newblock In \emph{IEEE/CVF International Conference on Computer Vision
  (ICCV)}, 2025.

\bibitem[Yue et~al.(2020)Yue, Zhao, Zhang, and Meng]{yue2020danet}
Zongsheng Yue, Qian Zhao, Lei Zhang, and Deyu Meng.
\newblock {Dual Adversarial Network: Toward Real-world Noise Removal and Noise
  Generation}.
\newblock In \emph{European Conference on Computer Vision (ECCV)}, 2020.

\bibitem[Zamir et~al.(2020)Zamir, Arora, Khan, Hayat, Khan, Yang, and
  Shao]{zamir2020cycleisp}
Syed~Waqas Zamir, Aditya Arora, Salman Khan, Munawar Hayat, Fahad~Shahbaz Khan,
  Ming-Hsuan Yang, and Ling Shao.
\newblock {CycleISP: Real Image Restoration via Improved Data Synthesis}.
\newblock In \emph{IEEE/CVF Conference on Computer Vision and Pattern
  Recognition (CVPR)}, 2020.

\bibitem[Zhang et~al.(2017)Zhang, Zuo, Chen, Meng, and Zhang]{zhang2017beyond}
Kai Zhang, Wangmeng Zuo, Yunjin Chen, Deyu Meng, and Lei Zhang.
\newblock Beyond a {G}aussian denoiser: Residual learning of deep {CNN} for
  image denoising.
\newblock \emph{IEEE Transactions on Image Processing}, 26\penalty0
  (7):\penalty0 3142--3155, 2017.

\bibitem[Zhang et~al.(2023)Zhang, Wang, Tian, and Pappas]{zhang2023grain}
Kaixuan Zhang, Jingxian Wang, Daizong Tian, and Thrasyvoulos~N. Pappas.
\newblock Film grain rendering and parameter estimation.
\newblock \emph{ACM Transactions on Graphics}, 42\penalty0 (4), 2023.

\bibitem[Zhang et~al.(2018)Zhang, Isola, Efros, Shechtman, and
  Wang]{zhang2018lpips}
Richard Zhang, Phillip Isola, Alexei~A. Efros, Eli Shechtman, and Oliver Wang.
\newblock The unreasonable effectiveness of deep features as a perceptual
  metric.
\newblock In \emph{IEEE/CVF Conference on Computer Vision and Pattern
  Recognition (CVPR)}, pages 586--595, 2018.

\bibitem[Zheng et~al.(2024)Zheng, Zou, Zhang, and Bao]{zheng2024senm}
Dihan Zheng, Yihang Zou, Xiaowen Zhang, and Chenglong Bao.
\newblock {SeNM-VAE: Semi-Supervised Noise Modeling with Hierarchical
  Variational Autoencoder}.
\newblock In \emph{IEEE/CVF Conference on Computer Vision and Pattern
  Recognition (CVPR)}, pages 25889--25899, 2024.

\bibitem[Zou and Fu(2022)]{zou2022noise}
Yunhao Zou and Ying Fu.
\newblock {Estimating Fine-Grained Noise Model via Contrastive Learning}.
\newblock In \emph{IEEE/CVF Conference on Computer Vision and Pattern
  Recognition (CVPR)}, pages 12682--12691, 2022.

\end{thebibliography}
}

\clearpage
\appendix
\setcounter{figure}{0}
\setcounter{table}{0}
\setcounter{algorithm}{0}
\setcounter{equation}{0}
\renewcommand{\thesection}{\Alph{section}}
\renewcommand{\thefigure}{S\arabic{figure}}
\renewcommand{\thetable}{S\arabic{table}}
\renewcommand{\thealgorithm}{S\arabic{algorithm}}
\renewcommand{\theequation}{S\arabic{equation}}
\def\theHsection{supp.\Alph{section}}
\def\theHfigure{supp.S\arabic{figure}}
\def\theHtable{supp.S\arabic{table}}
\def\theHalgorithm{supp.S\arabic{algorithm}}
\def\theHequation{supp.S\arabic{equation}}
\renewcommand{\floatpagefraction}{0.8}
\renewcommand{\dblfloatpagefraction}{0.8}
\makeatletter
\setlength{\@fptop}{0pt}
\setlength{\@fpsep}{12pt plus 6pt}
\setlength{\@fpbot}{0pt plus 1fil}
\setlength{\@dblfptop}{0pt}
\setlength{\@dblfpsep}{12pt plus 6pt}
\setlength{\@dblfpbot}{0pt plus 1fil}
\makeatother
\maketitlesupplementary

\pdfbookmark[0]{Explanation and evidence}{supp-evidence}
\begin{strip}
\noindent
\begin{minipage}[t]{0.48\textwidth}
\textbf{Explanation and evidence}\par\smallskip
\small
\hyperref[sec:supp-equations]{\ref*{sec:supp-equations}. PGNM in equations}\hfill\pageref{sec:supp-equations}\par
\hyperref[sec:supp-ablation]{\ref*{sec:supp-ablation}. Ablation details}\hfill\pageref{sec:supp-ablation}\par
\hyperref[sec:supp-sidd-breakdown]{\ref*{sec:supp-sidd-breakdown}. Results by camera, ISO, and grain radius}\hfill\pageref{sec:supp-sidd-breakdown}\par
\hyperref[sec:supp-analyses]{\ref*{sec:supp-analyses}. Further analyses}\hfill\pageref{sec:supp-analyses}\par
\hyperref[sec:supp-denoising]{\ref*{sec:supp-denoising}. Denoising}\hfill\pageref{sec:supp-denoising}\par
\hyperref[sec:supp-bandwidth]{\ref*{sec:supp-bandwidth}. Video delivery}\hfill\pageref{sec:supp-bandwidth}\par
\hyperref[sec:supp-sidd-visual]{\ref*{sec:supp-sidd-visual}. Visual comparisons on SIDD}\hfill\pageref{sec:supp-sidd-visual}\par
\hyperref[sec:supp-film-visual]{\ref*{sec:supp-film-visual}. Visual comparisons on film grain}\hfill\pageref{sec:supp-film-visual}\par
\end{minipage}\hfill
\begin{minipage}[t]{0.48\textwidth}
\textbf{\hyperref[part:supp-reference]{Reproducibility reference}}\par\smallskip
\small
\hyperref[sec:supp-spec]{\ref*{sec:supp-spec}. Implementation specification}\hfill\pageref{sec:supp-spec}\par
\hyperref[sec:supp-metrics]{\ref*{sec:supp-metrics}. Metric estimators}\hfill\pageref{sec:supp-metrics}\par
\hyperref[sec:supp-baselines]{\ref*{sec:supp-baselines}. Baseline protocols}\hfill\pageref{sec:supp-baselines}\par
\hyperref[sec:supp-cost]{\ref*{sec:supp-cost}. Cost measurement setup}\hfill\pageref{sec:supp-cost}\par
\smallskip
For exact reimplementation, jump to this part for algorithms, numerical
conventions, data splits, random seeds, and evaluation procedures.
\end{minipage}
\end{strip}

\paragraph{Reading the scores.}
Conditional $W_1$ compares residual distributions at similar clean intensities;
spatial and color correlation errors compare dependence; tile KL compares local
residual histograms. Film DISTS and LPIPS compare overall image appearance.
Lower values are better for these scores. Exact estimators are in
Sec.~\ref{sec:supp-metrics}.
Numbers with the prefix S and lettered sections refer to this supplement, plain
numbers to the main paper. Image crops are shown without interpolation; please
zoom in.

\section{PGNM in equations}
\label{sec:supp-equations}

This section restates PGNM as plain formulas, one step per paragraph;
Sec.~\ref{sec:supp-spec} gives the same steps as exact algorithms. Images are in
$[0,1]$, $x$ is the clean image, $y$ the noisy one, $r=y-x$ the residual, $c$ a
color channel and $\mathbf p$ a pixel location.

\paragraph{Synthesis.}
Independent Gaussian values are filtered by a full RGB convolution and then mapped,
pixel by pixel, to signed residuals:
\begin{align}
    t_c &= \sum_{i\in\{R,G,B\}} A_{ci}\ast z_i,\qquad z_i(\mathbf p)\sim\mathcal N(0,1),\label{eq:supp-field}\\
    1 &= \sum_{i}\sum_{u,v}A_{ci}(u,v)^2\quad\text{for each }c,\label{eq:supp-energy}\\
    \hat y_c(\mathbf p) &= \operatorname{clip}\bigl(x_c(\mathbf p)+f_c\bigl(x_c(\mathbf p),t_c(\mathbf p)\bigr),\,0,\,1\bigr),\label{eq:supp-output}\\
    f_c(x,t) &= F_{c,x}^{-1}\bigl(\Phi(t)\bigr).\label{eq:supp-lookup}
\end{align}
The energy constraint~\eqref{eq:supp-energy} makes every $t_c(\mathbf p)$ standard Gaussian, so $\Phi(t_c)$ is
uniform and the residual at clean intensity $x$ follows the fitted distribution
$F_{c,x}$ whatever the filters, up to interpolation and clipping. In practice $t$ is read from a random cyclic
window of one of four precomputed fields (Alg.~\ref{alg:supp-render}).

\paragraph{Residual lookup.}
$F_{c,x}$ is the empirical distribution of the integer residual codes
$\operatorname{rint}(255\,r_c)$ of channel $c$ over \emph{all} training pixels whose
clean code lies within $\pm\rho$ codes of $\operatorname{rint}(255\,x)$, with $\rho\ge2$
the smallest radius that collects at least 2,000 samples (or spans all codes). The
lookup stores $f_c$ at the 256 clean codes and at 25 Gaussian values
$t\in\{-3,-2.75,\dots,3\}$, treating each residual code as a unit-width bin with
uniform mass (Alg.~\ref{alg:supp-lookup}); it is evaluated by bilinear interpolation,
with linear extrapolation in $t$ (Alg.~\ref{alg:supp-render}).

\paragraph{Flat pixels.}
The filters are fitted only where residuals are not mixed with image structure. With
$\kappa(\mathbf p)=\max_c\operatorname{sd}_7(x_c)(\mathbf p)$ the largest local standard
deviation over the channels in a $7\times7$ window,
\begin{align}
    \mathcal M_{\mathrm{flat}}&=\{\mathbf p:\kappa(\mathbf p)\le\tau\},\\
    \tau&=\operatorname{clip}\bigl(2\,q_{2\%}(\kappa),\,0.003,\,0.03\bigr),
\end{align}
where $q_{2\%}$ is the second percentile over the training images.

\begin{figure}[tp]
\centering
\includegraphics[width=\linewidth]{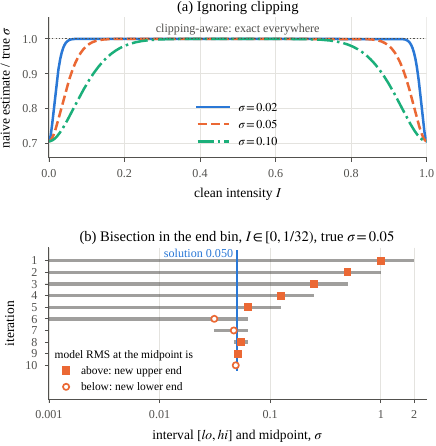}
\caption{\textbf{Clipping-aware pilot strength}
(Eqs.~\ref{eq:supp-pilot-moment}--\ref{eq:supp-pilot}), computed with the fitting code on
illustrative values. (a)~For a Gaussian residual clipped to $[0,1]$, the naive strength
$\sqrt{\operatorname{mean}r^2}$ falls below the true $\sigma$ near black and white, by 29\%
at the boundary itself and over a wider range for stronger noise; the clipping-aware
pilot is exact at every intensity. (b)~Solving Eq.~\ref{eq:supp-pilot} in the end bin
$k=0$, with clean intensities spread over $[0,1/32)$ and a true $\sigma$ of 0.05
(observed RMS 0.037, the naive estimate). Each midpoint whose model RMS exceeds the
observed RMS becomes the new upper end of the interval, any other the new lower end, so
the interval halves at every step (60 steps in the fitter).}
\label{fig:supp-pilot}
\end{figure}

\begin{figure}[tp]
\centering
\includegraphics[width=\linewidth]{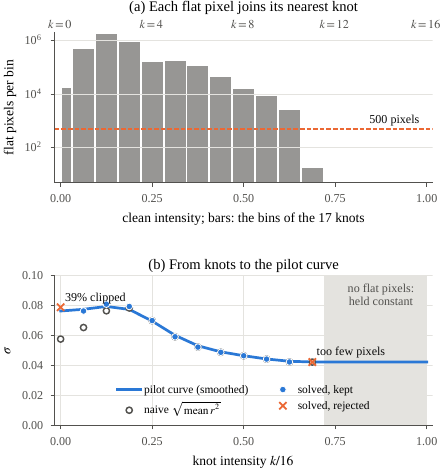}
\caption{\textbf{Building the pilot curve} on real data: SIDD GP ISO\,3200, green
channel, with the fitter's own flatness threshold and accumulator on the class's
fitting crops. (a)~Flat pixels per bin: each pixel joins its nearest of the 17
knots, so a bin spans 16 codes and an end bin 8. (b)~At the dark knots $k\le3$, the
naive estimate falls short of the clipping-aware solution. Knot~0 is rejected because
39\% of its bin's noise values are predicted to clip and knot~11 because its bin holds
18 pixels; bins 12--16 have no flat pixels. Their values come from the nearest kept
knots before one smoothing pass.}
\label{fig:supp-pilot-knots}
\end{figure}

\paragraph{Clipping-aware pilot strength.}
To put residuals from different intensities on one scale before measuring their
correlation, we need the strength $\sigma_c(x)$ of the noise \emph{before} the
image was clipped to $[0,1]$. Near black or white, clipping pins part of the noise
at the boundary and the observed spread underestimates $\sigma$
(Fig.~\ref{fig:supp-pilot}a). Left uncorrected, residuals there would be inflated by the
normalization below, and the clipping guard would admit pixels whose noise does clip. We therefore model
a flat pixel of clean intensity $I$ as $y=\operatorname{clip}(I+\sigma\varepsilon,0,1)$ with
$\varepsilon\sim\mathcal N(0,1)$ and use the exact expected squared residual of
this model. The noise is clipped at 0 when $\varepsilon<a=-I/\sigma$ and at 1 when
$\varepsilon>b=(1-I)/\sigma$; with $\phi$ and $\Phi$ the standard
Gaussian density and distribution function,
\begin{equation}
\begin{aligned}
    m(\sigma,I)&=\mathbb E\bigl[(y-I)^2\bigr]
    =\int_{-\infty}^{\infty}(y-I)^2\,\phi(\varepsilon)\,d\varepsilon\\
    &=\underbrace{\int_{-\infty}^{a}I^2\,\phi(\varepsilon)\,d\varepsilon}_{\text{clipped at }0}
    +\underbrace{\int_{a}^{b}\sigma^2\varepsilon^2\,\phi(\varepsilon)\,d\varepsilon}_{\text{not clipped}}\\
    &\quad+\underbrace{\int_{b}^{\infty}(1-I)^2\,\phi(\varepsilon)\,d\varepsilon}_{\text{clipped at }1}\\
    &=I^2\,\Phi(a)+\sigma^2\bigl[\Phi(b)-\Phi(a)+a\phi(a)-b\phi(b)\bigr]\\
    &\quad+(1-I)^2\bigl(1-\Phi(b)\bigr),
\end{aligned}
\label{eq:supp-pilot-moment}
\end{equation}
where the outer terms are constants times the Gaussian tail probabilities
$\Phi(a)$ and $1-\Phi(b)$. For the middle term, the standard Gaussian density
satisfies $\phi'(\varepsilon)=-\varepsilon\,\phi(\varepsilon)$, so integration by parts gives
\begin{equation*}
\begin{aligned}
    \int_a^b\varepsilon^2\phi(\varepsilon)\,d\varepsilon
    &=\int_a^b\varepsilon\bigl(-\phi'(\varepsilon)\bigr)\,d\varepsilon\\
    &=\bigl[-\varepsilon\,\phi(\varepsilon)\bigr]_a^b+\int_a^b\phi(\varepsilon)\,d\varepsilon\\
    &=\Phi(b)-\Phi(a)+a\,\phi(a)-b\,\phi(b).
\end{aligned}
\end{equation*}
Far from the boundaries $m\approx\sigma^2$; near them $m$ is clearly smaller, and
$m$ always increases with $\sigma$.

Fitting $\sigma$ with Eq.~\ref{eq:supp-pilot-moment} rests on two ideas. A single squared
residual says little about $\sigma$, so the pilot pools many flat pixels into a few
intensity bins, separately per channel. The pixels of a bin keep their own clean
intensities, however, and near black and white these decide how much of their noise
clips; the model's prediction for a bin therefore averages
Eq.~\ref{eq:supp-pilot-moment} over the pixels' own intensities, each pixel counting
equally, rather than evaluating it once at the bin's position. What follows
spells out the bins, the two bin averages and how the fitter computes them.

The pilot uses 17 knots at $k/16$,
$k=0,\dots,16$ (codes 0, 15.9, 31.9, \dots, 239.1, 255), and each flat pixel joins its
nearest knot. The knots' bins are therefore the clean codes $[0,8)$, $[8,24)$,
$[24,40)$, \dots, $[232,248)$ and $[248,256)$: 16 codes each, and 8 for the two end
knots, which sit at 0 and 255 (Fig.~\ref{fig:supp-pilot-knots}a). With $P_{c,k}$ the flat
pixels of bin $k$ in channel $c$, the bin's observation is its mean squared residual,
\begin{equation}
    \overline{r^2}_{c,k}=\frac{1}{|P_{c,k}|}\sum_{\mathbf p\in P_{c,k}}r_c(\mathbf p)^2.
    \label{eq:supp-pilot-observed}
\end{equation}
The model's prediction for the same pixels is the average of
Eq.~\ref{eq:supp-pilot-moment} at their clean intensities. The fitter streams over
millions of pixels and cannot keep each one, so it splits every bin into 32 equal
subbins of about half a code (the end bins use 16) and keeps, for subbin $j$, the
number $n_{c,k,j}$ of the bin's flat pixels in it and their mean clean intensity
$\bar x_{c,k,j}$. The prediction is then
\begin{equation}
\begin{aligned}
    \bar m_{c,k}(\sigma)&=\sum_{j}\frac{n_{c,k,j}}{|P_{c,k}|}\,m\bigl(\sigma,\bar x_{c,k,j}\bigr),\\
    |P_{c,k}|&=\sum_{j}n_{c,k,j},
\end{aligned}
\label{eq:supp-pilot-predicted}
\end{equation}
so each subbin counts with its share of the bin's pixels. For 8-bit clean images,
each occupied subbin holds a single code and $\bar x_{c,k,j}$ is that code, so
Eq.~\ref{eq:supp-pilot-predicted} equals the per-pixel average exactly.
Alg.~\ref{alg:supp-pilot} rounds the shares to multiples of $1/512$, at least one per
occupied subbin; on all 16 SIDD classes, this changes the kept knot strengths by at
most 0.32\%. The subbins are finer than 8-bit codes because clean images need not be
8-bit: for the continuous blurred film images $\mu$ (Sec.~\ref{sec:supp-spec-experiments}),
$\bar x_{c,k,j}$ averages nearby intensities, for example to 7.5 codes, where
Eq.~\ref{eq:supp-pilot-moment} is evaluated directly. Because $m$ increases with
$\sigma$, the knot's strength $\sigma_{c,k}$ is the single solution of
\begin{equation}
    \bar m_{c,k}(\sigma_{c,k})=\overline{r^2}_{c,k},
    \label{eq:supp-pilot}
\end{equation}
found by bisection over $0<\sigma\le2$ with 60 halvings of the interval (Fig.~\ref{fig:supp-pilot}b).

Each pixel enters at its own intensity because clipping changes quickly near black and
white. In bin 0 (codes 0 to 7), Eq.~\ref{eq:supp-pilot-moment} clips half of the noise
values at code 0 but only a few at code 7 for typical noise. Evaluating all of the
bin's pixels at the knot value 0 would treat the pixels at code 7 as half clipped and
predict too little residual energy for them; to match the observation, the solve would
raise $\sigma$, inflating the pilot near black, and likewise near white (for equally
many pixels at codes 0 and 7 and $\sigma=0.02$, it returns 0.023 instead of 0.020). Away from the boundaries $m\approx\sigma^2$, and the intensities within a bin
barely matter.

A knot's solution is kept only if three conditions hold. First, its bin has at least
500 flat pixels. Second, the solution is below the upper end $\sigma=2$ of the
search, twice the whole intensity range $[0,1]$; a strength this large means that
clipped Gaussian noise does not explain the bin, for instance when misregistration
leaves image structure in the residuals. Third, at most 35\% of the bin's noise values are predicted to clip. Rejected knots take the value interpolated linearly between kept knots, or
the nearest kept value beyond the outermost ones; the 17 values are then smoothed
once with weights $(\frac14,\frac12,\frac14)$, end values held, and $\sigma_c(x)$ is
linear between knots (Fig.~\ref{fig:supp-pilot-knots}b; Alg.~\ref{alg:supp-pilot}). Without the pooling, with
one knot per clean code, the final spatial correlation error is 1.8\% higher
(Sec.~\ref{sec:supp-ablation}). The pilot only prepares the
correlation fit: the lookup describes the observed, clipped residuals directly
and needs no correction.

\paragraph{Correlation.}
Pixels whose noise is likely to clip are excluded with a guard on the clean image,
and residuals are normalized by the pilot and centered within each image:
\begin{align}
    \mathcal M&=\mathcal M_{\mathrm{flat}}\cap\bigcap_c\bigl\{\mathbf p:2.5\,\sigma_c(x_c)<x_c<1-2.5\,\sigma_c(x_c)\bigr\},\\
    u_c&=\frac{r_c}{\sigma_c(x_c)}-\operatorname{mean}_{\mathcal M}\frac{r_c}{\sigma_c(x_c)}.
\end{align}
Pooled over all training images, the measured correlation of channel $a$ at $\mathbf p$
with channel $b$ at $\mathbf p+\mathbf d$, $\mathbf d\in[-16,16]^2$, is
$\widehat R_{ab}(\mathbf d)=\widehat C_{ab}(\mathbf d)/\sqrt{\widehat C_{aa}(\mathbf 0)\,\widehat C_{bb}(\mathbf 0)}$ with
$\widehat C_{ab}(\mathbf d)$ the mean of $u_a(\mathbf p)\,u_b(\mathbf p+\mathbf d)$ over pairs inside $\mathcal M$
(Eqs.~\mainref{eq:spatial_covariance}--\mainref{eq:makesure} of the main paper).
If $\mathcal M$ is empty in the whole training set, $\mathcal M_{\mathrm{flat}}$
is used instead.

\begin{figure*}[tp]
\centering
\includegraphics[width=\linewidth]{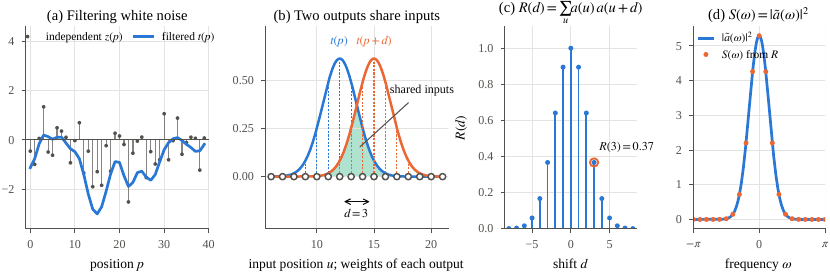}
\caption{\textbf{Why filtering correlates noise}, in one dimension. (a)~Independent
Gaussian samples $z(p)$ and their output $t$ through a 9-tap Gaussian filter:
neighboring outputs become similar. (b)~The filter's weights for the outputs at $p$
and $p+d$ (dashed: the weight each input receives); the inputs under both windows
(shaded) are shared, which is why $t(p)$ and $t(p+d)$ are correlated. (c)~The
correlation is the overlap of the filter with its shift
(Eq.~\ref{eq:supp-1d-correlation}). (d)~Its Fourier transform $S(\omega)$ equals
$|\tilde a(\omega)|^2$ at every frequency (Eq.~\ref{eq:supp-1d-spectrum}), so a filter with
a given correlation follows from $\sqrt S$.}
\label{fig:supp-spectral}
\end{figure*}

\paragraph{Spectral factorization.}
A covariance describes how variables vary together; a linear transformation
describes how to construct that dependence from shared inputs. If a zero-mean
vector $\mathbf z$ has covariance $I$, then $\mathbf t=L\mathbf z$ has covariance
\begin{equation*}
    \operatorname{Cov}(\mathbf t)
    =L\operatorname{Cov}(\mathbf z)L^{\mathsf T}=LL^{\mathsf T}.
\end{equation*}
Constructing variables with a desired covariance $C$ therefore amounts to finding
a factor $L$ such that $C=LL^{\mathsf T}$. For a stationary spatial field, covariance
depends only on the displacement between locations. Fourier coordinates separate
this translation-invariant construction by frequency, replacing a large spatial
factorization with one scalar or small channel-matrix factorization at each
frequency. This is the general idea behind spectral factorization. Here it turns
the measured correlations $\widehat R$ into the convolution filters of
Eq.~\ref{eq:supp-field}.

To see how a filter creates dependence, consider independent, zero-mean,
unit-variance inputs $z(p)$ and let
\begin{equation*}
    t(p)=\frac{z(p-1)+z(p)}{\sqrt2}.
\end{equation*}
The neighboring outputs $t(p)$ and $t(p+1)$ share $z(p)$: a large value of that input
raises both outputs. Each output has variance one, while their covariance is
$1/2$; outputs two or more positions apart share no inputs and have covariance
zero. More generally, convolution forms $t(p)=\sum_u a(u)z(p-u)$, and its output
covariance is
\begin{equation}
\begin{aligned}
    R(d)&:=\mathbb E[t(p)t(p+d)]\\
        &=\sum_{u,v}a(u)a(v)\,
          \mathbb E[z(p-u)z(p+d-v)]\\
        &=\sum_u a(u)a(u+d).
\end{aligned}
\label{eq:supp-1d-correlation}
\end{equation}
Only terms referring to the same input location survive the expectation. Thus
$R(d)$ is the overlap of the filter with a copy shifted by $d$
(Fig.~\ref{fig:supp-spectral}b,c). At zero displacement,
$R(0)=\sum_u a(u)^2$; with unit filter energy, covariance and normalized correlation
coincide. Equation~\ref{eq:supp-1d-correlation} is the forward problem: a filter
determines a covariance. Fitting solves the inverse problem: construct a filter
from a desired covariance.

In spatial coordinates, this inverse problem couples filter coefficients through
their products. To separate it by frequency, define the filter's frequency
response and the covariance's power spectrum as
\begin{equation*}
    \tilde a(\omega)=\sum_u a(u)e^{-\mathrm i\omega u},\qquad
    S(\omega)=\sum_d R(d)e^{-\mathrm i\omega d},
\end{equation*}
where $\mathrm i^2=-1$ and $\omega\in[-\pi,\pi]$ is angular frequency. The complex
gain $\tilde a(\omega)$ describes how the filter scales and shifts a wave of
frequency $\omega$; $S(\omega)$ describes output power at that frequency.
Substitution of Eq.~\ref{eq:supp-1d-correlation}, followed by $v=u+d$, gives
\begin{equation}
\begin{aligned}
    S(\omega)
    &=\sum_{u,v}a(u)a(v)e^{-\mathrm i\omega(v-u)}\\
    &=\left(\sum_u a(u)e^{\mathrm i\omega u}\right)
      \left(\sum_v a(v)e^{-\mathrm i\omega v}\right)\\
    &=\overline{\tilde a(\omega)}\,\tilde a(\omega)
      =|\tilde a(\omega)|^2.
\end{aligned}
\label{eq:supp-1d-spectrum}
\end{equation}
The bar denotes complex conjugation. The Wiener--Khinchin relation identifies the
Fourier transform of the covariance as the power
spectrum~\cite{wiener1930generalized,khintchine1934korrelationstheorie}; the
calculation above expresses that spectrum in terms of the filter. Filtering
multiplies the amplitude at each frequency by $\tilde a(\omega)$ and hence its power
by $|\tilde a(\omega)|^2$; unit-variance white inputs have unit power
spectrum~\cite{papoulis2002probability}.

A valid covariance has a nonnegative spectrum. The desired filter can therefore
be constructed one frequency at a time, then transformed back to spatial
coordinates:
\begin{equation}
\begin{aligned}
    \tilde a(\omega)&=\sqrt{S(\omega)},\\
    a(u)&=\frac1{2\pi}\int_{-\pi}^{\pi}
          \sqrt{S(\omega)}e^{\mathrm i\omega u}\,d\omega.
\end{aligned}
\label{eq:supp-1d-root}
\end{equation}
For example, the two-tap filter above gives $R(0)=1$, $R(\pm1)=1/2$, and hence
$S(\omega)=1+\cos\omega$. This example illustrates why emphasizing low frequencies
smooths a signal, as in Fig.~\ref{fig:supp-spectral}a.

The covariance determines the magnitude of $\tilde a$, but leaves its phase free:
multiplying by a unit-magnitude phase factor leaves
Eq.~\ref{eq:supp-1d-spectrum} unchanged. Phases must also respect conjugate symmetry
for the spatial filter to remain real. The nonnegative scalar root chooses zero
phase and gives a real, symmetric filter. It is one valid construction, and the
covariance alone cannot identify a unique original filter. This choice need not
have finite support even when a finite filter could produce the same covariance.

For multiple channels, each output mixes several independent input fields through
$t_c(\mathbf p)=\sum_i\sum_{\mathbf k}A_{ci}(\mathbf k)z_i(\mathbf p-\mathbf k)$.
In two dimensions, $\mathbf p$, $\mathbf k$, and $\mathbf d$ are spatial vectors and
$\boldsymbol\omega=(\omega_y,\omega_x)$. At each frequency, the filters form a
matrix $\widetilde A$, with rows for output channels and columns for input channels.
Using the same lag convention as in the measured correlations,
\begin{equation}
\begin{aligned}
    \widetilde A_{ci}(\boldsymbol\omega)
        &=\sum_{\mathbf k}A_{ci}(\mathbf k)
          e^{-\mathrm i\boldsymbol\omega\cdot\mathbf k},\\
    R_{ab}(\mathbf d)&:=\mathbb E[t_a(\mathbf p)t_b(\mathbf p+\mathbf d)],\\
    S_{ab}(\boldsymbol\omega)
        &=\sum_{\mathbf d}R_{ab}(\mathbf d)
          e^{-\mathrm i\boldsymbol\omega\cdot\mathbf d}\\
        &=\sum_i\overline{\widetilde A_{ai}(\boldsymbol\omega)}
                       \widetilde A_{bi}(\boldsymbol\omega).
\end{aligned}
\label{eq:supp-2d-spectrum}
\end{equation}
The last equality follows from Eq.~\mainref{eq:model_correlation} of the main paper
just as Eq.~\ref{eq:supp-1d-spectrum} follows from
Eq.~\ref{eq:supp-1d-correlation}. Thus
$S=\overline{\widetilde A}\,\widetilde A^{\mathsf T}$ is a Hermitian positive
semidefinite matrix at each frequency: $S=S^{\mathsf H}$ and all its eigenvalues are
nonnegative. This is the matrix counterpart of $S=|\tilde a|^2$, so a scalar square
root becomes a matrix square root~\cite{shinozuka1971simulation}. Write
$S=V\Lambda V^{\mathsf H}$, where $\mathsf H$ denotes conjugate transpose. Its
Hermitian square root gives a valid factor:
\begin{equation*}
    W=V\sqrt\Lambda V^{\mathsf H},\qquad \widetilde A=\overline W.
\end{equation*}
The eigenvectors identify channel mixtures with separate powers; taking the
square root of each power gives the corresponding filter gain. Transforming back
to the original channel coordinates yields a matrix square root. The conjugation
follows from our definition of $R_{ab}(\mathbf d)$: it ensures
$\overline{\widetilde A}\,\widetilde A^{\mathsf T}=WW^{\mathsf H}=S$ and preserves the
specified lag direction. For RGB, the matrix is $3\times3$ because there are three
channels; this size is independent of the filters' spatial support.

The fit applies this construction in three steps (Alg.~\ref{alg:supp-corr}). First,
the measured correlations $\widehat R$ are symmetrized to satisfy
$R_{ab}(\mathbf d)=R_{ba}(-\mathbf d)$ and tapered at distant lags:
\begin{equation}
\begin{aligned}
    \widehat R^{\mathrm s}_{ab}(\mathbf d)
        &=\tfrac12\bigl[\widehat R_{ab}(\mathbf d)
                     +\widehat R_{ba}(-\mathbf d)\bigr]\eta(\mathbf d),\\
    \eta(\mathbf d)
        &=\cos^2\!\left[\frac\pi2\operatorname{clip}\!\left(
          \frac{\|\mathbf d\|_2-8}{8},0,1\right)\right].
\end{aligned}
\label{eq:supp-taper}
\end{equation}
The taper leaves lags within radius 8 unchanged and smoothly reduces them to zero
at radius 16. Second, $\widehat R^{\mathrm s}$ is transformed to $\widehat S$ and each
frequency matrix is factored:
\begin{equation}
\begin{aligned}
    \widehat S&=V\Lambda V^{\mathsf H},\qquad
        \Lambda_+=\max(\Lambda,0),\\
    S_+&=V\Lambda_+V^{\mathsf H},\\
    W&=V\Lambda_+^{1/2}V^{\mathsf H},\qquad
        \widetilde A=\overline W.
\end{aligned}
\label{eq:supp-spectral-factor}
\end{equation}
The maximum and square root act on the diagonal eigenvalues. Negative
eigenvalues, which can arise in an estimated and tapered spectrum, are set to zero
to obtain a valid spectral matrix, so
$\overline{\widetilde A}\,\widetilde A^{\mathsf T}=WW^{\mathsf H}=S_+$.
Third, the frequency responses are inverse-transformed, cropped to
$\mathcal K=\{-12,\ldots,12\}^2$, and normalized per output channel. For
$\mathbf k\in\mathcal K$, the retained coefficients are
\begin{equation}
\begin{aligned}
    \Gamma_{ci}(\mathbf k)
        &=\mathcal F^{-1}[\widetilde A_{ci}](\mathbf k),\\
    A_{ci}(\mathbf k)
        &=\frac{\Gamma_{ci}(\mathbf k)}
          {\left(\sum_j\sum_{\boldsymbol\ell\in\mathcal K}
            \Gamma_{cj}(\boldsymbol\ell)^2\right)^{1/2}}.
\end{aligned}
\label{eq:supp-filters}
\end{equation}
Here $\mathcal F^{-1}$ is the two-dimensional inverse Fourier transform. The
transforms are computed on a $64\times64$ frequency grid. The result is nine
$25\times25$ filters satisfying the unit-energy constraint of
Eq.~\ref{eq:supp-energy}. The covariance construction requires only uncorrelated,
zero-mean, unit-variance inputs; choosing independent standard Gaussian inputs
also makes the filtered field Gaussian, with standard normal marginals for the
lookup of Eq.~\ref{eq:supp-lookup}. Tapering, spectral projection, cropping, and
renormalization make the covariance fit approximate; the subsequent nonlinear
lookup can further change residual correlations.

\FloatBarrier
\section{Ablation details}
\label{sec:supp-ablation}

\providecommand{\abltie}[1]{\textcolor{black!45}{#1}}
\providecommand{\ablworse}[1]{\textcolor[HTML]{B2182B}{#1}}
\providecommand{\ablbetter}[1]{\textcolor[HTML]{1B7837}{#1}}
\begin{table*}[tp]
\sidecaptioned{\footnotesize
\setlength{\tabcolsep}{6pt}
\begin{tabular}{@{}lccccc@{}}
\toprule
Variant & KL, $32^2$ tiles & KL, $256^2$ blocks & $W_1$ & Spatial & Color \\
\midrule
Full PGNM & 0.0219 & 0.0108 & 1.319 & 0.0094 & 0.0144 \\
\midrule
No fitting masks & \ablworse{\makecell{$+$1.2\%\\[-1pt]{\scriptsize[$+$0.6, $+$1.8]}}} & \abltie{\makecell{0.0\%\\[-1pt]{\scriptsize[0.0, $+$0.1]}}} & \abltie{\makecell{0.0\%\\[-1pt]{\scriptsize[0.0, $+$0.1]}}} & \abltie{\makecell{$-$1.7\%\\[-1pt]{\scriptsize[$-$11.8, $+$6.8]}}} & \ablworse{\makecell{$+$46.3\%\\[-1pt]{\scriptsize[$+$14.9, $+$71.0]}}} \\[2pt]
Flat mask only & \abltie{\makecell{$+$0.1\%\\[-1pt]{\scriptsize[$-$0.6, $+$0.8]}}} & \abltie{\makecell{0.0\%\\[-1pt]{\scriptsize[$-$0.1, 0.0]}}} & \abltie{\makecell{0.0\%\\[-1pt]{\scriptsize[0.0, $+$0.1]}}} & \abltie{\makecell{$-$4.3\%\\[-1pt]{\scriptsize[$-$16.3, $+$7.3]}}} & \abltie{\makecell{$+$3.2\%\\[-1pt]{\scriptsize[$-$8.8, $+$15.6]}}} \\[2pt]
Clipping guard only & \ablworse{\makecell{$+$0.6\%\\[-1pt]{\scriptsize[$+$0.4, $+$0.7]}}} & \abltie{\makecell{0.0\%\\[-1pt]{\scriptsize[0.0, $+$0.1]}}} & \abltie{\makecell{0.0\%\\[-1pt]{\scriptsize[0.0, $+$0.1]}}} & \abltie{\makecell{$+$1.9\%\\[-1pt]{\scriptsize[$-$1.6, $+$6.3]}}} & \ablworse{\makecell{$+$33.8\%\\[-1pt]{\scriptsize[$+$12.9, $+$54.0]}}} \\[2pt]
Kernel $1\times1$ & \ablbetter{\makecell{$-$5.4\%\\[-1pt]{\scriptsize[$-$6.2, $-$4.7]}}} & \abltie{\makecell{$-$0.2\%\\[-1pt]{\scriptsize[$-$0.6, $+$0.1]}}} & \abltie{\makecell{0.0\%\\[-1pt]{\scriptsize[$-$0.2, $+$0.3]}}} & \ablworse{\makecell{$+$586\%\\[-1pt]{\scriptsize[$+$570, $+$603]}}} & \ablworse{\makecell{$+$770\%\\[-1pt]{\scriptsize[$+$743, $+$801]}}} \\[2pt]
Kernel $3\times3$ & \ablbetter{\makecell{$-$0.6\%\\[-1pt]{\scriptsize[$-$1.3, 0.0]}}} & \abltie{\makecell{$+$0.1\%\\[-1pt]{\scriptsize[$-$0.3, $+$0.5]}}} & \abltie{\makecell{$+$0.2\%\\[-1pt]{\scriptsize[$-$0.1, $+$0.4]}}} & \ablworse{\makecell{$+$45.8\%\\[-1pt]{\scriptsize[$+$37.8, $+$54.2]}}} & \abltie{\makecell{$+$47.3\%\\[-1pt]{\scriptsize[$-$4.5, $+$88.5]}}} \\[2pt]
Kernel $5\times5$ & \ablbetter{\makecell{$-$0.8\%\\[-1pt]{\scriptsize[$-$1.4, $-$0.2]}}} & \abltie{\makecell{0.0\%\\[-1pt]{\scriptsize[$-$0.3, $+$0.3]}}} & \abltie{\makecell{$+$0.2\%\\[-1pt]{\scriptsize[$-$0.1, $+$0.4]}}} & \abltie{\makecell{$-$6.5\%\\[-1pt]{\scriptsize[$-$14.3, $+$0.3]}}} & \abltie{\makecell{$+$0.1\%\\[-1pt]{\scriptsize[$-$2.5, $+$3.2]}}} \\[2pt]
Kernel $11\times11$ & \ablbetter{\makecell{$-$0.9\%\\[-1pt]{\scriptsize[$-$1.4, $-$0.5]}}} & \abltie{\makecell{0.0\%\\[-1pt]{\scriptsize[$-$0.3, $+$0.4]}}} & \abltie{\makecell{$+$0.1\%\\[-1pt]{\scriptsize[$-$0.1, $+$0.3]}}} & \ablbetter{\makecell{$-$8.5\%\\[-1pt]{\scriptsize[$-$14.4, $-$3.3]}}} & \abltie{\makecell{$+$1.2\%\\[-1pt]{\scriptsize[$-$1.2, $+$4.1]}}} \\[2pt]
Gaussian conditional$^\ddagger$ & \ablworse{\makecell{$+$138\%\\[-1pt]{\scriptsize[$+$122, $+$151]}}} & \ablworse{\makecell{$+$294\%\\[-1pt]{\scriptsize[$+$263, $+$321]}}} & \ablworse{\makecell{$+$108\%\\[-1pt]{\scriptsize[$+$97, $+$119]}}} & \abltie{\makecell{$+$4.5\%\\[-1pt]{\scriptsize[$-$1.4, $+$12.7]}}} & \abltie{\makecell{$+$2.0\%\\[-1pt]{\scriptsize[$-$4.5, $+$6.6]}}} \\[2pt]
Gaussian conditional, 33 sources$^\S$ & \ablworse{\makecell{$+$133\%\\[-1pt]{\scriptsize[$+$112, $+$147]}}} & \ablworse{\makecell{$+$276\%\\[-1pt]{\scriptsize[$+$232, $+$308]}}} & \ablworse{\makecell{$+$102\%\\[-1pt]{\scriptsize[$+$80, $+$123]}}} & \abltie{\makecell{$+$4.7\%\\[-1pt]{\scriptsize[$-$1.5, $+$15.1]}}} & \abltie{\makecell{$+$4.3\%\\[-1pt]{\scriptsize[$-$2.9, $+$9.4]}}} \\
\midrule
Pilot, one knot per code$^\P$ & \ablworse{\makecell{$+$0.1\%\\[-1pt]{\scriptsize[$+$0.1, $+$0.2]}}} & \abltie{\makecell{0.0\%\\[-1pt]{\scriptsize[0.0, 0.0]}}} & \abltie{\makecell{0.0\%\\[-1pt]{\scriptsize[0.0, 0.0]}}} & \ablworse{\makecell{$+$1.8\%\\[-1pt]{\scriptsize[$+$0.5, $+$3.3]}}} & \abltie{\makecell{0.0\%\\[-1pt]{\scriptsize[$-$0.9, $+$0.8]}}} \\
\bottomrule
\end{tabular}}
{\caption{\textbf{SIDD ablation, all variants} (39 sources, 8 scenes, five generator runs). Full PGNM's scores, then each variant's change relative to them with the 95\% scene-bootstrap interval (100,000 draws resampling the physical scenes): \ablworse{worse}, \ablbetter{better}, or \abltie{tie} when the interval includes zero. Each variant changes one fitting step and is rendered from the same random streams; the $25\times25$ refit of full PGNM reproduces the main evaluation's presets, outputs and scores exactly. $^\ddagger$The moment-matched Gaussian lookup exceeds the residual range $|g|\le1$ in two classes; there its table is clipped to $\pm1$ (Sec.~\ref{sec:supp-ablation}). $^\S$The same comparison restricted to the classes where the unclipped Gaussian is valid (excludes S6\,ISO\,3200, S6\,ISO\,800). $^\P$Added while preparing this supplement, after the other variants were scored: the pilot strength is solved at one knot per 8-bit clean code instead of 17 knots (Sec.~\ref{sec:supp-equations}), with the same rules.}
\label{tab:supp-ablation-sidd}}
\end{table*}

\begin{table*}[tp]
\centering
\footnotesize
\setlength{\tabcolsep}{3.2pt}
\begin{tabular}{@{}llrrrrrrrrrrrrrrrr@{}}
\toprule
\multicolumn{2}{@{}l}{} & \multicolumn{4}{c}{$W_1$} & \multicolumn{4}{c}{Spatial} & \multicolumn{4}{c}{DISTS} & \multicolumn{4}{c}{LPIPS} \\
\cmidrule(l){3-6} \cmidrule(l){7-10} \cmidrule(l){11-14} \cmidrule(l){15-18}
Radius & Full & $k{=}1$ & $k{=}3$ & $k{=}5$ & $k{=}11$ & $k{=}1$ & $k{=}3$ & $k{=}5$ & $k{=}11$ & $k{=}1$ & $k{=}3$ & $k{=}5$ & $k{=}11$ & $k{=}1$ & $k{=}3$ & $k{=}5$ & $k{=}11$ \\
\midrule
$0.025$ & 0.107 & \ablbetter{$-$3.3} & \ablbetter{$-$1.1} & \ablbetter{$-$0.9} & \abltie{$-$0.1} & \ablworse{$+$792} & \ablworse{$+$76.4} & \ablbetter{$-$23.0} & \ablbetter{$-$16.3} & \ablworse{$+$160} & \ablworse{$+$2.2} & \ablbetter{$-$0.1} & \ablworse{$+$0.5} & \ablworse{$+$58.8} & \ablworse{$+$8.0} & \ablworse{$+$0.4} & \ablworse{$+$0.3} \\
$0.05$ & 0.138 & \ablbetter{$-$7.0} & \ablbetter{$-$2.3} & \ablbetter{$-$0.6} & \abltie{0.0} & \ablworse{$+$1247} & \ablworse{$+$103} & \ablbetter{$-$7.4} & \ablbetter{$-$5.5} & \ablworse{$+$200} & \ablworse{$+$2.2} & \ablworse{$+$0.9} & \ablbetter{$+$0.1} & \ablworse{$+$137} & \ablworse{$+$3.2} & \ablbetter{$-$0.6} & \ablbetter{$-$0.2} \\
$0.075$ & 0.168 & \ablbetter{$-$8.8} & \ablworse{$-$0.4} & \ablworse{$+$1.1} & \ablworse{$+$0.8} & \ablworse{$+$1298} & \ablworse{$+$97.1} & \ablbetter{$-$6.5} & \ablbetter{$-$3.5} & \ablworse{$+$222} & \ablworse{$+$2.2} & \abltie{$-$0.3} & \ablworse{$+$0.3} & \ablworse{$+$161} & \ablworse{$+$2.7} & \ablbetter{$-$0.3} & \ablworse{$+$0.1} \\
$0.1$ & 0.206 & \ablbetter{$-$11.8} & \ablbetter{$-$2.2} & \ablbetter{$-$0.4} & \abltie{$-$0.2} & \ablworse{$+$1314} & \ablworse{$+$94.6} & \ablbetter{$-$5.4} & \ablbetter{$-$2.7} & \ablworse{$+$232} & \ablworse{$+$2.2} & \ablworse{$+$0.1} & \ablworse{$+$0.7} & \ablworse{$+$173} & \ablworse{$+$2.7} & \ablbetter{$-$0.2} & \ablbetter{$-$0.2} \\
$0.2$ & 0.419 & \ablbetter{$-$11.6} & \ablbetter{$-$2.3} & \abltie{$-$0.4} & \ablworse{0.0} & \ablworse{$+$1350} & \ablworse{$+$101} & \ablbetter{$-$2.4} & \ablbetter{$-$1.4} & \ablworse{$+$233} & \ablworse{$+$2.3} & \ablworse{$+$0.3} & \ablworse{$+$0.5} & \ablworse{$+$198} & \ablworse{$+$2.0} & \ablbetter{$-$0.2} & \ablbetter{$-$0.1} \\
$0.4$ & 0.852 & \ablbetter{$-$12.2} & \ablbetter{$-$2.7} & \ablbetter{$-$1.0} & \ablbetter{$-$0.8} & \ablworse{$+$1269} & \ablworse{$+$114} & \ablbetter{$-$6.3} & \ablbetter{$-$4.3} & \ablworse{$+$207} & \ablworse{$+$3.3} & \ablworse{$+$0.2} & \ablworse{$+$0.4} & \ablworse{$+$212} & \ablworse{$+$1.2} & \ablbetter{0.0} & \ablbetter{$-$0.1} \\
$0.6$ & 1.305 & \ablbetter{$-$11.3} & \ablbetter{$-$2.3} & \ablworse{0.0} & \ablbetter{$-$0.7} & \ablworse{$+$1120} & \ablworse{$+$128} & \ablbetter{$-$10.6} & \ablbetter{$-$6.6} & \ablworse{$+$179} & \ablworse{$+$4.9} & \ablworse{$+$0.4} & \ablbetter{$-$0.4} & \ablworse{$+$244} & \ablworse{$+$2.8} & \ablworse{$+$0.1} & \abltie{$+$0.2} \\
$0.8$ & 1.810 & \ablbetter{$-$10.2} & \ablbetter{$-$2.2} & \ablworse{$+$0.2} & \abltie{$+$0.2} & \ablworse{$+$1109} & \ablworse{$+$180} & \ablbetter{$-$2.3} & \ablbetter{$-$3.1} & \ablworse{$+$154} & \ablworse{$+$4.5} & \ablworse{$+$0.1} & \ablbetter{$+$0.3} & \ablworse{$+$282} & \ablworse{$+$4.2} & \ablbetter{$-$0.5} & \ablworse{$+$0.1} \\
\midrule
All radii & 0.626 & \ablbetter{$-$10.7} & \abltie{$-$2.2} & \abltie{$-$0.2} & \abltie{$-$0.2} & \ablworse{$+$1164} & \ablworse{$+$113} & \ablbetter{$-$8.6} & \ablbetter{$-$5.9} & \ablworse{$+$193} & \ablworse{$+$3.5} & \abltie{$+$0.2} & \abltie{$+$0.2} & \ablworse{$+$230} & \ablworse{$+$2.8} & \abltie{$-$0.2} & \abltie{0.0} \\
$r\le0.2$ & 0.208 & \ablbetter{$-$9.7} & \abltie{$-$1.9} & \abltie{$-$0.2} & \abltie{$+$0.1} & \ablworse{$+$1167} & \ablworse{$+$92.8} & \ablbetter{$-$10.1} & \ablbetter{$-$6.7} & \ablworse{$+$219} & \ablworse{$+$2.3} & \abltie{$+$0.2} & \abltie{$+$0.4} & \ablworse{$+$176} & \ablworse{$+$2.6} & \abltie{$-$0.2} & \abltie{$-$0.1} \\
\bottomrule
\end{tabular}
\caption{\textbf{Film-grain kernel support, all sizes and radii.} Change of each metric (\%) when the $25\times25$ support of full PGNM is cropped to $k\times k$ in the fit, before normalization; Full: full PGNM's conditional $W_1$ in 8-bit codes. Per radius, a change is \ablworse{worse} or \ablbetter{better} when more than half of the 60 test images move in that direction, else \abltie{tie}; in the last two rows, when this holds at every radius of the scope (the rule of Tab.~\mainref{tab:ablation} of the main paper). Each image averages 25 output--reference pairs. The $25\times25$ refit equals the evaluated preset at every radius.}
\label{tab:supp-ablation-film}
\end{table*}

\paragraph{Design.}
Each variant changes one step of \textsc{Fit} (Alg.~\ref{alg:supp-fit}) and is
otherwise refitted on the same training pairs, rendered by the unchanged
synthesis from the same random streams, and scored like the main results. The
variants, and which of them the main paper shows, were fixed before any was scored,
except the last one below, and all are reported. A refit with the full recipe reproduces the main evaluation's
presets, outputs and scores exactly on SIDD and its presets at every film radius, so
every difference is due to the changed step. The variants are:
\begin{itemize}
\item \emph{No fitting masks}: the pilot uses all pixels and the correlation fit
      uses neither the flat mask nor the clipping guard.
\item \emph{Flat mask only} / \emph{clipping guard only}: one of the two masks for
      the correlation fit (the guard-only variant also fits the pilot on all pixels).
\item \emph{Kernel $k\times k$}: the factor is cropped to $k\times k$ instead of
      $25\times25$ before normalization.
\item \emph{Gaussian conditional}: the lookup is replaced by a Gaussian with the
      mean and standard deviation of the same pooled residual histogram at each
      clean intensity, stored as a lookup linear in $t$.
\item \emph{Pilot, one knot per code}: the pilot strength is solved at one knot per
      8-bit clean code instead of at 17 knots (Sec.~\ref{sec:supp-equations}), with the
      same rules for keeping, filling and smoothing knots. This variant was added
      while preparing this supplement, after the others had been scored.
\end{itemize}
The Gaussian lookup exceeds the residual range $|g|\le1$ for the dark red and
blue channels of S6 ISO\,800 and ISO\,3200, whose residuals there have means of 69
to 124 codes and standard deviations of 60 to 78 codes. The main row therefore clips
those table entries to $\pm1$ (29 entries in the two presets, changing fewer than
$10^{-5}$ of the channel values rendered on their training data), a choice made after
some Gaussian scores of the other classes existed but before any was looked at; an
extra row compares the unclipped variant with full PGNM on the 33 sources of the
other classes.

\paragraph{Results.}
Tab.~\ref{tab:supp-ablation-sidd} lists every SIDD variant with its interval. The
flat mask alone recovers full PGNM on every metric, whereas the clipping guard
alone leaves the color correlation error 34\% higher, close to the 46\% of no
masks: textured pixels in the correlation fit weaken the fitted cross-channel
dependence between neighboring pixels. A $3\times3$ kernel removes most of the
$1\times1$ loss on SIDD but is still 46\% worse in spatial correlation; from
$5\times5$ on, no correlation error is worse than full PGNM's ($11\times11$ is
8.5\% better in spatial correlation). Smaller kernels
lower the tile KL slightly, by 0.6--0.9\% from $3\times3$ to $11\times11$ and by
5.4\% for $1\times1$, where both correlation errors grow several-fold. Tab.~\ref{tab:supp-ablation-film} gives the film-grain kernel
sweep by radius: a $1\times1$ support loses the grain structure at every radius,
a $3\times3$ support raises the spatial correlation error by 76--180\% at every
radius, and from $5\times5$ on the changes are small and mostly improvements.

\begin{figure}[tp]
\centering
\includegraphics[width=\linewidth]{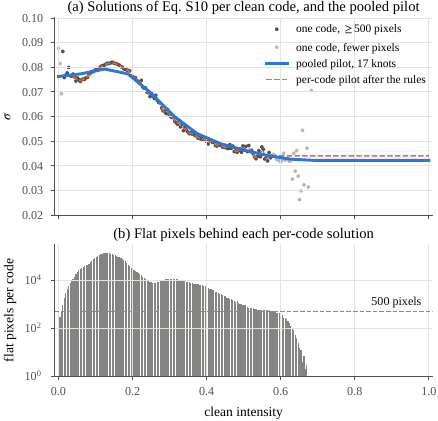}
\caption{\textbf{Pooled and per-code pilot} for SIDD GP ISO\,3200, green channel (the
class of Fig.~\ref{fig:supp-pilot-knots}). (a)~Eq.~\ref{eq:supp-pilot} solved at every clean
code with flat pixels (points; light: fewer than 500 pixels), the pooled 17-knot pilot
of PGNM, and the per-code pilot after the same rules for keeping, filling and
smoothing knots. (b)~Flat pixels per code. The per-code strengths follow the pooled
pilot where flat pixels are plentiful and scatter where they are few, in the first
codes and at the brightest flat intensities.}
\label{fig:supp-pilot-grid}
\end{figure}

Solving the pilot at one knot per clean code instead of pooling at 17 knots raises the
spatial correlation error by 1.8\% and the tile KL by 0.15\%, with the other measures
tied: the per-code strengths agree with the pooled pilot where flat pixels are
plentiful and scatter where they are few (Fig.~\ref{fig:supp-pilot-grid}).

\FloatBarrier
\section{SIDD sources, results by camera and ISO, and film grain by radius}
\label{sec:supp-sidd-breakdown}

\begin{table*}[tp]
\centering
\small
\setlength{\tabcolsep}{3.5pt}
\begin{tabular}{@{}rlrrrrlccc@{}}
\toprule
Image & SIDD scene instance & Phone & ISO & Exposure (s) & Illuminant (K) & Light & Scene & Evaluated & 19-source \\
\midrule
0 & \texttt{0009\_001\_S6\_00800\_00350\_3200\_L} & S6 & 800 & 1/350 & 3200 & L & 001 & \checkmark & \checkmark \\
1 & \texttt{0021\_001\_GP\_10000\_05000\_5500\_N} & GP & 10000 & 1/5000 & 5500 & N & 001 & \checkmark &  \\
2 & \texttt{0024\_001\_N6\_03200\_01500\_5500\_N} & N6 & 3200 & 1/1500 & 5500 & N & 001 & \checkmark &  \\
3 & \texttt{0026\_001\_G4\_00400\_00160\_5500\_L} & G4 & 400 & 1/160 & 5500 & L & 001 & \checkmark &  \\
4 & \texttt{0031\_001\_IP\_01600\_02000\_3200\_N} & IP & 1600 & 1/2000 & 3200 & N & 001 & \checkmark & \checkmark \\
5 & \texttt{0037\_002\_GP\_06400\_03200\_3200\_L} & GP & 6400 & 1/3200 & 3200 & L & 002 & \checkmark &  \\
6 & \texttt{0041\_002\_IP\_01600\_04000\_5500\_L} & IP & 1600 & 1/4000 & 5500 & L & 002 & \checkmark & \checkmark \\
7 & \texttt{0046\_002\_G4\_00400\_00350\_3200\_L} & G4 & 400 & 1/350 & 3200 & L & 002 & \checkmark &  \\
8 & \texttt{0049\_002\_N6\_00800\_00800\_5500\_L} & N6 & 800 & 1/800 & 5500 & L & 002 & \checkmark & \checkmark \\
9 & \texttt{0053\_002\_S6\_03200\_02000\_5500\_N} & S6 & 3200 & 1/2000 & 5500 & N & 002 & \checkmark & \checkmark \\
10 & \texttt{0056\_003\_N6\_03200\_04000\_5500\_N} & N6 & 3200 & 1/4000 & 5500 & N & 003 & \checkmark &  \\
11 & \texttt{0058\_003\_G4\_00400\_00500\_5500\_L} & G4 & 400 & 1/500 & 5500 & L & 003 & \checkmark &  \\
12 & \texttt{0061\_003\_S6\_00800\_00500\_4400\_L} & S6 & 800 & 1/500 & 4400 & L & 003 & \checkmark & \checkmark \\
13 & \texttt{0067\_003\_GP\_01600\_01500\_3200\_L} & GP & 1600 & 1/1500 & 3200 & L & 003 & \checkmark &  \\
14 & \texttt{0071\_003\_IP\_02000\_04000\_5500\_L} & IP & 2000 & 1/4000 & 5500 & L & 003 & \checkmark &  \\
15 & \texttt{0074\_004\_N6\_00400\_00040\_3200\_L} & N6 & 400 & 1/40 & 3200 & L & 004 & \checkmark &  \\
16 & \texttt{0079\_004\_G4\_00800\_00160\_3200\_N} & G4 & 800 & 1/160 & 3200 & N & 004 & \checkmark & \checkmark \\
17 & \texttt{0082\_004\_S6\_03200\_00500\_4400\_L} & S6 & 3200 & 1/500 & 4400 & L & 004 & \checkmark & \checkmark \\
18 & \texttt{0085\_004\_GP\_06400\_02000\_4400\_N} & GP & 6400 & 1/2000 & 4400 & N & 004 & \checkmark &  \\
19 & \texttt{0093\_004\_IP\_01250\_00250\_3200\_L} & IP & 1250 & 1/250 & 3200 & L & 004 & excluded &  \\
20 & \texttt{0095\_005\_N6\_00400\_00200\_3200\_L} & N6 & 400 & 1/200 & 3200 & L & 005 & \checkmark &  \\
21 & \texttt{0100\_005\_G4\_00800\_00400\_3200\_N} & G4 & 800 & 1/400 & 3200 & N & 005 & \checkmark & \checkmark \\
22 & \texttt{0103\_005\_S6\_01600\_00800\_4400\_L} & S6 & 1600 & 1/800 & 4400 & L & 005 & \checkmark & \checkmark \\
23 & \texttt{0109\_005\_GP\_10000\_08000\_4400\_N} & GP & 10000 & 1/8000 & 4400 & N & 005 & \checkmark &  \\
24 & \texttt{0112\_005\_IP\_01000\_07500\_5500\_L} & IP & 1000 & 1/7500 & 5500 & L & 005 & \checkmark &  \\
25 & \texttt{0119\_006\_N6\_00400\_00100\_3200\_L} & N6 & 400 & 1/100 & 3200 & L & 006 & \checkmark &  \\
26 & \texttt{0124\_006\_G4\_00800\_00350\_3200\_N} & G4 & 800 & 1/350 & 3200 & N & 006 & \checkmark & \checkmark \\
27 & \texttt{0128\_006\_S6\_03200\_01600\_4400\_L} & S6 & 3200 & 1/1600 & 4400 & L & 006 & \checkmark & \checkmark \\
28 & \texttt{0131\_006\_GP\_01600\_01250\_4400\_N} & GP & 1600 & 1/1250 & 4400 & N & 006 & \checkmark &  \\
29 & \texttt{0141\_006\_IP\_01600\_01620\_3200\_L} & IP & 1600 & 1/1620 & 3200 & L & 006 & \checkmark & \checkmark \\
30 & \texttt{0143\_007\_N6\_00800\_00800\_4400\_N} & N6 & 800 & 1/800 & 4400 & N & 007 & \checkmark & \checkmark \\
31 & \texttt{0148\_007\_G4\_00400\_00400\_4400\_L} & G4 & 400 & 1/400 & 4400 & L & 007 & \checkmark &  \\
32 & \texttt{0153\_007\_S6\_03200\_03200\_5500\_L} & S6 & 3200 & 1/3200 & 5500 & L & 007 & \checkmark & \checkmark \\
33 & \texttt{0158\_007\_GP\_03200\_03200\_5500\_N} & GP & 3200 & 1/3200 & 5500 & N & 007 & \checkmark &  \\
34 & \texttt{0162\_007\_IP\_01600\_01600\_3200\_L} & IP & 1600 & 1/1600 & 3200 & L & 007 & \checkmark & \checkmark \\
35 & \texttt{0171\_008\_N6\_03200\_01600\_4400\_L} & N6 & 3200 & 1/1600 & 4400 & L & 008 & \checkmark &  \\
36 & \texttt{0174\_008\_G4\_00800\_00800\_4400\_N} & G4 & 800 & 1/800 & 4400 & N & 008 & \checkmark & \checkmark \\
37 & \texttt{0176\_008\_S6\_00400\_00100\_5500\_L} & S6 & 400 & 1/100 & 5500 & L & 008 & \checkmark & \checkmark \\
38 & \texttt{0183\_008\_GP\_06400\_06400\_5500\_N} & GP & 6400 & 1/6400 & 5500 & N & 008 & \checkmark &  \\
39 & \texttt{0187\_008\_IP\_01600\_01600\_3200\_L} & IP & 1600 & 1/1600 & 3200 & L & 008 & \checkmark & \checkmark \\
\bottomrule
\end{tabular}
\caption{\textbf{SIDD Validation sources.} Validation image $i$ (32 blocks each) is the $i$-th SIDD benchmark scene instance in sorted order; its fields follow SIDD's naming (exposure $1/\text{shutter}$; light: low light L or normal brightness N). The iPhone~7 ISO\,1250 source is excluded because no SIDD-Medium training data share its camera--ISO class. 19-source: covered by the sRGB-Flow and NM-FlowGAN checkpoints. Evaluated sources by phone (G4 8, GP 8, IP 7, N6 8, S6 8); by illuminant (3200\,K 14, 4400\,K 11, 5500\,K 14); by brightness (low 24, normal 15); by scene (001: 5, 002: 5, 003: 5, 004: 4, 005: 5, 006: 5, 007: 5, 008: 5); 16 camera--ISO classes with 1--5 sources each. Phones: G4: LG G4; GP: Google Pixel; IP: iPhone 7; N6: Nexus 6; S6: Galaxy S6 Edge.}
\label{tab:supp-sidd-sources}
\end{table*}

\begin{table*}[tp]
\centering
\small
\setlength{\tabcolsep}{5pt}
\begin{tabular}{@{}lcccccccccc@{}}
\toprule
& \multicolumn{8}{c}{Radius $r$} & \multicolumn{2}{c}{Mean over radii} \\
\cmidrule(lr){2-9}\cmidrule(l){10-11}
Method & $0.025$ & $0.05$ & $0.075$ & $0.1$ & $0.2$ & $0.4$ & $0.6$ & $0.8$ & All & $\le0.2$ \\
\midrule
\multicolumn{11}{@{}l}{\textit{Conditional $W_1$ (8-bit codes)}} \\
\textbf{PGNM} & 0.107 & 0.138 & \textbf{0.168} & \textbf{0.206} & \textbf{0.419} & \textbf{0.852} & \textbf{1.305} & \textbf{1.810} & \textbf{0.626} & \textbf{0.208} \\
Zhang23$^*$ & \textbf{0.070} & \textbf{0.118} & 0.174 & 0.235 & 0.564 & 1.662 & 3.226 & 5.167 & 1.402 & 0.232 \\
GrainNet & 3.652 & 3.456 & 3.321 & 3.224 & 2.913 & 2.434 & 2.473 & 3.102 & 3.072 & 3.313 \\
Ind. Newson render & 0.040 & 0.080 & 0.121 & 0.158 & 0.314 & 0.618 & 0.892 & 1.167 & 0.424 & 0.143 \\
\midrule
\multicolumn{11}{@{}l}{\textit{Spatial correlation error}} \\
\textbf{PGNM} & 0.0104 & 0.0074 & 0.0072 & 0.0072 & 0.0071 & 0.0077 & 0.0090 & 0.0097 & \textbf{0.0082} & 0.0079 \\
Zhang23$^*$ & \textbf{0.0090} & \textbf{0.0069} & \textbf{0.0069} & \textbf{0.0068} & \textbf{0.0069} & \textbf{0.0071} & \textbf{0.0074} & 0.1115 & 0.0203 & \textbf{0.0073} \\
GrainNet & 0.0596 & 0.0329 & 0.0209 & 0.0153 & 0.0088 & 0.0076 & 0.0078 & \textbf{0.0083} & 0.0202 & 0.0275 \\
Ind. Newson render & 0.0062 & 0.0065 & 0.0067 & 0.0067 & 0.0069 & 0.0069 & 0.0073 & 0.0074 & 0.0068 & 0.0066 \\
\midrule
\multicolumn{11}{@{}l}{\textit{DISTS}} \\
\textbf{PGNM} & 0.0204 & 0.0339 & 0.0436 & \textbf{0.0512} & 0.0720 & 0.0991 & 0.1214 & 0.1429 & 0.0731 & \textbf{0.0442} \\
Zhang23$^*$ & \textbf{0.0192} & \textbf{0.0334} & \textbf{0.0435} & 0.0514 & 0.0735 & 0.1064 & 0.1361 & 0.3475 & 0.1014 & \textbf{0.0442} \\
GrainNet & 0.0273 & 0.0378 & 0.0466 & 0.0531 & \textbf{0.0718} & \textbf{0.0939} & \textbf{0.1068} & \textbf{0.1148} & \textbf{0.0690} & 0.0473 \\
Ind. Newson render & 0.0187 & 0.0329 & 0.0429 & 0.0504 & 0.0699 & 0.0919 & 0.1047 & 0.1113 & 0.0653 & 0.0430 \\
\midrule
\multicolumn{11}{@{}l}{\textit{LPIPS}} \\
\textbf{PGNM} & 0.0052 & \textbf{0.0161} & \textbf{0.0279} & \textbf{0.0395} & 0.0803 & 0.1387 & 0.1658 & 0.1732 & 0.0809 & \textbf{0.0338} \\
Zhang23$^*$ & \textbf{0.0051} & \textbf{0.0161} & \textbf{0.0279} & \textbf{0.0395} & 0.0803 & 0.1401 & 0.1690 & 0.6392 & 0.1397 & \textbf{0.0338} \\
GrainNet & 0.0099 & 0.0189 & 0.0296 & 0.0404 & \textbf{0.0797} & \textbf{0.1351} & \textbf{0.1601} & \textbf{0.1651} & \textbf{0.0799} & 0.0357 \\
Ind. Newson render & 0.0050 & 0.0159 & 0.0276 & 0.0391 & 0.0792 & 0.1355 & 0.1606 & 0.1649 & 0.0785 & 0.0333 \\
\bottomrule
\end{tabular}
\caption{\textbf{Simulated film grain by radius} (lower is better). Breakdown of Tab.~\mainref{tab:film} of the main paper: each value averages the 25 output--reference pairs of each test image, then the 60 test images; the last two columns average the radii of each scope with equal weights and reproduce the main table. Lowest fitted method of each column in bold, values equal at the printed precision together; the independent Newson render compares two simulator runs and is not bolded. Descriptive only, without intervals. $^*$Our implementation of the published model, with the shared blur $B$; its small-grain approximation applies up to $r=0.2$.}
\label{tab:supp-film-radius}
\end{table*}

\begin{table}[tp]
\centering
\scriptsize
\setlength{\tabcolsep}{2.4pt}
\begin{tabular}{@{}lrcccccc@{}}
\toprule
Class & $n$ & \textbf{PGNM} & NAFlow & GMDCN$^*$ & DANet & sRGB-Flow & NM-FlowG. \\
\midrule
G4 400 & 4 & 0.026 & 0.018 & \textbf{0.015} & 0.106 & -- & -- \\
G4 800 & 4 & 0.023 & 0.027 & \textbf{0.020} & 0.189 & 0.041 & 0.033 \\
\textit{G4} & 8 & 0.024 & 0.022 & \textbf{0.018} & 0.148 & -- & -- \\
\addlinespace[2pt]
GP 1600 & 2 & \textbf{0.018} & 0.036 & 0.020 & 0.175 & -- & -- \\
GP 3200 & 1 & \textbf{0.015} & 0.049 & 0.018 & 0.470 & -- & -- \\
GP 6400 & 3 & \textbf{0.028} & 0.035 & 0.031 & 0.047 & -- & -- \\
GP 10000 & 2 & \textbf{0.025} & 0.044 & 0.039 & 0.149 & -- & -- \\
\textit{GP} & 8 & \textbf{0.021} & 0.041 & 0.027 & 0.210 & -- & -- \\
\addlinespace[2pt]
IP 1000 & 1 & \textbf{0.021} & 0.028 & 0.028 & 0.120 & -- & -- \\
IP 1600 & 5 & \textbf{0.023} & 0.037 & 0.042 & 0.068 & 0.042 & 0.033 \\
IP 2000 & 1 & \textbf{0.009} & 0.040 & 0.066 & 0.078 & -- & -- \\
\textit{IP} & 7 & \textbf{0.018} & 0.035 & 0.045 & 0.089 & -- & -- \\
\addlinespace[2pt]
N6 400 & 3 & \textbf{0.019} & 0.034 & \textbf{0.019} & 0.095 & -- & -- \\
N6 800 & 2 & \textbf{0.012} & 0.027 & 0.025 & 0.216 & 0.067 & 0.035 \\
N6 3200 & 3 & \textbf{0.018} & 0.041 & 0.036 & 0.114 & -- & -- \\
\textit{N6} & 8 & \textbf{0.017} & 0.034 & 0.026 & 0.142 & -- & -- \\
\addlinespace[2pt]
S6 400 & 1 & 0.031 & 0.044 & \textbf{0.021} & 0.136 & 0.030 & 0.043 \\
S6 800 & 2 & \textbf{0.025} & 0.028 & 0.032 & 0.164 & 0.053 & 0.039 \\
S6 1600 & 1 & \textbf{0.029} & 0.035 & 0.044 & 0.082 & 0.057 & 0.042 \\
S6 3200 & 4 & \textbf{0.031} & 0.056 & 0.061 & 0.118 & 0.080 & 0.071 \\
\textit{S6} & 8 & \textbf{0.029} & 0.040 & 0.039 & 0.125 & 0.055 & 0.049 \\
\midrule
All & 39 & \textbf{0.022} & 0.035 & 0.031 & 0.143 & -- & -- \\
\bottomrule
\end{tabular}
\caption{\textbf{SIDD by camera and camera--ISO class: $D_{\mathrm{KL}}$, $32\times32$ tiles} (lower is better). Mean over five generator runs and over the $n$ source images of each class; italic rows weight the camera's classes equally, and All weights cameras equally, which reproduces Tab.~\mainref{tab:sidd} of the main paper. Lowest value of each row in bold; descriptive only, without intervals. --: class or camera not fully covered by the checkpoint (sRGB-Flow and NM-FlowGAN: 19-source block only). $^*$Our implementation of the published model.}
\label{tab:supp-sidd-kl-all-tiles32}
\end{table}

\begin{table}[tp]
\centering
\scriptsize
\setlength{\tabcolsep}{2.4pt}
\begin{tabular}{@{}lrcccccc@{}}
\toprule
Class & $n$ & \textbf{PGNM} & NAFlow & GMDCN$^*$ & DANet & sRGB-Flow & NM-FlowG. \\
\midrule
G4 400 & 4 & 0.0183 & 0.0074 & \textbf{0.0049} & 0.0455 & -- & -- \\
G4 800 & 4 & 0.0116 & 0.0121 & \textbf{0.0071} & 0.1073 & 0.0290 & 0.0165 \\
\textit{G4} & 8 & 0.0150 & 0.0098 & \textbf{0.0060} & 0.0764 & -- & -- \\
\addlinespace[2pt]
GP 1600 & 2 & 0.0080 & 0.0224 & \textbf{0.0069} & 0.1224 & -- & -- \\
GP 3200 & 1 & \textbf{0.0011} & 0.0359 & 0.0036 & 0.3664 & -- & -- \\
GP 6400 & 3 & 0.0125 & 0.0162 & \textbf{0.0124} & 0.0207 & -- & -- \\
GP 10000 & 2 & \textbf{0.0054} & 0.0238 & 0.0203 & 0.0728 & -- & -- \\
\textit{GP} & 8 & \textbf{0.0068} & 0.0246 & 0.0108 & 0.1456 & -- & -- \\
\addlinespace[2pt]
IP 1000 & 1 & \textbf{0.0134} & 0.0197 & 0.0190 & 0.0628 & -- & -- \\
IP 1600 & 5 & \textbf{0.0124} & 0.0266 & 0.0294 & 0.0342 & 0.0356 & 0.0218 \\
IP 2000 & 1 & \textbf{0.0023} & 0.0289 & 0.0524 & 0.0420 & -- & -- \\
\textit{IP} & 7 & \textbf{0.0094} & 0.0251 & 0.0336 & 0.0463 & -- & -- \\
\addlinespace[2pt]
N6 400 & 3 & 0.0094 & 0.0220 & \textbf{0.0065} & 0.0417 & -- & -- \\
N6 800 & 2 & \textbf{0.0022} & 0.0144 & 0.0104 & 0.1633 & 0.0562 & 0.0195 \\
N6 3200 & 3 & \textbf{0.0031} & 0.0241 & 0.0152 & 0.0493 & -- & -- \\
\textit{N6} & 8 & \textbf{0.0049} & 0.0202 & 0.0107 & 0.0847 & -- & -- \\
\addlinespace[2pt]
S6 400 & 1 & 0.0257 & 0.0364 & \textbf{0.0134} & 0.0750 & 0.0222 & 0.0358 \\
S6 800 & 2 & \textbf{0.0162} & 0.0163 & 0.0215 & 0.1081 & 0.0438 & 0.0281 \\
S6 1600 & 1 & \textbf{0.0161} & 0.0198 & 0.0272 & 0.0595 & 0.0425 & 0.0265 \\
S6 3200 & 4 & \textbf{0.0147} & 0.0324 & 0.0301 & 0.0626 & 0.0580 & 0.0415 \\
\textit{S6} & 8 & \textbf{0.0182} & 0.0262 & 0.0230 & 0.0763 & 0.0416 & 0.0330 \\
\midrule
All & 39 & \textbf{0.0108} & 0.0212 & 0.0168 & 0.0859 & -- & -- \\
\bottomrule
\end{tabular}
\caption{\textbf{SIDD by camera and camera--ISO class: $D_{\mathrm{KL}}$, $256\times256$ blocks} (lower is better). Weighting and notation as in Tab.~\ref{tab:supp-sidd-kl-all-tiles32}.}
\label{tab:supp-sidd-kl-block256}
\end{table}

\begin{table}[tp]
\centering
\scriptsize
\setlength{\tabcolsep}{2.4pt}
\begin{tabular}{@{}lrcccccc@{}}
\toprule
Class & $n$ & \textbf{PGNM} & NAFlow & GMDCN$^*$ & DANet & sRGB-Flow & NM-FlowG. \\
\midrule
G4 400 & 4 & 1.459 & 0.795 & \textbf{0.705} & 1.272 & -- & -- \\
G4 800 & 4 & 1.348 & 1.406 & \textbf{1.174} & 2.485 & 3.804 & 1.636 \\
\textit{G4} & 8 & 1.403 & 1.100 & \textbf{0.940} & 1.879 & -- & -- \\
\addlinespace[2pt]
GP 1600 & 2 & \textbf{0.962} & 1.864 & 1.240 & 2.835 & -- & -- \\
GP 3200 & 1 & \textbf{0.583} & 2.844 & 1.106 & 6.542 & -- & -- \\
GP 6400 & 3 & 2.738 & 3.373 & \textbf{2.600} & 3.297 & -- & -- \\
GP 10000 & 2 & \textbf{1.894} & 4.137 & 3.220 & 4.404 & -- & -- \\
\textit{GP} & 8 & \textbf{1.544} & 3.055 & 2.041 & 4.269 & -- & -- \\
\addlinespace[2pt]
IP 1000 & 1 & 0.992 & 1.681 & \textbf{0.899} & 1.722 & -- & -- \\
IP 1600 & 5 & \textbf{0.941} & 1.684 & 1.556 & 1.263 & 2.442 & 1.409 \\
IP 2000 & 1 & \textbf{0.371} & 1.551 & 1.460 & 1.212 & -- & -- \\
\textit{IP} & 7 & \textbf{0.768} & 1.638 & 1.305 & 1.399 & -- & -- \\
\addlinespace[2pt]
N6 400 & 3 & \textbf{0.861} & 1.350 & 0.896 & 1.619 & -- & -- \\
N6 800 & 2 & \textbf{0.592} & 1.445 & 1.058 & 2.536 & 3.299 & 1.732 \\
N6 3200 & 3 & \textbf{1.576} & 4.047 & 2.649 & 2.883 & -- & -- \\
\textit{N6} & 8 & \textbf{1.009} & 2.281 & 1.534 & 2.346 & -- & -- \\
\addlinespace[2pt]
S6 400 & 1 & 1.800 & 1.865 & \textbf{1.029} & 1.669 & 3.334 & 2.355 \\
S6 800 & 2 & 1.394 & \textbf{1.383} & 1.774 & 2.247 & 5.695 & 2.297 \\
S6 1600 & 1 & \textbf{2.024} & 2.201 & 2.029 & 5.035 & 7.605 & 3.079 \\
S6 3200 & 4 & \textbf{2.261} & 4.353 & 2.956 & 3.259 & 12.333 & 5.123 \\
\textit{S6} & 8 & \textbf{1.870} & 2.451 & 1.947 & 3.052 & 7.242 & 3.214 \\
\midrule
All & 39 & \textbf{1.319} & 2.105 & 1.553 & 2.589 & -- & -- \\
\bottomrule
\end{tabular}
\caption{\textbf{SIDD by camera and camera--ISO class: conditional $W_1$ (8-bit codes)} (lower is better). Weighting and notation as in Tab.~\ref{tab:supp-sidd-kl-all-tiles32}.}
\label{tab:supp-sidd-full-conditional-w1}
\end{table}

\begin{table}[tp]
\centering
\scriptsize
\setlength{\tabcolsep}{2.4pt}
\begin{tabular}{@{}lrcccccc@{}}
\toprule
Class & $n$ & \textbf{PGNM} & NAFlow & GMDCN$^*$ & DANet & sRGB-Flow & NM-FlowG. \\
\midrule
G4 400 & 4 & \textbf{0.0079} & 0.0205 & 0.0108 & 0.0190 & -- & -- \\
G4 800 & 4 & \textbf{0.0046} & 0.0185 & 0.0125 & 0.0149 & 0.0247 & 0.0084 \\
\textit{G4} & 8 & \textbf{0.0062} & 0.0195 & 0.0117 & 0.0170 & -- & -- \\
\addlinespace[2pt]
GP 1600 & 2 & 0.0114 & 0.0219 & \textbf{0.0103} & 0.0132 & -- & -- \\
GP 3200 & 1 & \textbf{0.0046} & 0.0216 & 0.0134 & 0.0173 & -- & -- \\
GP 6400 & 3 & \textbf{0.0063} & 0.0168 & 0.0120 & 0.0120 & -- & -- \\
GP 10000 & 2 & \textbf{0.0091} & 0.0185 & 0.0134 & 0.0173 & -- & -- \\
\textit{GP} & 8 & \textbf{0.0078} & 0.0197 & 0.0123 & 0.0150 & -- & -- \\
\addlinespace[2pt]
IP 1000 & 1 & 0.0191 & 0.0259 & 0.0152 & \textbf{0.0149} & -- & -- \\
IP 1600 & 5 & 0.0252 & 0.0225 & 0.0189 & 0.0234 & 0.0215 & \textbf{0.0146} \\
IP 2000 & 1 & \textbf{0.0086} & 0.0243 & 0.0237 & 0.0191 & -- & -- \\
\textit{IP} & 7 & \textbf{0.0176} & 0.0242 & 0.0193 & 0.0192 & -- & -- \\
\addlinespace[2pt]
N6 400 & 3 & \textbf{0.0074} & 0.0215 & 0.0139 & 0.0195 & -- & -- \\
N6 800 & 2 & \textbf{0.0076} & 0.0160 & 0.0122 & 0.0149 & 0.0246 & 0.0098 \\
N6 3200 & 3 & \textbf{0.0095} & 0.0140 & \textbf{0.0095} & 0.0106 & -- & -- \\
\textit{N6} & 8 & \textbf{0.0082} & 0.0172 & 0.0119 & 0.0150 & -- & -- \\
\addlinespace[2pt]
S6 400 & 1 & \textbf{0.0079} & 0.0192 & 0.0132 & 0.0172 & 0.0202 & 0.0128 \\
S6 800 & 2 & 0.0058 & 0.0175 & 0.0102 & 0.0143 & 0.0214 & \textbf{0.0051} \\
S6 1600 & 1 & 0.0081 & 0.0158 & 0.0073 & 0.0117 & 0.0208 & \textbf{0.0042} \\
S6 3200 & 4 & 0.0076 & 0.0126 & 0.0083 & 0.0111 & 0.0233 & \textbf{0.0066} \\
\textit{S6} & 8 & 0.0073 & 0.0163 & 0.0097 & 0.0136 & 0.0214 & \textbf{0.0072} \\
\midrule
All & 39 & \textbf{0.0094} & 0.0194 & 0.0130 & 0.0159 & -- & -- \\
\bottomrule
\end{tabular}
\caption{\textbf{SIDD by camera and camera--ISO class: spatial correlation error} (lower is better). Weighting and notation as in Tab.~\ref{tab:supp-sidd-kl-all-tiles32}.}
\label{tab:supp-sidd-full-acf-error}
\end{table}

\begin{table}[tp]
\centering
\scriptsize
\setlength{\tabcolsep}{2.4pt}
\begin{tabular}{@{}lrcccccc@{}}
\toprule
Class & $n$ & \textbf{PGNM} & NAFlow & GMDCN$^*$ & DANet & sRGB-Flow & NM-FlowG. \\
\midrule
G4 400 & 4 & \textbf{0.0110} & 0.0479 & 0.0276 & 0.0568 & -- & -- \\
G4 800 & 4 & \textbf{0.0135} & 0.0565 & 0.0439 & 0.0607 & 0.0414 & 0.0354 \\
\textit{G4} & 8 & \textbf{0.0122} & 0.0522 & 0.0358 & 0.0588 & -- & -- \\
\addlinespace[2pt]
GP 1600 & 2 & \textbf{0.0213} & 0.0480 & 0.0316 & 0.0535 & -- & -- \\
GP 3200 & 1 & \textbf{0.0062} & 0.0545 & 0.0286 & 0.0687 & -- & -- \\
GP 6400 & 3 & \textbf{0.0196} & 0.0592 & 0.0241 & 0.0753 & -- & -- \\
GP 10000 & 2 & \textbf{0.0113} & 0.0407 & 0.0346 & 0.0546 & -- & -- \\
\textit{GP} & 8 & \textbf{0.0146} & 0.0506 & 0.0297 & 0.0630 & -- & -- \\
\addlinespace[2pt]
IP 1000 & 1 & 0.0287 & 0.0607 & 0.0304 & \textbf{0.0285} & -- & -- \\
IP 1600 & 5 & \textbf{0.0142} & 0.0593 & 0.0369 & 0.0456 & 0.0615 & 0.0231 \\
IP 2000 & 1 & \textbf{0.0151} & 0.0706 & 0.0498 & 0.0590 & -- & -- \\
\textit{IP} & 7 & \textbf{0.0193} & 0.0635 & 0.0390 & 0.0444 & -- & -- \\
\addlinespace[2pt]
N6 400 & 3 & \textbf{0.0117} & 0.0258 & 0.0356 & 0.0299 & -- & -- \\
N6 800 & 2 & \textbf{0.0135} & 0.0277 & 0.0277 & 0.0286 & 0.0357 & 0.0597 \\
N6 3200 & 3 & \textbf{0.0121} & 0.0342 & 0.0204 & 0.0237 & -- & -- \\
\textit{N6} & 8 & \textbf{0.0124} & 0.0293 & 0.0279 & 0.0274 & -- & -- \\
\addlinespace[2pt]
S6 400 & 1 & 0.0264 & 0.0281 & 0.0516 & 0.0516 & \textbf{0.0255} & 0.0401 \\
S6 800 & 2 & \textbf{0.0109} & 0.0445 & 0.0217 & 0.0368 & 0.0398 & 0.0285 \\
S6 1600 & 1 & \textbf{0.0047} & 0.0526 & 0.0148 & 0.0198 & 0.0458 & 0.0151 \\
S6 3200 & 4 & \textbf{0.0124} & 0.0397 & 0.0143 & 0.0252 & 0.0595 & 0.0151 \\
\textit{S6} & 8 & \textbf{0.0136} & 0.0412 & 0.0256 & 0.0334 & 0.0427 & 0.0247 \\
\midrule
All & 39 & \textbf{0.0144} & 0.0474 & 0.0316 & 0.0454 & -- & -- \\
\bottomrule
\end{tabular}
\caption{\textbf{SIDD by camera and camera--ISO class: color correlation error} (lower is better). Weighting and notation as in Tab.~\ref{tab:supp-sidd-kl-all-tiles32}.}
\label{tab:supp-sidd-full-rgb-crosslag-error}
\end{table}

\paragraph{Sources.}
Tab.~\ref{tab:supp-sidd-sources} lists the 40 SIDD Validation source images. The
camera--ISO labels follow the order of the SIDD benchmark scene instances:
Validation image $i$ is the $i$-th benchmark instance in sorted order, which we
read from the benchmark archive's index and confirmed for all 40 images by
matching their content to SIDD-Medium images of the same scene and camera. The
benchmark instances are not part of SIDD-Medium; they come from eight of its ten
scenes and its five cameras but are separate captures.

\paragraph{Results by camera and ISO.}
Tabs.~\ref{tab:supp-sidd-kl-all-tiles32}--\ref{tab:supp-sidd-full-rgb-crosslag-error}
break down Tab.~\mainref{tab:sidd} of the main paper by camera and camera--ISO
class, one table per metric, and add the $256\times256$ block KL
(Tab.~\ref{tab:supp-sidd-kl-block256}). Each value averages the five generator
runs and the class's source images; camera rows weight the camera's classes
equally, and the last row reproduces the main table. A class contains one to five
source images, so the tables are descriptive and carry no intervals.

PGNM has the lowest value in 13 of the 16 classes for the tile KL, 10 for the
block KL, 10 for conditional $W_1$, 10 for spatial and 14 for color correlation
error; values equal at the printed precision count for every method concerned.
GMDCN has all remaining lowest KL values and five of the six remaining lowest
conditional $W_1$, including both G4 classes and S6 ISO\,400 on all three of these
metrics; NAFlow has the lowest
conditional $W_1$ for S6 ISO\,800, and NM-FlowGAN the lowest spatial correlation
error in four of the seven classes it covers.

\paragraph{Film grain by radius.}
Tab.~\ref{tab:supp-film-radius} breaks down Tab.~\mainref{tab:film} of the main
paper by grain radius, with the main table's weighting. Zhang23 has the lowest
conditional $W_1$ at $r=0.025$ and $0.05$, and PGNM from $r=0.075$ on. In spatial
correlation error, Zhang23 is lowest up to $r=0.6$ and GrainNet at $r=0.8$, where
Zhang23's error rises to 0.11; PGNM's is above GrainNet's from $r=0.4$ on. In DISTS
and LPIPS, PGNM and Zhang23 differ by at most 0.0013 up to $r=0.1$, and GrainNet is
lowest from $r=0.2$ on.

\section{Further analyses}
\label{sec:supp-analyses}

\begin{table*}[tp]
\centering
\footnotesize
\setlength{\tabcolsep}{6pt}
\begin{tabular}{@{}lcccc@{\hspace{14pt}}ccc@{}}
\toprule
& \multicolumn{4}{c}{Per-bin residual distribution$\downarrow$} & \multicolumn{3}{c}{Dependence of standardized residuals} \\
\cmidrule(r{12pt}){2-5}\cmidrule{6-8}
Method & $W_1$ (codes) & Bias (codes) & Strength (codes) & Shape ($\sigma$) & RGB, same pixel & Lag 1, horizontal & Lag 1, vertical \\
\midrule
\multicolumn{8}{@{}l}{\textit{All 39 sources (16 classes)}} \\
Real & & & & & 0.357 & 0.440 & 0.441 \\
\textbf{PGNM} & \textbf{1.319} & \textbf{0.92} & 1.84 & \textbf{0.041} & 0.341 & 0.429 & 0.430 \\
NAFlow & 2.105 & 1.85 & 2.17 & 0.059 & 0.370 & 0.485 & 0.490 \\
GMDCN$^*$ & 1.553 & 1.12 & \textbf{1.26} & 0.059 & 0.361 & 0.440 & 0.443 \\
DANet & 2.589 & 1.09 & 2.95 & 0.089 & 0.350 & 0.443 & 0.439 \\
\midrule
\multicolumn{8}{@{}l}{\textit{19 sources covered by sRGB-Flow and NM-FlowGAN}} \\
Real & & & & & 0.329 & 0.451 & 0.451 \\
\textbf{PGNM} & \textbf{1.187} & \textbf{0.76} & 1.64 & \textbf{0.041} & 0.308 & 0.438 & 0.439 \\
NAFlow & 1.746 & 1.55 & 1.71 & 0.057 & 0.365 & 0.488 & 0.492 \\
GMDCN$^*$ & 1.434 & 1.09 & \textbf{1.03} & 0.064 & 0.337 & 0.442 & 0.446 \\
DANet & 2.334 & 1.01 & 2.64 & 0.103 & 0.356 & 0.444 & 0.438 \\
sRGB-Flow & 4.197 & 4.27 & 2.35 & 0.091 & 0.393 & 0.324 & 0.320 \\
NM-FlowGAN & 1.998 & 1.53 & 2.08 & 0.083 & 0.288 & 0.446 & 0.444 \\
\bottomrule
\end{tabular}
\caption{\textbf{Where the SIDD residual errors come from.} Descriptive per-bin error measures beside conditional $W_1$ on the scored outputs (all pixels, five runs, weighted as in Tab.~\mainref{tab:sidd} of the main paper). In each clean-intensity bin: bias, the difference of residual means; strength, the difference of standard deviations (both root mean square over bins, weighted by pixel count, averaged over RGB); shape, the mean absolute difference of 256 quantiles after each residual is standardized by its own bin mean and standard deviation (bins with a standard deviation below 0.5 codes are skipped). Right: correlations of residuals standardized per bin, same pixel across channels and to the next pixel within a channel; the Real rows give the target. Lowest error in bold. $^*$Our implementation of the published model.}
\label{tab:supp-decomposition}
\end{table*}

\begin{table*}[tp]
\sidecaptioned{\footnotesize
\setlength{\tabcolsep}{8pt}
\begin{tabular}{@{}lccccc@{}}
\toprule
& \multicolumn{2}{c}{$W_1$ (codes)} & \multicolumn{2}{c}{KL, $32^2$ tiles} & PGNM lower \\
\cmidrule(lr){2-3}\cmidrule(lr){4-5}
Oracle correction & PGNM & GMDCN$^*$ & PGNM & GMDCN$^*$ & ($W_1$ / KL) \\
\midrule
As scored & 1.318 & 1.554 & 0.0218 & 0.0312 & 11 / 12 of 16 \\
One gain per channel & 1.045 & 1.432 & 0.0241 & 0.0290 & 15 / 12 of 16 \\
Strength per bin & 0.847 & 1.346 & 0.0231 & 0.0284 & 15 / 13 of 16 \\
Mean and strength per bin & 0.489 & 0.885 & 0.0199 & 0.0237 & 15 / 13 of 16 \\
\bottomrule
\end{tabular}}
{\caption{\textbf{Oracle strength check} (diagnostic only; it reads the scored real image). Each generated residual is rescaled about its own mean so that its standard deviation equals the real one, per channel or per clean-intensity bin; the last row also moves each bin mean to the real one. Integer residuals are dequantized with uniform dither first. First two generator runs, all 39 sources, weighted as in the main table; last column: camera--ISO classes in which PGNM's score is lower than GMDCN's.}
\label{tab:supp-oracle}}
\end{table*}

These analyses were computed after the main evaluation, on its scored outputs or
with its fits and checkpoints unchanged; they describe the reported results and are
not additional comparisons.

\subsection{Where the SIDD errors come from}
\label{sec:supp-decomposition}

Tab.~\ref{tab:supp-decomposition} complements conditional $W_1$ with three per-bin
error measures, the difference of residual means (bias), of standard deviations
(strength) and of standardized quantiles (shape), and lists the correlations of
residuals standardized per bin. On all 39 sources, PGNM has the lowest bias (0.92 codes) and
shape error ($0.041\sigma$), whereas GMDCN matches the per-bin strength best (1.26
against 1.84 codes for PGNM); with PGNM's lower color correlation error
(Tab.~\mainref{tab:sidd} of the main paper), this is the comparison summarized in
Sec.~\mainref{sec:analysis}. That error covers neighboring pixels only; at the same
pixel, GMDCN's channel correlation is closer to the real one (0.361 against 0.357,
PGNM 0.341). The dependence columns show the two directions of error in
neighboring-pixel correlation named there: NAFlow's is too high (0.49 against 0.44
for real noise) and sRGB-Flow's too low (0.32 against 0.45 on the
19-source block); NM-FlowGAN's same-pixel color correlation is the lowest (0.29
against 0.33).

Because strength is PGNM's largest remaining error, we checked how much of the
$W_1$ gap to GMDCN it explains with an oracle that rescales each generated
residual to the standard deviation of the scored real image
(Tab.~\ref{tab:supp-oracle}). The oracle reads the evaluation data and is not a
method. With one gain per channel, PGNM's conditional $W_1$ falls from 1.32 to 1.05
codes and is lower than GMDCN's in 15 of the 16 classes, against 11 as scored;
per-bin corrections keep this ranking. The tile KL changes little under any
correction.

\subsection{A clean image from outside SIDD}
\label{sec:supp-unseen}

\begin{table*}[tp]
\centering
\footnotesize
\setlength{\tabcolsep}{2.5pt}
\begin{tabular}{@{}lccccccc@{}}
\toprule
\multicolumn{8}{@{}l}{\textit{Residual standard deviation (codes): SIDD $\to$ Kodak (change)}} \\
Class & Real & \textbf{PGNM} & GMDCN$^*$ & NM-FlowGAN & NAFlow & sRGB-Flow & DANet \\
\midrule
G4 800 & 14.6 & 13.7 $\to$ 12.2 ($-$11\%) & 14.1 $\to$ 13.4 ($-$5\%) & 15.2 $\to$ 14.4 ($-$5\%) & 14.2 $\to$ 21.7 ($+$52\%) & 14.8 $\to$ 15.8 ($+$7\%) & 15.3 $\to$ 7.8 ($-$49\%) \\
GP 6400 & 31.8 & 34.6 $\to$ 31.0 ($-$10\%) & 31.0 $\to$ 29.1 ($-$6\%) & -- & 30.0 $\to$ 37.8 ($+$26\%) & -- & 33.6 $\to$ 8.3 ($-$75\%) \\
IP 1600 & 9.6 & 10.2 $\to$ 7.7 ($-$24\%) & 9.6 $\to$ 8.7 ($-$10\%) & 9.6 $\to$ 8.9 ($-$7\%) & 9.6 $\to$ 17.1 ($+$78\%) & 10.0 $\to$ 10.8 ($+$8\%) & 9.0 $\to$ 8.0 ($-$11\%) \\
N6 800 & 14.3 & 14.5 $\to$ 12.6 ($-$13\%) & 14.5 $\to$ 14.0 ($-$3\%) & 14.4 $\to$ 13.9 ($-$3\%) & 12.9 $\to$ 21.1 ($+$64\%) & 11.6 $\to$ 12.2 ($+$5\%) & 13.5 $\to$ 7.8 ($-$42\%) \\
S6 800 & 17.5 & 19.0 $\to$ 17.2 ($-$9\%) & 19.1 $\to$ 16.1 ($-$16\%) & 19.8 $\to$ 20.3 ($+$2\%) & 16.7 $\to$ 25.7 ($+$54\%) & 19.9 $\to$ 20.8 ($+$4\%) & 18.4 $\to$ 8.3 ($-$55\%) \\
\midrule
\multicolumn{8}{@{}l}{\textit{Detail slope: SIDD $\to$ Kodak}} \\
Class & Real & \textbf{PGNM} & GMDCN$^*$ & NM-FlowGAN & NAFlow & sRGB-Flow & DANet \\
\midrule
G4 800 & $+$0.61 & $-$0.02 $\to$ $+$0.01 & $+$0.45 $\to$ $+$0.25 & $+$0.49 $\to$ $+$0.23 & $-$0.34 $\to$ $-$0.35 & $-$0.19 $\to$ $-$0.11 & $+$0.26 $\to$ $+$0.03 \\
GP 6400 & $+$0.03 & $-$0.19 $\to$ $-$0.01 & $+$0.03 $\to$ $+$0.14 & -- & $-$0.71 $\to$ $-$0.47 & -- & $-$0.06 $\to$ $+$0.04 \\
IP 1600 & $-$0.11 & $-$0.30 $\to$ $+$0.02 & $-$0.25 $\to$ $+$0.04 & $-$0.25 $\to$ $+$0.20 & $-$1.02 $\to$ $-$0.35 & $-$0.38 $\to$ $-$0.07 & $-$0.15 $\to$ $+$0.04 \\
N6 800 & $+$0.53 & $-$0.01 $\to$ $+$0.01 & $+$0.32 $\to$ $+$0.27 & $+$0.14 $\to$ $+$0.23 & $-$0.90 $\to$ $-$0.35 & $-$0.19 $\to$ $-$0.08 & $-$0.05 $\to$ $+$0.03 \\
S6 800 & $+$0.68 & $+$0.04 $\to$ 0.00 & $+$0.41 $\to$ $+$0.28 & $+$0.33 $\to$ $+$0.30 & $-$0.72 $\to$ $-$0.37 & $-$0.25 $\to$ $-$0.15 & $+$0.03 $\to$ $+$0.04 \\
\bottomrule
\end{tabular}
\caption{\textbf{Every SIDD generator on a clean image from outside SIDD} (Kodak \emph{kodim19}, the evaluated fits and checkpoints, one draw per method and class). SIDD values: the same method's first generator run on the class's Validation sources; Real: the real SIDD residual. Detail slope: least-squares slope (through the origin) of the residual on the clean image's detail, the clean image minus its Gaussian blur ($\sigma=1$ pixel), per channel; positive means the output is sharper than the clean image. --: class outside the 19-source block. $^*$Our implementation of the published model.}
\label{tab:supp-unseen}
\end{table*}

\begin{table*}[tp]
\sidecaptioned{\footnotesize
\setlength{\tabcolsep}{7pt}
\begin{tabular}{@{}lrrrrrrrr@{}}
\toprule
Radius & $0.025$ & $0.05$ & $0.075$ & $0.1$ & $0.2$ & $0.4$ & $0.6$ & $0.8$ \\
\midrule
Newson (codes) & 1.70 & 3.29 & 4.91 & 6.53 & 12.95 & 25.51 & 37.07 & 47.12 \\
\textbf{PGNM}\,+\,$B$ & $+$3.6\% & $+$1.5\% & $+$1.1\% & $+$0.3\% & $+$0.9\% & $+$0.7\% & $+$0.4\% & $+$0.9\% \\
Zhang23\,+\,$B$ & $-$0.4\% & $+$0.3\% & $+$0.6\% & $+$1.1\% & $+$1.5\% & $+$2.2\% & $+$0.7\% & $-$1.1\% \\
GrainNet & $+$60.5\% & $+$9.4\% & $+$3.2\% & $+$2.1\% & $+$1.6\% & $+$0.2\% & $+$0.2\% & $+$0.4\% \\
Ind. Newson render & $-$1.0\% & $+$0.1\% & $-$0.2\% & $+$0.1\% & $+$0.1\% & $+$0.5\% & $-$0.2\% & $-$0.1\% \\
\bottomrule
\end{tabular}}
{\caption{\textbf{Film grain on the gray Kodak image.} Residual standard deviation relative to $\mu=B*x$: the Newson reference in 8-bit codes, then each method's difference from it, with the evaluated film fits and settings (one draw).}
\label{tab:supp-unseen-film}}
\end{table*}

Tab.~\ref{tab:supp-unseen} gives the numbers behind the Kodak check of
Sec.~\mainref{sec:analysis}, and Fig.~\ref{fig:supp-unseen} shows one class. Each
SIDD fit and checkpoint was applied unchanged to \emph{kodim19}, whose most
textured windows have more than twice the local contrast of the most textured SIDD
window, for five camera--ISO classes, one per camera. Residual
strength depends on the clean intensities present, so every method changes
somewhat between images. NAFlow's strength rises by 26--78\% above its own SIDD
level; DANet, which receives no camera or ISO input, produces 7.8--8.3 codes in
every class, while its SIDD level varies from 9.0 to 33.6; the other methods
change by at most 24\%.

The detail slope measures whether a residual follows the clean image's fine
structure. Real SIDD residuals have a positive slope in the G4, N6 and S6 classes
($+0.53$ to $+0.68$): the noisy captures are slightly sharper than their reference
images. GMDCN and NM-FlowGAN reproduce this on SIDD and keep a positive slope on
Kodak in these classes ($+0.23$ to $+0.30$), which we read as learned behavior
rather than a failure; NAFlow's slope is negative on both. PGNM's slope stays
within about $\pm0.02$ on Kodak,
as expected from a model that uses the clean image only through its intensity.

The same check with the film fits on a gray version of \emph{kodim19}
(Tab.~\ref{tab:supp-unseen-film}) leaves PGNM's grain strength within 4\% of the
Newson reference at every radius ($+0.3\%$ to $+3.6\%$), comparable to Zhang23
($-1.1\%$ to $+2.2\%$) and to the independent Newson render ($-1.0\%$ to $+0.5\%$);
GrainNet's finest grain is 60\% stronger.

\subsection{Offset and strength of film-grain outputs}
\label{sec:supp-film-offset}

\begin{table*}[tp]
\centering
\footnotesize
\setlength{\tabcolsep}{4.5pt}
\begin{tabular}{@{}lccccc@{\hspace{14pt}}ccccc@{}}
\toprule
& \multicolumn{5}{c}{Offset (codes)} & \multicolumn{5}{c}{Strength (codes)} \\
\cmidrule(r{12pt}){2-6}\cmidrule{7-11}
Radius & Newson & \textbf{PGNM}\,+\,$B$ & Zhang23\,+\,$B$ & GrainNet & \makecell{Ind. Newson\\render} & Newson & \textbf{PGNM}\,+\,$B$ & Zhang23\,+\,$B$ & GrainNet & \makecell{Ind. Newson\\render} \\
\midrule
$0.025$ & 0.05 & 0.09 & 0.02 & 3.61 & 0.05 & 1.49 & 1.59 & 1.50 & 2.15 & 1.49 \\
$0.05$ & 0.07 & 0.11 & 0.05 & 3.46 & 0.07 & 2.89 & 2.96 & 2.93 & 3.10 & 2.89 \\
$0.075$ & 0.09 & 0.13 & 0.07 & 3.34 & 0.09 & 4.32 & 4.38 & 4.38 & 4.43 & 4.31 \\
$0.1$ & 0.11 & 0.14 & 0.09 & 3.24 & 0.11 & 5.73 & 5.81 & 5.82 & 5.84 & 5.73 \\
$0.2$ & 0.20 & 0.23 & 0.18 & 2.91 & 0.20 & 11.37 & 11.54 & 11.59 & 11.47 & 11.37 \\
$0.4$ & 0.37 & 0.43 & 0.45 & 2.34 & 0.37 & 22.28 & 22.67 & 22.75 & 22.26 & 22.30 \\
$0.6$ & 0.54 & 0.58 & 0.92 & 2.26 & 0.54 & 32.22 & 32.87 & 32.29 & 32.37 & 32.22 \\
$0.8$ & 0.73 & 0.76 & 1.51 & 2.77 & 0.71 & 41.02 & 41.94 & 40.06 & 41.28 & 41.03 \\
\bottomrule
\end{tabular}
\caption{\textbf{Offset and strength of film-grain outputs} relative to the expected image $\mu=B*x$, inside the 3-pixel border, on the scored outputs (60 test images, five runs). In each merged clean-intensity bin of $\mu$ (as for conditional $W_1$): offset, the pixel-weighted mean of $|$bin mean residual$|$; strength, the pixel-weighted mean of the bin standard deviation.}
\label{tab:supp-film-offset}
\end{table*}

\begin{table}[tp]
\centering
\footnotesize
\setlength{\tabcolsep}{3pt}
\begin{tabular}{@{}lccccc@{}}
\toprule
& \multicolumn{4}{c}{Largest $|$bin mean$|$} & GrainNet \\
\cmidrule(lr){2-5}
Radius & Newson & \textbf{PGNM} & Zhang23 & \makecell{Ind. Newson\\render} & range \\
\midrule
$0.025$ & 0.21 & 0.25 & 0.03 & 0.21 & $-$6.3 to $+$6.9 \\
$0.05$ & 0.22 & 0.30 & 0.01 & 0.22 & $-$6.0 to $+$6.2 \\
$0.075$ & 0.23 & 0.30 & 0.01 & 0.22 & $-$5.8 to $+$5.5 \\
$0.1$ & 0.20 & 0.27 & 0.03 & 0.19 & $-$5.6 to $+$4.9 \\
$0.2$ & 0.28 & 0.27 & 0.16 & 0.29 & $-$5.2 to $+$3.1 \\
$0.4$ & 0.40 & 0.69 & 1.32 & 0.41 & $-$5.3 to $+$2.4 \\
$0.6$ & 0.38 & 0.83 & 2.82 & 0.36 & $-$6.2 to $+$3.0 \\
$0.8$ & 0.39 & 0.75 & 4.33 & 0.41 & $-$6.7 to $+$3.2 \\
\bottomrule
\end{tabular}
\caption{\textbf{Mean film-grain residual by intensity}, in 16 equal-width bins of $\mu$, pooled over the 60 test images and five runs (8-bit codes): the largest magnitude over bins for each method and the range of GrainNet's bin means.}
\label{tab:supp-film-bins}
\end{table}

Tab.~\ref{tab:supp-film-offset} measures, on the scored film outputs, how far each
method's mean residual departs from the expected image $\mu=B*x$ and how strong its
grain is, and Tab.~\ref{tab:supp-film-bins} resolves the mean residual by
intensity. Both Newson runs keep every bin mean within $\pm0.41$ codes. PGNM stays
within $\pm0.30$ codes up to $r=0.2$ and within $\pm0.83$ codes beyond, with a
pixel-weighted offset within 0.06 codes of the reference at every radius. Zhang23
stays within $\pm0.16$ codes up to $r=0.2$ and reaches 4.3 codes at $r=0.8$, where
its approximation no longer holds. GrainNet's mean residual depends on intensity at
every radius, from up to $+7$ codes in mid-tones at the finest radius ($+2.4$ to
$+3.2$ from $r=0.4$ on) to about $-6$ codes near white; its pixel-weighted offset is
2.3--3.6 codes against 0.05--0.73 for the reference. Its grain is 44\% stronger than
the reference at $r=0.025$ and 7\% at $r=0.05$, and within 3\% from $r=0.075$ on.
These offsets enter the residual errors of Tab.~\mainref{tab:film} of the main
paper, while the appearance distances largely tolerate them.

\FloatBarrier
\section{Denoising details}
\label{sec:supp-denoising}

\begin{table*}[tp]
\sidecaptioned{\small
\setlength{\tabcolsep}{7pt}
\begin{tabular}{@{}lcccccc@{}}
\toprule
& \multicolumn{2}{c}{All 40} & \multicolumn{2}{c}{19 in training classes} & \multicolumn{2}{c}{21 in other classes} \\
\cmidrule(lr){2-3}\cmidrule(lr){4-5}\cmidrule(l){6-7}
Training noise & PSNR$\uparrow$ & SSIM$\uparrow$ & PSNR$\uparrow$ & SSIM$\uparrow$ & PSNR$\uparrow$ & SSIM$\uparrow$ \\
\midrule
Real noise (reference) & 37.94 & 0.896 & 38.16 & 0.910 & 37.74 & 0.883 \\ \midrule
GMDCN & 36.15 & 0.872 & 36.22 & 0.884 & 36.08 & 0.862 \\
NM-FlowGAN & 35.15 & 0.875 & 35.31 & 0.888 & 35.00 & 0.862 \\
NAFlow & 36.72 & 0.878 & 36.80 & 0.889 & 36.66 & 0.868 \\
NAFlow-NAS & 36.88 & 0.883 & 36.96 & 0.893 & 36.81 & 0.873 \\
\textbf{PGNM} & \textbf{37.22} & \textbf{0.889} & \textbf{37.39} & \textbf{0.903} & \textbf{37.06} & \textbf{0.877} \\
\midrule
\multicolumn{7}{@{}l}{\textit{Reference outside the comparison}} \\
NAFlow's released DnCNN & 37.23 & 0.888 & 37.37 & 0.900 & 37.10 & 0.878 \\
\bottomrule
\end{tabular}}
{\caption{\textbf{Denoising on SIDD Validation by subset} (training seed 1, epoch 30). Training classes: the 15 camera--ISO classes of the training data; the other 21 images belong to other classes. Bold: best synthetic noise. NAFlow-NAS also reads the real noisy training patch. Reference: NAFlow's own released denoiser, trained by its authors with noise-aware sampling on all of SIDD-Medium. The main paper reports the All~40 columns.}
\label{tab:supp-denoise-val}}
\end{table*}

\begin{table*}[tp]
\sidecaptioned{\small
\setlength{\tabcolsep}{8pt}
\begin{tabular}{@{}lccc@{}}
\toprule
PGNM $-$ row (dB) & All 40 & 19 in training classes & 21 in other classes \\
\midrule
Real noise (reference) & $-$0.72 {\scriptsize[$-$0.89, $-$0.57]} & $-$0.77 {\scriptsize[$-$1.01, $-$0.52]} & $-$0.68 {\scriptsize[$-$0.81, $-$0.55]} \\ \midrule
GMDCN & $+$1.07 {\scriptsize[$+$0.85, $+$1.29]} & $+$1.17 {\scriptsize[$+$0.74, $+$1.56]} & $+$0.98 {\scriptsize[$+$0.77, $+$1.15]} \\
NM-FlowGAN & $+$2.06 {\scriptsize[$+$1.62, $+$2.49]} & $+$2.07 {\scriptsize[$+$1.39, $+$2.65]} & $+$2.05 {\scriptsize[$+$1.70, $+$2.56]} \\
NAFlow & $+$0.49 {\scriptsize[$+$0.27, $+$0.70]} & $+$0.59 {\scriptsize[$+$0.22, $+$0.92]} & $+$0.40 {\scriptsize[$+$0.12, $+$0.66]} \\
NAFlow-NAS & $+$0.33 {\scriptsize[$+$0.12, $+$0.54]} & $+$0.43 {\scriptsize[$+$0.08, $+$0.74]} & $+$0.25 {\scriptsize[$-$0.10, $+$0.56]} \\
\bottomrule
\end{tabular}}
{\caption{\textbf{Paired PSNR differences on SIDD Validation}: PGNM's denoiser minus the row's, mean over blocks, with a 95\% interval from resampling the 8 physical scenes (100,000 draws, seed 11). An interval containing zero is a tie.}
\label{tab:supp-denoise-diff}}
\end{table*}

\begin{figure*}[tp]
\centering
\includegraphics[width=\textwidth]{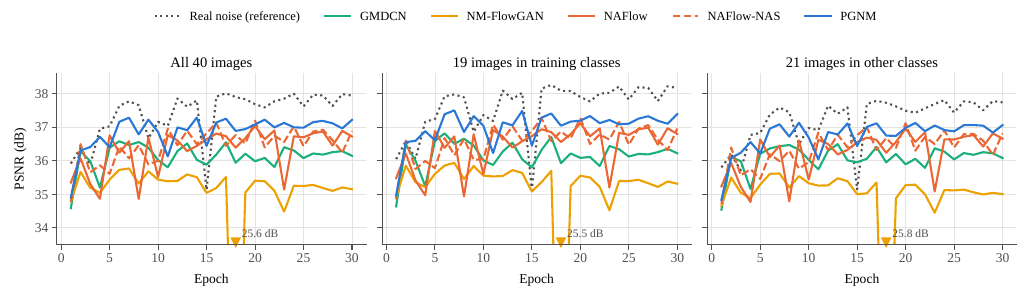}
\caption{\textbf{SIDD Validation PSNR after every training epoch} (seed 1). Descriptive monitoring only: every reported number uses the epoch-30 checkpoint, and no checkpoint, epoch or setting was chosen from these curves. The learning rate halves after every fifth epoch. A triangle at the lower edge marks a value below the plotted range.}
\label{fig:supp-denoise-curves}
\end{figure*}

This section gives the setup behind Tab.~\mainref{tab:denoise_sidd} of the main
paper and the complete results. The setup was fixed before any denoiser was
trained.

\paragraph{Training data and noise sources.}
We use NM-FlowGAN's denoiser split~\cite{han2023nmflowgan}: 123 SIDD-Medium scene
instances with both captures (246 sRGB pairs), which are all Medium instances of the
15 camera--ISO classes NM-FlowGAN supports (G4 ISO\,100/800; GP 100/800; IP
100/400/800/1600; N6 100/800; S6 100/400/800/1600/3200), so every generator is
conditioned natively. Each image is cut into $96\times96$ patches at stride 48,
with the last row and column clamped to the image edge, giving 1,473,956 patches
per epoch. Every epoch draws a fresh permutation and augmentation (rotation by a
multiple of $90^\circ$ and a horizontal flip, applied to the clean patch and its
real noisy patch before renoising), and every arm draws fresh noise for every patch
and epoch, conditioned on the image's camera and ISO:
\begin{itemize}
\item \emph{Real}: the patch's own real noisy observation, as the upper reference.
\item \emph{PGNM}: one preset per class, fitted by Alg.~\ref{alg:supp-fit} on every
      $256\times256$ crop of the same 246 captures, so PGNM sees no other data. Each
      patch gets an independent latent field of its own size instead of a bank
      window (Sec.~\ref{sec:supp-spec-experiments}), computed on CPU in the data
      loader.
\item \emph{GMDCN}: the checkpoint of Tab.~\mainref{tab:sidd}, with its camera--ISO
      condition map.
\item \emph{NAFlow}~\cite{kim2024naflow}: class-conditional sampling of the released
      checkpoint.
\item \emph{NAFlow-NAS}: the same checkpoint with its published noise-aware
      sampling, which encodes the patch's real noisy observation, weights all class
      Gaussians by their likelihood, and samples from the mixture. It receives
      information the other synthetic arms do not.
\item \emph{NM-FlowGAN}: the released sampler with its native camera and ISO inputs.
\end{itemize}
All generators use their native settings. Every
noisy patch, real or synthetic, is clipped to $[0,1]$ and rounded half up to 8 bits.
Neural generators run in float32 with TF32 disabled.

\paragraph{Denoiser and schedule.}
Every arm trains the DnCNN of NM-FlowGAN's denoiser configuration: 18
convolution layers of 64 features, batch normalization (momentum 0.9,
$\epsilon=10^{-4}$) on the 16 middle layers, and a residual output. Inputs are 8-bit
codes divided by 256, and the loss is the mean squared error in code units. Adam
(learning rate $10^{-4}$, $\beta=(0.9,0.999)$) halves its learning rate after every
fifth epoch, for 30 epochs at batch 128 (11,516 iterations per epoch), in float32
with TF32 disabled, on one RTX 3090 per run. Three training seeds were planned;
only seed~1 was run, for all six arms. For a given seed, all arms share the
initial weights, patch order and augmentation. The reported model is the epoch-30
checkpoint; no checkpoint, epoch or setting was chosen from the evaluation curves
(Fig.~\ref{fig:supp-denoise-curves}).

\paragraph{Test set.}
SIDD Validation: all 1,280 blocks of the 40 images, the 19 images whose
camera--ISO class is among the 15 training classes, and the other 21 images. Scores and their computation follow Sec.~\ref{sec:supp-metrics}; paired
intervals resample the eight physical scenes.

\paragraph{Reference outside the comparison.}
NAFlow's own released denoiser, trained by its authors with noise-aware sampling
on all of SIDD-Medium, scores 37.23\,dB on SIDD Validation; it is outside the
comparison because its training data and schedule differ.

\paragraph{Results.}
On SIDD Validation (Tabs.~\ref{tab:supp-denoise-val} and~\ref{tab:supp-denoise-diff}),
PGNM's denoiser is 0.72\,dB below the real-pair reference on all 40 images (0.77 and
0.68\,dB on the 19 and 21 images in and outside the training classes) and above every
other synthetic arm on all three sets; the 95\% interval of each lead excludes zero
except against NAFlow-NAS on the 21 images, which is a tie.

\section{Video delivery details}
\label{sec:supp-bandwidth}

\begin{figure*}[tp]
\centering
\includegraphics[width=\textwidth]{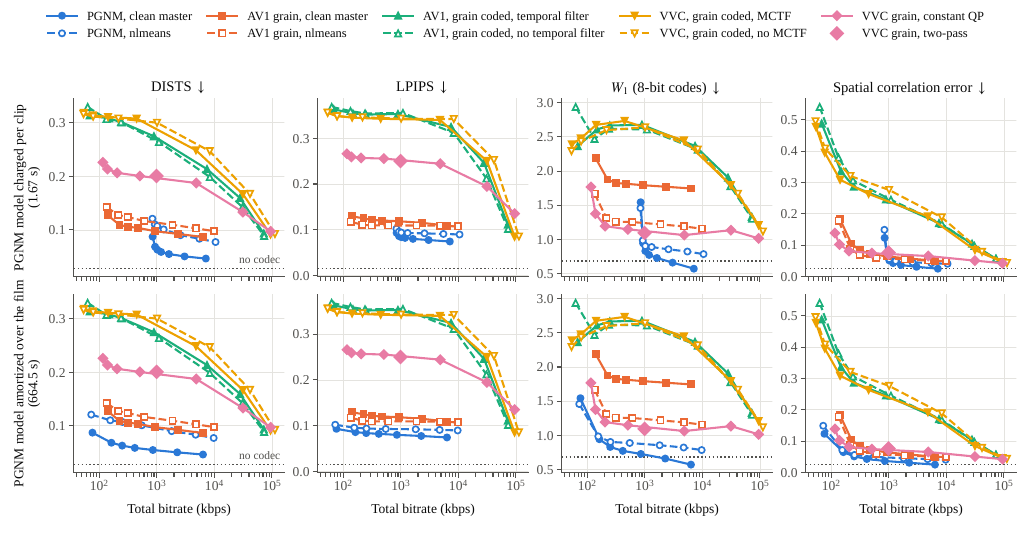}
\caption{\textbf{Rate--distortion curves of video delivery} (\emph{Nocturne}, 100 frames; lower is better). Each curve runs over its encoder's rate ladder (libaom Q\,15--63, VVenC QP\,17--47; lossless encodes omitted); the single diamond is VVC grain with two-pass rate control. Total bitrate includes PGNM's 161\,KB model, charged in full to this 1.67\,s clip (top, conservative) or amortized over the 664.5\,s film, which one grain setting covers (bottom); only PGNM's curves move between the rows. Grain coded: the grain version encoded directly, with the encoder's temporal filter on (solid) or off (dashed). Dotted: the grain version after 4:2:0 conversion alone. Values in Tabs.~\ref{tab:supp-bandwidth-libaom} and~\ref{tab:supp-bandwidth-vvc}.}
\label{fig:supp-bandwidth-rd}
\end{figure*}

\begin{table*}[tp]
\sidecaptioned{\scriptsize
\setlength{\tabcolsep}{4.5pt}
\begin{tabular}{@{}llrrccccccr@{}}
\toprule
& & \multicolumn{2}{c}{kbps} & \multicolumn{2}{c}{Appearance$\downarrow$} & \multicolumn{3}{c}{Grain fidelity$\downarrow$} & & \\
\cmidrule(lr){3-4}\cmidrule(lr){5-6}\cmidrule(lr){7-9}
Arm & Q & clip & film & DISTS & LPIPS & $W_1$ & Spatial & Color & Strength (CV) & PSNR \\
\midrule
No codec (4:2:0 only) & -- & -- & -- & 0.028 & 0.015 & 0.69 & 0.026 & 0.100 & 0.86 (0.005) & 39.45 \\
\addlinespace[2pt]
Grain coded & 15 & 72{,}098 & 72{,}098 & 0.093 & 0.111 & 1.33 & 0.056 & 0.208 & 0.77 (0.036) & 34.51 \\
 & 23 & 27{,}903 & 27{,}903 & 0.160 & 0.246 & 1.90 & 0.105 & 0.281 & 0.72 (0.053) & 32.87 \\
 & 31 & 7{,}462 & 7{,}462 & 0.213 & 0.325 & 2.37 & 0.170 & 0.328 & 0.68 (0.077) & 31.88 \\
 & 39 & 894 & 894 & 0.275 & 0.353 & 2.68 & 0.246 & 0.334 & 0.65 (0.113) & 31.47 \\
 & 47 & 247 & 247 & 0.301 & 0.352 & 2.67 & 0.287 & 0.345 & 0.66 (0.135) & 31.33 \\
 & 55 & 148 & 148 & 0.308 & 0.354 & 2.61 & 0.336 & 0.377 & 0.70 (0.144) & 31.15 \\
 & 63 & 69 & 69 & 0.314 & 0.363 & 2.36 & 0.488 & 0.436 & 0.85 (0.152) & 30.39 \\
 & lossless & 642{,}047 & 642{,}047 & 0.028 & 0.015 & 0.69 & 0.026 & 0.100 & 0.86 (0.005) & 39.45 \\
\addlinespace[2pt]
Grain coded, no temporal filter & 15 & 73{,}038 & 73{,}038 & 0.089 & 0.102 & 1.30 & 0.056 & 0.209 & 0.78 (0.026) & 34.56 \\
 & 23 & 30{,}735 & 30{,}735 & 0.143 & 0.212 & 1.78 & 0.097 & 0.282 & 0.73 (0.041) & 33.01 \\
 & 31 & 8{,}328 & 8{,}328 & 0.199 & 0.312 & 2.31 & 0.168 & 0.315 & 0.69 (0.070) & 31.93 \\
 & 39 & 1{,}090 & 1{,}090 & 0.264 & 0.355 & 2.61 & 0.245 & 0.339 & 0.66 (0.104) & 31.44 \\
 & 47 & 240 & 240 & 0.301 & 0.354 & 2.62 & 0.307 & 0.348 & 0.68 (0.115) & 31.24 \\
 & 55 & 134 & 134 & 0.307 & 0.360 & 2.47 & 0.380 & 0.387 & 0.74 (0.117) & 30.97 \\
 & 63 & 63 & 63 & 0.330 & 0.369 & 2.94 & 0.542 & 0.454 & 1.01 (0.098) & 29.73 \\
\addlinespace[2pt]
\textbf{PGNM}, clean master & 15 & 7{,}118 & 6{,}346 & 0.046 & 0.074 & 0.57 & 0.025 & 0.023 & 1.10 (0.017) & 29.17 \\
 & 23 & 3{,}041 & 2{,}269 & 0.051 & 0.077 & 0.66 & 0.031 & 0.022 & 1.12 (0.024) & 29.07 \\
 & 31 & 1{,}628 & 856 & 0.055 & 0.080 & 0.73 & 0.037 & 0.024 & 1.14 (0.032) & 29.00 \\
 & 39 & 1{,}188 & 416 & 0.059 & 0.082 & 0.77 & 0.043 & 0.027 & 1.15 (0.039) & 28.94 \\
 & 47 & 1{,}023 & 251 & 0.063 & 0.084 & 0.83 & 0.051 & 0.033 & 1.16 (0.045) & 28.88 \\
 & 55 & 935 & 163 & 0.069 & 0.086 & 0.94 & 0.065 & 0.045 & 1.18 (0.053) & 28.79 \\
 & 63 & 849 & 76 & 0.087 & 0.093 & 1.54 & 0.123 & 0.097 & 1.26 (0.070) & 28.42 \\
 & lossless & 327{,}771 & 326{,}999 & 0.039 & 0.068 & 0.23 & 0.015 & 0.055 & 1.02 (0.005) & 29.47 \\
\addlinespace[2pt]
AV1 grain, clean master & 15 & 6{,}412 & 6{,}412 & 0.087 & 0.109 & 1.75 & 0.050 & 0.163 & 0.80 (0.049) & 30.19 \\
 & 23 & 2{,}332 & 2{,}332 & 0.093 & 0.115 & 1.77 & 0.057 & 0.170 & 0.83 (0.057) & 30.06 \\
 & 31 & 919 & 919 & 0.098 & 0.118 & 1.79 & 0.064 & 0.176 & 0.85 (0.066) & 29.96 \\
 & 39 & 478 & 478 & 0.104 & 0.120 & 1.81 & 0.073 & 0.184 & 0.87 (0.074) & 29.88 \\
 & 47 & 313 & 313 & 0.106 & 0.121 & 1.83 & 0.084 & 0.194 & 0.88 (0.081) & 29.81 \\
 & 55 & 226 & 226 & 0.109 & 0.125 & 1.88 & 0.104 & 0.210 & 0.91 (0.089) & 29.69 \\
 & 63 & 141 & 141 & 0.127 & 0.129 & 2.18 & 0.184 & 0.270 & 1.01 (0.103) & 29.23 \\
 & lossless & 327{,}065 & 327{,}065 & 0.066 & 0.091 & 1.81 & 0.045 & 0.144 & 0.69 (0.042) & 30.57 \\
\addlinespace[2pt]
\textbf{PGNM}, nlmeans & 15 & 10{,}494 & 9{,}726 & 0.077 & 0.089 & 0.79 & 0.041 & 0.176 & 0.99 (0.032) & 29.94 \\
 & 23 & 5{,}502 & 4{,}734 & 0.084 & 0.091 & 0.82 & 0.042 & 0.180 & 1.00 (0.035) & 29.82 \\
 & 31 & 2{,}584 & 1{,}815 & 0.090 & 0.092 & 0.86 & 0.045 & 0.183 & 1.01 (0.039) & 29.72 \\
 & 39 & 1{,}317 & 549 & 0.101 & 0.093 & 0.89 & 0.049 & 0.185 & 1.01 (0.044) & 29.63 \\
 & 47 & 1{,}022 & 254 & 0.107 & 0.094 & 0.90 & 0.056 & 0.191 & 1.02 (0.051) & 29.56 \\
 & 55 & 924 & 156 & 0.110 & 0.096 & 0.98 & 0.071 & 0.203 & 1.05 (0.058) & 29.46 \\
 & 63 & 841 & 73 & 0.121 & 0.102 & 1.46 & 0.149 & 0.254 & 1.14 (0.076) & 29.00 \\
 & lossless & 245{,}548 & 244{,}780 & 0.064 & 0.085 & 0.72 & 0.039 & 0.164 & 0.99 (0.027) & 30.28 \\
\addlinespace[2pt]
AV1 grain, nlmeans & 15 & 9{,}789 & 9{,}789 & 0.098 & 0.107 & 1.16 & 0.049 & 0.197 & 0.92 (0.041) & 30.32 \\
 & 23 & 4{,}795 & 4{,}795 & 0.103 & 0.109 & 1.19 & 0.051 & 0.202 & 0.93 (0.043) & 30.20 \\
 & 31 & 1{,}876 & 1{,}876 & 0.110 & 0.109 & 1.22 & 0.055 & 0.206 & 0.93 (0.047) & 30.08 \\
 & 39 & 610 & 610 & 0.117 & 0.109 & 1.25 & 0.060 & 0.208 & 0.94 (0.053) & 29.99 \\
 & 47 & 314 & 314 & 0.124 & 0.109 & 1.26 & 0.068 & 0.215 & 0.95 (0.060) & 29.92 \\
 & 55 & 216 & 216 & 0.128 & 0.111 & 1.31 & 0.088 & 0.228 & 0.98 (0.067) & 29.80 \\
 & 63 & 135 & 135 & 0.142 & 0.117 & 1.67 & 0.178 & 0.284 & 1.07 (0.084) & 29.31 \\
 & lossless & 244{,}844 & 244{,}844 & 0.082 & 0.101 & 1.09 & 0.047 & 0.183 & 0.91 (0.038) & 30.70 \\
\bottomrule
\end{tabular}}
{\caption{\textbf{Video delivery with libaom AV1, every arm and quantizer} (\emph{Nocturne}, 100 frames, $1920\times1080$, 60\,fps). kbps: clip, PGNM's 161\,KB model charged to this 1.67\,s clip (the main paper's rule); film, the same model amortized over the 664.5\,s film. Appearance against the grain version (25 frames); grain fidelity of the residual from the clean master (all 100 frames; $W_1$ in 8-bit codes). Strength: RMS of the delivered residual over the grain version's, mean over frames (target 1), with its coefficient of variation across frames. PSNR (dB) against the grain version is reported but penalizes an independent grain realization by design. No temporal filter: alt-ref and keyframe filtering off. nlmeans: the grain version denoised with the 3GPP setting.}
\label{tab:supp-bandwidth-libaom}}
\end{table*}

\begin{table*}[tp]
\sidecaptioned{\scriptsize
\setlength{\tabcolsep}{4.5pt}
\begin{tabular}{@{}llrccccccr@{}}
\toprule
& & & \multicolumn{2}{c}{Appearance$\downarrow$} & \multicolumn{3}{c}{Grain fidelity$\downarrow$} & & \\
\cmidrule(lr){4-5}\cmidrule(lr){6-8}
Arm & QP & kbps & DISTS & LPIPS & $W_1$ & Spatial & Color & Strength (CV) & PSNR \\
\midrule
Grain coded & 17 & 94{,}536 & 0.091 & 0.085 & 1.21 & 0.046 & 0.206 & 0.78 (0.025) & 34.53 \\
 & 22 & 31{,}305 & 0.167 & 0.250 & 1.79 & 0.084 & 0.316 & 0.72 (0.051) & 32.64 \\
 & 27 & 4{,}791 & 0.249 & 0.340 & 2.45 & 0.192 & 0.329 & 0.66 (0.095) & 31.68 \\
 & 32 & 448 & 0.308 & 0.343 & 2.73 & 0.264 & 0.323 & 0.65 (0.140) & 31.38 \\
 & 37 & 143 & 0.311 & 0.345 & 2.68 & 0.309 & 0.348 & 0.68 (0.159) & 31.23 \\
 & 42 & 78 & 0.314 & 0.349 & 2.48 & 0.396 & 0.409 & 0.74 (0.173) & 30.92 \\
 & 47 & 54 & 0.319 & 0.357 & 2.39 & 0.478 & 0.473 & 0.84 (0.191) & 30.44 \\
\addlinespace[2pt]
Grain coded, no MCTF & 17 & 112{,}464 & 0.092 & 0.085 & 1.12 & 0.043 & 0.190 & 0.80 (0.031) & 35.01 \\
 & 22 & 41{,}521 & 0.167 & 0.252 & 1.67 & 0.079 & 0.301 & 0.73 (0.058) & 32.96 \\
 & 27 & 8{,}325 & 0.247 & 0.342 & 2.32 & 0.189 & 0.335 & 0.68 (0.092) & 31.78 \\
 & 32 & 1{,}015 & 0.300 & 0.343 & 2.64 & 0.276 & 0.325 & 0.66 (0.130) & 31.37 \\
 & 37 & 216 & 0.308 & 0.346 & 2.60 & 0.321 & 0.346 & 0.68 (0.151) & 31.20 \\
 & 42 & 79 & 0.312 & 0.348 & 2.43 & 0.409 & 0.401 & 0.75 (0.169) & 30.90 \\
 & 47 & 54 & 0.316 & 0.356 & 2.29 & 0.496 & 0.464 & 0.85 (0.185) & 30.41 \\
\addlinespace[2pt]
VVC grain & 17 & 94{,}598 & 0.097 & 0.135 & 1.01 & 0.042 & 0.124 & 1.13 (0.018) & 31.17 \\
 & 22 & 31{,}366 & 0.133 & 0.195 & 1.14 & 0.051 & 0.144 & 1.16 (0.014) & 29.78 \\
 & 27 & 4{,}853 & 0.188 & 0.244 & 1.06 & 0.065 & 0.154 & 1.15 (0.025) & 29.17 \\
 & 32 & 510 & 0.201 & 0.256 & 1.15 & 0.074 & 0.158 & 1.15 (0.034) & 28.93 \\
 & 37 & 205 & 0.207 & 0.257 & 1.19 & 0.081 & 0.165 & 1.17 (0.049) & 28.84 \\
 & 42 & 140 & 0.214 & 0.259 & 1.38 & 0.102 & 0.183 & 1.21 (0.063) & 28.65 \\
 & 47 & 116 & 0.226 & 0.266 & 1.77 & 0.139 & 0.210 & 1.29 (0.082) & 28.31 \\
\addlinespace[2pt]
VVC grain, two-pass & 935\,k target & 994 & 0.201 & 0.250 & 1.09 & 0.076 & 0.160 & 1.13 (0.040) & 29.05 \\
\bottomrule
\end{tabular}}
{\caption{\textbf{Video delivery with VVC} (VVenC~1.14 medium preset, VVdeC~3.2; constant QP except the two-pass row), measured as in Tab.~\ref{tab:supp-bandwidth-libaom}. No MCTF: VVenC's motion-compensated temporal filter off. VVC grain: VVenC's film-grain analysis of the grain version, synthesized by VVdeC; the two-pass row targets 935\,kbps, the total rate of PGNM at Q\,55.}
\label{tab:supp-bandwidth-vvc}}
\end{table*}

\begin{table*}[tp]
\sidecaptioned{\scriptsize
\setlength{\tabcolsep}{6pt}
\begin{tabular}{@{}llrcccccc@{}}
\toprule
& & & \multicolumn{4}{c}{Saving against each direct encode (\%, clip)} & \multicolumn{2}{c}{Smallest (\%)} \\
\cmidrule(lr){4-7}\cmidrule(l){8-9}
Grain synthesis & Q & kbps & AV1 & AV1\textsuperscript{nf} & VVC & VVC\textsuperscript{nf} & clip & film \\
\midrule
\textbf{PGNM}, master & 15 & 7{,}118 & $\ge$90.1 & $\ge$90.3 & $\ge$92.5 & $\ge$93.7 & $\ge$90.1 & $\ge$91.2 \\
 & 23 & 3{,}041 & $\ge$95.8 & $\ge$95.8 & $\ge$96.8 & $\ge$97.3 & $\ge$95.8 & $\ge$96.9 \\
 & 31 & 1{,}628 & $\ge$97.7 & $\ge$97.8 & $\ge$98.3 & $\ge$98.6 & $\ge$97.7 & $\ge$98.8 \\
 & 39 & 1{,}188 & $\ge$98.4 & $\ge$98.4 & $\ge$98.7 & $\ge$98.9 & $\ge$98.4 & $\ge$99.4 \\
 & 47 & 1{,}023 & $\ge$98.6 & $\ge$98.6 & $\ge$98.9 & $\ge$99.1 & $\ge$98.6 & $\ge$99.7 \\
 & 55 & 935 & $\ge$98.7 & $\ge$98.7 & $\ge$99.0 & $\ge$99.2 & $\ge$98.7 & $\ge$99.8 \\
 & 63 & 849 & $\ge$98.8 & $\ge$98.8 & $\ge$99.1 & $\ge$99.2 & $\ge$98.8 & $\ge$99.9 \\
\addlinespace[2pt]
AV1 grain, master & 15 & 6{,}412 & $\ge$91.1 & $\ge$91.2 & $\ge$93.2 & $\ge$94.3 & $\ge$91.1 & $\ge$91.1 \\
 & 23 & 2{,}332 & $\ge$96.8 & 96.6 & 97.5 & 97.9 & 96.6 & 96.6 \\
 & 31 & 919 & 98.6 & 98.5 & 98.9 & 99.1 & 98.5 & 98.5 \\
 & 39 & 478 & 99.2 & 99.2 & 99.4 & 99.5 & 99.2 & 99.2 \\
 & 47 & 313 & 99.5 & 99.4 & 99.6 & 99.7 & 99.4 & 99.4 \\
 & 55 & 226 & 99.6 & 99.6 & 99.7 & 99.7 & 99.6 & 99.6 \\
 & 63 & 141 & 99.7 & 99.6 & 99.7 & 99.8 & 99.6 & 99.6 \\
\addlinespace[2pt]
\textbf{PGNM}, nlmeans & 15 & 10{,}494 & $\ge$85.4 & $\ge$85.6 & $\ge$88.9 & $\ge$90.7 & $\ge$85.4 & $\ge$86.5 \\
 & 23 & 5{,}502 & $\ge$92.4 & $\ge$92.5 & $\ge$94.2 & $\ge$95.1 & $\ge$92.4 & $\ge$93.4 \\
 & 31 & 2{,}584 & $\ge$96.4 & 96.4 & $\ge$97.3 & $\ge$97.7 & 96.4 & 97.5 \\
 & 39 & 1{,}317 & 98.0 & 97.8 & 98.4 & 98.7 & 97.8 & 99.1 \\
 & 47 & 1{,}022 & 98.3 & 98.1 & 98.6 & 98.9 & 98.1 & 99.5 \\
 & 55 & 924 & 98.4 & 98.2 & 98.7 & 99.0 & 98.2 & 99.7 \\
 & 63 & 841 & 98.3 & 98.1 & 98.6 & 98.9 & 98.1 & 99.8 \\
\addlinespace[2pt]
AV1 grain, nlmeans & 15 & 9{,}789 & 85.5 & 84.6 & 88.6 & 90.7 & 84.6 & 84.6 \\
 & 23 & 4{,}795 & 92.4 & 91.8 & 94.0 & 95.1 & 91.8 & 91.8 \\
 & 31 & 1{,}876 & 96.7 & 96.4 & 97.4 & 97.9 & 96.4 & 96.4 \\
 & 39 & 610 & 98.8 & 98.7 & 99.1 & 99.3 & 98.7 & 98.7 \\
 & 47 & 314 & 99.3 & 99.2 & 99.5 & 99.6 & 99.2 & 99.2 \\
 & 55 & 216 & 99.5 & 99.4 & 99.6 & 99.7 & 99.4 & 99.4 \\
 & 63 & 135 & 99.6 & 99.6 & 99.7 & 99.8 & 99.6 & 99.6 \\
\addlinespace[2pt]
VVC grain & 17 & 94{,}598 & $-$39.1 & $-$48.4 & $-$9.3 & 10.1 & $-$48.4 & $-$48.4 \\
 & 22 & 31{,}366 & 22.9 & 12.2 & 38.8 & 51.9 & 12.2 & 12.2 \\
 & 27 & 4{,}853 & 65.5 & 55.2 & 75.1 & 82.4 & 55.2 & 55.2 \\
 & 32 & 510 & 95.0 & 93.5 & 96.5 & 97.6 & 93.5 & 93.5 \\
 & 37 & 205 & 97.7 & 96.9 & 98.4 & 98.9 & 96.9 & 96.9 \\
 & 42 & 140 & 98.1 & 97.3 & 98.7 & 99.1 & 97.3 & 97.3 \\
 & 47 & 116 & 97.6 & 96.7 & 98.6 & 99.1 & 96.7 & 96.7 \\
\addlinespace[2pt]
VVC grain, two-pass & 935\,k & 994 & 90.3 & 87.3 & 93.1 & 95.3 & 87.3 & 87.3 \\
\bottomrule
\end{tabular}}
{\caption{\textbf{DISTS-matched bitrate savings}, the headline measure. For each grain-synthesis rate, the direct encode's bitrate reaching the same DISTS is interpolated linearly in log-rate between adjacent ladder points (first crossing from low rate); saving $=1-$ rate$/$matched direct rate. $\ge$: the direct encode never reaches the arm's DISTS, so its highest rate gives a lower bound. AV1/VVC: the grain-coded encodes; \textsuperscript{nf}: temporal filter off. Smallest: over the four direct encodes, with PGNM's model charged per clip or amortized over the film; the main paper cites the per-clip range.}
\label{tab:supp-bandwidth-savings}}
\end{table*}

\begin{table*}[tp]
\sidecaptioned{\footnotesize
\setlength{\tabcolsep}{6pt}
\begin{tabular}{@{}lccccc@{}}
\toprule
Grain synthesis & DISTS & LPIPS & $W_1$ & Spatial & Color \\
\midrule
\textbf{PGNM}, master & $\ge$90.1 to $\ge$98.8 & $\ge$90.1 to $\ge$98.8 & $\ge$90.1 to $\ge$98.7 & $\ge$90.1 to $\ge$98.6 & $\ge$90.1 to $\ge$98.8 \\
AV1 grain, master & $\ge$91.1 to 99.6 & 90.7 to 99.8 & 80.4 to 99.1 & $\ge$91.1 to 99.0 & $\ge$91.1 to 99.7 \\
\textbf{PGNM}, nlmeans & $\ge$85.4 to 98.2 & $\ge$85.4 to $\ge$98.8 & $\ge$85.4 to $\ge$98.7 & $\ge$85.4 to 98.6 & $\ge$85.4 to $\ge$98.7 \\
AV1 grain, nlmeans & 84.6 to 99.6 & 86.0 to 99.8 & $\ge$86.4 to 99.7 & $\ge$86.4 to 99.4 & $\ge$86.4 to 99.6 \\
VVC grain & $-$48.4 to 97.3 & $-$68.3 to 99.2 & $\ge$$-$31.2 to 99.8 & $\ge$$-$31.2 to 99.4 & $\ge$$-$31.2 to 99.8 \\
VVC grain, two-pass & 87.3 & 94.7 & $\ge$98.6 & 97.5 & $\ge$98.6 \\
\bottomrule
\end{tabular}}
{\caption{\textbf{Matched savings for every measurement} (secondary): the smallest per-clip saving over the four direct encodes, as a range over each arm's rates (\%). Matching as in Tab.~\ref{tab:supp-bandwidth-savings}, with each measurement in place of DISTS.}
\label{tab:supp-bandwidth-matched}}
\end{table*}

\begin{figure*}[tp]
\centering
\setlength{\tabcolsep}{0.8pt}
\begin{tabular}{ccccc}
\begin{tabular}[t]{@{}c@{}}\includegraphics[width=0.192\textwidth]{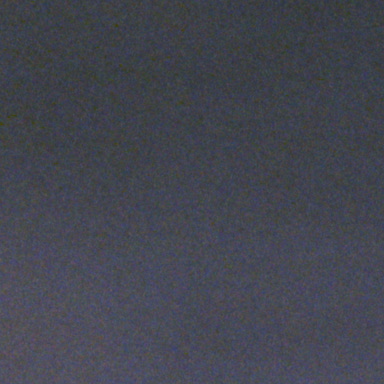}\\{\scriptsize Grain version}\end{tabular} &
\begin{tabular}[t]{@{}c@{}}\includegraphics[width=0.192\textwidth]{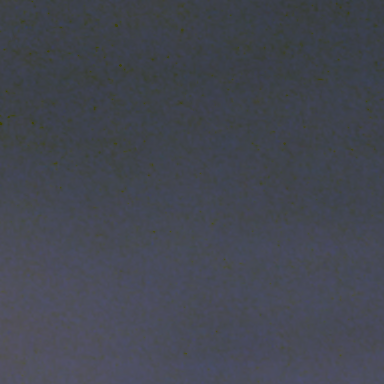}\\{\scriptsize Clean master}\end{tabular} &
\begin{tabular}[t]{@{}c@{}}\includegraphics[width=0.192\textwidth]{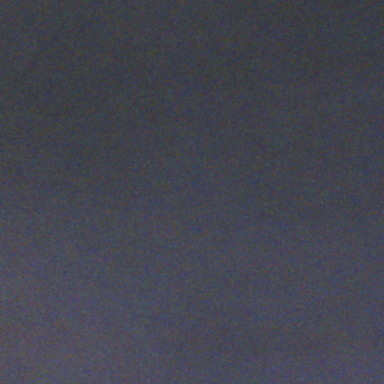}\\{\scriptsize \textbf{PGNM}, master, Q\,31}\\{\tiny 1{,}628\,kbps}\end{tabular} &
\begin{tabular}[t]{@{}c@{}}\includegraphics[width=0.192\textwidth]{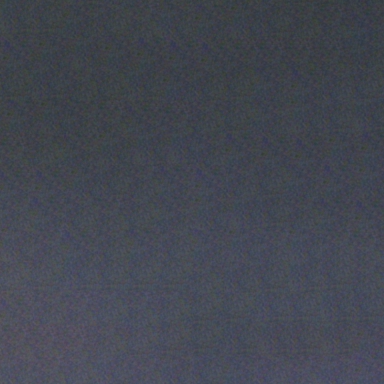}\\{\scriptsize AV1 grain, master, Q\,31}\\{\tiny 919\,kbps}\end{tabular} &
\begin{tabular}[t]{@{}c@{}}\includegraphics[width=0.192\textwidth]{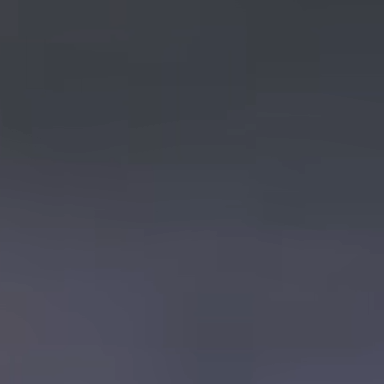}\\{\scriptsize AV1, Q\,39}\\{\tiny 894\,kbps}\end{tabular} \\[3pt]
\begin{tabular}[t]{@{}c@{}}\includegraphics[width=0.192\textwidth]{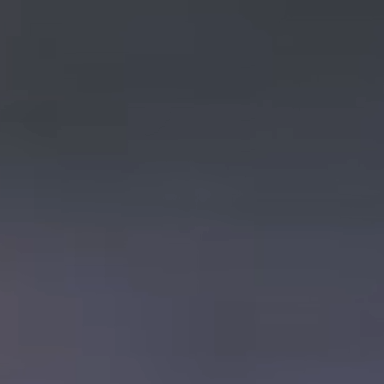}\\{\scriptsize AV1 no filter, Q\,39}\\{\tiny 1{,}090\,kbps}\end{tabular} &
\begin{tabular}[t]{@{}c@{}}\includegraphics[width=0.192\textwidth]{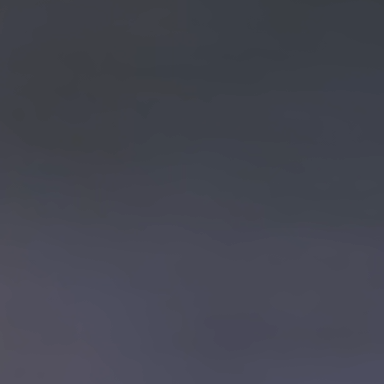}\\{\scriptsize VVC, QP\,32}\\{\tiny 448\,kbps}\end{tabular} &
\begin{tabular}[t]{@{}c@{}}\includegraphics[width=0.192\textwidth]{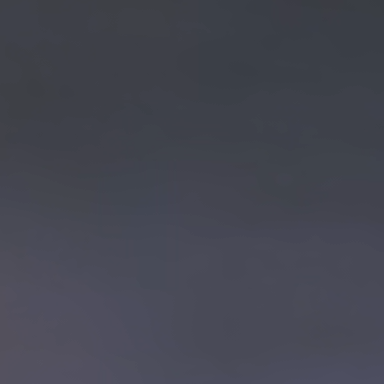}\\{\scriptsize VVC no MCTF, QP\,32}\\{\tiny 1{,}015\,kbps}\end{tabular} &
\begin{tabular}[t]{@{}c@{}}\includegraphics[width=0.192\textwidth]{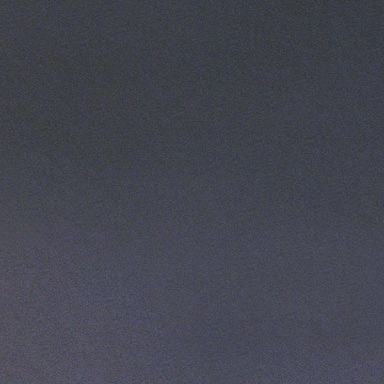}\\{\scriptsize VVC grain, QP\,32}\\{\tiny 510\,kbps}\end{tabular} &
\begin{tabular}[t]{@{}c@{}}\includegraphics[width=0.192\textwidth]{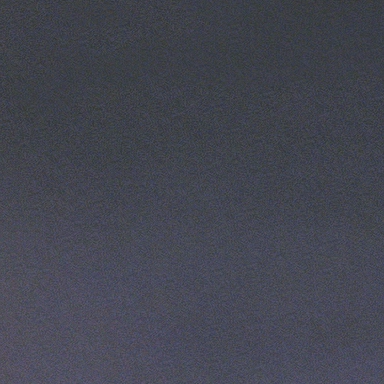}\\{\scriptsize VVC grain, 935\,k}\\{\tiny 994\,kbps}\end{tabular} \\[3pt]
\end{tabular}
\caption{\textbf{Delivered frames.} Center $384\times384$ crop of clip frame 50 (top-left corner at row 348, column 768 of the $1920\times1080$ crop), chosen by rule before viewing: PGNM and AV1 grain at Q\,31, each direct encode at the ladder rate closest to PGNM's total rate, and VVC grain at QP\,32 and its two-pass run. kbps: total rate, PGNM's model charged per clip. Please zoom in; grain is at native 4K size. \emph{Nocturne} \textcopyright{} Netflix, Inc., CC BY 4.0.}
\label{fig:supp-bandwidth}
\end{figure*}

This section completes Tab.~\mainref{tab:bandwidth} of the main paper with every
arm and rate and the setup. It illustrates delivery on one clip and is not a codec
benchmark. The setup was fixed before any shot was selected or encoded, and every
arm and rate is reported.

\paragraph{Source and clip.}
Netflix's \emph{Nocturne}~\cite{netflixnocturne} (2018, CC BY 4.0) provides a clean
master (JPEG~2000, 12-bit RGB, PQ, P3-D65, $3840\times2160$, 60\,fps) and a grain
version with Netflix's synthetic Filmbox grain (Vision3 stock, strength 1.2) on the
same master (ProRes 422, 10-bit). The pair is the ground truth: the residual
(grain version minus clean master) also contains the grain file's ProRes and 4:2:2 coding
error, which no arm can remove; the grain version is also slightly darker than the
master (by 0.5--1.8 codes in 8 bits, already present in PQ). Among the eight \emph{Nocturne} shots of the AOM
test material, we took the one whose first frame has median BT.709 luma closest
to 128 (master frames 2370--2539, median 82; the other shots are darker) and its
first 100 frames (1.67\,s). The clean frames are read from the AOM test archive of
the shot, whose first frame is identical to master frame 2370, and every grain frame
was checked to match the clean frame of the same index. Both files are cropped to the center
$1920\times1080$, keeping the grain at its native 4K size, and converted to 8-bit
SDR by one fixed pipeline: the ST\,2084 transfer function, P3-D65 to BT.709, division
by the 203\,cd/m$^2$ reference white, a Hable filmic curve applied to $\max(R,G,B)$
with the RGB ratio kept, clipping, and the inverse BT.1886 curve. The codec domain is
BT.709 limited-range 8-bit Y$'$CbCr 4:2:0, with the same converter for every arm.

\paragraph{Arms.}
\emph{Grain coded} encodes the grain version, with and without libaom's temporal
filtering (alt-ref and keyframe filtering), which otherwise removes grain before
coding. The grain-synthesis arms encode
either the clean master (a perfect denoiser, so any saving comes from modeling the
grain) or the grain version denoised with ffmpeg \texttt{nlmeans=3.5:5:3:9:7}, the
3GPP setting for these film-grain sequences, and add grain after decoding. PGNM is
fitted by the unchanged \textsc{Fit} (Alg.~\ref{alg:supp-fit}) on clip frames 0,
10, \dots, 90, on the 8-bit RGB pairs (clean or denoised, grain version), and
renders frame $t$ with the unchanged renderer, bank seeds 6100--6103 and window
seed $9000+t$, identical across rates and arms. AV1's grain table comes from
libaom's \texttt{noise\_model} with its defaults (autoregression lag 3, $32\times32$
flat-block finder) on the same frames, and aomdec applies it. VVC's grain arm uses
VVenC's film-grain analysis, which estimates grain from its own temporally filtered
input and accepts no clean or denoised input, so it is not matched to the
master or nlmeans inputs; VVdeC synthesizes the grain.

\paragraph{Encoders and rates.}
libaom v3.6.0 runs libaom's own film-grain example settings (\texttt{--good
--cpu-used=4 --end-usage=q --lag-in-frames=25 --auto-alt-ref=2}, two threads,
8-bit 4:2:0) at the quantizers $Q\in\{15,23,31,39,47,55,63\}$ of Netflix's AOM
\emph{Nocturne} encodes, plus a lossless row. VVenC v1.14.0 uses its medium preset
with two threads at QP $\in\{17,22,\dots,47\}$, and VVdeC v3.2.0 decodes. The
bitrate is the bitstream payload over 1.67\,s. PGNM's model is charged as the
zlib-compressed float64 arrays the renderer uses: 161,269 bytes, or 774\,kbps on
this clip (the main paper's rule; 160,469 bytes and 770\,kbps for the nlmeans fit) and 1.9\,kbps amortized over the 664.5\,s film,
since one grain setting covers the whole film. AV1's grain parameters are inside
its bitstream; the rate difference between the AV1-grain and PGNM encodes of the
clean master, about 65\,kbps at Q\,31, makes PGNM's model the cheaper side
information beyond about 20\,s of video.

\paragraph{Measurements.}
Grain fidelity uses the SIDD metric code of Sec.~\ref{sec:supp-metrics} unchanged,
with the clean master as clean image and the grain version as real image, on all
100 frames. DISTS and LPIPS use GrainNet's definitions on the full crop of frames
0, 4, \dots, 96, against the grain version. The main arms were scored on CPU; the
supplementary groups were scored on GPU after each group's first frame matched the
CPU values within $10^{-4}$ for every output.

\paragraph{Main-table rows and later arms.}
The main paper's rows were fixed by rate before the final scores: each grain-coded
encoder at its highest lossy rate with the closer temporal-filter setting, AV1 grain
and PGNM at Q\,31, and PGNM at the Q whose total rate is closest to AV1 grain at
Q\,31 (Q\,55). Three arms were added after results existed and are reported like the
others: the grain-coded arm without libaom's temporal filter (after part of the main
scores), the VVC arms (after a preview), and VVC grain with two-pass rate control at
935\,kbps, the rate of PGNM at Q\,55, because VVC grain had no constant-QP point near
it (after the final reports; its row then joined the main table).

\paragraph{Results.}
Fig.~\ref{fig:supp-bandwidth-rd} plots the rate--distortion curves of every arm,
with PGNM's model charged per clip and amortized over the film;
Tabs.~\ref{tab:supp-bandwidth-libaom} and~\ref{tab:supp-bandwidth-vvc} list every
arm and rate, Tabs.~\ref{tab:supp-bandwidth-savings}
and~\ref{tab:supp-bandwidth-matched} the matched savings, and
Fig.~\ref{fig:supp-bandwidth} the delivered frames. Charged per clip, the model
adds about 770\,kbps to every PGNM point, so PGNM's curves start near 850\,kbps; amortized,
it adds 1.9\,kbps and PGNM on the clean master reaches DISTS 0.087 at 76\,kbps,
below the DISTS of every lossy grain-coded encode (at least 0.089, at 73\,Mbps).
Coding the grain costs 9--12 times the bits of the clean master at Q\,15--31 (up to
14 times without libaom's temporal filter); from Q\,47 on, the encoder drops
the grain and the two rates meet. Against even the closest of the four grain-coded
baselines, PGNM saves 90.1--98.8\% of the bitrate at matched DISTS when
synthesizing on the clean master and 85.4--98.2\% on nlmeans-denoised video. AV1
grain synthesis reaches similar savings, but each is matched at its own DISTS,
which is higher than PGNM's at every quantizer (0.087--0.127 against 0.046--0.087
on the clean master); VVC's own grain pipeline stays near DISTS 0.20 at both 0.51
and 0.99\,Mbps. On the clean master at the lossy rates, PGNM's delivered grain is 10--26\%
stronger in RMS than the grain version's, and AV1's ranges from 20\% weaker to 1\%
stronger; the 4:2:0 round trip
of the grain version alone keeps 86\% of its RMS. PGNM's signed lookup can represent
the grain version's darker offset and AV1's zero-mean grain cannot: delivered mean
offsets at Q\,31 are $(0.16,-0.01,0.17)$ codes for PGNM and $(0.74,1.89,1.71)$ for AV1
grain on the clean master. Strength, PSNR and offsets are reported for information
only.

\FloatBarrier
\section{Visual comparisons on SIDD}
\label{sec:supp-sidd-visual}

\begin{table*}[tp]
\sidecaptioned{\small
\setlength{\tabcolsep}{6pt}
\begin{tabular}{@{}llrrr@{}}
\toprule
Class & SIDD source (Validation index) & Block & $x$ & $y$ \\
\midrule
G4 ISO\,400 & \texttt{0148\_007\_G4\_00400\_00400\_4400\_L} (31) & 20 & 34 & 65 \\
G4 ISO\,800 & \texttt{0100\_005\_G4\_00800\_00400\_3200\_N} (21) & 1 & 2 & 9 \\
GP ISO\,1600 & \texttt{0067\_003\_GP\_01600\_01500\_3200\_L} (13) & 26 & 117 & 83 \\
GP ISO\,3200 & \texttt{0158\_007\_GP\_03200\_03200\_5500\_N} (33) & 16 & 125 & 78 \\
GP ISO\,6400 & \texttt{0183\_008\_GP\_06400\_06400\_5500\_N} (38) & 20 & 72 & 70 \\
GP ISO\,10000 & \texttt{0109\_005\_GP\_10000\_08000\_4400\_N} (23) & 8 & 86 & 105 \\
IP ISO\,1000 & \texttt{0112\_005\_IP\_01000\_07500\_5500\_L} (24) & 17 & 98 & 4 \\
IP ISO\,1600 & \texttt{0162\_007\_IP\_01600\_01600\_3200\_L} (34) & 27 & 11 & 22 \\
IP ISO\,2000 & \texttt{0071\_003\_IP\_02000\_04000\_5500\_L} (14) & 27 & 69 & 2 \\
N6 ISO\,400 & \texttt{0074\_004\_N6\_00400\_00040\_3200\_L} (15) & 9 & 54 & 62 \\
N6 ISO\,800 & \texttt{0049\_002\_N6\_00800\_00800\_5500\_L} (8) & 0 & 16 & 0 \\
N6 ISO\,3200 & \texttt{0024\_001\_N6\_03200\_01500\_5500\_N} (2) & 21 & 83 & 67 \\
S6 ISO\,400$^\star$ & \texttt{0176\_008\_S6\_00400\_00100\_5500\_L} (37) & 8 & 98 & 79 \\
S6 ISO\,800 & \texttt{0009\_001\_S6\_00800\_00350\_3200\_L} (0) & 14 & 103 & 128 \\
S6 ISO\,1600 & \texttt{0103\_005\_S6\_01600\_00800\_4400\_L} (22) & 31 & 88 & 48 \\
S6 ISO\,3200 & \texttt{0153\_007\_S6\_03200\_03200\_5500\_L} (32) & 20 & 88 & 108 \\
\midrule
Radius & Test image (size $h\times w$) & & $x$ & $y$ \\
\midrule
$r=0.025$ & 59 ($499\times333$) & & 173 & 284 \\
$r=0.05$ & 26 ($333\times500$) & & 220 & 45 \\
$r=0.075$ & 45 ($334\times500$) & & 98 & 164 \\
$r=0.1$ & 25 ($334\times500$) & & 311 & 72 \\
$r=0.2$ & 32 ($336\times500$) & & 189 & 121 \\
$r=0.4$$^\star$ & 28 ($331\times500$) & & 190 & 136 \\
$r=0.6$ & 17 ($375\times500$) & & 279 & 240 \\
$r=0.8$$^\star$ & 29 ($333\times499$) & & 57 & 13 \\
\bottomrule
\end{tabular}}
{\caption{\textbf{Crop windows of the uncurated comparisons} (Figs.~\ref{fig:supp-sidd-1}--\ref{fig:supp-sidd-4} and~\ref{fig:supp-film-1}--\ref{fig:supp-film-2}): top-left corner $(x,y)$ of each $128\times128$ window; SIDD blocks are $256\times256$, block indices from 0. $^\star$Also shown in the main paper.}
\label{tab:supp-crops}}
\end{table*}

\begin{table*}[tp]
\sidecaptioned{\small
\setlength{\tabcolsep}{6pt}
\begin{tabular}{@{}llrrrr@{}}
\toprule
Class & SIDD source (Validation index) & Block & $x$ & $y$ & Tex. \\
\midrule
G4 ISO\,400 & \texttt{0058\_003\_G4\_00400\_00500\_5500\_L} (11) & 22 & 128 & 128 & 6.6 \\
G4 ISO\,800 & \texttt{0174\_008\_G4\_00800\_00800\_4400\_N} (36) & 5 & 16 & 128 & 14.3 \\
GP ISO\,1600 & \texttt{0131\_006\_GP\_01600\_01250\_4400\_N} (28) & 14 & 24 & 80 & 7.5 \\
GP ISO\,3200 & \texttt{0158\_007\_GP\_03200\_03200\_5500\_N} (33) & 13 & 8 & 96 & 9.8 \\
GP ISO\,6400$^\star$ & \texttt{0085\_004\_GP\_06400\_02000\_4400\_N} (18) & 19 & 120 & 80 & 16.9 \\
GP ISO\,10000 & \texttt{0109\_005\_GP\_10000\_08000\_4400\_N} (23) & 3 & 128 & 0 & 14.3 \\
IP ISO\,1000 & \texttt{0112\_005\_IP\_01000\_07500\_5500\_L} (24) & 30 & 80 & 24 & 3.3 \\
IP ISO\,1600 & \texttt{0031\_001\_IP\_01600\_02000\_3200\_N} (4) & 20 & 40 & 8 & 5.0 \\
IP ISO\,2000 & \texttt{0071\_003\_IP\_02000\_04000\_5500\_L} (14) & 22 & 8 & 24 & 3.1 \\
N6 ISO\,400 & \texttt{0074\_004\_N6\_00400\_00040\_3200\_L} (15) & 19 & 96 & 120 & 12.8 \\
N6 ISO\,800 & \texttt{0049\_002\_N6\_00800\_00800\_5500\_L} (8) & 22 & 104 & 104 & 7.8 \\
N6 ISO\,3200 & \texttt{0056\_003\_N6\_03200\_04000\_5500\_N} (10) & 29 & 48 & 96 & 12.3 \\
S6 ISO\,400 & \texttt{0176\_008\_S6\_00400\_00100\_5500\_L} (37) & 26 & 0 & 8 & 15.5 \\
S6 ISO\,800 & \texttt{0061\_003\_S6\_00800\_00500\_4400\_L} (12) & 27 & 96 & 56 & 13.0 \\
S6 ISO\,1600 & \texttt{0103\_005\_S6\_01600\_00800\_4400\_L} (22) & 22 & 128 & 0 & 9.1 \\
S6 ISO\,3200 & \texttt{0053\_002\_S6\_03200\_02000\_5500\_N} (9) & 22 & 24 & 96 & 13.9 \\
\midrule
Radius & Test image (size $h\times w$) & & $x$ & $y$ & Tex. \\
\midrule
All radii & 38 ($333\times499$) & & 115 & 19 & 62.2 \\
\bottomrule
\end{tabular}}
{\caption{\textbf{Crop windows of the highly textured comparisons} (Figs.~\ref{fig:supp-sidd-textured-1}--\ref{fig:supp-sidd-textured-4} and~\ref{fig:supp-film-textured-1}--\ref{fig:supp-film-textured-2}): top-left corner $(x,y)$ of each $128\times128$ window and its texture (Tex.), the median $7\times7$ local standard deviation of the clean window in 8-bit code values. $^\star$Also shown in the main paper.}
\label{tab:supp-crops-textured}}
\end{table*}

This section extends Fig.~\mainref{fig:main_sidd_vis} of the main paper to every
evaluated camera--ISO class of SIDD~\cite{abdelhamed2018sidd}. Every crop comes from the
evaluation scored in Tab.~\mainref{tab:sidd} of the main paper: each method shows the first of its
five generator runs, with the fit or checkpoint setting used there. NM-FlowGAN and
sRGB-Flow are shown only for the classes their checkpoints cover. Residual rows
show the output minus the clean image, amplified by one gain per class that is set
from the real residual alone, so every method in a row is shown at the same scale.

\begin{figure*}[tp]
\centering
\vcsetup{8}%
\begin{tabular}{@{}c*{8}{c}@{}}
  & \vchead{Clean} & \vchead{Real} & \vchead{\textbf{PGNM}} & \vchead{GMDCN} & \vchead{NM-FlowGAN} & \vchead{NAFlow} & \vchead{sRGB-Flow} & \vchead{DANet} \\[2pt]
\vcrow{G4, ISO\,400} & \includegraphics[width=\vccell]{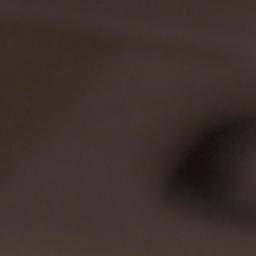} & \includegraphics[width=\vccell]{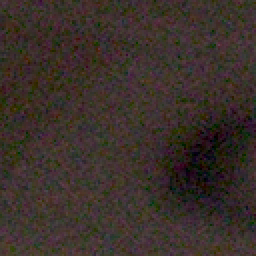} & \includegraphics[width=\vccell]{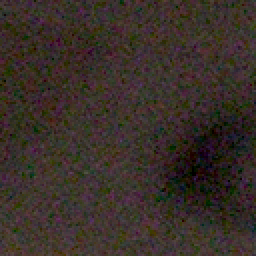} & \includegraphics[width=\vccell]{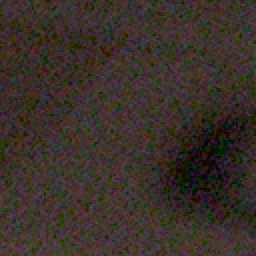} & \vcna & \includegraphics[width=\vccell]{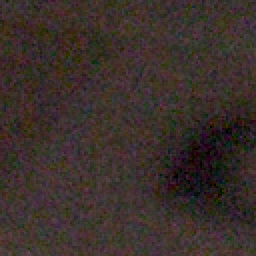} & \vcna & \includegraphics[width=\vccell]{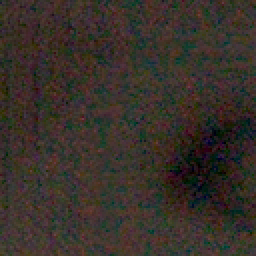} \\[0.5pt]
\vcrow{residual $\times$3} & \includegraphics[width=\vccell]{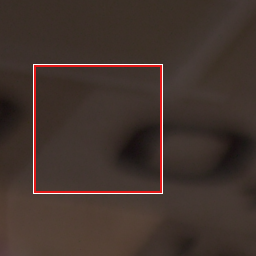} & \includegraphics[width=\vccell]{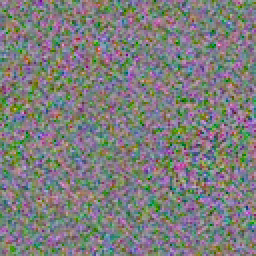} & \includegraphics[width=\vccell]{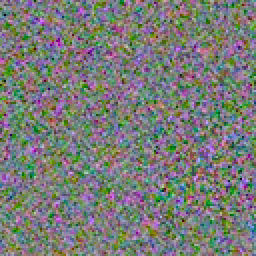} & \includegraphics[width=\vccell]{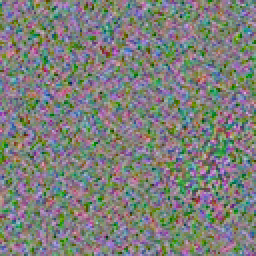} & \vcna & \includegraphics[width=\vccell]{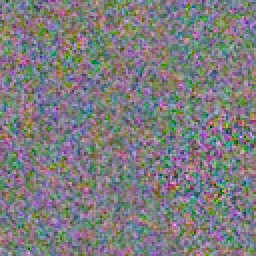} & \vcna & \includegraphics[width=\vccell]{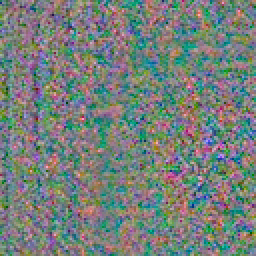} \\[3pt]
\vcrow{G4, ISO\,800} & \includegraphics[width=\vccell]{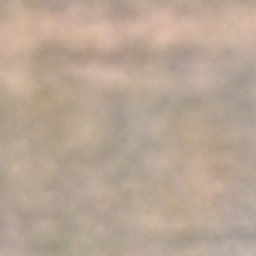} & \includegraphics[width=\vccell]{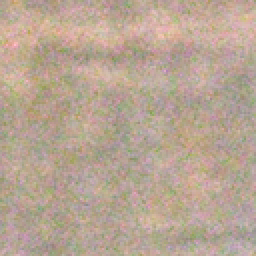} & \includegraphics[width=\vccell]{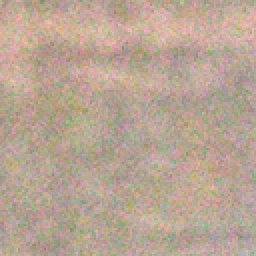} & \includegraphics[width=\vccell]{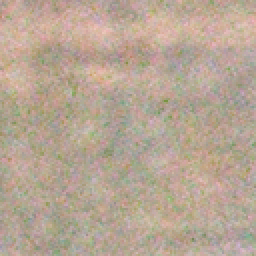} & \includegraphics[width=\vccell]{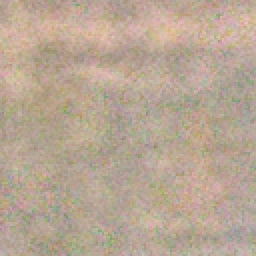} & \includegraphics[width=\vccell]{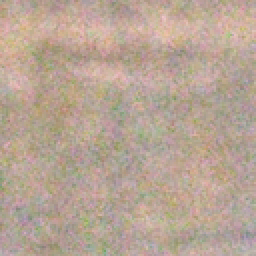} & \includegraphics[width=\vccell]{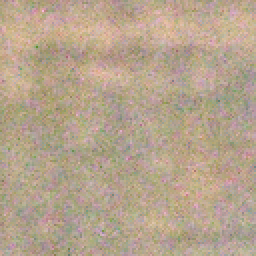} & \includegraphics[width=\vccell]{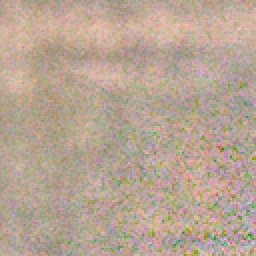} \\[0.5pt]
\vcrow{residual $\times$3} & \includegraphics[width=\vccell]{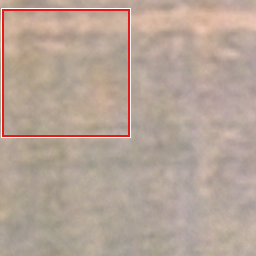} & \includegraphics[width=\vccell]{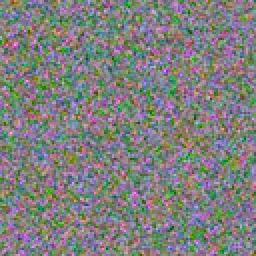} & \includegraphics[width=\vccell]{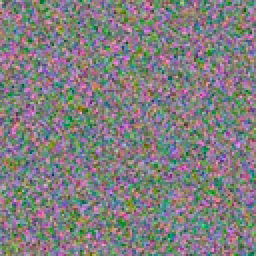} & \includegraphics[width=\vccell]{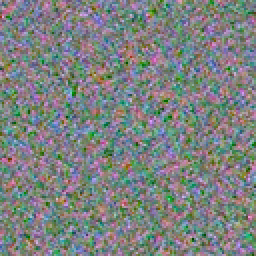} & \includegraphics[width=\vccell]{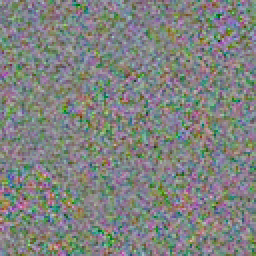} & \includegraphics[width=\vccell]{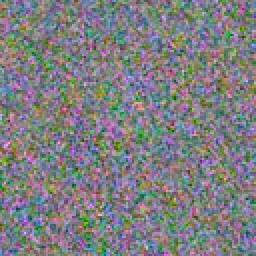} & \includegraphics[width=\vccell]{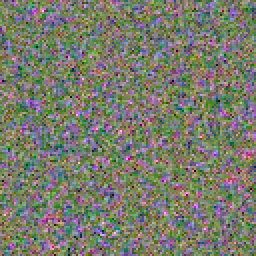} & \includegraphics[width=\vccell]{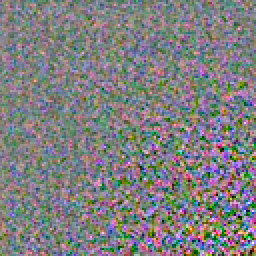} \\[3pt]
\vcrow{GP, ISO\,1600} & \includegraphics[width=\vccell]{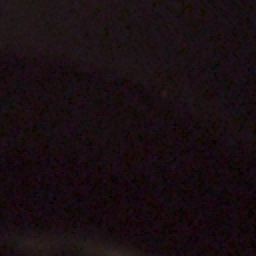} & \includegraphics[width=\vccell]{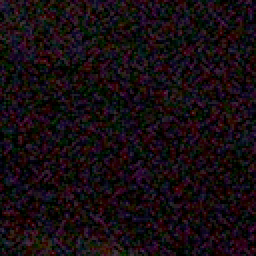} & \includegraphics[width=\vccell]{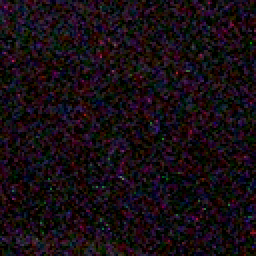} & \includegraphics[width=\vccell]{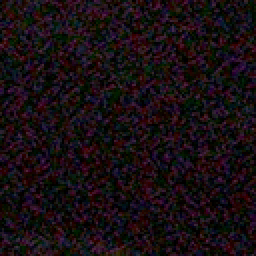} & \vcna & \includegraphics[width=\vccell]{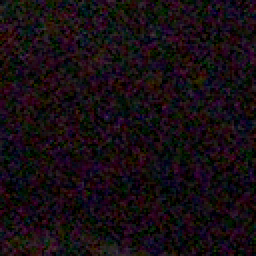} & \vcna & \includegraphics[width=\vccell]{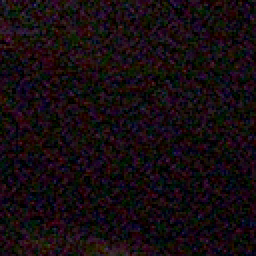} \\[0.5pt]
\vcrow{residual $\times$2.5} & \includegraphics[width=\vccell]{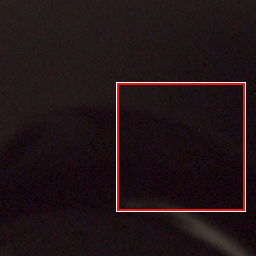} & \includegraphics[width=\vccell]{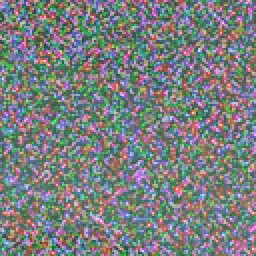} & \includegraphics[width=\vccell]{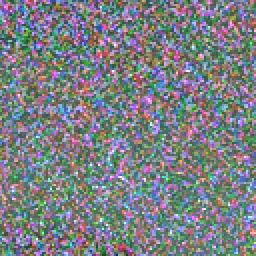} & \includegraphics[width=\vccell]{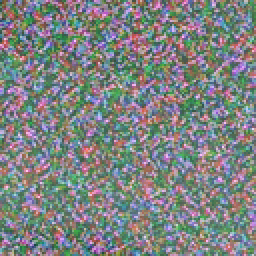} & \vcna & \includegraphics[width=\vccell]{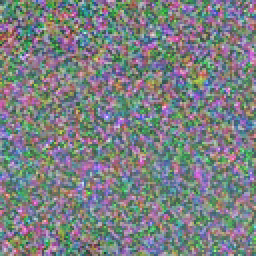} & \vcna & \includegraphics[width=\vccell]{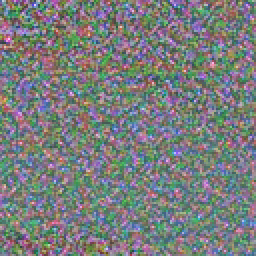} \\[3pt]
\vcrow{GP, ISO\,3200} & \includegraphics[width=\vccell]{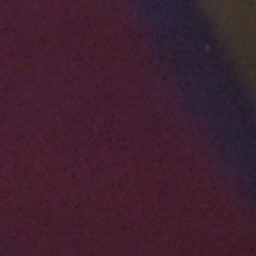} & \includegraphics[width=\vccell]{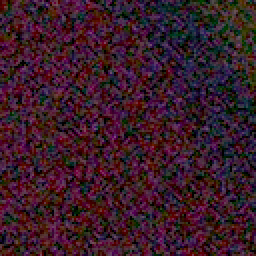} & \includegraphics[width=\vccell]{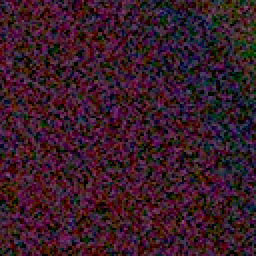} & \includegraphics[width=\vccell]{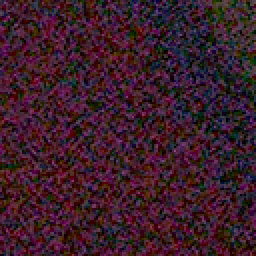} & \vcna & \includegraphics[width=\vccell]{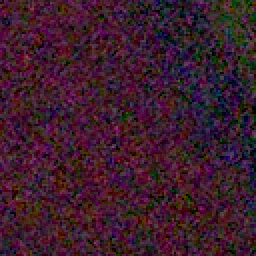} & \vcna & \includegraphics[width=\vccell]{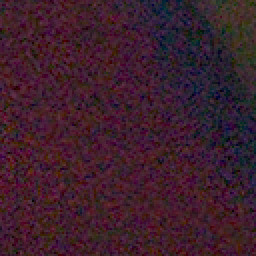} \\[0.5pt]
\vcrow{residual $\times$1.5} & \includegraphics[width=\vccell]{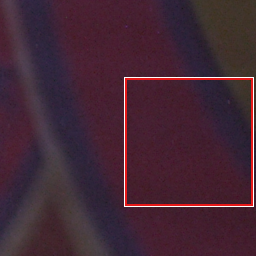} & \includegraphics[width=\vccell]{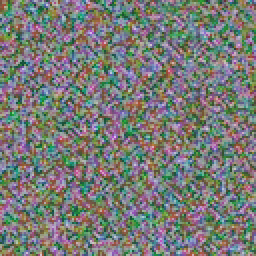} & \includegraphics[width=\vccell]{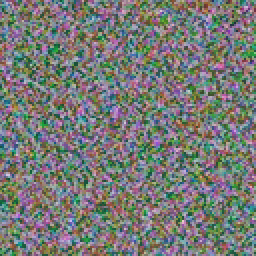} & \includegraphics[width=\vccell]{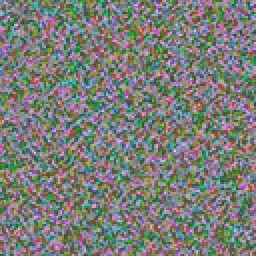} & \vcna & \includegraphics[width=\vccell]{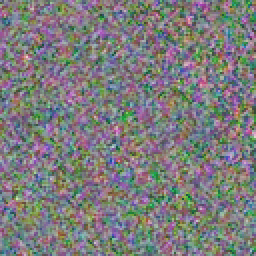} & \vcna & \includegraphics[width=\vccell]{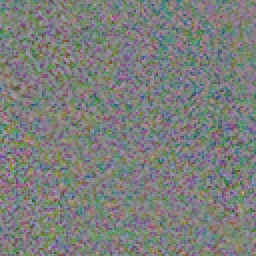} \\
\end{tabular}
\caption{\textbf{Uncurated SIDD comparison, every evaluated camera--ISO class (1/4).} Each source, block and $128\times128$ window was drawn with a fixed pseudo-random seed from the clean image only; a window was redrawn only when more than 25\% of its clean pixels were near saturation, and no draw was replaced after viewing the outputs. Every method shows its first generator run. sRGB-Flow and NM-FlowGAN are shown only for the 19-source block covered by their checkpoints (n/a otherwise). Second row of each pair: residual relative to the clean image, amplified by the factor at left (set from the real residual alone; gray is zero); under the clean column, the full $256\times256$ block with the crop window marked. Crops are shown at native resolution; zoom in for detail. }
\label{fig:supp-sidd-1}
\end{figure*}

\begin{figure*}[tp]
\centering
\vcsetup{8}%
\begin{tabular}{@{}c*{8}{c}@{}}
  & \vchead{Clean} & \vchead{Real} & \vchead{\textbf{PGNM}} & \vchead{GMDCN} & \vchead{NM-FlowGAN} & \vchead{NAFlow} & \vchead{sRGB-Flow} & \vchead{DANet} \\[2pt]
\vcrow{GP, ISO\,6400} & \includegraphics[width=\vccell]{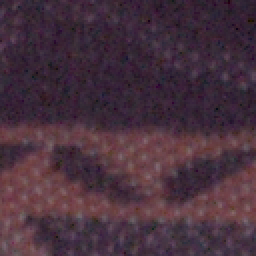} & \includegraphics[width=\vccell]{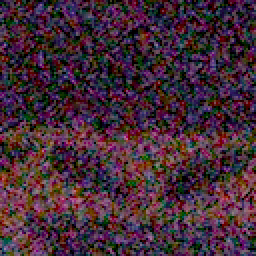} & \includegraphics[width=\vccell]{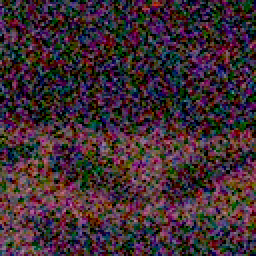} & \includegraphics[width=\vccell]{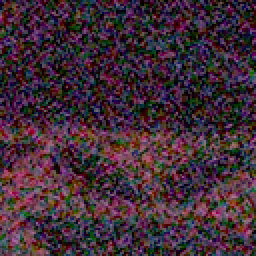} & \vcna & \includegraphics[width=\vccell]{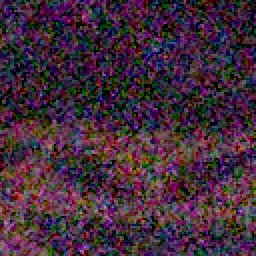} & \vcna & \includegraphics[width=\vccell]{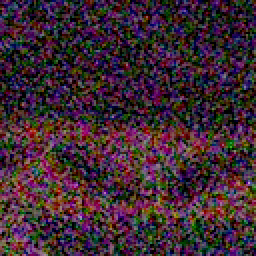} \\[0.5pt]
\vcrow{residual $\times$1} & \includegraphics[width=\vccell]{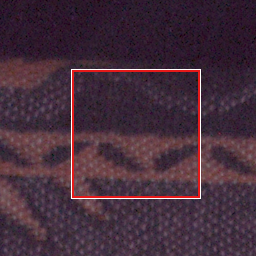} & \includegraphics[width=\vccell]{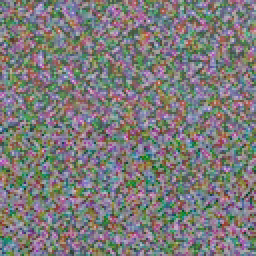} & \includegraphics[width=\vccell]{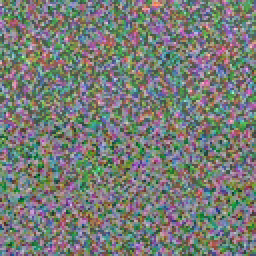} & \includegraphics[width=\vccell]{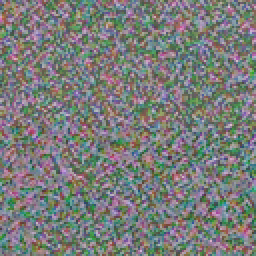} & \vcna & \includegraphics[width=\vccell]{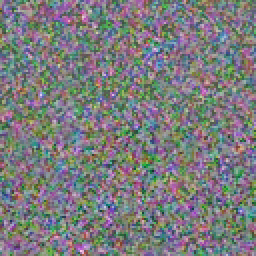} & \vcna & \includegraphics[width=\vccell]{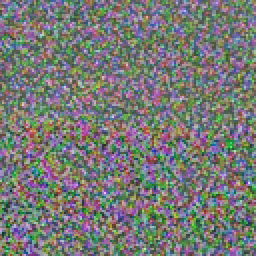} \\[3pt]
\vcrow{GP, ISO\,10000} & \includegraphics[width=\vccell]{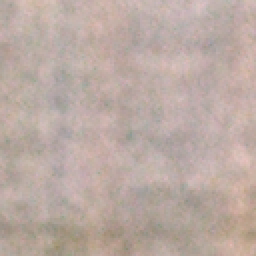} & \includegraphics[width=\vccell]{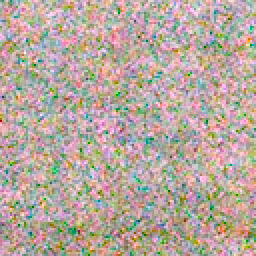} & \includegraphics[width=\vccell]{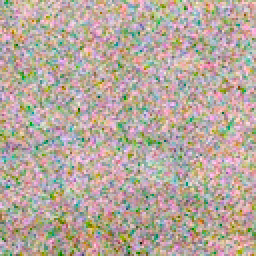} & \includegraphics[width=\vccell]{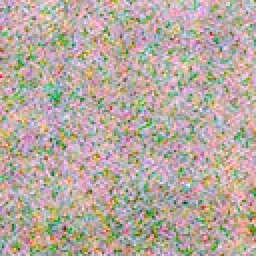} & \vcna & \includegraphics[width=\vccell]{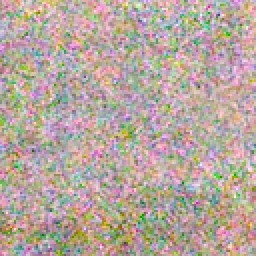} & \vcna & \includegraphics[width=\vccell]{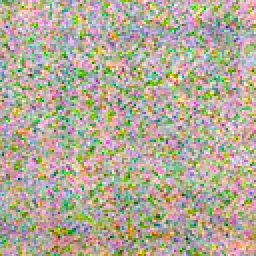} \\[0.5pt]
\vcrow{residual $\times$1} & \includegraphics[width=\vccell]{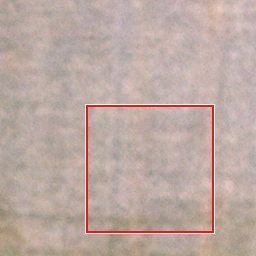} & \includegraphics[width=\vccell]{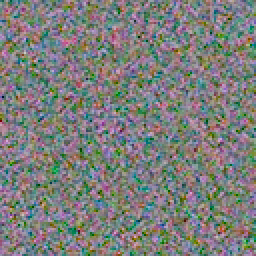} & \includegraphics[width=\vccell]{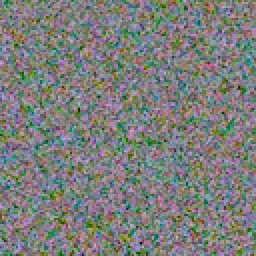} & \includegraphics[width=\vccell]{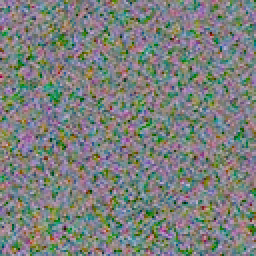} & \vcna & \includegraphics[width=\vccell]{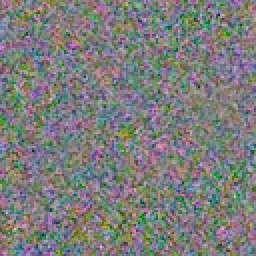} & \vcna & \includegraphics[width=\vccell]{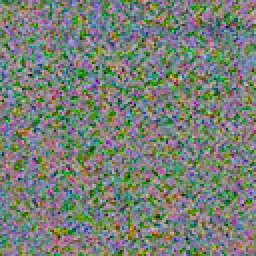} \\[3pt]
\vcrow{IP, ISO\,1000} & \includegraphics[width=\vccell]{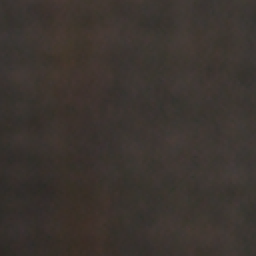} & \includegraphics[width=\vccell]{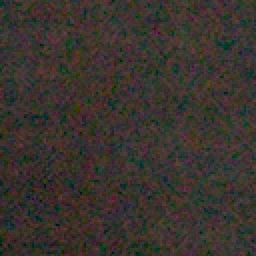} & \includegraphics[width=\vccell]{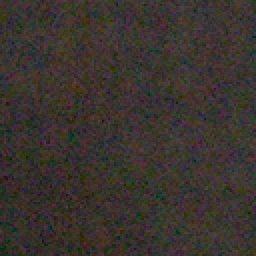} & \includegraphics[width=\vccell]{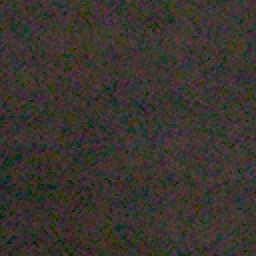} & \vcna & \includegraphics[width=\vccell]{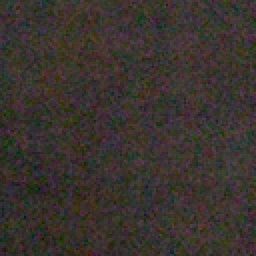} & \vcna & \includegraphics[width=\vccell]{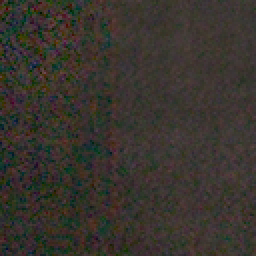} \\[0.5pt]
\vcrow{residual $\times$2.5} & \includegraphics[width=\vccell]{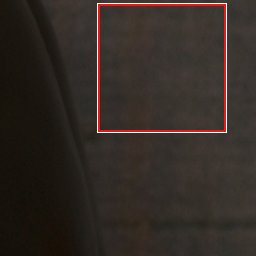} & \includegraphics[width=\vccell]{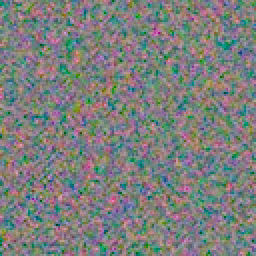} & \includegraphics[width=\vccell]{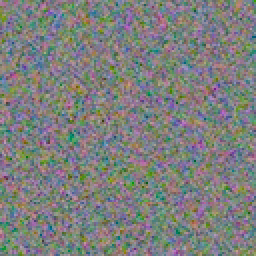} & \includegraphics[width=\vccell]{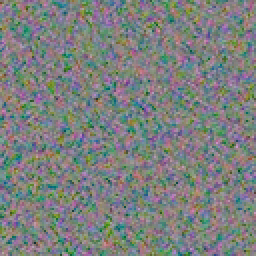} & \vcna & \includegraphics[width=\vccell]{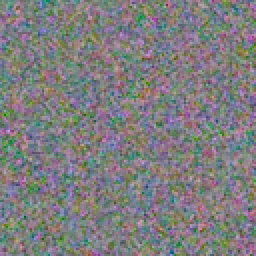} & \vcna & \includegraphics[width=\vccell]{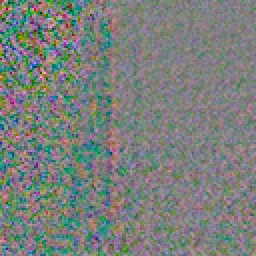} \\[3pt]
\vcrow{IP, ISO\,1600} & \includegraphics[width=\vccell]{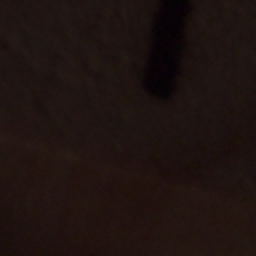} & \includegraphics[width=\vccell]{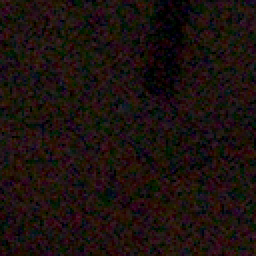} & \includegraphics[width=\vccell]{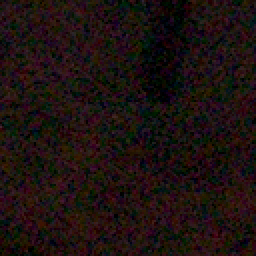} & \includegraphics[width=\vccell]{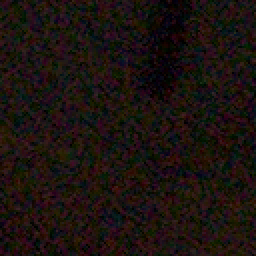} & \includegraphics[width=\vccell]{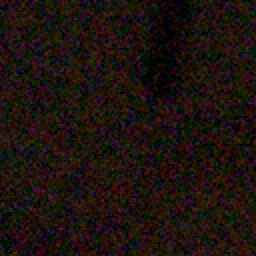} & \includegraphics[width=\vccell]{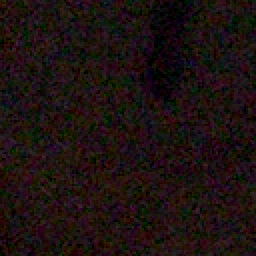} & \includegraphics[width=\vccell]{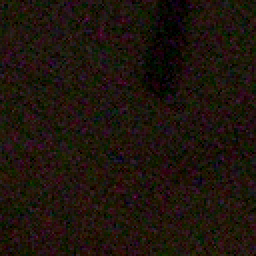} & \includegraphics[width=\vccell]{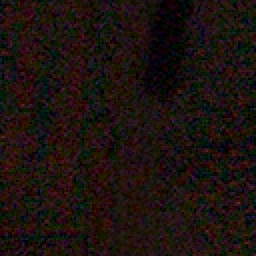} \\[0.5pt]
\vcrow{residual $\times$3} & \includegraphics[width=\vccell]{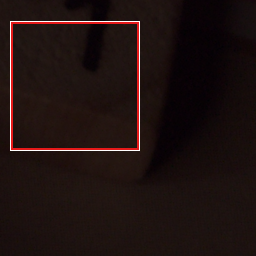} & \includegraphics[width=\vccell]{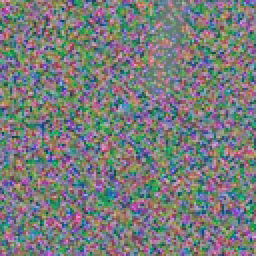} & \includegraphics[width=\vccell]{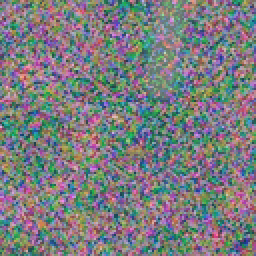} & \includegraphics[width=\vccell]{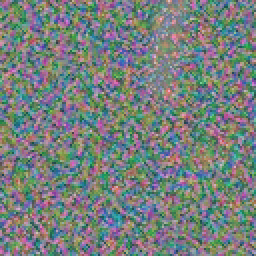} & \includegraphics[width=\vccell]{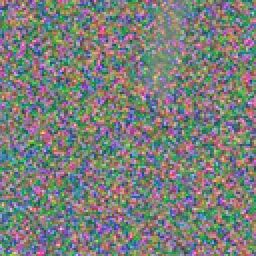} & \includegraphics[width=\vccell]{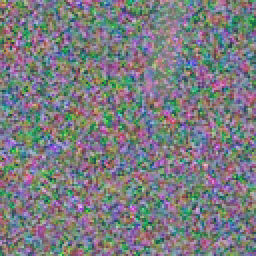} & \includegraphics[width=\vccell]{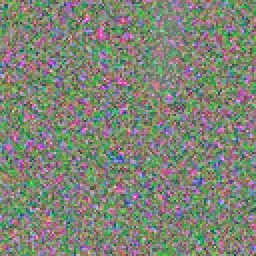} & \includegraphics[width=\vccell]{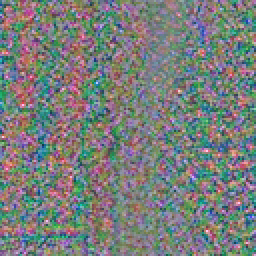} \\
\end{tabular}
\caption{\textbf{Uncurated SIDD comparison, every evaluated camera--ISO class (2/4).} Each source, block and $128\times128$ window was drawn with a fixed pseudo-random seed from the clean image only; a window was redrawn only when more than 25\% of its clean pixels were near saturation, and no draw was replaced after viewing the outputs. Every method shows its first generator run. sRGB-Flow and NM-FlowGAN are shown only for the 19-source block covered by their checkpoints (n/a otherwise). Second row of each pair: residual relative to the clean image, amplified by the factor at left (set from the real residual alone; gray is zero); under the clean column, the full $256\times256$ block with the crop window marked. Crops are shown at native resolution; zoom in for detail. }
\label{fig:supp-sidd-2}
\end{figure*}

\begin{figure*}[tp]
\centering
\vcsetup{8}%
\begin{tabular}{@{}c*{8}{c}@{}}
  & \vchead{Clean} & \vchead{Real} & \vchead{\textbf{PGNM}} & \vchead{GMDCN} & \vchead{NM-FlowGAN} & \vchead{NAFlow} & \vchead{sRGB-Flow} & \vchead{DANet} \\[2pt]
\vcrow{IP, ISO\,2000} & \includegraphics[width=\vccell]{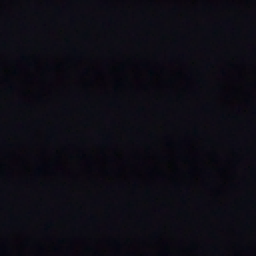} & \includegraphics[width=\vccell]{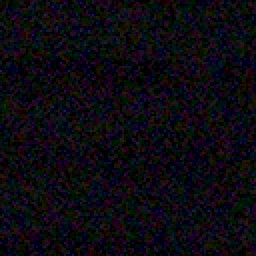} & \includegraphics[width=\vccell]{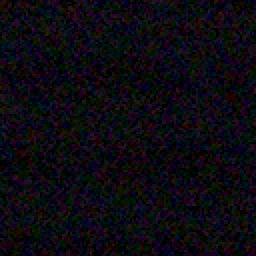} & \includegraphics[width=\vccell]{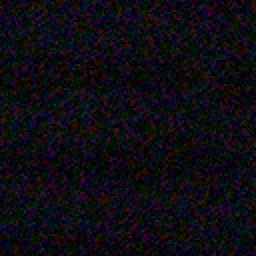} & \vcna & \includegraphics[width=\vccell]{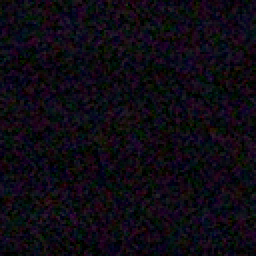} & \vcna & \includegraphics[width=\vccell]{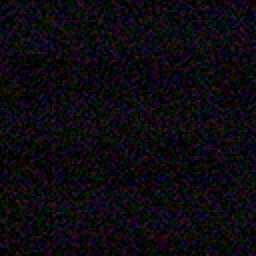} \\[0.5pt]
\vcrow{residual $\times$4} & \includegraphics[width=\vccell]{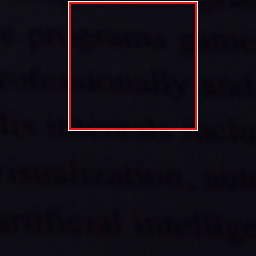} & \includegraphics[width=\vccell]{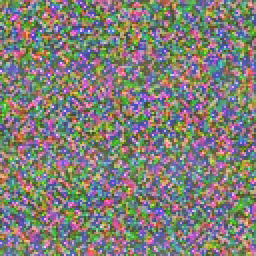} & \includegraphics[width=\vccell]{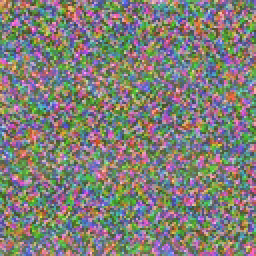} & \includegraphics[width=\vccell]{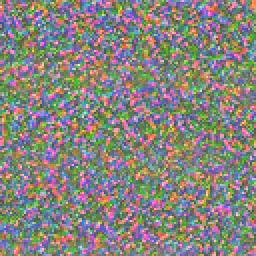} & \vcna & \includegraphics[width=\vccell]{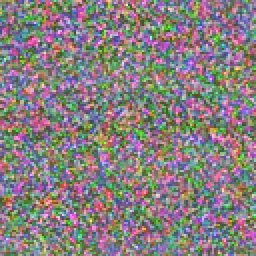} & \vcna & \includegraphics[width=\vccell]{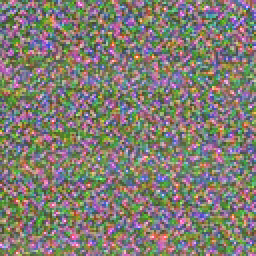} \\[3pt]
\vcrow{N6, ISO\,400} & \includegraphics[width=\vccell]{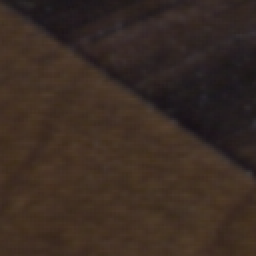} & \includegraphics[width=\vccell]{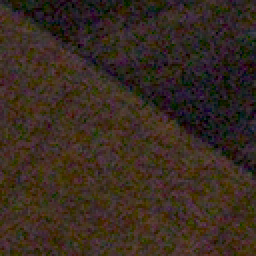} & \includegraphics[width=\vccell]{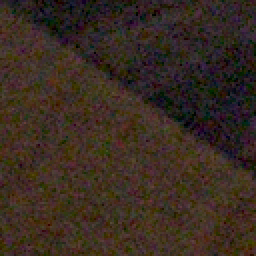} & \includegraphics[width=\vccell]{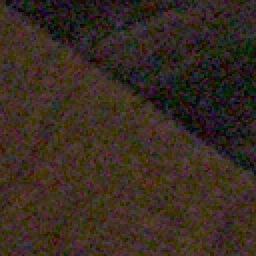} & \vcna & \includegraphics[width=\vccell]{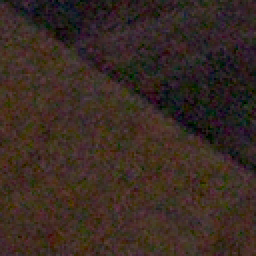} & \vcna & \includegraphics[width=\vccell]{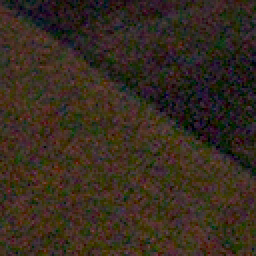} \\[0.5pt]
\vcrow{residual $\times$2.5} & \includegraphics[width=\vccell]{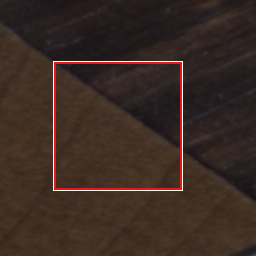} & \includegraphics[width=\vccell]{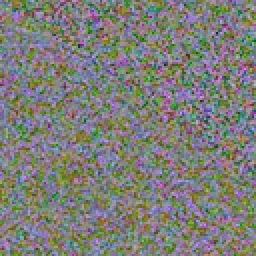} & \includegraphics[width=\vccell]{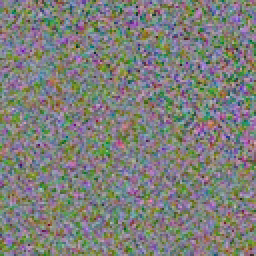} & \includegraphics[width=\vccell]{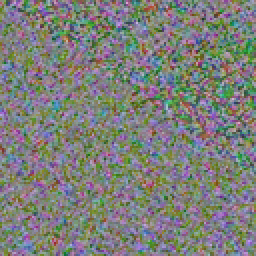} & \vcna & \includegraphics[width=\vccell]{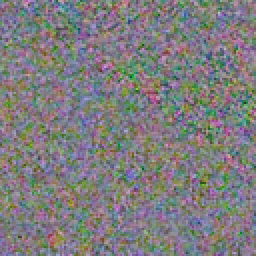} & \vcna & \includegraphics[width=\vccell]{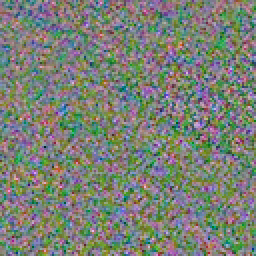} \\[3pt]
\vcrow{N6, ISO\,800} & \includegraphics[width=\vccell]{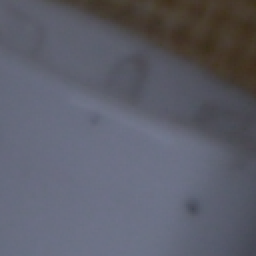} & \includegraphics[width=\vccell]{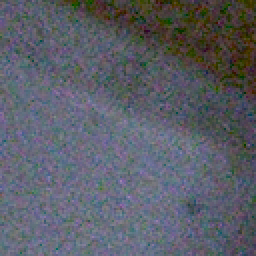} & \includegraphics[width=\vccell]{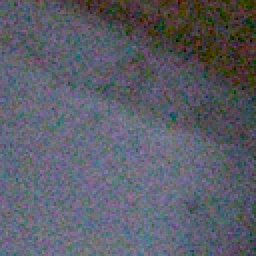} & \includegraphics[width=\vccell]{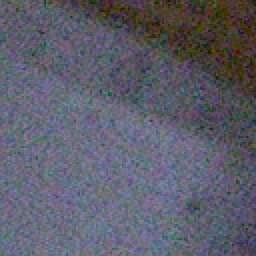} & \includegraphics[width=\vccell]{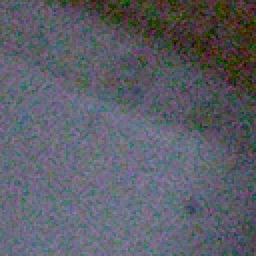} & \includegraphics[width=\vccell]{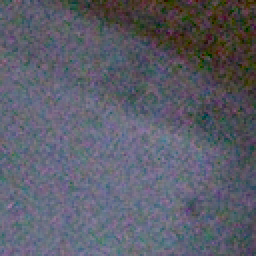} & \includegraphics[width=\vccell]{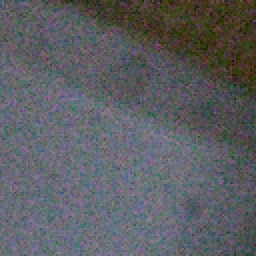} & \includegraphics[width=\vccell]{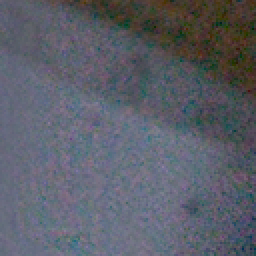} \\[0.5pt]
\vcrow{residual $\times$2.5} & \includegraphics[width=\vccell]{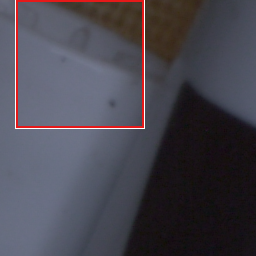} & \includegraphics[width=\vccell]{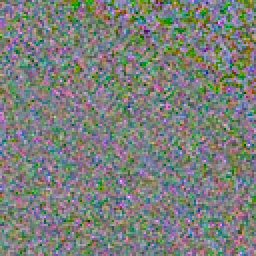} & \includegraphics[width=\vccell]{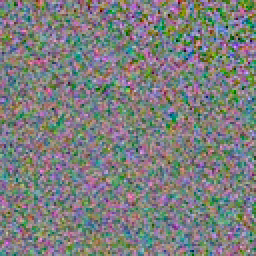} & \includegraphics[width=\vccell]{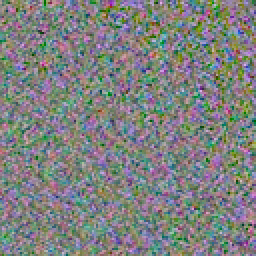} & \includegraphics[width=\vccell]{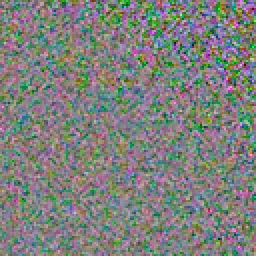} & \includegraphics[width=\vccell]{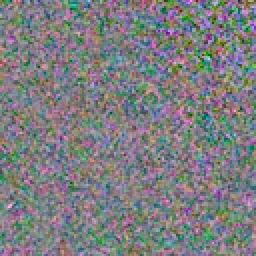} & \includegraphics[width=\vccell]{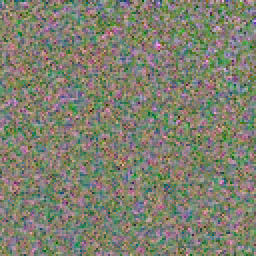} & \includegraphics[width=\vccell]{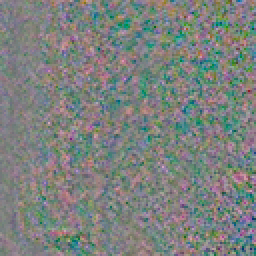} \\[3pt]
\vcrow{N6, ISO\,3200} & \includegraphics[width=\vccell]{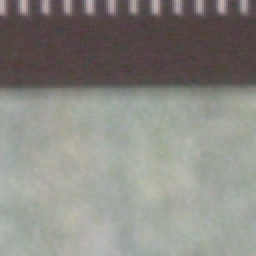} & \includegraphics[width=\vccell]{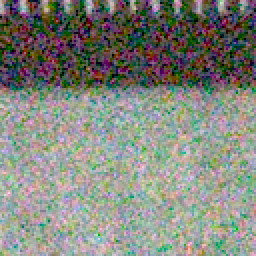} & \includegraphics[width=\vccell]{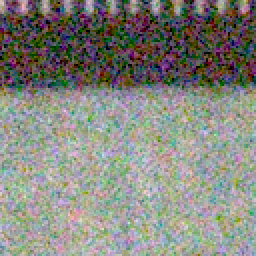} & \includegraphics[width=\vccell]{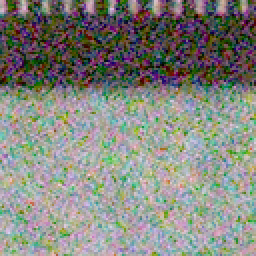} & \vcna & \includegraphics[width=\vccell]{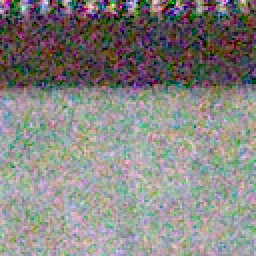} & \vcna & \includegraphics[width=\vccell]{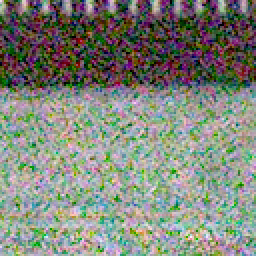} \\[0.5pt]
\vcrow{residual $\times$1} & \includegraphics[width=\vccell]{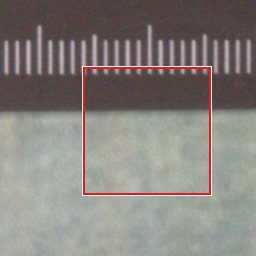} & \includegraphics[width=\vccell]{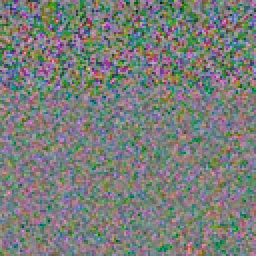} & \includegraphics[width=\vccell]{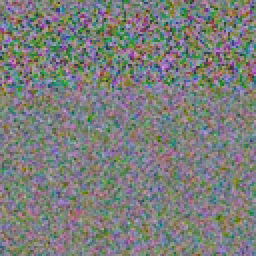} & \includegraphics[width=\vccell]{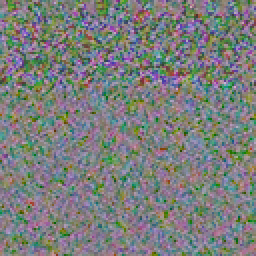} & \vcna & \includegraphics[width=\vccell]{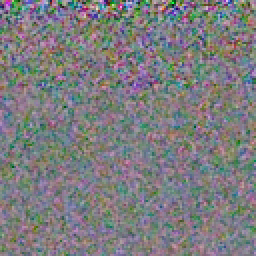} & \vcna & \includegraphics[width=\vccell]{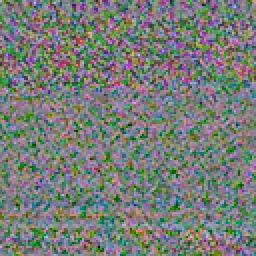} \\
\end{tabular}
\caption{\textbf{Uncurated SIDD comparison, every evaluated camera--ISO class (3/4).} Each source, block and $128\times128$ window was drawn with a fixed pseudo-random seed from the clean image only; a window was redrawn only when more than 25\% of its clean pixels were near saturation, and no draw was replaced after viewing the outputs. Every method shows its first generator run. sRGB-Flow and NM-FlowGAN are shown only for the 19-source block covered by their checkpoints (n/a otherwise). Second row of each pair: residual relative to the clean image, amplified by the factor at left (set from the real residual alone; gray is zero); under the clean column, the full $256\times256$ block with the crop window marked. Crops are shown at native resolution; zoom in for detail. }
\label{fig:supp-sidd-3}
\end{figure*}

\begin{figure*}[tp]
\centering
\vcsetup{8}%
\begin{tabular}{@{}c*{8}{c}@{}}
  & \vchead{Clean} & \vchead{Real} & \vchead{\textbf{PGNM}} & \vchead{GMDCN} & \vchead{NM-FlowGAN} & \vchead{NAFlow} & \vchead{sRGB-Flow} & \vchead{DANet} \\[2pt]
\vcrow{S6, ISO\,400$^\star$} & \includegraphics[width=\vccell]{supp/visual_comparison/sidd/S6-400/clean.png} & \includegraphics[width=\vccell]{supp/visual_comparison/sidd/S6-400/real.png} & \includegraphics[width=\vccell]{supp/visual_comparison/sidd/S6-400/pgnm.png} & \includegraphics[width=\vccell]{supp/visual_comparison/sidd/S6-400/cfg_nin.png} & \includegraphics[width=\vccell]{supp/visual_comparison/sidd/S6-400/nmflowgan.png} & \includegraphics[width=\vccell]{supp/visual_comparison/sidd/S6-400/naflow.png} & \includegraphics[width=\vccell]{supp/visual_comparison/sidd/S6-400/flow_srgb.png} & \includegraphics[width=\vccell]{supp/visual_comparison/sidd/S6-400/danet.png} \\[0.5pt]
\vcrow{residual $\times$2.5} & \includegraphics[width=\vccell]{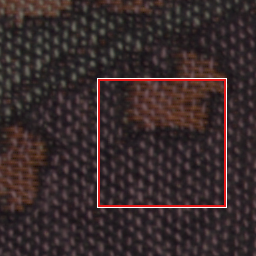} & \includegraphics[width=\vccell]{supp/visual_comparison/sidd/S6-400/real-res.png} & \includegraphics[width=\vccell]{supp/visual_comparison/sidd/S6-400/pgnm-res.png} & \includegraphics[width=\vccell]{supp/visual_comparison/sidd/S6-400/cfg_nin-res.png} & \includegraphics[width=\vccell]{supp/visual_comparison/sidd/S6-400/nmflowgan-res.png} & \includegraphics[width=\vccell]{supp/visual_comparison/sidd/S6-400/naflow-res.png} & \includegraphics[width=\vccell]{supp/visual_comparison/sidd/S6-400/flow_srgb-res.png} & \includegraphics[width=\vccell]{supp/visual_comparison/sidd/S6-400/danet-res.png} \\[3pt]
\vcrow{S6, ISO\,800} & \includegraphics[width=\vccell]{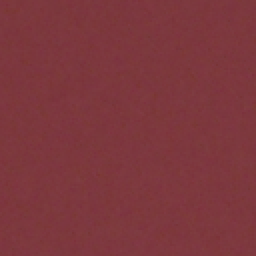} & \includegraphics[width=\vccell]{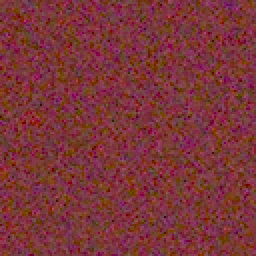} & \includegraphics[width=\vccell]{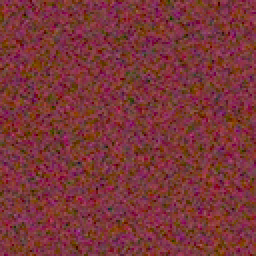} & \includegraphics[width=\vccell]{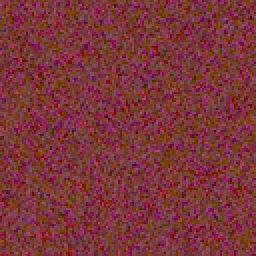} & \includegraphics[width=\vccell]{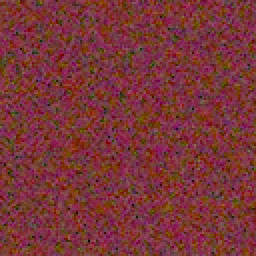} & \includegraphics[width=\vccell]{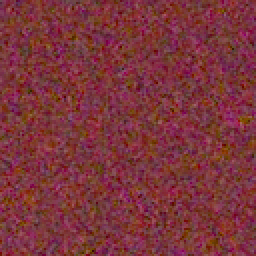} & \includegraphics[width=\vccell]{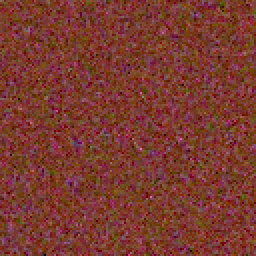} & \includegraphics[width=\vccell]{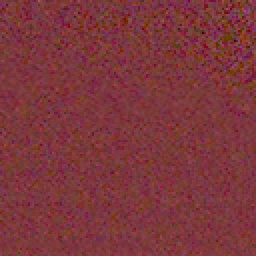} \\[0.5pt]
\vcrow{residual $\times$1.5} & \includegraphics[width=\vccell]{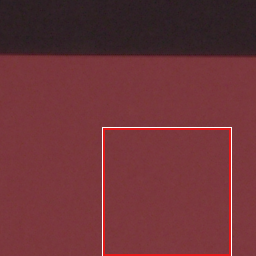} & \includegraphics[width=\vccell]{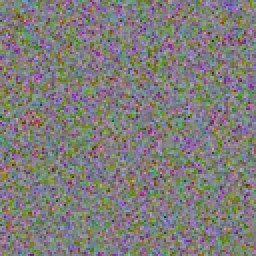} & \includegraphics[width=\vccell]{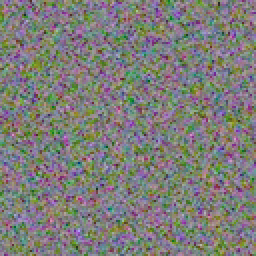} & \includegraphics[width=\vccell]{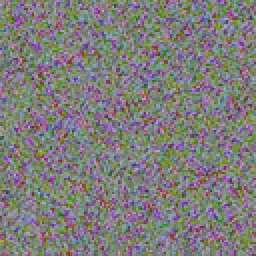} & \includegraphics[width=\vccell]{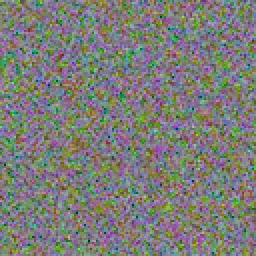} & \includegraphics[width=\vccell]{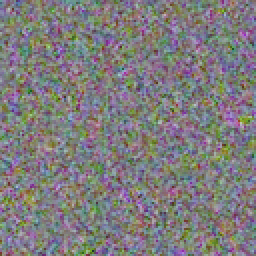} & \includegraphics[width=\vccell]{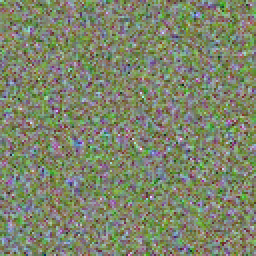} & \includegraphics[width=\vccell]{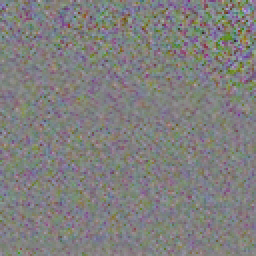} \\[3pt]
\vcrow{S6, ISO\,1600} & \includegraphics[width=\vccell]{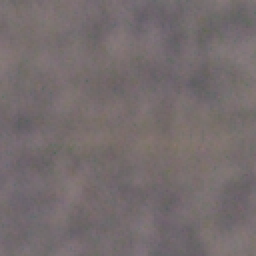} & \includegraphics[width=\vccell]{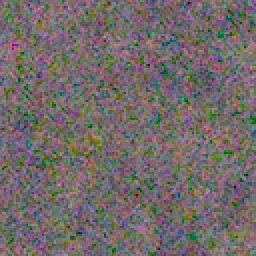} & \includegraphics[width=\vccell]{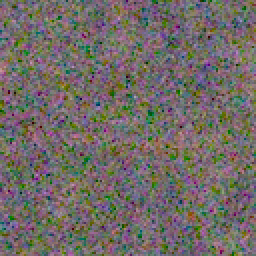} & \includegraphics[width=\vccell]{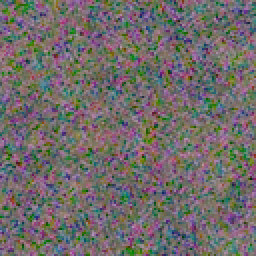} & \includegraphics[width=\vccell]{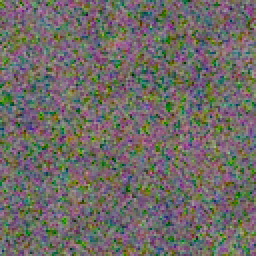} & \includegraphics[width=\vccell]{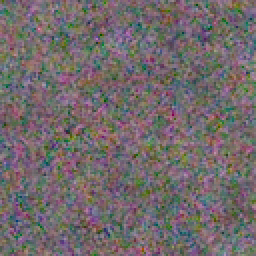} & \includegraphics[width=\vccell]{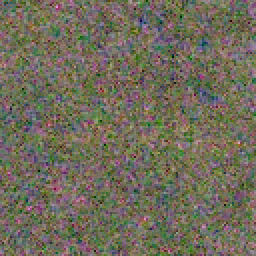} & \includegraphics[width=\vccell]{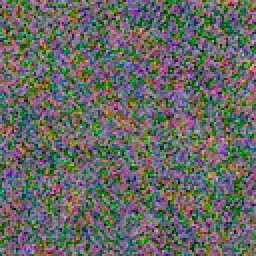} \\[0.5pt]
\vcrow{residual $\times$1} & \includegraphics[width=\vccell]{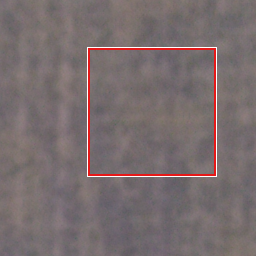} & \includegraphics[width=\vccell]{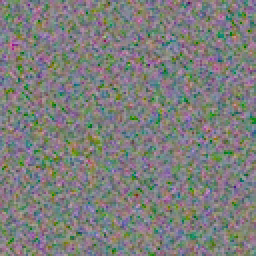} & \includegraphics[width=\vccell]{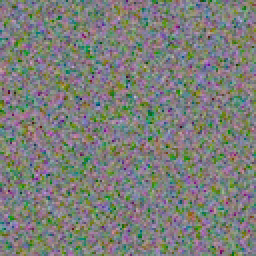} & \includegraphics[width=\vccell]{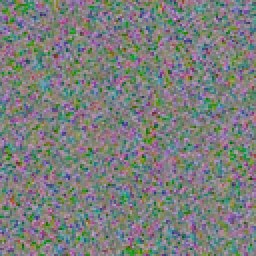} & \includegraphics[width=\vccell]{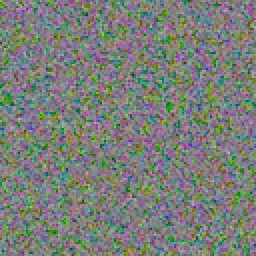} & \includegraphics[width=\vccell]{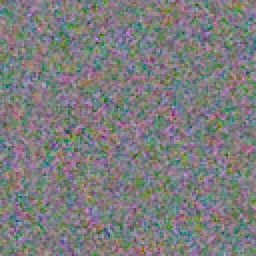} & \includegraphics[width=\vccell]{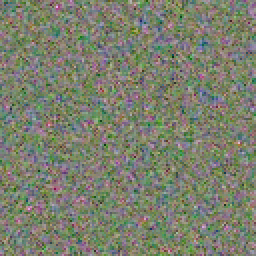} & \includegraphics[width=\vccell]{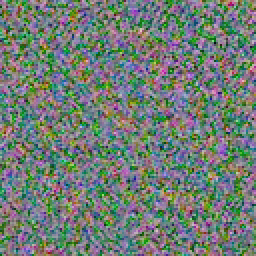} \\[3pt]
\vcrow{S6, ISO\,3200} & \includegraphics[width=\vccell]{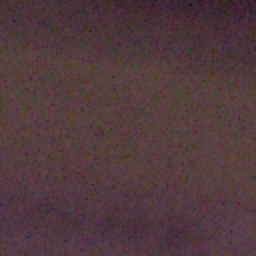} & \includegraphics[width=\vccell]{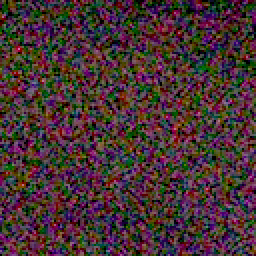} & \includegraphics[width=\vccell]{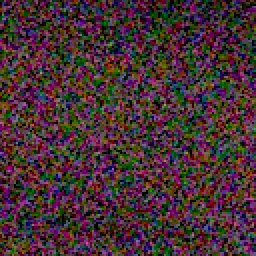} & \includegraphics[width=\vccell]{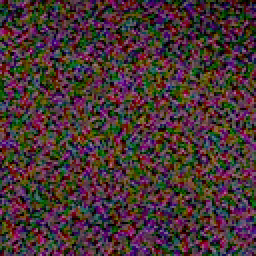} & \includegraphics[width=\vccell]{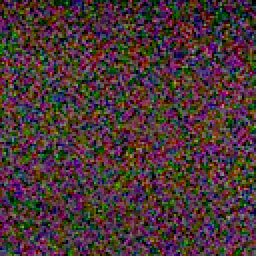} & \includegraphics[width=\vccell]{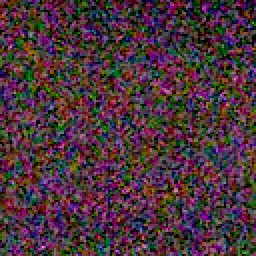} & \includegraphics[width=\vccell]{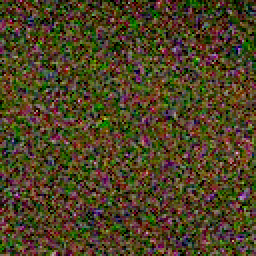} & \includegraphics[width=\vccell]{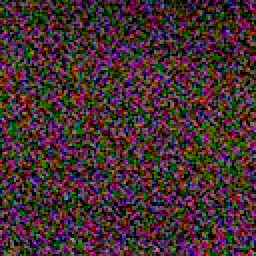} \\[0.5pt]
\vcrow{residual $\times$1} & \includegraphics[width=\vccell]{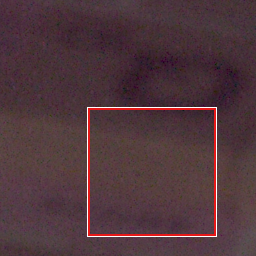} & \includegraphics[width=\vccell]{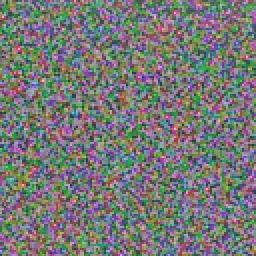} & \includegraphics[width=\vccell]{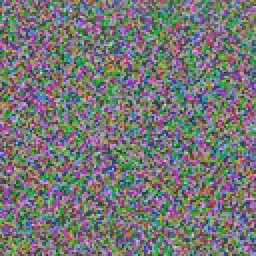} & \includegraphics[width=\vccell]{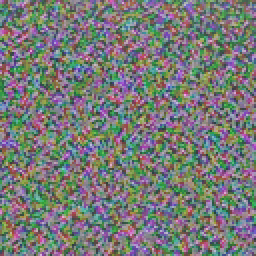} & \includegraphics[width=\vccell]{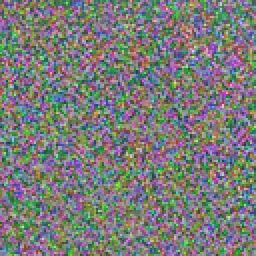} & \includegraphics[width=\vccell]{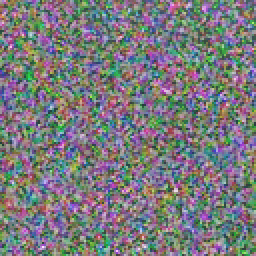} & \includegraphics[width=\vccell]{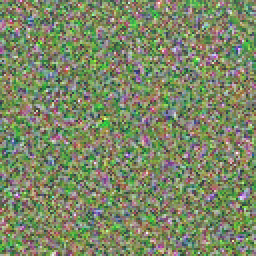} & \includegraphics[width=\vccell]{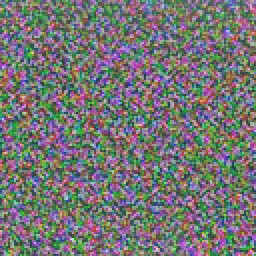} \\
\end{tabular}
\caption{\textbf{Uncurated SIDD comparison, every evaluated camera--ISO class (4/4).} Each source, block and $128\times128$ window was drawn with a fixed pseudo-random seed from the clean image only; a window was redrawn only when more than 25\% of its clean pixels were near saturation, and no draw was replaced after viewing the outputs. Every method shows its first generator run. sRGB-Flow and NM-FlowGAN are shown only for the 19-source block covered by their checkpoints (n/a otherwise). Second row of each pair: residual relative to the clean image, amplified by the factor at left (set from the real residual alone; gray is zero); under the clean column, the full $256\times256$ block with the crop window marked. Crops are shown at native resolution; zoom in for detail. $^\star$Also shown in the main paper (Fig.~\mainref{fig:main_sidd_vis}). }
\label{fig:supp-sidd-4}
\end{figure*}

\paragraph{Uncurated crops.}
For every class, one source image, block and $128\times128$ window were drawn with
a fixed seed from the clean image only, before any output was viewed
(Figs.~\ref{fig:supp-sidd-1}--\ref{fig:supp-sidd-4}; windows in
Tab.~\ref{tab:supp-crops}). The S6 ISO\,400 crop of the main paper is one of them.

\begin{figure*}[tp]
\centering
\vcsetup{8}%
\begin{tabular}{@{}c*{8}{c}@{}}
  & \vchead{Clean} & \vchead{Real} & \vchead{\textbf{PGNM}} & \vchead{GMDCN} & \vchead{NM-FlowGAN} & \vchead{NAFlow} & \vchead{sRGB-Flow} & \vchead{DANet} \\[2pt]
\vcrow{G4, ISO\,400} & \includegraphics[width=\vccell]{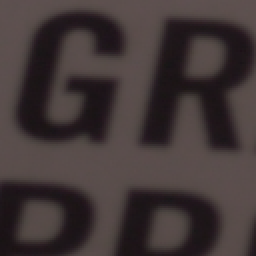} & \includegraphics[width=\vccell]{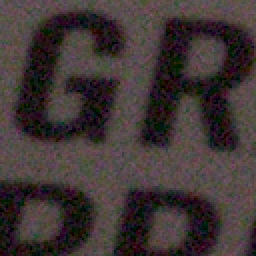} & \includegraphics[width=\vccell]{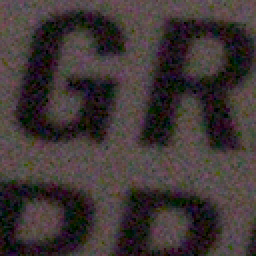} & \includegraphics[width=\vccell]{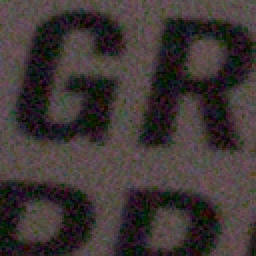} & \vcna & \includegraphics[width=\vccell]{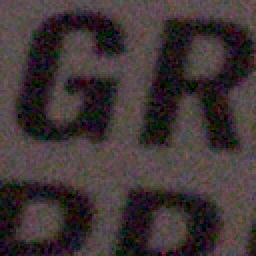} & \vcna & \includegraphics[width=\vccell]{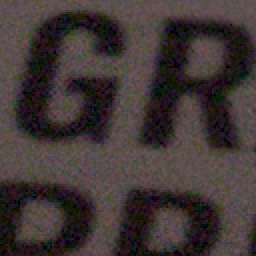} \\[0.5pt]
\vcrow{residual $\times$3} & \includegraphics[width=\vccell]{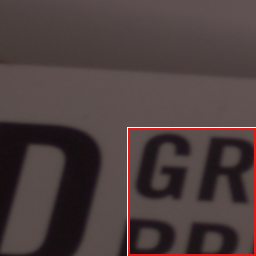} & \includegraphics[width=\vccell]{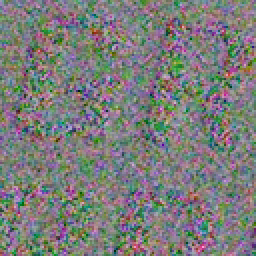} & \includegraphics[width=\vccell]{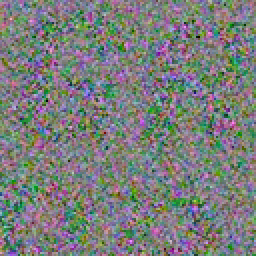} & \includegraphics[width=\vccell]{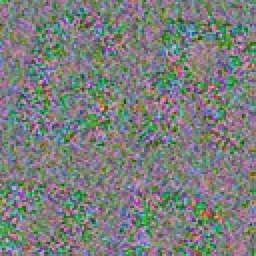} & \vcna & \includegraphics[width=\vccell]{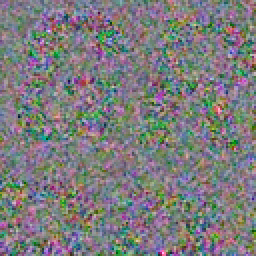} & \vcna & \includegraphics[width=\vccell]{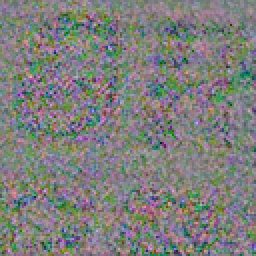} \\[3pt]
\vcrow{G4, ISO\,800} & \includegraphics[width=\vccell]{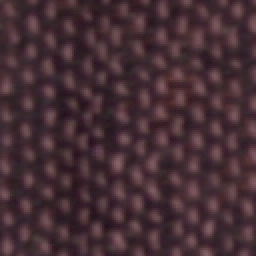} & \includegraphics[width=\vccell]{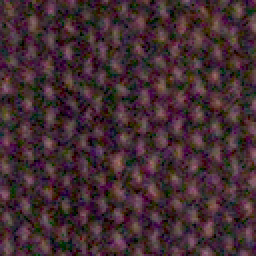} & \includegraphics[width=\vccell]{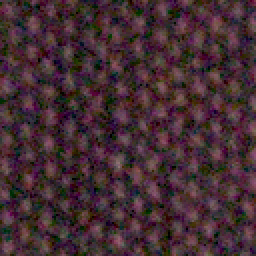} & \includegraphics[width=\vccell]{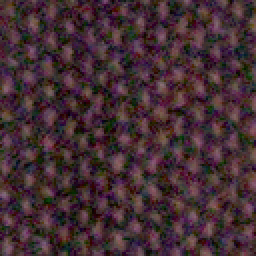} & \includegraphics[width=\vccell]{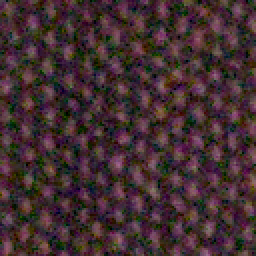} & \includegraphics[width=\vccell]{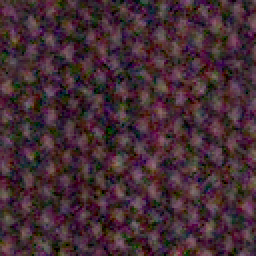} & \includegraphics[width=\vccell]{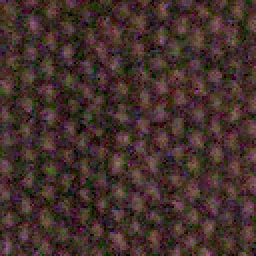} & \includegraphics[width=\vccell]{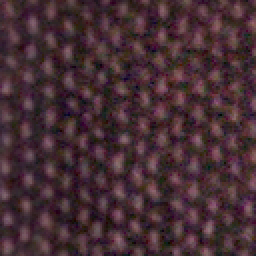} \\[0.5pt]
\vcrow{residual $\times$2} & \includegraphics[width=\vccell]{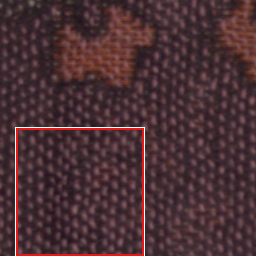} & \includegraphics[width=\vccell]{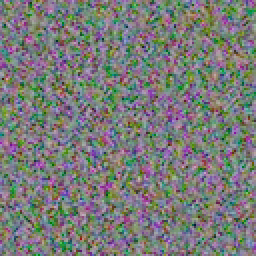} & \includegraphics[width=\vccell]{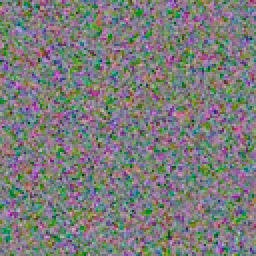} & \includegraphics[width=\vccell]{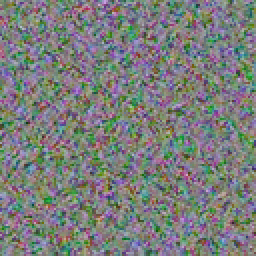} & \includegraphics[width=\vccell]{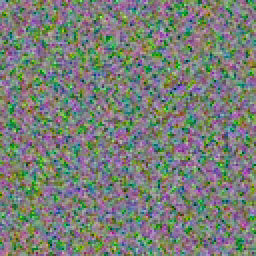} & \includegraphics[width=\vccell]{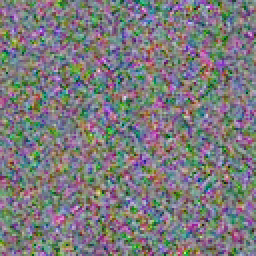} & \includegraphics[width=\vccell]{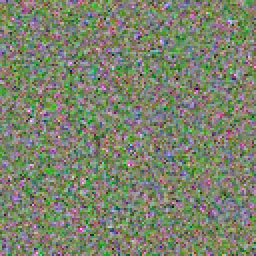} & \includegraphics[width=\vccell]{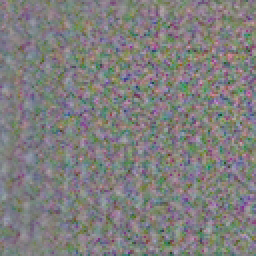} \\[3pt]
\vcrow{GP, ISO\,1600} & \includegraphics[width=\vccell]{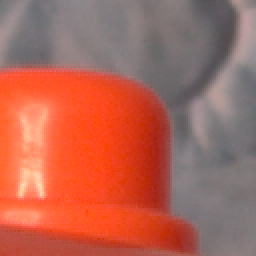} & \includegraphics[width=\vccell]{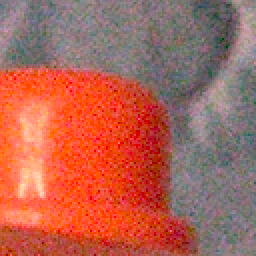} & \includegraphics[width=\vccell]{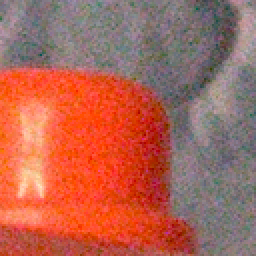} & \includegraphics[width=\vccell]{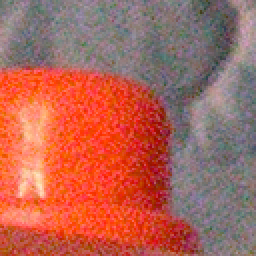} & \vcna & \includegraphics[width=\vccell]{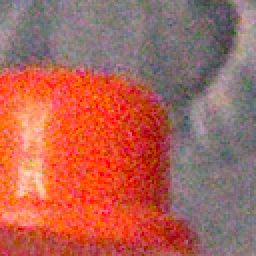} & \vcna & \includegraphics[width=\vccell]{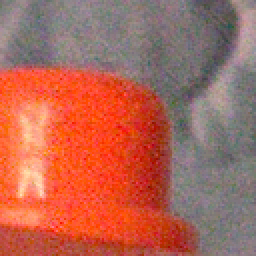} \\[0.5pt]
\vcrow{residual $\times$1.5} & \includegraphics[width=\vccell]{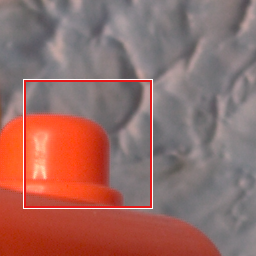} & \includegraphics[width=\vccell]{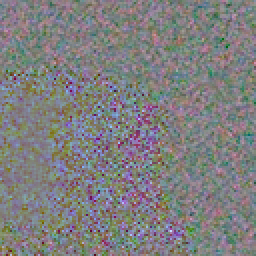} & \includegraphics[width=\vccell]{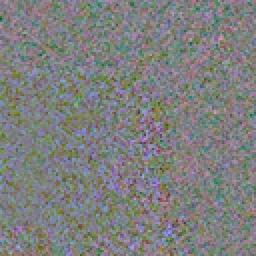} & \includegraphics[width=\vccell]{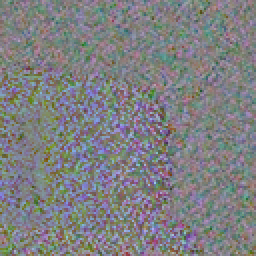} & \vcna & \includegraphics[width=\vccell]{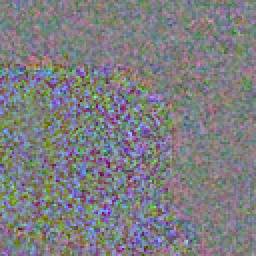} & \vcna & \includegraphics[width=\vccell]{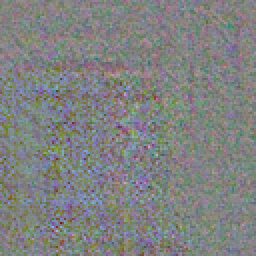} \\[3pt]
\vcrow{GP, ISO\,3200} & \includegraphics[width=\vccell]{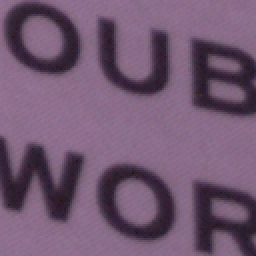} & \includegraphics[width=\vccell]{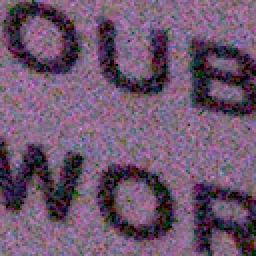} & \includegraphics[width=\vccell]{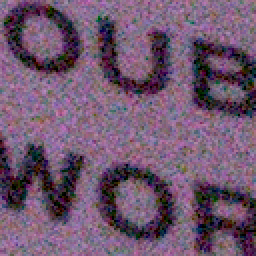} & \includegraphics[width=\vccell]{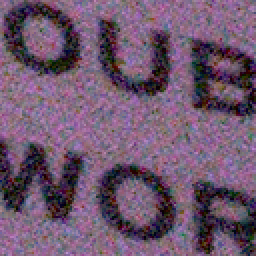} & \vcna & \includegraphics[width=\vccell]{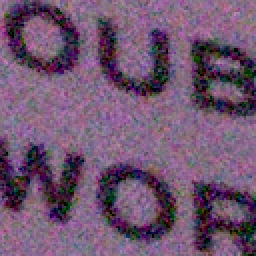} & \vcna & \includegraphics[width=\vccell]{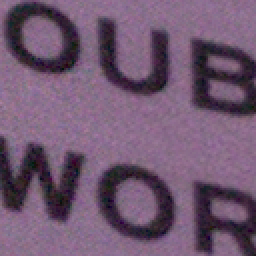} \\[0.5pt]
\vcrow{residual $\times$1.5} & \includegraphics[width=\vccell]{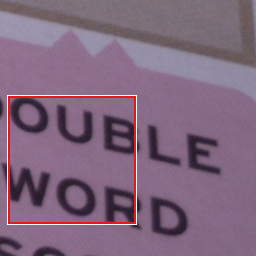} & \includegraphics[width=\vccell]{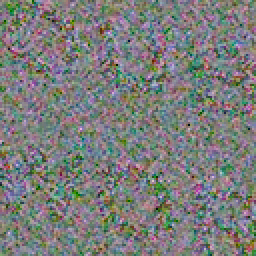} & \includegraphics[width=\vccell]{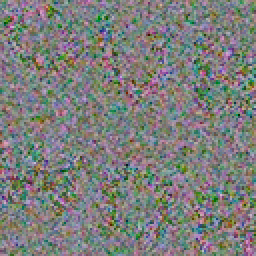} & \includegraphics[width=\vccell]{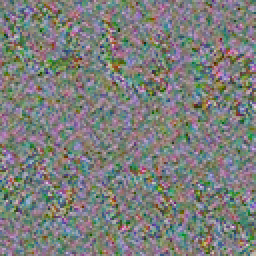} & \vcna & \includegraphics[width=\vccell]{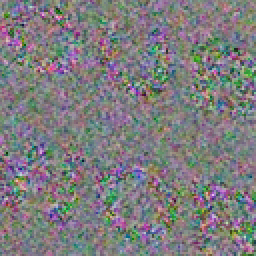} & \vcna & \includegraphics[width=\vccell]{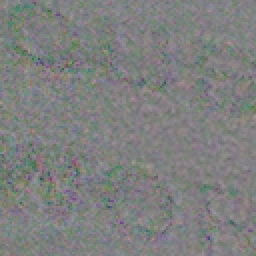} \\
\end{tabular}
\caption{\textbf{SIDD comparison on highly textured regions, every evaluated camera--ISO class (1/4).} In each class, the $128\times128$ window with the highest median $7\times7$ local standard deviation of the clean image (maximum over RGB, the contrast of our flatness mask) was taken over every block of the class's evaluated sources on an 8-pixel grid, excluding windows with more than 25\% near-saturated clean pixels; it was fixed from the clean images before any output was viewed. Every method shows its first generator run. sRGB-Flow and NM-FlowGAN are shown only for the 19-source block covered by their checkpoints (n/a otherwise). Second row of each pair: residual relative to the clean image, amplified by the factor at left (set from the real residual alone; gray is zero); under the clean column, the full $256\times256$ block with the crop window marked. Crops are shown at native resolution; zoom in for detail. }
\label{fig:supp-sidd-textured-1}
\end{figure*}

\begin{figure*}[tp]
\centering
\vcsetup{8}%
\begin{tabular}{@{}c*{8}{c}@{}}
  & \vchead{Clean} & \vchead{Real} & \vchead{\textbf{PGNM}} & \vchead{GMDCN} & \vchead{NM-FlowGAN} & \vchead{NAFlow} & \vchead{sRGB-Flow} & \vchead{DANet} \\[2pt]
\vcrow{GP, ISO\,6400$^\star$} & \includegraphics[width=\vccell]{supp/visual_comparison/sidd-textured/GP-6400/clean.png} & \includegraphics[width=\vccell]{supp/visual_comparison/sidd-textured/GP-6400/real.png} & \includegraphics[width=\vccell]{supp/visual_comparison/sidd-textured/GP-6400/pgnm.png} & \includegraphics[width=\vccell]{supp/visual_comparison/sidd-textured/GP-6400/cfg_nin.png} & \vcna & \includegraphics[width=\vccell]{supp/visual_comparison/sidd-textured/GP-6400/naflow.png} & \vcna & \includegraphics[width=\vccell]{supp/visual_comparison/sidd-textured/GP-6400/danet.png} \\[0.5pt]
\vcrow{residual $\times$1} & \includegraphics[width=\vccell]{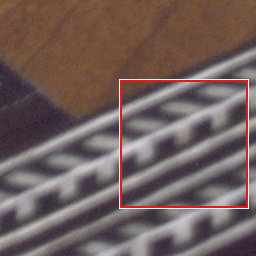} & \includegraphics[width=\vccell]{supp/visual_comparison/sidd-textured/GP-6400/real-res.png} & \includegraphics[width=\vccell]{supp/visual_comparison/sidd-textured/GP-6400/pgnm-res.png} & \includegraphics[width=\vccell]{supp/visual_comparison/sidd-textured/GP-6400/cfg_nin-res.png} & \vcna & \includegraphics[width=\vccell]{supp/visual_comparison/sidd-textured/GP-6400/naflow-res.png} & \vcna & \includegraphics[width=\vccell]{supp/visual_comparison/sidd-textured/GP-6400/danet-res.png} \\[3pt]
\vcrow{GP, ISO\,10000} & \includegraphics[width=\vccell]{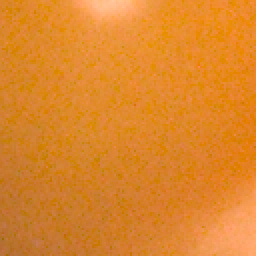} & \includegraphics[width=\vccell]{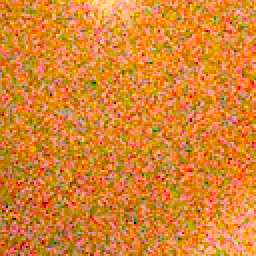} & \includegraphics[width=\vccell]{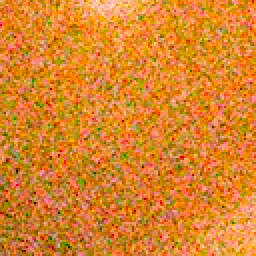} & \includegraphics[width=\vccell]{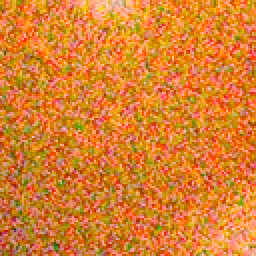} & \vcna & \includegraphics[width=\vccell]{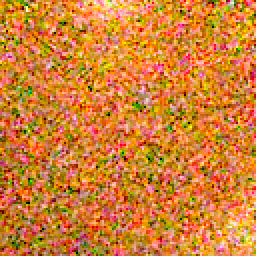} & \vcna & \includegraphics[width=\vccell]{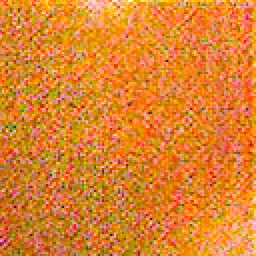} \\[0.5pt]
\vcrow{residual $\times$1} & \includegraphics[width=\vccell]{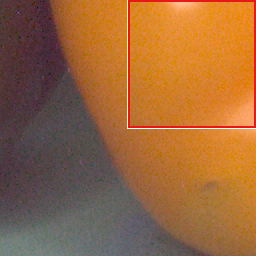} & \includegraphics[width=\vccell]{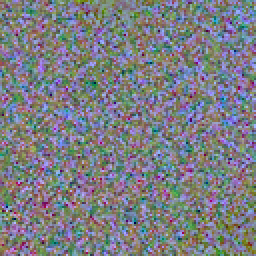} & \includegraphics[width=\vccell]{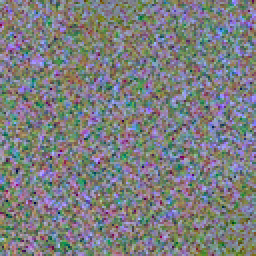} & \includegraphics[width=\vccell]{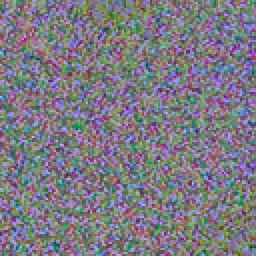} & \vcna & \includegraphics[width=\vccell]{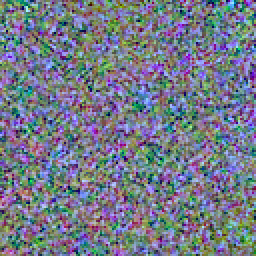} & \vcna & \includegraphics[width=\vccell]{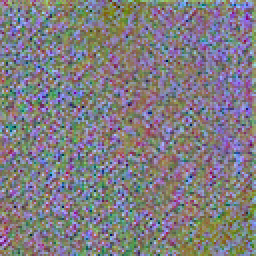} \\[3pt]
\vcrow{IP, ISO\,1000} & \includegraphics[width=\vccell]{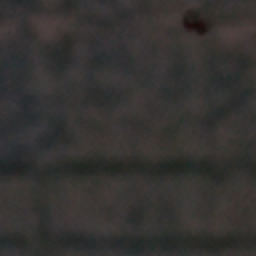} & \includegraphics[width=\vccell]{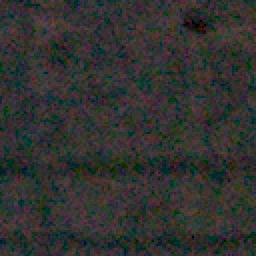} & \includegraphics[width=\vccell]{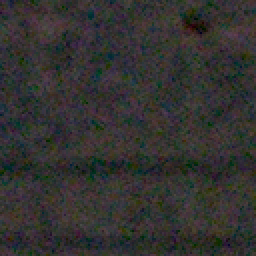} & \includegraphics[width=\vccell]{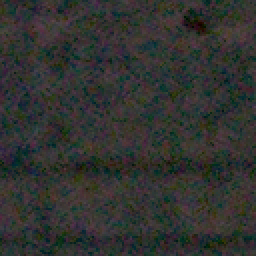} & \vcna & \includegraphics[width=\vccell]{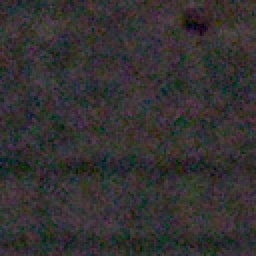} & \vcna & \includegraphics[width=\vccell]{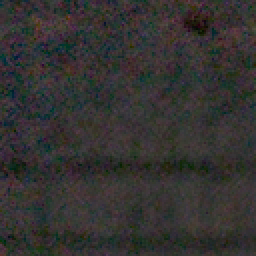} \\[0.5pt]
\vcrow{residual $\times$2} & \includegraphics[width=\vccell]{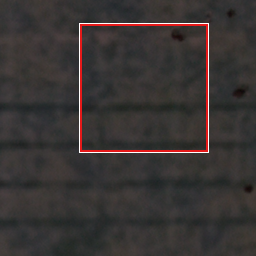} & \includegraphics[width=\vccell]{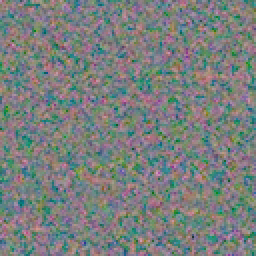} & \includegraphics[width=\vccell]{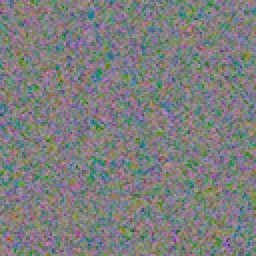} & \includegraphics[width=\vccell]{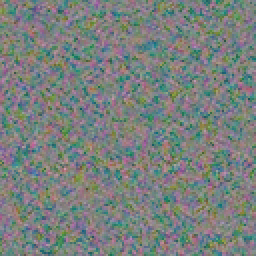} & \vcna & \includegraphics[width=\vccell]{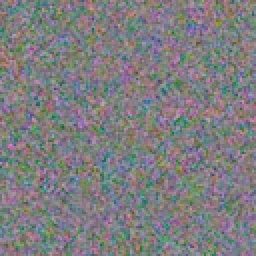} & \vcna & \includegraphics[width=\vccell]{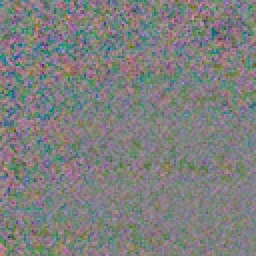} \\[3pt]
\vcrow{IP, ISO\,1600} & \includegraphics[width=\vccell]{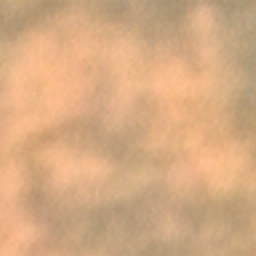} & \includegraphics[width=\vccell]{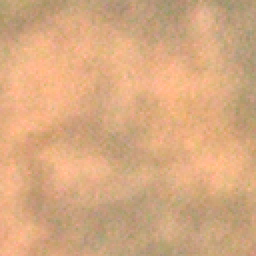} & \includegraphics[width=\vccell]{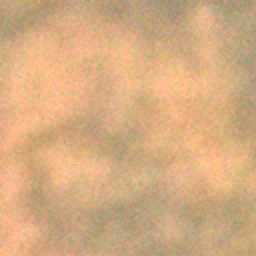} & \includegraphics[width=\vccell]{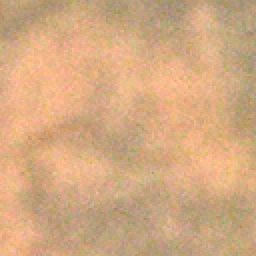} & \includegraphics[width=\vccell]{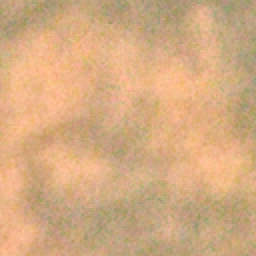} & \includegraphics[width=\vccell]{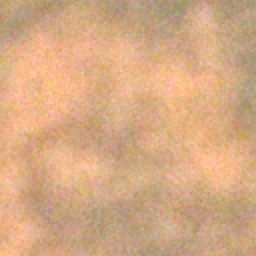} & \includegraphics[width=\vccell]{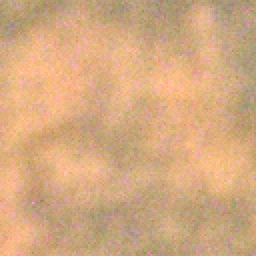} & \includegraphics[width=\vccell]{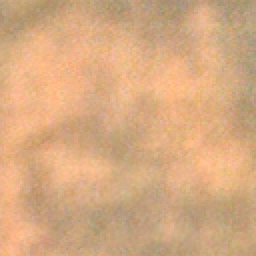} \\[0.5pt]
\vcrow{residual $\times$6.5} & \includegraphics[width=\vccell]{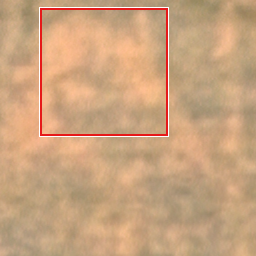} & \includegraphics[width=\vccell]{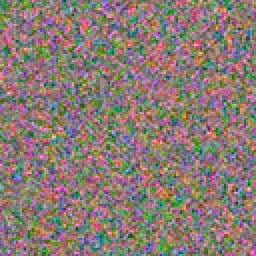} & \includegraphics[width=\vccell]{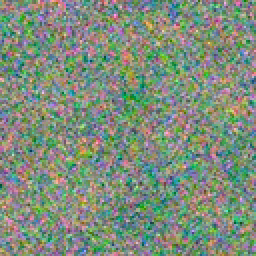} & \includegraphics[width=\vccell]{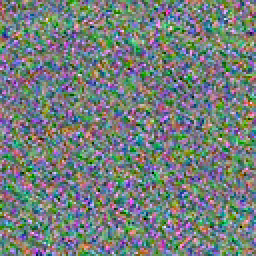} & \includegraphics[width=\vccell]{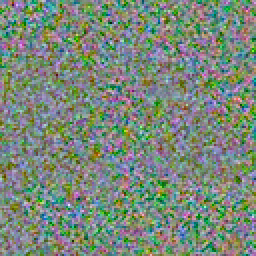} & \includegraphics[width=\vccell]{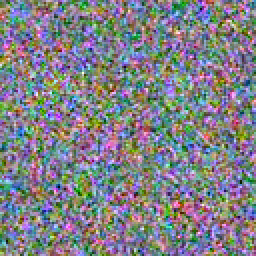} & \includegraphics[width=\vccell]{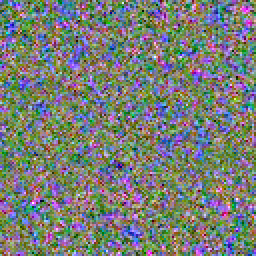} & \includegraphics[width=\vccell]{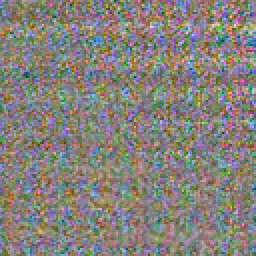} \\
\end{tabular}
\caption{\textbf{SIDD comparison on highly textured regions, every evaluated camera--ISO class (2/4).} In each class, the $128\times128$ window with the highest median $7\times7$ local standard deviation of the clean image (maximum over RGB, the contrast of our flatness mask) was taken over every block of the class's evaluated sources on an 8-pixel grid, excluding windows with more than 25\% near-saturated clean pixels; it was fixed from the clean images before any output was viewed. Every method shows its first generator run. sRGB-Flow and NM-FlowGAN are shown only for the 19-source block covered by their checkpoints (n/a otherwise). Second row of each pair: residual relative to the clean image, amplified by the factor at left (set from the real residual alone; gray is zero); under the clean column, the full $256\times256$ block with the crop window marked. Crops are shown at native resolution; zoom in for detail. $^\star$Also shown in the main paper (Fig.~\mainref{fig:main_sidd_vis}). }
\label{fig:supp-sidd-textured-2}
\end{figure*}

\begin{figure*}[tp]
\centering
\vcsetup{8}%
\begin{tabular}{@{}c*{8}{c}@{}}
  & \vchead{Clean} & \vchead{Real} & \vchead{\textbf{PGNM}} & \vchead{GMDCN} & \vchead{NM-FlowGAN} & \vchead{NAFlow} & \vchead{sRGB-Flow} & \vchead{DANet} \\[2pt]
\vcrow{IP, ISO\,2000} & \includegraphics[width=\vccell]{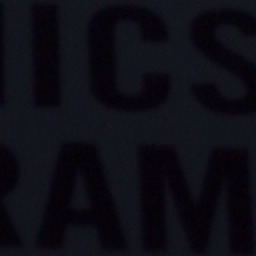} & \includegraphics[width=\vccell]{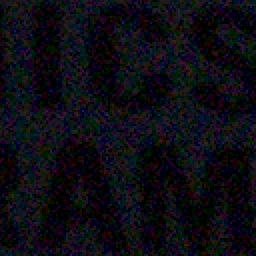} & \includegraphics[width=\vccell]{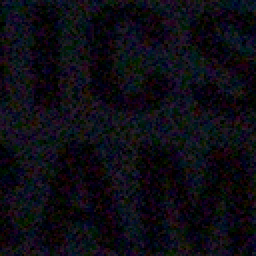} & \includegraphics[width=\vccell]{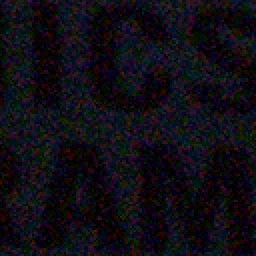} & \vcna & \includegraphics[width=\vccell]{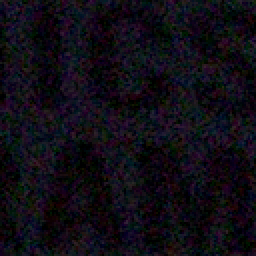} & \vcna & \includegraphics[width=\vccell]{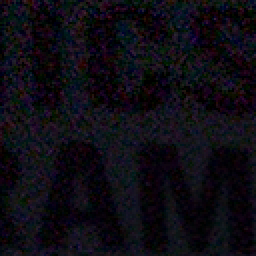} \\[0.5pt]
\vcrow{residual $\times$3.5} & \includegraphics[width=\vccell]{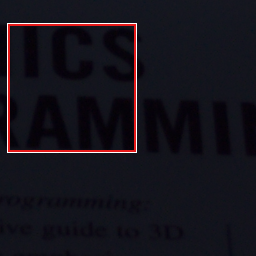} & \includegraphics[width=\vccell]{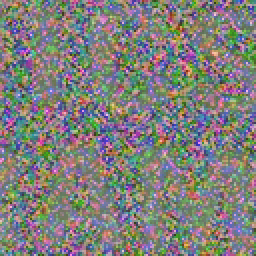} & \includegraphics[width=\vccell]{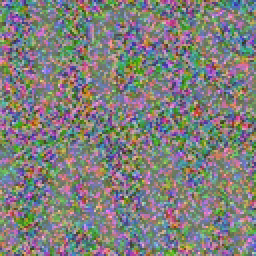} & \includegraphics[width=\vccell]{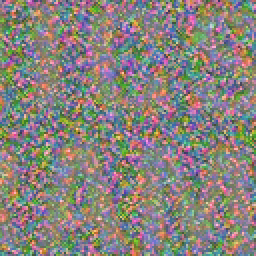} & \vcna & \includegraphics[width=\vccell]{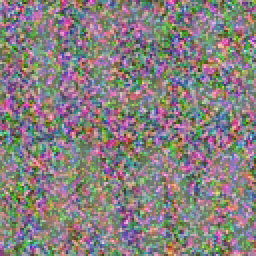} & \vcna & \includegraphics[width=\vccell]{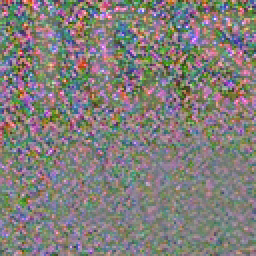} \\[3pt]
\vcrow{N6, ISO\,400} & \includegraphics[width=\vccell]{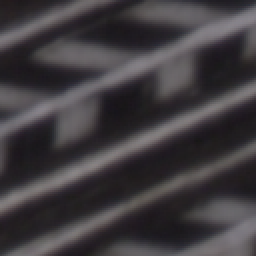} & \includegraphics[width=\vccell]{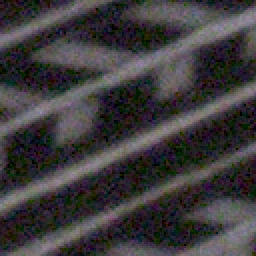} & \includegraphics[width=\vccell]{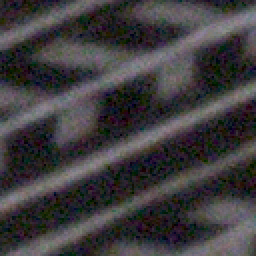} & \includegraphics[width=\vccell]{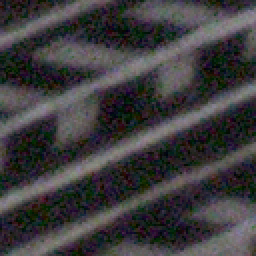} & \vcna & \includegraphics[width=\vccell]{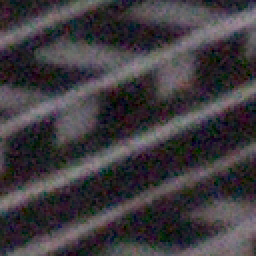} & \vcna & \includegraphics[width=\vccell]{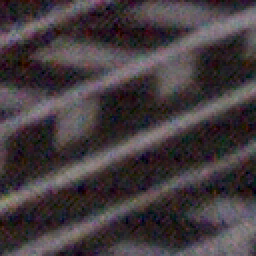} \\[0.5pt]
\vcrow{residual $\times$2.5} & \includegraphics[width=\vccell]{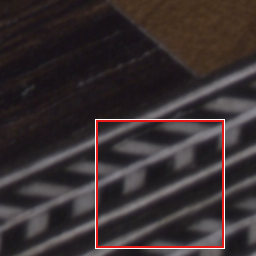} & \includegraphics[width=\vccell]{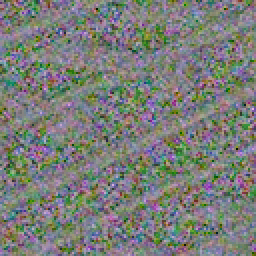} & \includegraphics[width=\vccell]{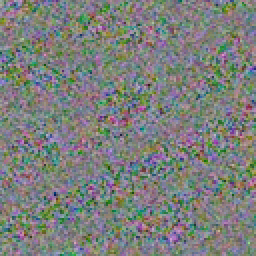} & \includegraphics[width=\vccell]{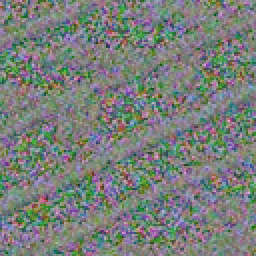} & \vcna & \includegraphics[width=\vccell]{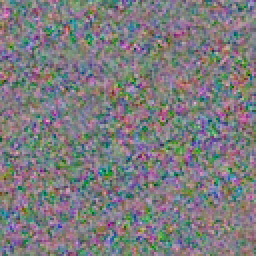} & \vcna & \includegraphics[width=\vccell]{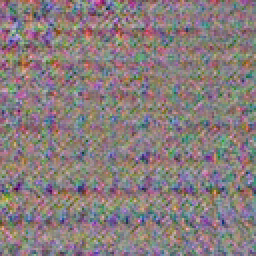} \\[3pt]
\vcrow{N6, ISO\,800} & \includegraphics[width=\vccell]{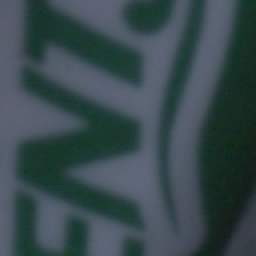} & \includegraphics[width=\vccell]{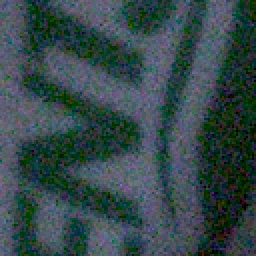} & \includegraphics[width=\vccell]{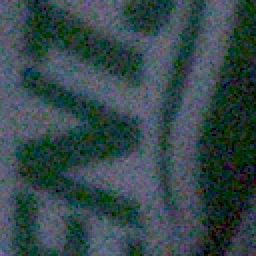} & \includegraphics[width=\vccell]{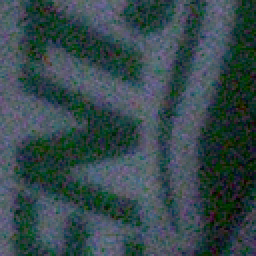} & \includegraphics[width=\vccell]{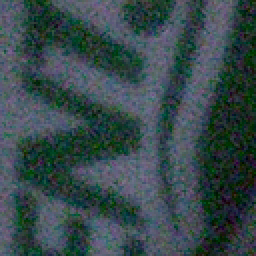} & \includegraphics[width=\vccell]{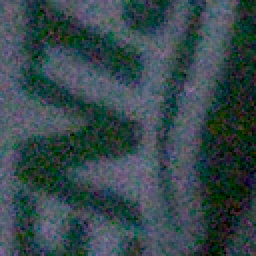} & \includegraphics[width=\vccell]{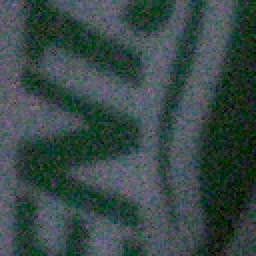} & \includegraphics[width=\vccell]{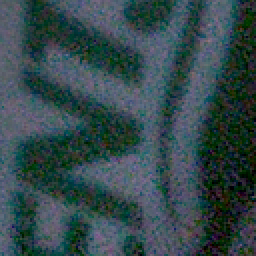} \\[0.5pt]
\vcrow{residual $\times$2} & \includegraphics[width=\vccell]{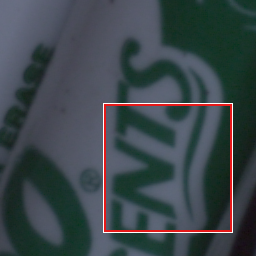} & \includegraphics[width=\vccell]{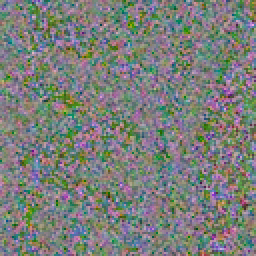} & \includegraphics[width=\vccell]{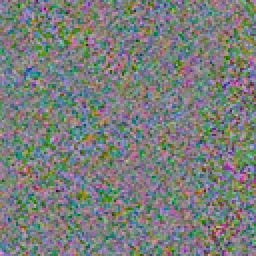} & \includegraphics[width=\vccell]{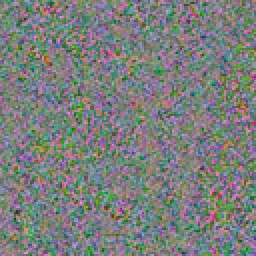} & \includegraphics[width=\vccell]{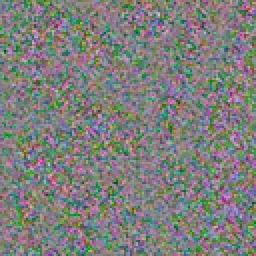} & \includegraphics[width=\vccell]{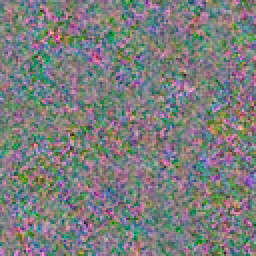} & \includegraphics[width=\vccell]{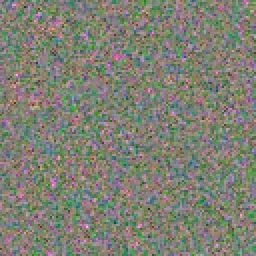} & \includegraphics[width=\vccell]{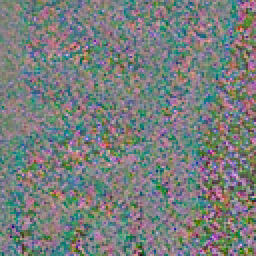} \\[3pt]
\vcrow{N6, ISO\,3200} & \includegraphics[width=\vccell]{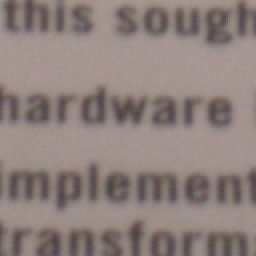} & \includegraphics[width=\vccell]{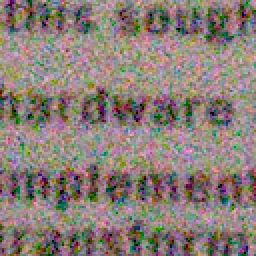} & \includegraphics[width=\vccell]{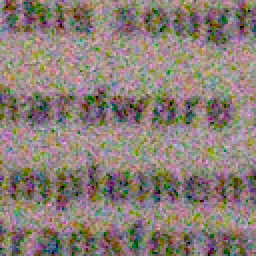} & \includegraphics[width=\vccell]{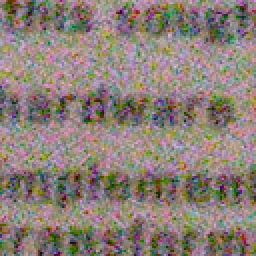} & \vcna & \includegraphics[width=\vccell]{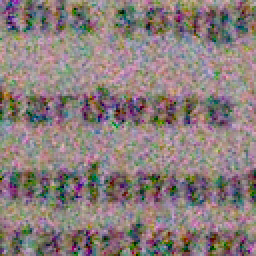} & \vcna & \includegraphics[width=\vccell]{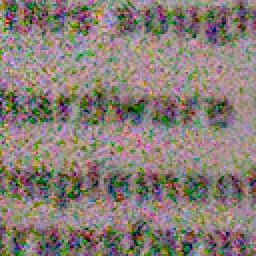} \\[0.5pt]
\vcrow{residual $\times$1} & \includegraphics[width=\vccell]{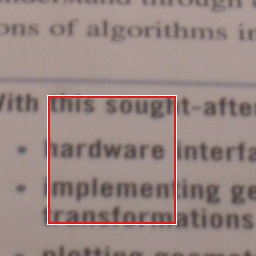} & \includegraphics[width=\vccell]{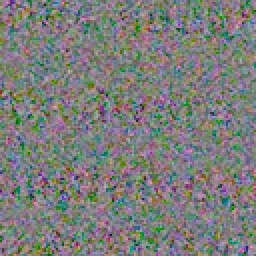} & \includegraphics[width=\vccell]{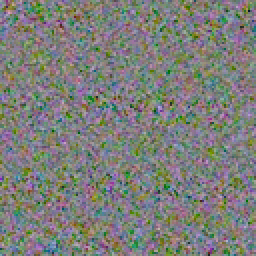} & \includegraphics[width=\vccell]{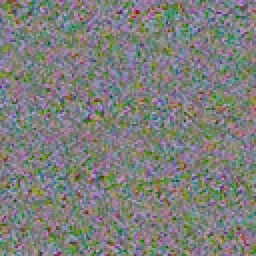} & \vcna & \includegraphics[width=\vccell]{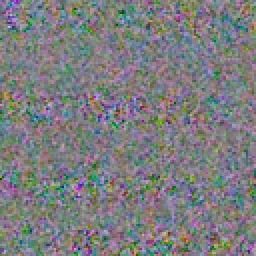} & \vcna & \includegraphics[width=\vccell]{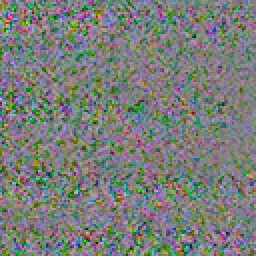} \\
\end{tabular}
\caption{\textbf{SIDD comparison on highly textured regions, every evaluated camera--ISO class (3/4).} In each class, the $128\times128$ window with the highest median $7\times7$ local standard deviation of the clean image (maximum over RGB, the contrast of our flatness mask) was taken over every block of the class's evaluated sources on an 8-pixel grid, excluding windows with more than 25\% near-saturated clean pixels; it was fixed from the clean images before any output was viewed. Every method shows its first generator run. sRGB-Flow and NM-FlowGAN are shown only for the 19-source block covered by their checkpoints (n/a otherwise). Second row of each pair: residual relative to the clean image, amplified by the factor at left (set from the real residual alone; gray is zero); under the clean column, the full $256\times256$ block with the crop window marked. Crops are shown at native resolution; zoom in for detail. }
\label{fig:supp-sidd-textured-3}
\end{figure*}

\begin{figure*}[tp]
\centering
\vcsetup{8}%
\begin{tabular}{@{}c*{8}{c}@{}}
  & \vchead{Clean} & \vchead{Real} & \vchead{\textbf{PGNM}} & \vchead{GMDCN} & \vchead{NM-FlowGAN} & \vchead{NAFlow} & \vchead{sRGB-Flow} & \vchead{DANet} \\[2pt]
\vcrow{S6, ISO\,400} & \includegraphics[width=\vccell]{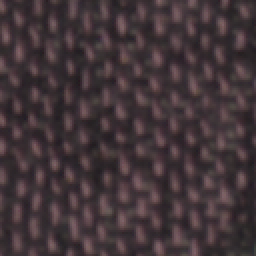} & \includegraphics[width=\vccell]{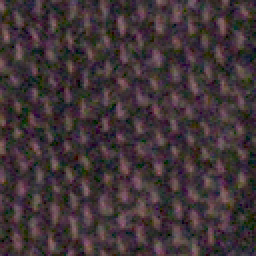} & \includegraphics[width=\vccell]{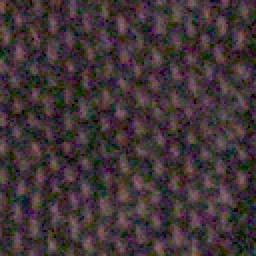} & \includegraphics[width=\vccell]{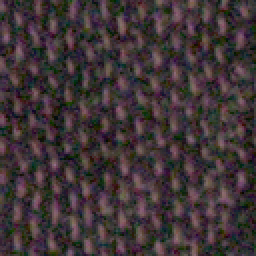} & \includegraphics[width=\vccell]{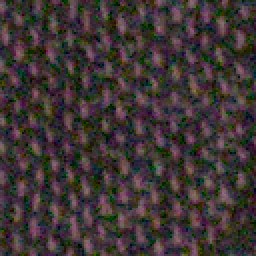} & \includegraphics[width=\vccell]{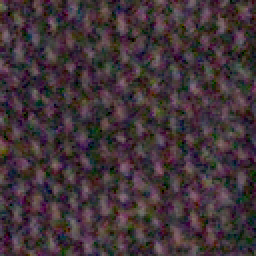} & \includegraphics[width=\vccell]{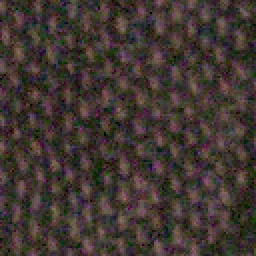} & \includegraphics[width=\vccell]{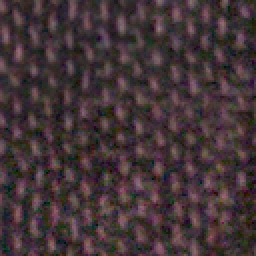} \\[0.5pt]
\vcrow{residual $\times$2.5} & \includegraphics[width=\vccell]{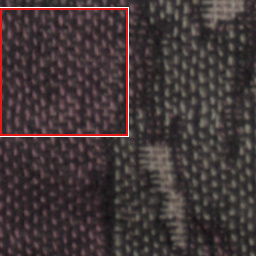} & \includegraphics[width=\vccell]{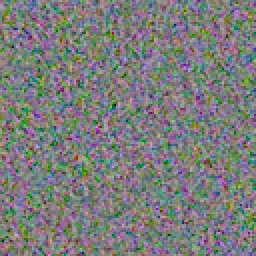} & \includegraphics[width=\vccell]{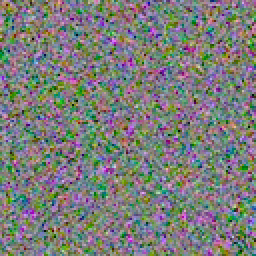} & \includegraphics[width=\vccell]{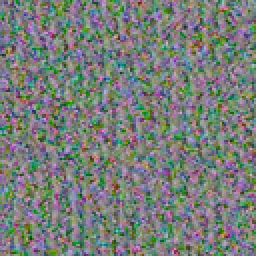} & \includegraphics[width=\vccell]{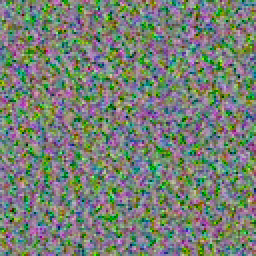} & \includegraphics[width=\vccell]{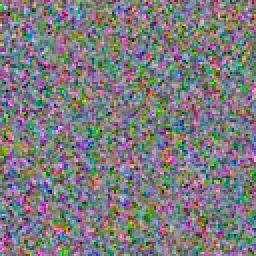} & \includegraphics[width=\vccell]{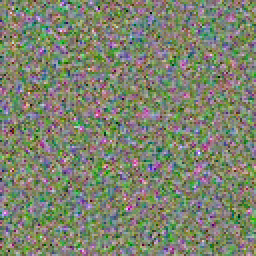} & \includegraphics[width=\vccell]{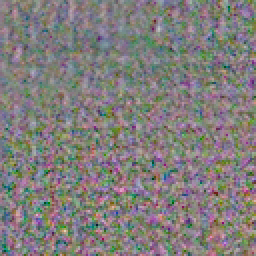} \\[3pt]
\vcrow{S6, ISO\,800} & \includegraphics[width=\vccell]{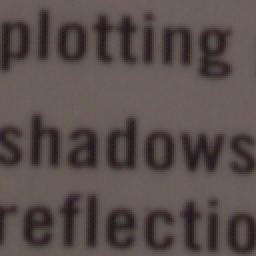} & \includegraphics[width=\vccell]{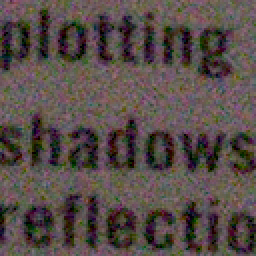} & \includegraphics[width=\vccell]{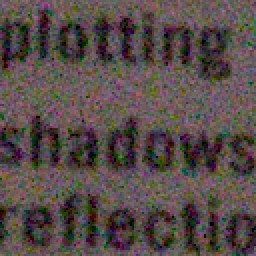} & \includegraphics[width=\vccell]{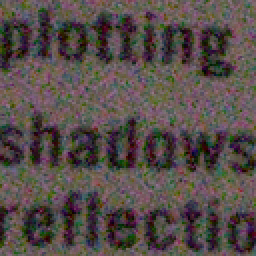} & \includegraphics[width=\vccell]{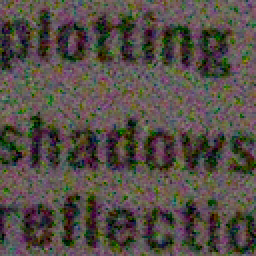} & \includegraphics[width=\vccell]{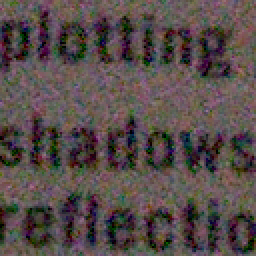} & \includegraphics[width=\vccell]{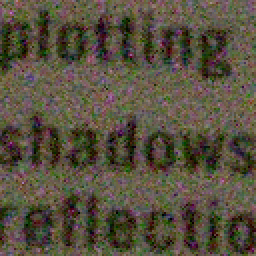} & \includegraphics[width=\vccell]{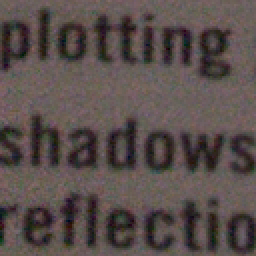} \\[0.5pt]
\vcrow{residual $\times$2} & \includegraphics[width=\vccell]{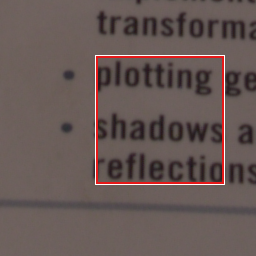} & \includegraphics[width=\vccell]{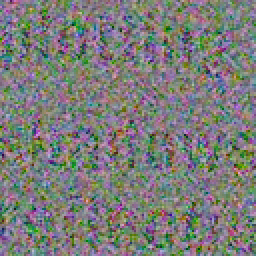} & \includegraphics[width=\vccell]{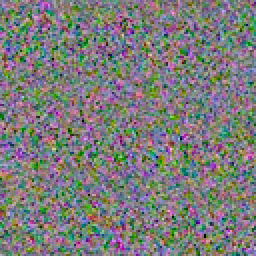} & \includegraphics[width=\vccell]{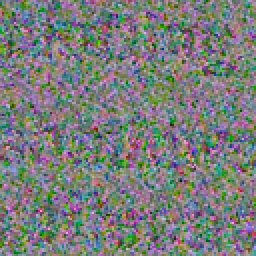} & \includegraphics[width=\vccell]{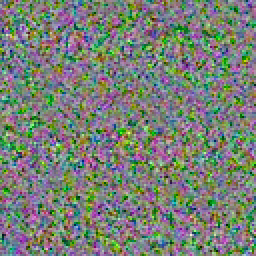} & \includegraphics[width=\vccell]{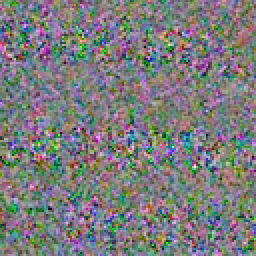} & \includegraphics[width=\vccell]{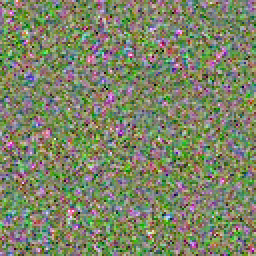} & \includegraphics[width=\vccell]{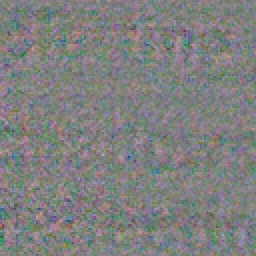} \\[3pt]
\vcrow{S6, ISO\,1600} & \includegraphics[width=\vccell]{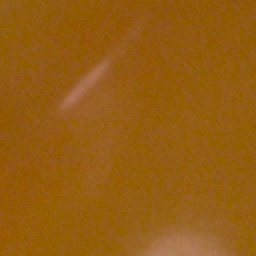} & \includegraphics[width=\vccell]{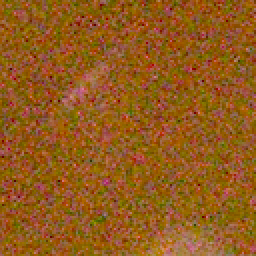} & \includegraphics[width=\vccell]{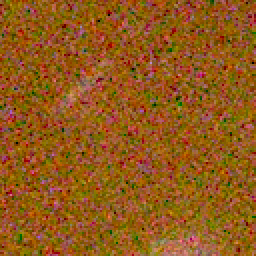} & \includegraphics[width=\vccell]{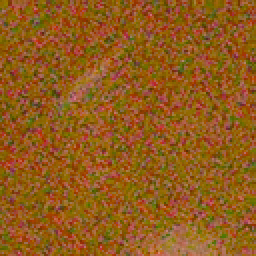} & \includegraphics[width=\vccell]{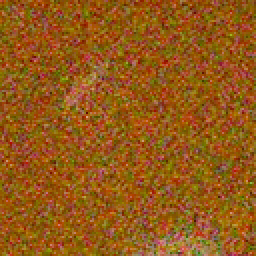} & \includegraphics[width=\vccell]{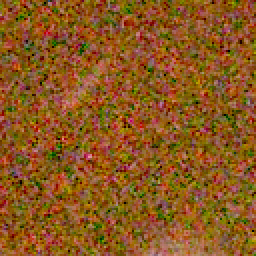} & \includegraphics[width=\vccell]{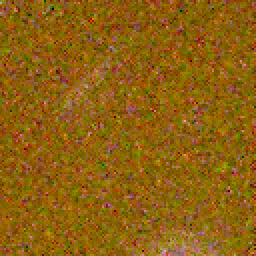} & \includegraphics[width=\vccell]{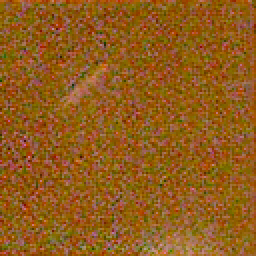} \\[0.5pt]
\vcrow{residual $\times$1} & \includegraphics[width=\vccell]{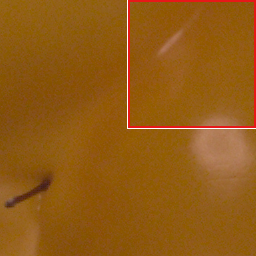} & \includegraphics[width=\vccell]{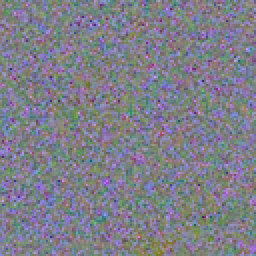} & \includegraphics[width=\vccell]{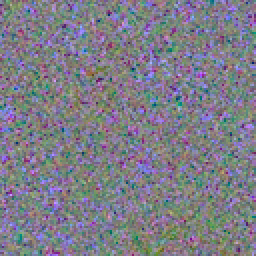} & \includegraphics[width=\vccell]{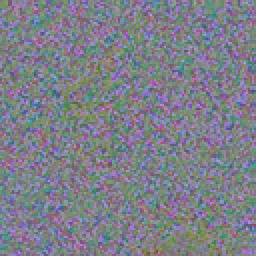} & \includegraphics[width=\vccell]{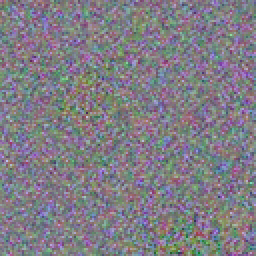} & \includegraphics[width=\vccell]{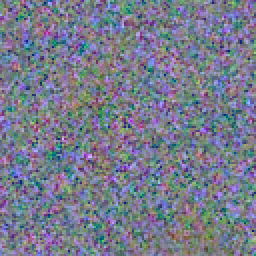} & \includegraphics[width=\vccell]{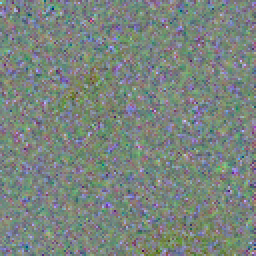} & \includegraphics[width=\vccell]{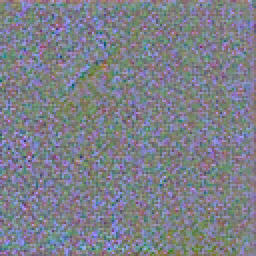} \\[3pt]
\vcrow{S6, ISO\,3200} & \includegraphics[width=\vccell]{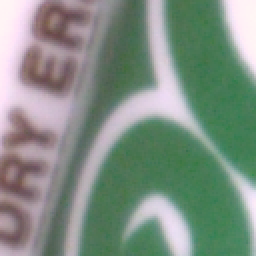} & \includegraphics[width=\vccell]{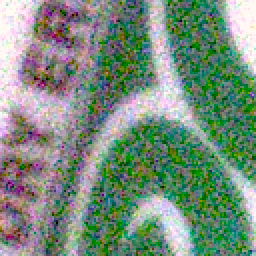} & \includegraphics[width=\vccell]{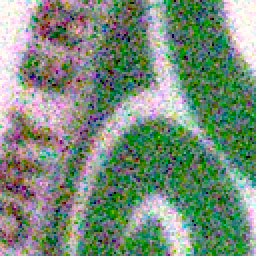} & \includegraphics[width=\vccell]{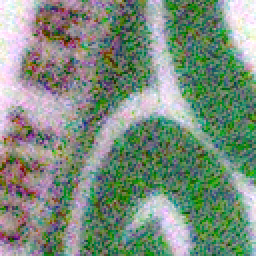} & \includegraphics[width=\vccell]{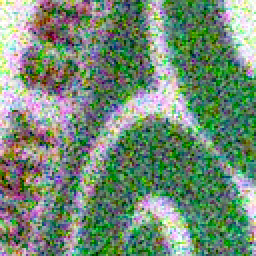} & \includegraphics[width=\vccell]{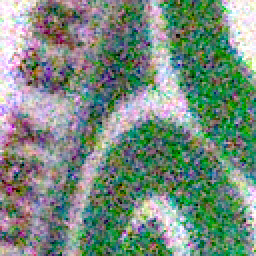} & \includegraphics[width=\vccell]{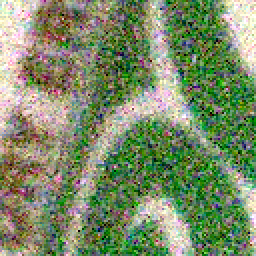} & \includegraphics[width=\vccell]{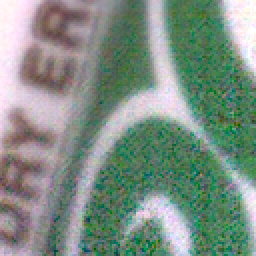} \\[0.5pt]
\vcrow{residual $\times$1} & \includegraphics[width=\vccell]{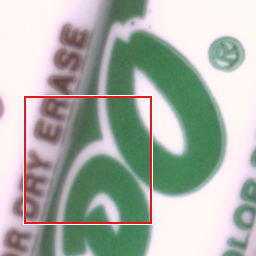} & \includegraphics[width=\vccell]{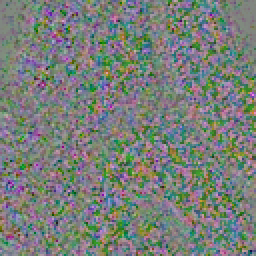} & \includegraphics[width=\vccell]{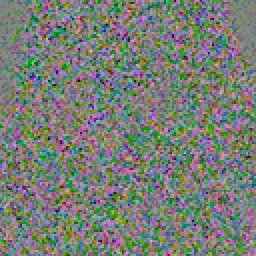} & \includegraphics[width=\vccell]{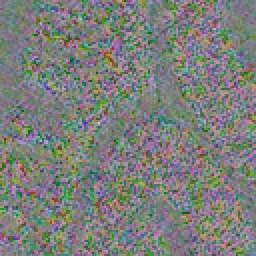} & \includegraphics[width=\vccell]{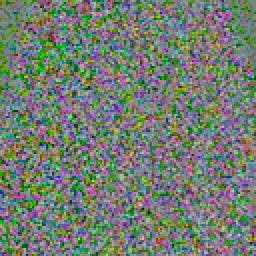} & \includegraphics[width=\vccell]{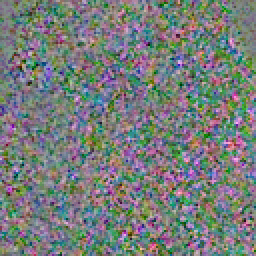} & \includegraphics[width=\vccell]{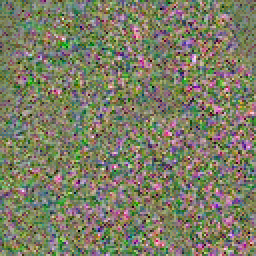} & \includegraphics[width=\vccell]{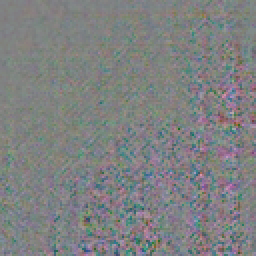} \\
\end{tabular}
\caption{\textbf{SIDD comparison on highly textured regions, every evaluated camera--ISO class (4/4).} In each class, the $128\times128$ window with the highest median $7\times7$ local standard deviation of the clean image (maximum over RGB, the contrast of our flatness mask) was taken over every block of the class's evaluated sources on an 8-pixel grid, excluding windows with more than 25\% near-saturated clean pixels; it was fixed from the clean images before any output was viewed. Every method shows its first generator run. sRGB-Flow and NM-FlowGAN are shown only for the 19-source block covered by their checkpoints (n/a otherwise). Second row of each pair: residual relative to the clean image, amplified by the factor at left (set from the real residual alone; gray is zero); under the clean column, the full $256\times256$ block with the crop window marked. Crops are shown at native resolution; zoom in for detail. }
\label{fig:supp-sidd-textured-4}
\end{figure*}

\paragraph{Highly textured regions.}
PGNM uses the clean image only through pixel intensity, whereas the neural
generators also see the surrounding image structure. For every class,
Figs.~\ref{fig:supp-sidd-textured-1}--\ref{fig:supp-sidd-textured-4} therefore
show the window with the most local contrast in the clean image, selected by a
rule fixed from the clean images alone (Tab.~\ref{tab:supp-crops-textured}). The
GP ISO\,6400 crop of the main paper is one of them.

\begin{figure*}[tp]
    \centering
    \scriptsize
    \setlength{\tabcolsep}{0.35pt}
    \renewcommand{\arraystretch}{0.5}
    \begin{tabular}{cccccccc}
        Clean & Real & \textbf{PGNM} & GMDCN & NM-FlowGAN & NAFlow & sRGB-Flow & DANet \\
        \includegraphics[width=0.122\textwidth]{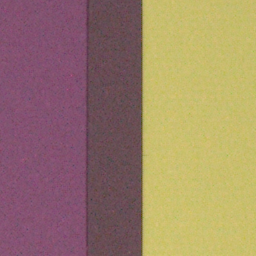} &
        \includegraphics[width=0.122\textwidth]{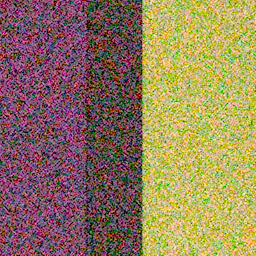} &
        \includegraphics[width=0.122\textwidth]{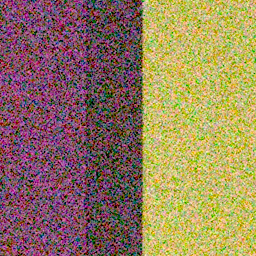} &
        \includegraphics[width=0.122\textwidth]{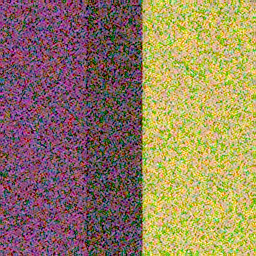} &
        \includegraphics[width=0.122\textwidth]{figures/na.pdf} &
        \includegraphics[width=0.122\textwidth]{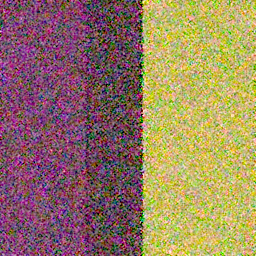} &
        \includegraphics[width=0.122\textwidth]{figures/na.pdf} &
        \includegraphics[width=0.122\textwidth]{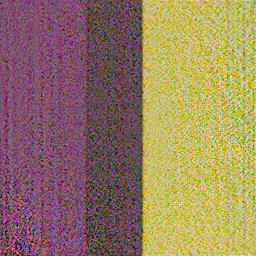} \\
        \raisebox{30pt}{GP 10000} &
        \includegraphics[width=0.122\textwidth]{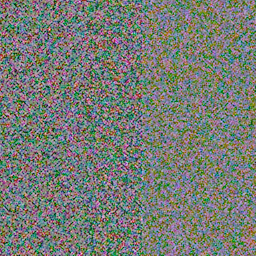} &
        \includegraphics[width=0.122\textwidth]{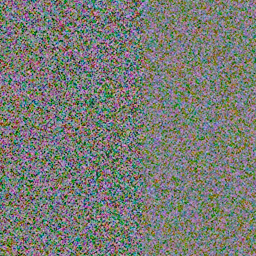} &
        \includegraphics[width=0.122\textwidth]{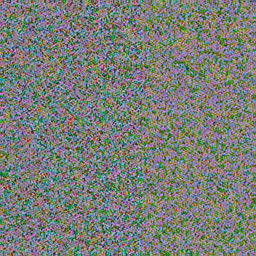} &
        \includegraphics[width=0.122\textwidth]{figures/na.pdf} &
        \includegraphics[width=0.122\textwidth]{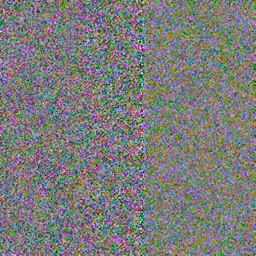} &
        \includegraphics[width=0.122\textwidth]{figures/na.pdf} &
        \includegraphics[width=0.122\textwidth]{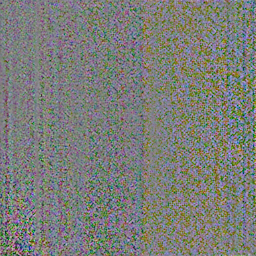} \\
         &  & KL 0.027 & KL 0.039 & -- & KL 0.051 & -- & KL 0.240 \\[3pt]
        \includegraphics[width=0.122\textwidth]{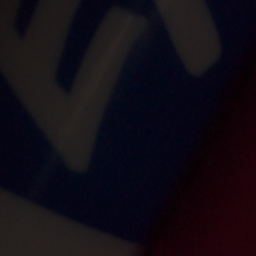} &
        \includegraphics[width=0.122\textwidth]{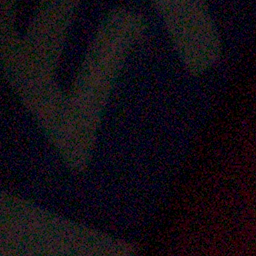} &
        \includegraphics[width=0.122\textwidth]{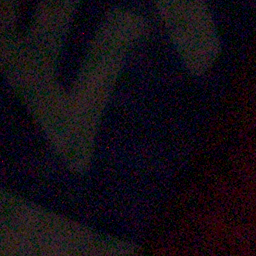} &
        \includegraphics[width=0.122\textwidth]{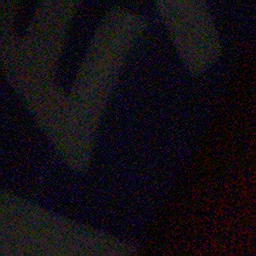} &
        \includegraphics[width=0.122\textwidth]{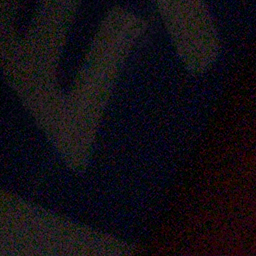} &
        \includegraphics[width=0.122\textwidth]{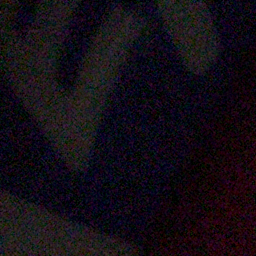} &
        \includegraphics[width=0.122\textwidth]{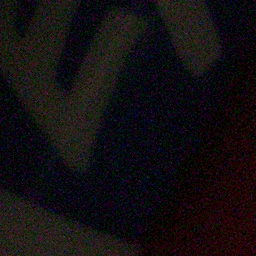} &
        \includegraphics[width=0.122\textwidth]{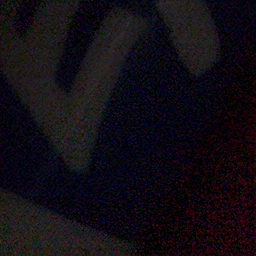} \\
        \raisebox{30pt}{IP 1600} &
        \includegraphics[width=0.122\textwidth]{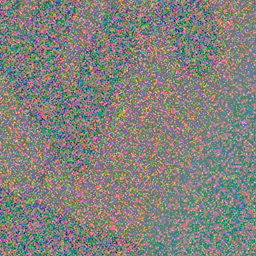} &
        \includegraphics[width=0.122\textwidth]{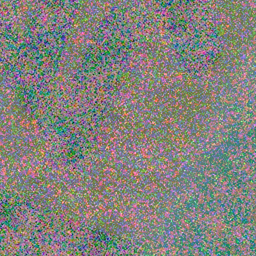} &
        \includegraphics[width=0.122\textwidth]{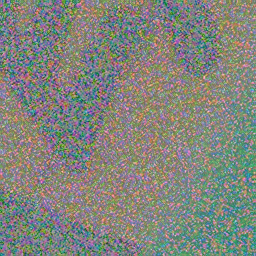} &
        \includegraphics[width=0.122\textwidth]{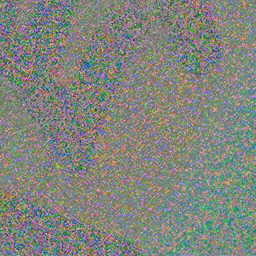} &
        \includegraphics[width=0.122\textwidth]{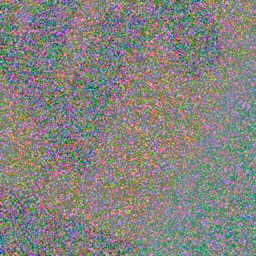} &
        \includegraphics[width=0.122\textwidth]{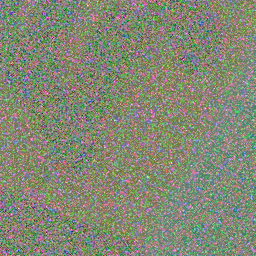} &
        \includegraphics[width=0.122\textwidth]{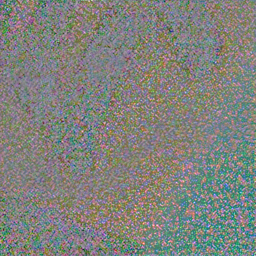} \\
         &  & KL 0.017 & KL 0.048 & KL 0.036 & KL 0.025 & KL 0.026 & KL 0.113 \\[3pt]
    \end{tabular}
    \caption{\textbf{The two SIDD Validation blocks shown by NAFlow}~\cite{kim2024naflow} (its Fig.~4; complete $256\times256$ blocks), rendered by every generator with its fit or checkpoint (first generator run, as scored). Second row of each pair: residual relative to the clean image, amplified by one gain per block set from the real residual (gray is zero). KL: small-scale KL of the displayed draw, averaged over the block's 64 $32\times32$ tiles. n/a: class not covered by the checkpoint.}
    \label{fig:naflow_blocks}
\end{figure*}

\paragraph{The blocks shown by NAFlow.}
NAFlow~\cite{kim2024naflow} illustrates its results with two complete
$256\times256$ Validation blocks. Matching their real noisy panels against all
1,280 Validation blocks identifies them as image~1, block~0 (GP ISO\,10000) and
image~6, block~4 (IP ISO\,1600). Fig.~\ref{fig:naflow_blocks} shows every
generator on these two blocks, which were selected by NAFlow's authors rather than
by us. On both, PGNM has the lowest small-scale KL, for the displayed draw and for
the mean over the five generator runs.

\begin{figure*}[tp]
    \centering
    \scriptsize
    \setlength{\tabcolsep}{0.5pt}
    \renewcommand{\arraystretch}{0.5}
    \begin{tabular}{ccccccc}
        Clean & \textbf{PGNM} & GMDCN & NM-FlowGAN & NAFlow & sRGB-Flow & DANet \\
        \includegraphics[width=0.14\textwidth]{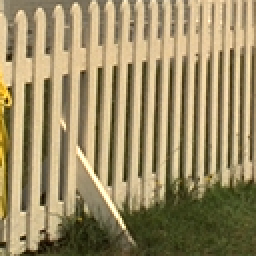} &
        \includegraphics[width=0.14\textwidth]{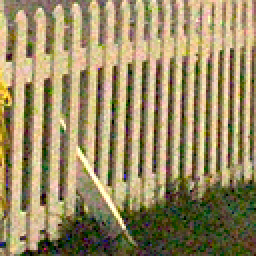} &
        \includegraphics[width=0.14\textwidth]{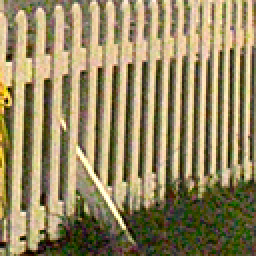} &
        \includegraphics[width=0.14\textwidth]{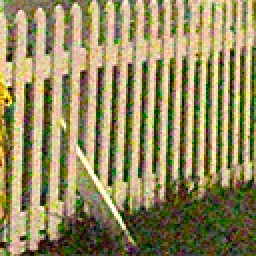} &
        \includegraphics[width=0.14\textwidth]{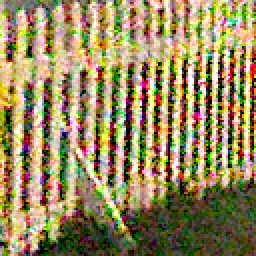} &
        \includegraphics[width=0.14\textwidth]{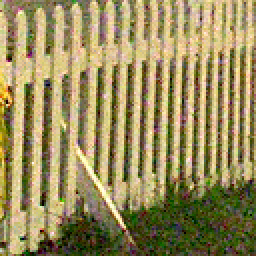} &
        \includegraphics[width=0.14\textwidth]{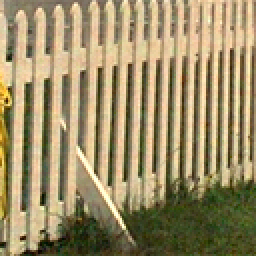} \\
        \includegraphics[width=0.14\textwidth]{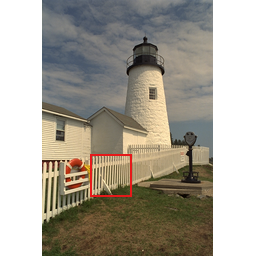} &
        \includegraphics[width=0.14\textwidth]{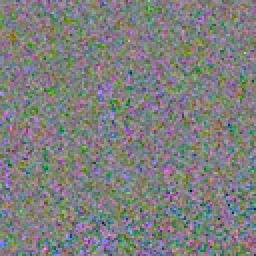} &
        \includegraphics[width=0.14\textwidth]{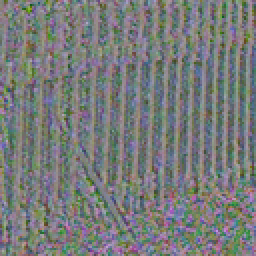} &
        \includegraphics[width=0.14\textwidth]{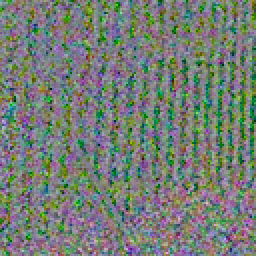} &
        \includegraphics[width=0.14\textwidth]{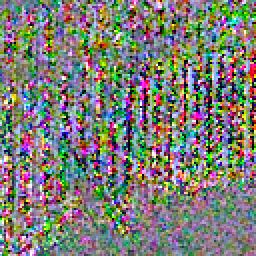} &
        \includegraphics[width=0.14\textwidth]{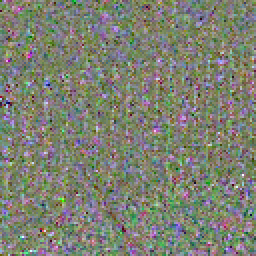} &
        \includegraphics[width=0.14\textwidth]{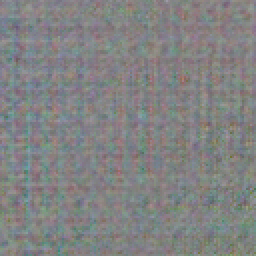}
    \end{tabular}
    \caption{\textbf{A clean image from outside SIDD.} Kodak \emph{kodim19} rendered by every SIDD-trained generator with its S6 ISO\,800 fit or checkpoint, unchanged (one draw each). Top: output; bottom: residual relative to the clean image, amplified $\times$1.5 (gray is zero); bottom left: the full image with the $128\times128$ window marked, the image's most textured window by the rule of Figs.~\ref{fig:supp-sidd-textured-1}--\ref{fig:supp-sidd-textured-4}, fixed before rendering.}
    \label{fig:supp-unseen}
\end{figure*}

\paragraph{A clean image from outside SIDD.}
Sec.~\mainref{sec:analysis} of the main paper applies every SIDD fit and
checkpoint, unchanged, to the Kodak image \emph{kodim19}, whose most textured
windows have more than twice the local contrast of the most textured SIDD window.
Fig.~\ref{fig:supp-unseen} shows the S6 ISO\,800 class, which every method
supports, in the image's most textured window.

\section{Visual comparisons on film grain}
\label{sec:supp-film-visual}

This section extends Fig.~\mainref{fig:main_newson_vis} of the main paper to all
eight radii of the Newson renderer~\cite{newson2017stochastic}. Every crop comes
from the evaluation scored in Tab.~\mainref{tab:film} of the main paper: each method shows its first
generator run, the reference is one Newson realization, and an independently
seeded Newson run shows the renderer's own sampling variability. PGNM and
Zhang23~\cite{zhang2023grain} add grain to the shared blur $B*x$ fitted on
training pairs. Residuals are taken relative to $\mu=B*x$ and amplified by one gain
per radius, set from the reference residual alone.

\begin{figure*}[tp]
\centering
\vcsetup{8}%
\begin{tabular}{@{}c*{6}{c}@{}}
  & \vchead{Clean} & \vchead{Newson} & \vchead{\textbf{PGNM}\,+\,$B$} & \vchead{Zhang23\,+\,$B$} & \vchead{GrainNet} & \vchead{Ind. Newson render} \\[2pt]
\vcrow{$r=0.025$} & \includegraphics[width=\vccell]{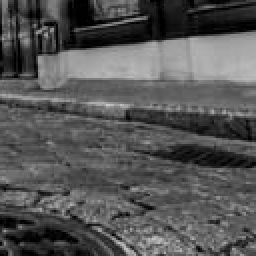} & \includegraphics[width=\vccell]{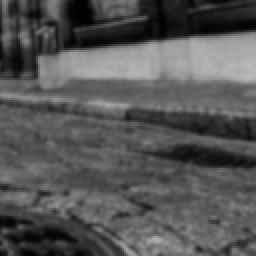} & \includegraphics[width=\vccell]{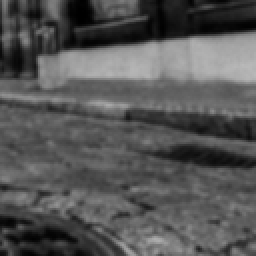} & \includegraphics[width=\vccell]{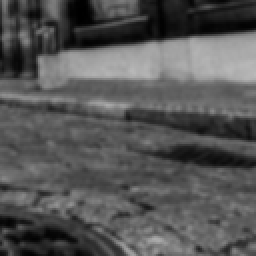} & \includegraphics[width=\vccell]{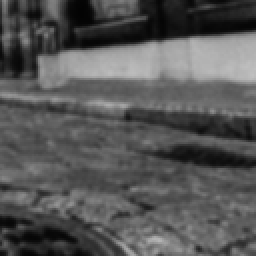} & \includegraphics[width=\vccell]{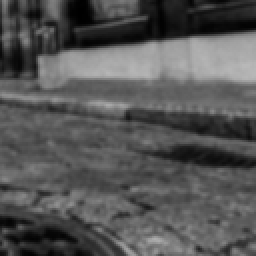} \\[0.5pt]
\vcrow{residual $\times$16} & \includegraphics[width=\vccell]{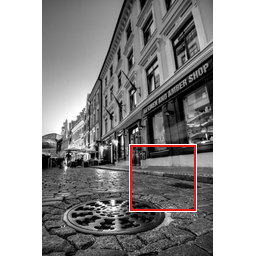} & \includegraphics[width=\vccell]{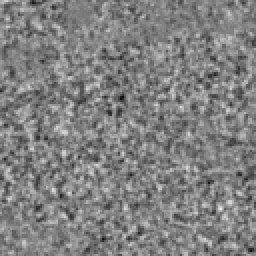} & \includegraphics[width=\vccell]{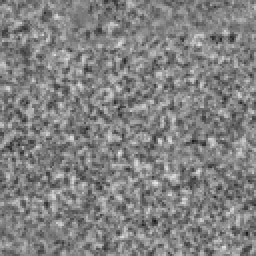} & \includegraphics[width=\vccell]{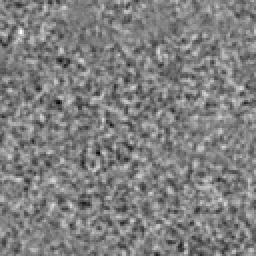} & \includegraphics[width=\vccell]{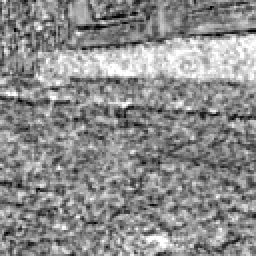} & \includegraphics[width=\vccell]{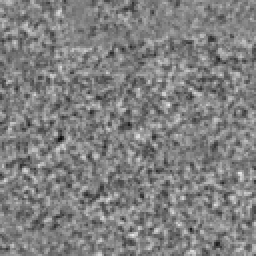} \\[3pt]
\vcrow{$r=0.05$} & \includegraphics[width=\vccell]{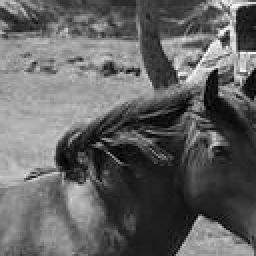} & \includegraphics[width=\vccell]{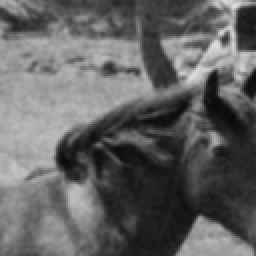} & \includegraphics[width=\vccell]{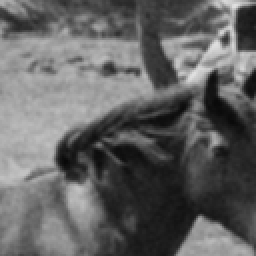} & \includegraphics[width=\vccell]{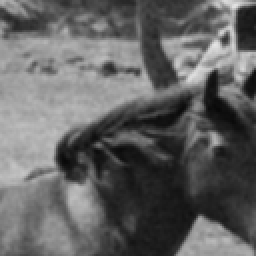} & \includegraphics[width=\vccell]{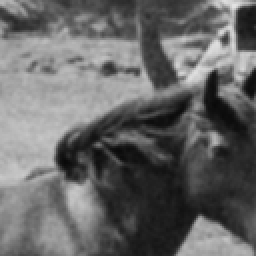} & \includegraphics[width=\vccell]{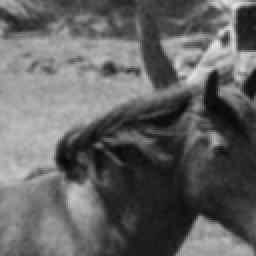} \\[0.5pt]
\vcrow{residual $\times$12} & \includegraphics[width=\vccell]{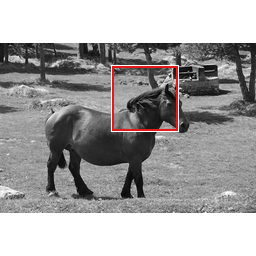} & \includegraphics[width=\vccell]{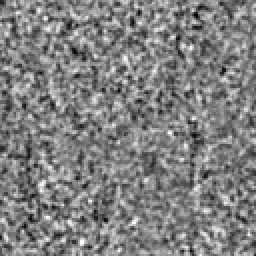} & \includegraphics[width=\vccell]{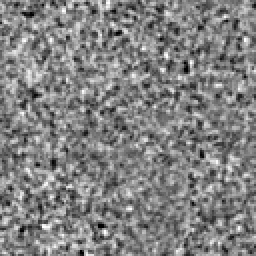} & \includegraphics[width=\vccell]{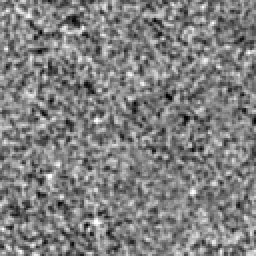} & \includegraphics[width=\vccell]{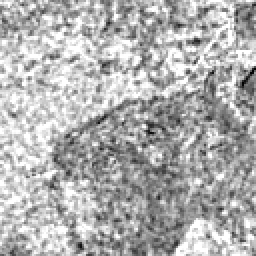} & \includegraphics[width=\vccell]{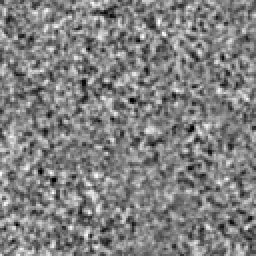} \\[3pt]
\vcrow{$r=0.075$} & \includegraphics[width=\vccell]{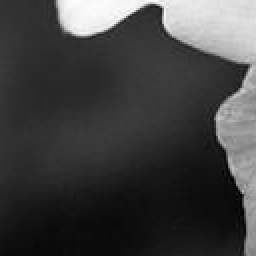} & \includegraphics[width=\vccell]{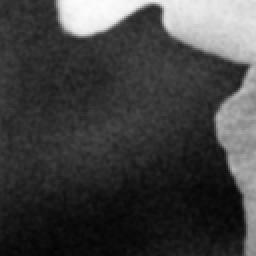} & \includegraphics[width=\vccell]{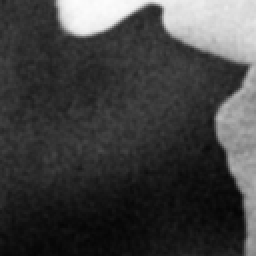} & \includegraphics[width=\vccell]{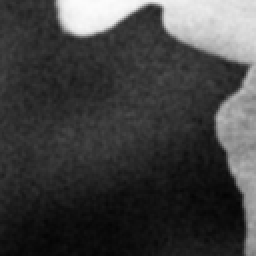} & \includegraphics[width=\vccell]{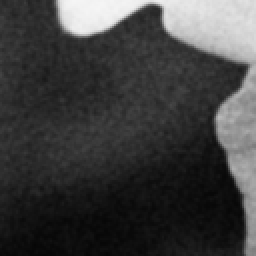} & \includegraphics[width=\vccell]{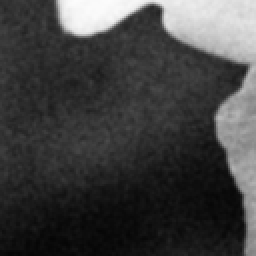} \\[0.5pt]
\vcrow{residual $\times$9} & \includegraphics[width=\vccell]{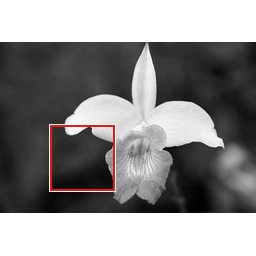} & \includegraphics[width=\vccell]{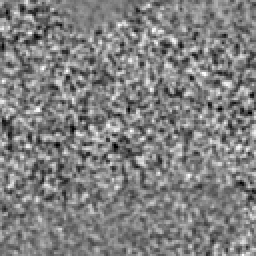} & \includegraphics[width=\vccell]{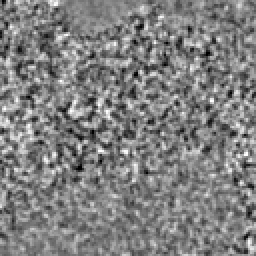} & \includegraphics[width=\vccell]{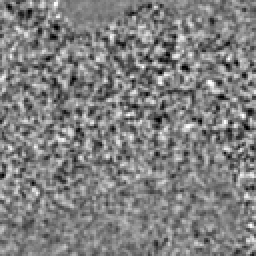} & \includegraphics[width=\vccell]{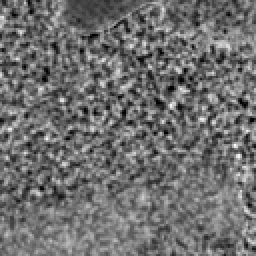} & \includegraphics[width=\vccell]{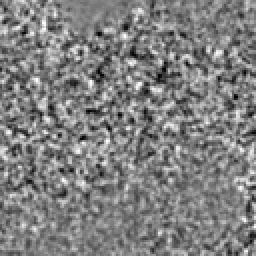} \\[3pt]
\vcrow{$r=0.1$} & \includegraphics[width=\vccell]{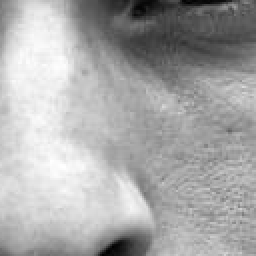} & \includegraphics[width=\vccell]{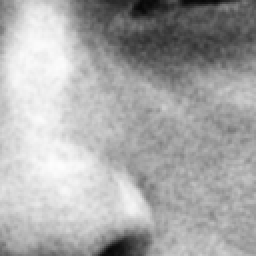} & \includegraphics[width=\vccell]{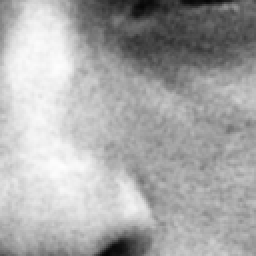} & \includegraphics[width=\vccell]{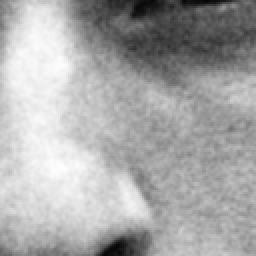} & \includegraphics[width=\vccell]{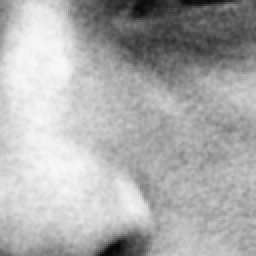} & \includegraphics[width=\vccell]{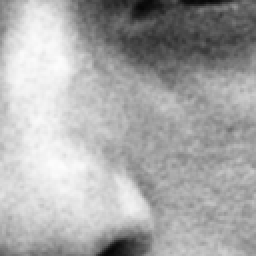} \\[0.5pt]
\vcrow{residual $\times$6.5} & \includegraphics[width=\vccell]{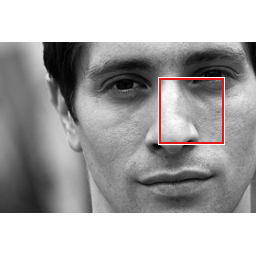} & \includegraphics[width=\vccell]{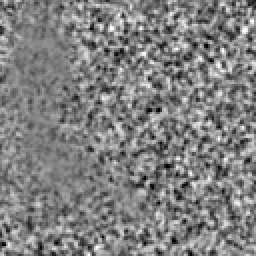} & \includegraphics[width=\vccell]{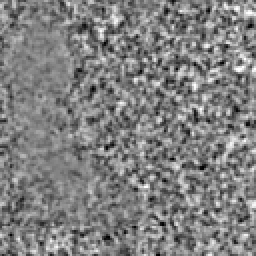} & \includegraphics[width=\vccell]{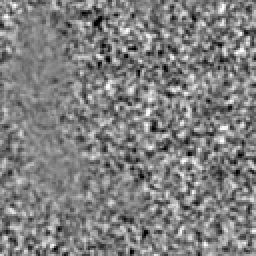} & \includegraphics[width=\vccell]{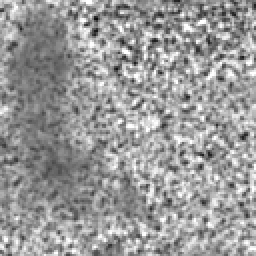} & \includegraphics[width=\vccell]{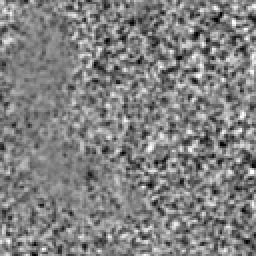} \\
\end{tabular}
\caption{\textbf{Uncurated film-grain comparison, every radius (1/2).} One test image per radius (without replacement) and a $128\times128$ window inside the 3-pixel scoring border were drawn with a fixed pseudo-random seed from the clean image only; a window was redrawn only when more than 25\% of its clean pixels were near 0 or 255, and no draw was replaced after viewing the outputs. Every method shows its first generator run; the reference is one Newson realization and the independent Newson render is an independently seeded simulator run, showing sampling variability. PGNM and Zhang23 add grain to the shared training-fitted blur $B*x$. Second row of each pair: residual relative to $\mu=B*x$, amplified by the factor at left (set from the Newson reference residual alone; gray is zero); under the clean column, the full image with the crop window marked. Crops are shown at native resolution; zoom in for detail. }
\label{fig:supp-film-1}
\end{figure*}

\begin{figure*}[tp]
\centering
\vcsetup{8}%
\begin{tabular}{@{}c*{6}{c}@{}}
  & \vchead{Clean} & \vchead{Newson} & \vchead{\textbf{PGNM}\,+\,$B$} & \vchead{Zhang23\,+\,$B$} & \vchead{GrainNet} & \vchead{Ind. Newson render} \\[2pt]
\vcrow{$r=0.2$} & \includegraphics[width=\vccell]{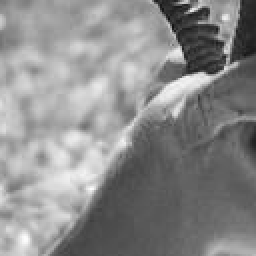} & \includegraphics[width=\vccell]{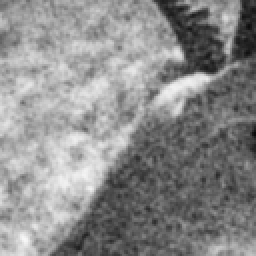} & \includegraphics[width=\vccell]{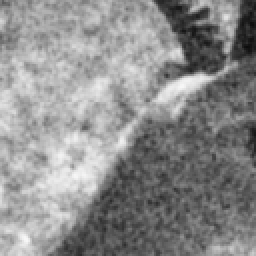} & \includegraphics[width=\vccell]{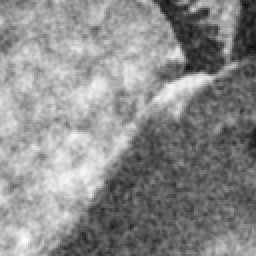} & \includegraphics[width=\vccell]{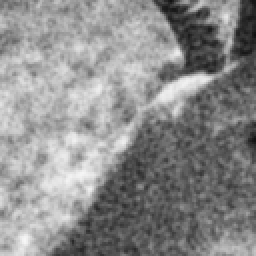} & \includegraphics[width=\vccell]{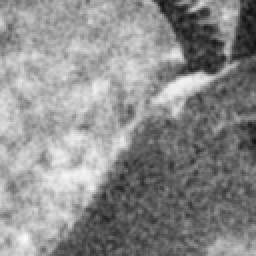} \\[0.5pt]
\vcrow{residual $\times$2.5} & \includegraphics[width=\vccell]{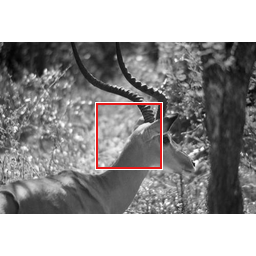} & \includegraphics[width=\vccell]{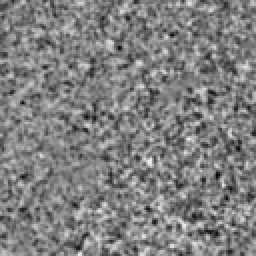} & \includegraphics[width=\vccell]{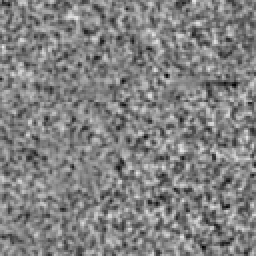} & \includegraphics[width=\vccell]{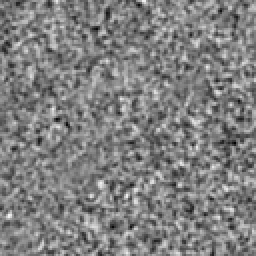} & \includegraphics[width=\vccell]{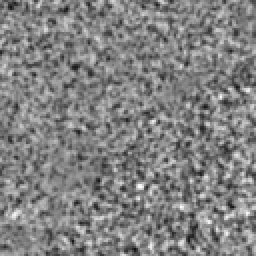} & \includegraphics[width=\vccell]{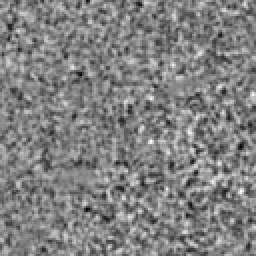} \\[3pt]
\vcrow{$r=0.4$$^\star$} & \includegraphics[width=\vccell]{supp/visual_comparison/film/r400/clean.png} & \includegraphics[width=\vccell]{supp/visual_comparison/film/r400/reference.png} & \includegraphics[width=\vccell]{supp/visual_comparison/film/r400/pgnm.png} & \includegraphics[width=\vccell]{supp/visual_comparison/film/r400/zhang_blur.png} & \includegraphics[width=\vccell]{supp/visual_comparison/film/r400/lesne.png} & \includegraphics[width=\vccell]{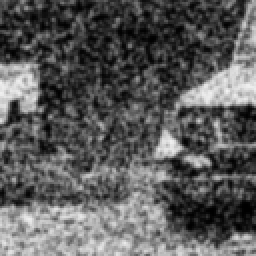} \\[0.5pt]
\vcrow{residual $\times$1.5} & \includegraphics[width=\vccell]{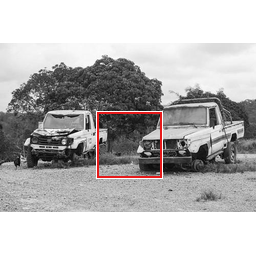} & \includegraphics[width=\vccell]{supp/visual_comparison/film/r400/reference-res.png} & \includegraphics[width=\vccell]{supp/visual_comparison/film/r400/pgnm-res.png} & \includegraphics[width=\vccell]{supp/visual_comparison/film/r400/zhang_blur-res.png} & \includegraphics[width=\vccell]{supp/visual_comparison/film/r400/lesne-res.png} & \includegraphics[width=\vccell]{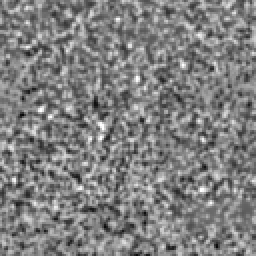} \\[3pt]
\vcrow{$r=0.6$} & \includegraphics[width=\vccell]{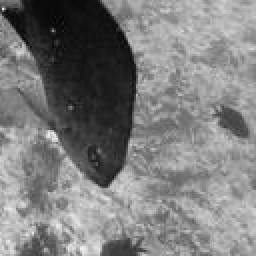} & \includegraphics[width=\vccell]{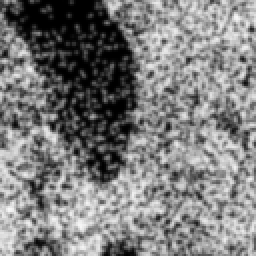} & \includegraphics[width=\vccell]{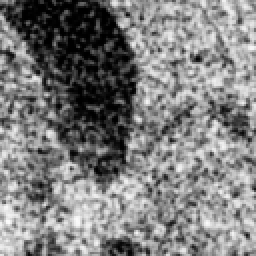} & \includegraphics[width=\vccell]{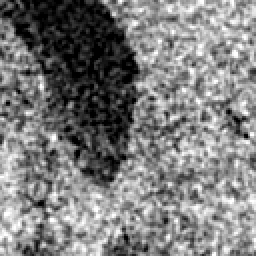} & \includegraphics[width=\vccell]{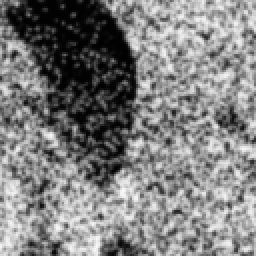} & \includegraphics[width=\vccell]{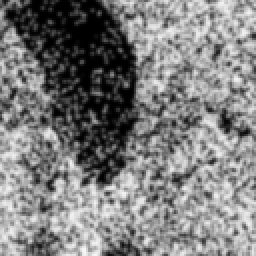} \\[0.5pt]
\vcrow{residual $\times$1} & \includegraphics[width=\vccell]{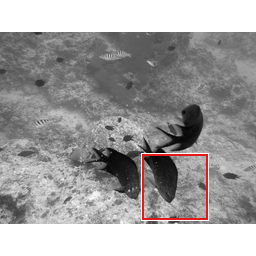} & \includegraphics[width=\vccell]{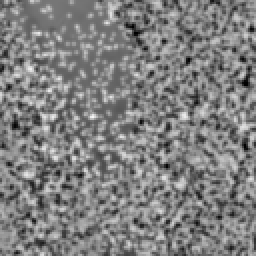} & \includegraphics[width=\vccell]{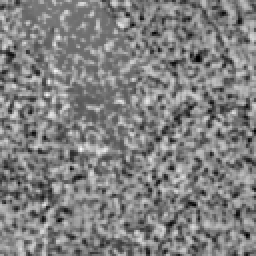} & \includegraphics[width=\vccell]{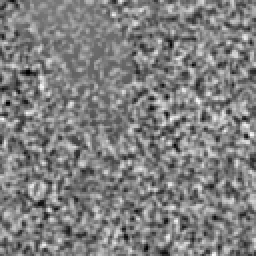} & \includegraphics[width=\vccell]{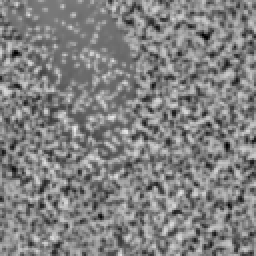} & \includegraphics[width=\vccell]{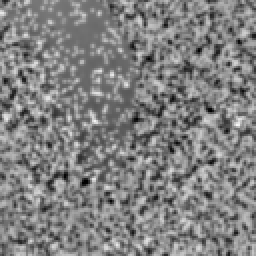} \\[3pt]
\vcrow{$r=0.8$$^\star$} & \includegraphics[width=\vccell]{supp/visual_comparison/film/r800/clean.png} & \includegraphics[width=\vccell]{supp/visual_comparison/film/r800/reference.png} & \includegraphics[width=\vccell]{supp/visual_comparison/film/r800/pgnm.png} & \includegraphics[width=\vccell]{supp/visual_comparison/film/r800/zhang_blur.png} & \includegraphics[width=\vccell]{supp/visual_comparison/film/r800/lesne.png} & \includegraphics[width=\vccell]{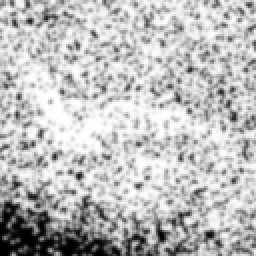} \\[0.5pt]
\vcrow{residual $\times$1} & \includegraphics[width=\vccell]{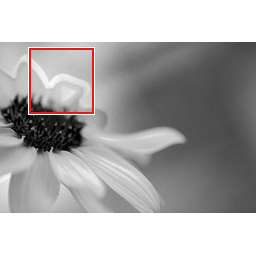} & \includegraphics[width=\vccell]{supp/visual_comparison/film/r800/reference-res.png} & \includegraphics[width=\vccell]{supp/visual_comparison/film/r800/pgnm-res.png} & \includegraphics[width=\vccell]{supp/visual_comparison/film/r800/zhang_blur-res.png} & \includegraphics[width=\vccell]{supp/visual_comparison/film/r800/lesne-res.png} & \includegraphics[width=\vccell]{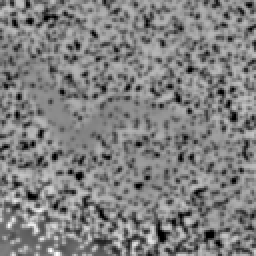} \\
\end{tabular}
\caption{\textbf{Uncurated film-grain comparison, every radius (2/2).} One test image per radius (without replacement) and a $128\times128$ window inside the 3-pixel scoring border were drawn with a fixed pseudo-random seed from the clean image only; a window was redrawn only when more than 25\% of its clean pixels were near 0 or 255, and no draw was replaced after viewing the outputs. Every method shows its first generator run; the reference is one Newson realization and the independent Newson render is an independently seeded simulator run, showing sampling variability. PGNM and Zhang23 add grain to the shared training-fitted blur $B*x$. Second row of each pair: residual relative to $\mu=B*x$, amplified by the factor at left (set from the Newson reference residual alone; gray is zero); under the clean column, the full image with the crop window marked. Crops are shown at native resolution; zoom in for detail. $^\star$Also shown in the main paper (Fig.~\mainref{fig:main_newson_vis}). }
\label{fig:supp-film-2}
\end{figure*}

\paragraph{Uncurated crops.}
For every radius, one test image and a $128\times128$ window were drawn with a
fixed seed from the clean images only, before any output was viewed
(Figs.~\ref{fig:supp-film-1} and~\ref{fig:supp-film-2}; windows in
Tab.~\ref{tab:supp-crops}). The crops for $r=0.4$ and $r=0.8$ in the main paper
are among them.

\begin{figure*}[tp]
\centering
\vcsetup{8}%
\begin{tabular}{@{}c*{6}{c}@{}}
  & \vchead{Clean} & \vchead{Newson} & \vchead{\textbf{PGNM}\,+\,$B$} & \vchead{Zhang23\,+\,$B$} & \vchead{GrainNet} & \vchead{Ind. Newson render} \\[2pt]
\vcrow{$r=0.025$} & \includegraphics[width=\vccell]{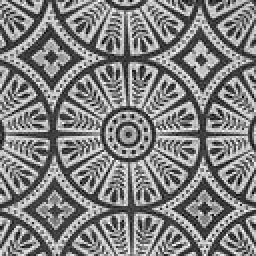} & \includegraphics[width=\vccell]{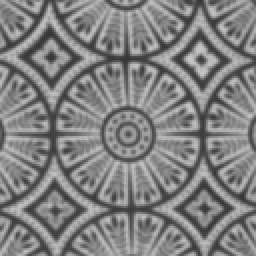} & \includegraphics[width=\vccell]{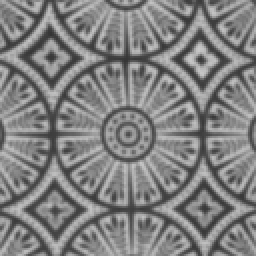} & \includegraphics[width=\vccell]{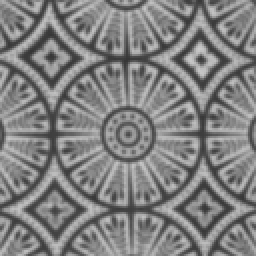} & \includegraphics[width=\vccell]{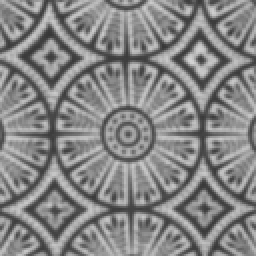} & \includegraphics[width=\vccell]{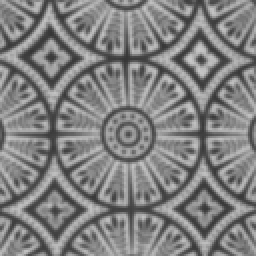} \\[0.5pt]
\vcrow{residual $\times$16} & \includegraphics[width=\vccell]{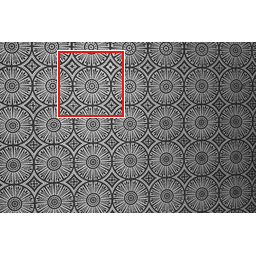} & \includegraphics[width=\vccell]{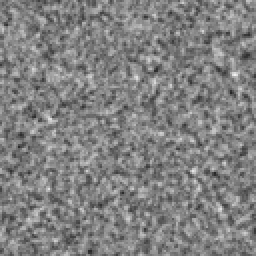} & \includegraphics[width=\vccell]{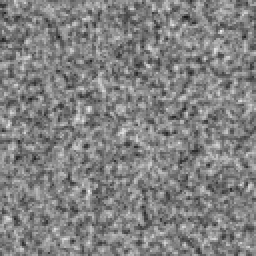} & \includegraphics[width=\vccell]{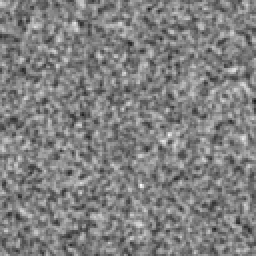} & \includegraphics[width=\vccell]{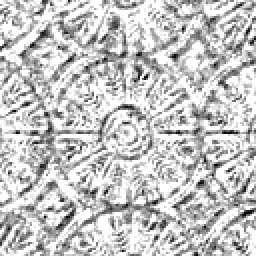} & \includegraphics[width=\vccell]{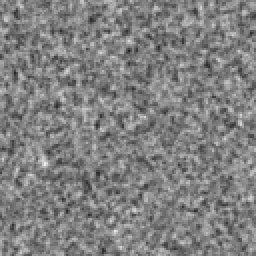} \\[3pt]
\vcrow{$r=0.05$} & \includegraphics[width=\vccell]{supp/visual_comparison/film-textured/r025/clean.png} & \includegraphics[width=\vccell]{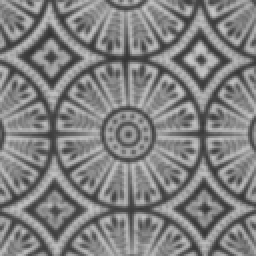} & \includegraphics[width=\vccell]{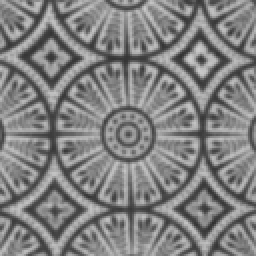} & \includegraphics[width=\vccell]{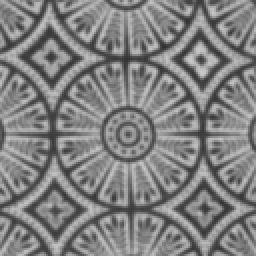} & \includegraphics[width=\vccell]{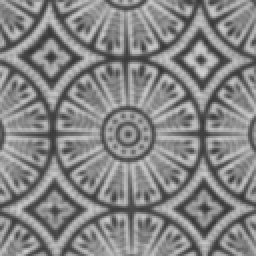} & \includegraphics[width=\vccell]{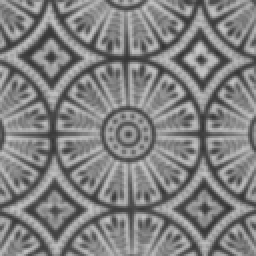} \\[0.5pt]
\vcrow{residual $\times$12.5} & \includegraphics[width=\vccell]{supp/visual_comparison/film-textured/r025/location.png} & \includegraphics[width=\vccell]{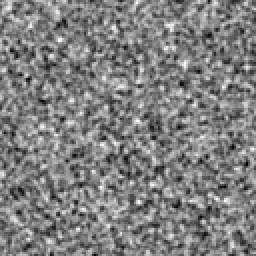} & \includegraphics[width=\vccell]{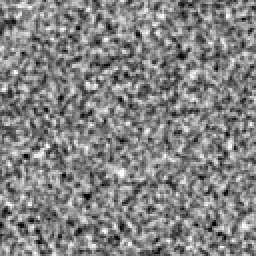} & \includegraphics[width=\vccell]{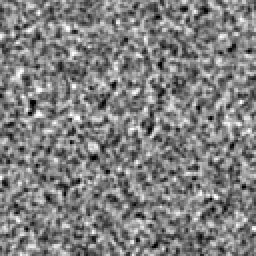} & \includegraphics[width=\vccell]{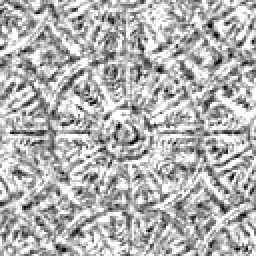} & \includegraphics[width=\vccell]{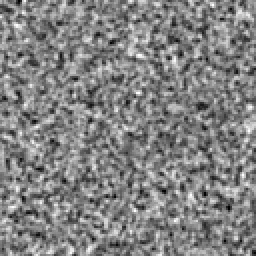} \\[3pt]
\vcrow{$r=0.075$} & \includegraphics[width=\vccell]{supp/visual_comparison/film-textured/r025/clean.png} & \includegraphics[width=\vccell]{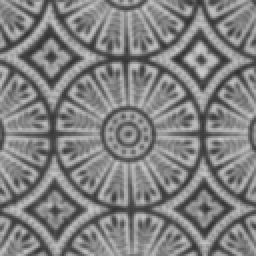} & \includegraphics[width=\vccell]{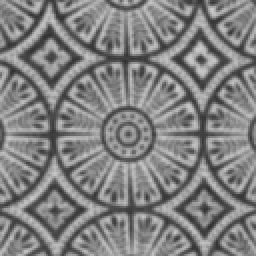} & \includegraphics[width=\vccell]{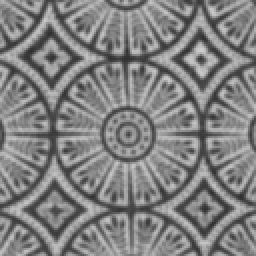} & \includegraphics[width=\vccell]{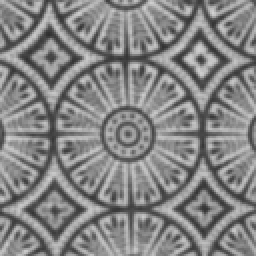} & \includegraphics[width=\vccell]{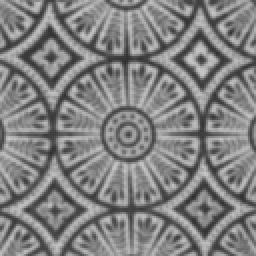} \\[0.5pt]
\vcrow{residual $\times$8.5} & \includegraphics[width=\vccell]{supp/visual_comparison/film-textured/r025/location.png} & \includegraphics[width=\vccell]{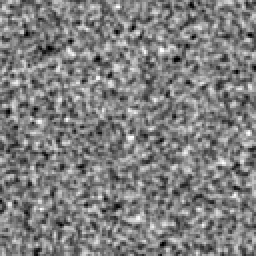} & \includegraphics[width=\vccell]{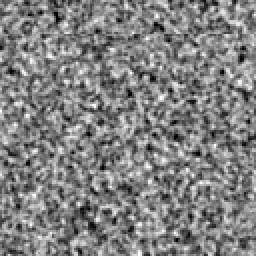} & \includegraphics[width=\vccell]{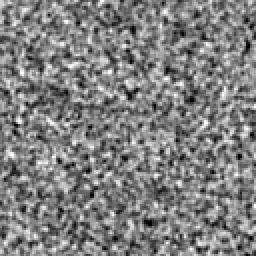} & \includegraphics[width=\vccell]{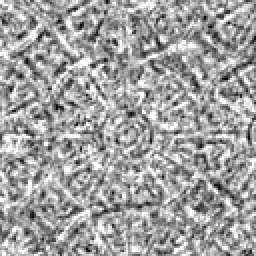} & \includegraphics[width=\vccell]{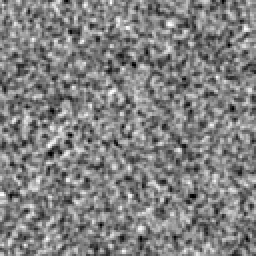} \\[3pt]
\vcrow{$r=0.1$} & \includegraphics[width=\vccell]{supp/visual_comparison/film-textured/r025/clean.png} & \includegraphics[width=\vccell]{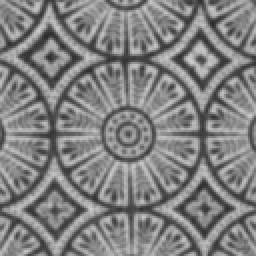} & \includegraphics[width=\vccell]{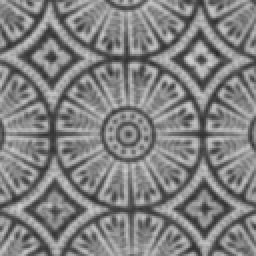} & \includegraphics[width=\vccell]{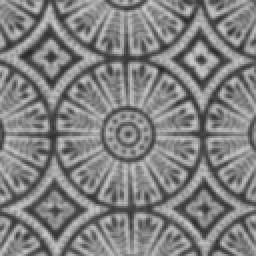} & \includegraphics[width=\vccell]{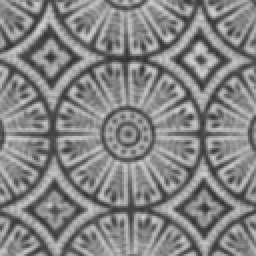} & \includegraphics[width=\vccell]{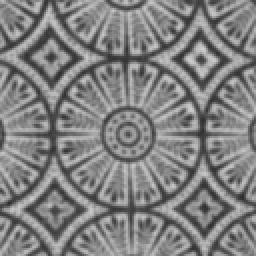} \\[0.5pt]
\vcrow{residual $\times$6} & \includegraphics[width=\vccell]{supp/visual_comparison/film-textured/r025/location.png} & \includegraphics[width=\vccell]{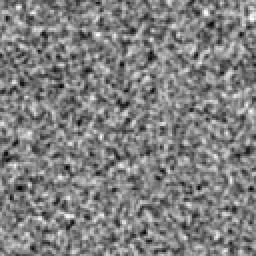} & \includegraphics[width=\vccell]{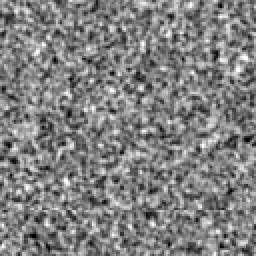} & \includegraphics[width=\vccell]{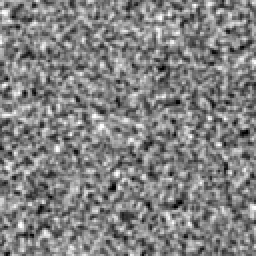} & \includegraphics[width=\vccell]{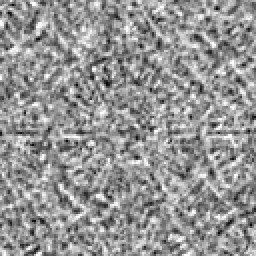} & \includegraphics[width=\vccell]{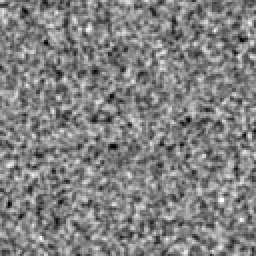} \\
\end{tabular}
\caption{\textbf{Film-grain comparison on a highly textured region, every radius (1/2).} The $128\times128$ window inside the 3-pixel scoring border with the highest median $7\times7$ local standard deviation of the clean image, over all 60 test images on an 8-pixel grid (excluding windows with more than 25\% of clean pixels near 0 or 255), is shown at every radius; it was fixed from the clean images before any output was viewed. Every method shows its first generator run; the reference is one Newson realization and the independent Newson render is an independently seeded simulator run, showing sampling variability. PGNM and Zhang23 add grain to the shared training-fitted blur $B*x$. Second row of each pair: residual relative to $\mu=B*x$, amplified by the factor at left (set from the Newson reference residual alone; gray is zero); under the clean column, the full image with the crop window marked. Crops are shown at native resolution; zoom in for detail. }
\label{fig:supp-film-textured-1}
\end{figure*}

\begin{figure*}[tp]
\centering
\vcsetup{8}%
\begin{tabular}{@{}c*{6}{c}@{}}
  & \vchead{Clean} & \vchead{Newson} & \vchead{\textbf{PGNM}\,+\,$B$} & \vchead{Zhang23\,+\,$B$} & \vchead{GrainNet} & \vchead{Ind. Newson render} \\[2pt]
\vcrow{$r=0.2$} & \includegraphics[width=\vccell]{supp/visual_comparison/film-textured/r025/clean.png} & \includegraphics[width=\vccell]{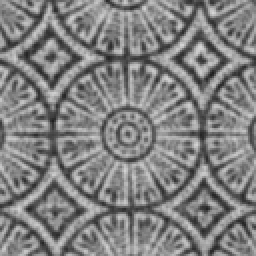} & \includegraphics[width=\vccell]{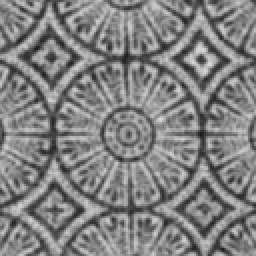} & \includegraphics[width=\vccell]{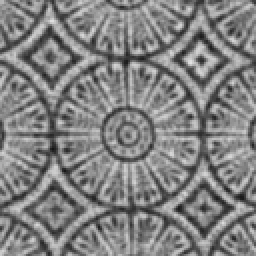} & \includegraphics[width=\vccell]{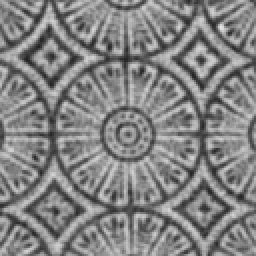} & \includegraphics[width=\vccell]{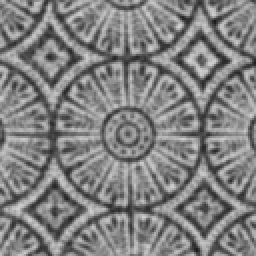} \\[0.5pt]
\vcrow{residual $\times$3} & \includegraphics[width=\vccell]{supp/visual_comparison/film-textured/r025/location.png} & \includegraphics[width=\vccell]{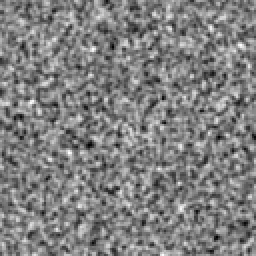} & \includegraphics[width=\vccell]{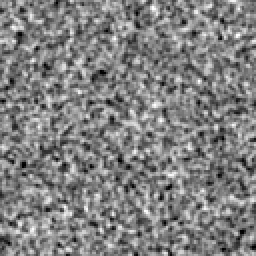} & \includegraphics[width=\vccell]{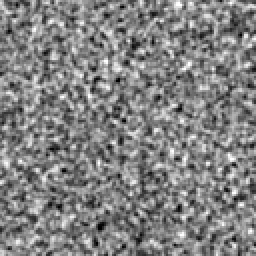} & \includegraphics[width=\vccell]{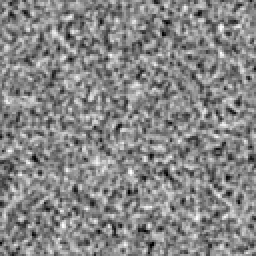} & \includegraphics[width=\vccell]{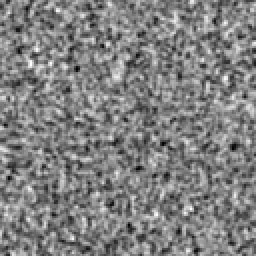} \\[3pt]
\vcrow{$r=0.4$} & \includegraphics[width=\vccell]{supp/visual_comparison/film-textured/r025/clean.png} & \includegraphics[width=\vccell]{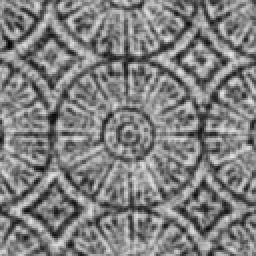} & \includegraphics[width=\vccell]{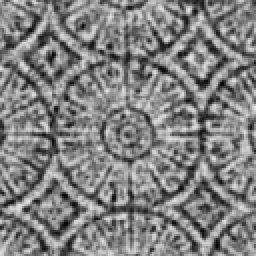} & \includegraphics[width=\vccell]{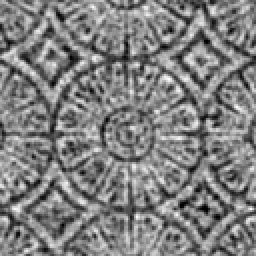} & \includegraphics[width=\vccell]{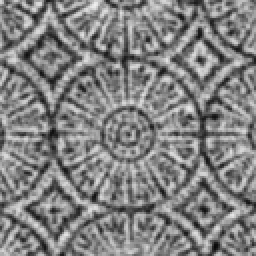} & \includegraphics[width=\vccell]{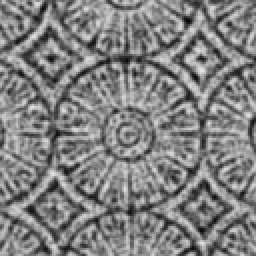} \\[0.5pt]
\vcrow{residual $\times$1.5} & \includegraphics[width=\vccell]{supp/visual_comparison/film-textured/r025/location.png} & \includegraphics[width=\vccell]{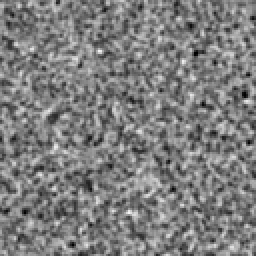} & \includegraphics[width=\vccell]{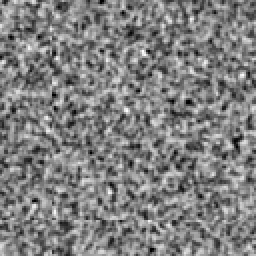} & \includegraphics[width=\vccell]{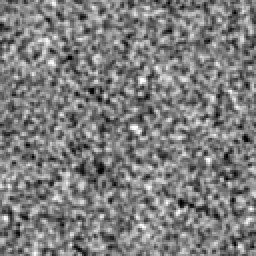} & \includegraphics[width=\vccell]{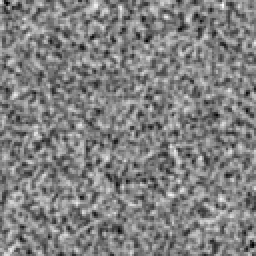} & \includegraphics[width=\vccell]{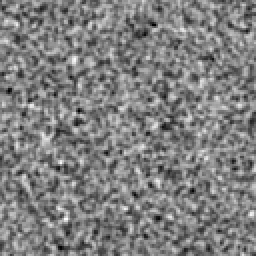} \\[3pt]
\vcrow{$r=0.6$} & \includegraphics[width=\vccell]{supp/visual_comparison/film-textured/r025/clean.png} & \includegraphics[width=\vccell]{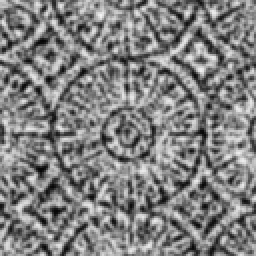} & \includegraphics[width=\vccell]{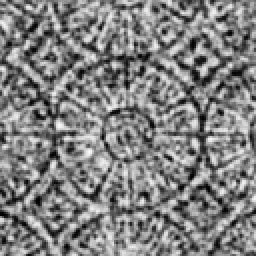} & \includegraphics[width=\vccell]{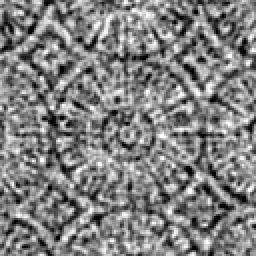} & \includegraphics[width=\vccell]{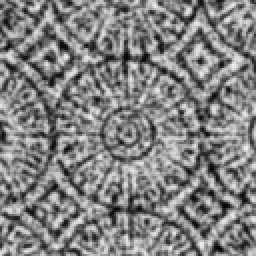} & \includegraphics[width=\vccell]{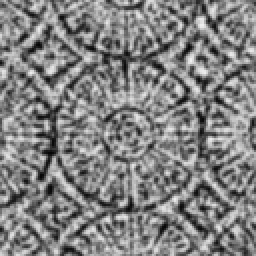} \\[0.5pt]
\vcrow{residual $\times$1} & \includegraphics[width=\vccell]{supp/visual_comparison/film-textured/r025/location.png} & \includegraphics[width=\vccell]{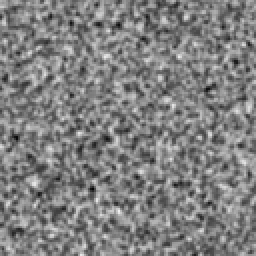} & \includegraphics[width=\vccell]{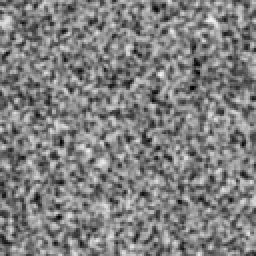} & \includegraphics[width=\vccell]{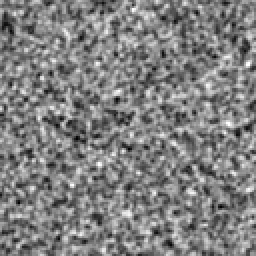} & \includegraphics[width=\vccell]{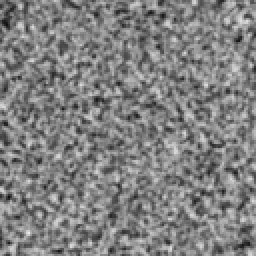} & \includegraphics[width=\vccell]{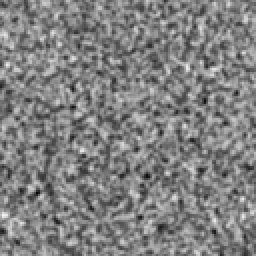} \\[3pt]
\vcrow{$r=0.8$} & \includegraphics[width=\vccell]{supp/visual_comparison/film-textured/r025/clean.png} & \includegraphics[width=\vccell]{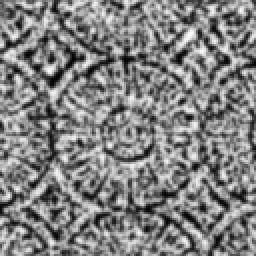} & \includegraphics[width=\vccell]{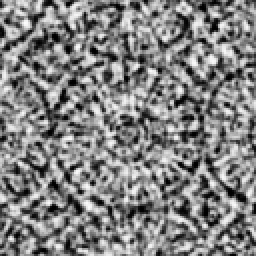} & \includegraphics[width=\vccell]{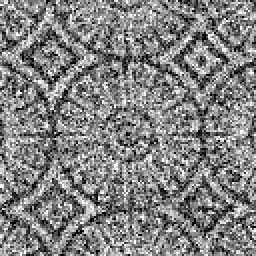} & \includegraphics[width=\vccell]{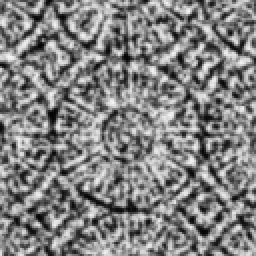} & \includegraphics[width=\vccell]{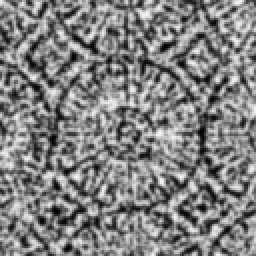} \\[0.5pt]
\vcrow{residual $\times$1} & \includegraphics[width=\vccell]{supp/visual_comparison/film-textured/r025/location.png} & \includegraphics[width=\vccell]{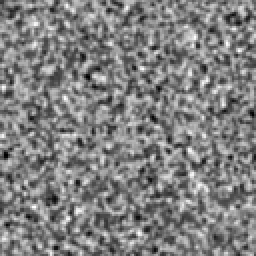} & \includegraphics[width=\vccell]{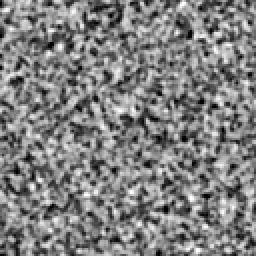} & \includegraphics[width=\vccell]{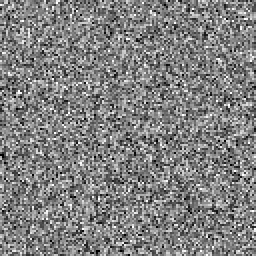} & \includegraphics[width=\vccell]{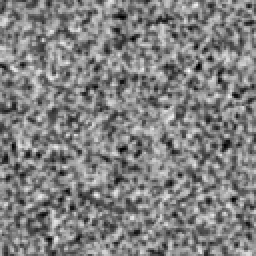} & \includegraphics[width=\vccell]{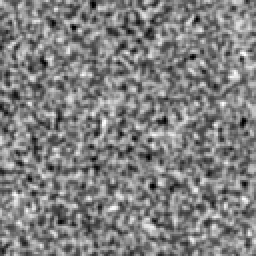} \\
\end{tabular}
\caption{\textbf{Film-grain comparison on a highly textured region, every radius (2/2).} The $128\times128$ window inside the 3-pixel scoring border with the highest median $7\times7$ local standard deviation of the clean image, over all 60 test images on an 8-pixel grid (excluding windows with more than 25\% of clean pixels near 0 or 255), is shown at every radius; it was fixed from the clean images before any output was viewed. Every method shows its first generator run; the reference is one Newson realization and the independent Newson render is an independently seeded simulator run, showing sampling variability. PGNM and Zhang23 add grain to the shared training-fitted blur $B*x$. Second row of each pair: residual relative to $\mu=B*x$, amplified by the factor at left (set from the Newson reference residual alone; gray is zero); under the clean column, the full image with the crop window marked. Crops are shown at native resolution; zoom in for detail. }
\label{fig:supp-film-textured-2}
\end{figure*}

\paragraph{A highly textured region.}
Figs.~\ref{fig:supp-film-textured-1} and~\ref{fig:supp-film-textured-2} show the
window with the most local contrast over all test images at every radius, selected
by the rule used for SIDD (Tab.~\ref{tab:supp-crops-textured}).

\clearpage
\twocolumn[{
        \phantomsection\label{part:supp-reference}
    \pdfbookmark[0]{Reproducibility reference}{supp-reference}
    \partheading{Reproducibility reference}{Exact algorithms, numerical conventions, and experimental settings for reproducing the reported results.}
}]
\section{Implementation specification}
\label{sec:supp-spec}

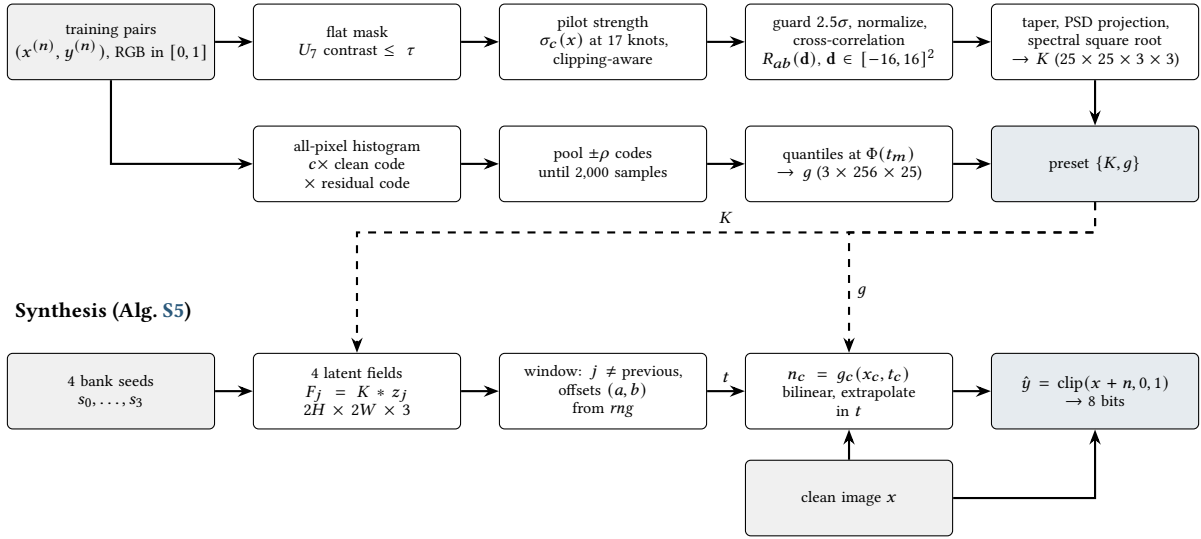
\begin{figure*}[tp]
\centering
\begin{tikzpicture}[
    node distance=4mm and 5mm,
    box/.style={draw, rounded corners=2pt, align=center, font=\scriptsize,
                minimum height=10mm, text width=26mm, inner sep=2pt},
    data/.style={box, fill=black!6},
    res/.style={box, fill=cvprblue!12},
    arr/.style={-{Stealth[length=2mm]}, thick},
    lane/.style={font=\small\bfseries, anchor=west}]
  \node[lane] (fitlabel) at (0,0) {Fitting (Algs.~\ref{alg:supp-fit}--\ref{alg:supp-lookup}), once per camera--ISO class or grain radius};
  \node[data, below=3mm of fitlabel.south west, anchor=north west] (pairs)
       {training pairs\\$(x^{(n)},y^{(n)})$, RGB in $[0,1]$};
  \node[box, right=of pairs] (flat) {flat mask\\$U_7$ contrast $\le\tau$};
  \node[box, right=of flat] (pilot) {pilot strength\\$\sigma_c(x)$ at 17 knots,\\clipping-aware};
  \node[box, right=of pilot] (corr) {guard $2.5\sigma$, normalize,\\cross-correlation\\$R_{ab}(\mathbf d)$, $\mathbf d\in[-16,16]^2$};
  \node[box, right=of corr] (factor) {taper, PSD projection,\\spectral square root\\$\to K$ ($25\times25\times3\times3$)};
  \node[box, below=6mm of flat] (hist) {all-pixel histogram\\$c\times$ clean code\\$\times$ residual code};
  \node[box, right=of hist] (pool) {pool $\pm\rho$ codes\\until 2,000 samples};
  \node[box, right=of pool] (quant) {quantiles at $\Phi(t_m)$\\$\to g$ ($3\times256\times25$)};
  \node[res, right=of quant] (preset) {preset $\{K,g\}$};
  \draw[arr] (pairs) -- (flat);
  \draw[arr] (flat) -- (pilot);
  \draw[arr] (pilot) -- (corr);
  \draw[arr] (corr) -- (factor);
  \draw[arr] (pairs.south) |- (hist.west);
  \draw[arr] (hist) -- (pool);
  \draw[arr] (pool) -- (quant);
  \draw[arr] (quant) -- (preset);
  \draw[arr] (factor) -- (preset);
  \node[lane, below=12mm of hist.south west -| fitlabel.west, anchor=north west] (synlabel)
       {Synthesis (Alg.~\ref{alg:supp-render})};
  \node[data, below=3mm of synlabel.south west, anchor=north west] (seeds) {4 bank seeds\\$s_0,\dots,s_3$};
  \node[box, right=of seeds] (bank) {4 latent fields\\$F_j=K\ast z_j$\\$2H\times2W\times3$};
  \node[box, right=of bank] (window) {window: $j\ne$ previous,\\offsets $(a,b)$\\from $\mathit{rng}$};
  \node[box, right=of window] (lookup) {$n_c=g_c(x_c,t_c)$\\bilinear, extrapolate\\in $t$};
  \node[res, right=of lookup] (output) {$\hat y=\operatorname{clip}(x+n,0,1)$\\$\to$ 8 bits};
  \node[data, below=4mm of lookup] (clean) {clean image $x$};
  \draw[arr] (seeds) -- (bank);
  \draw[arr] (bank) -- (window);
  \draw[arr] (window) -- node[above, font=\scriptsize] {$t$} (lookup);
  \draw[arr] (lookup) -- (output);
  \draw[arr] (clean) -- (lookup);
  \draw[arr] (clean.east) -| (output.south);
  \draw[arr, dashed] (preset.south) -- ++(0,-4mm) -| node[pos=0.25, above, font=\scriptsize] {$K$} (bank.north);
  \draw[arr, dashed] (preset.south) -- ++(0,-4mm) -| node[pos=0.75, right, font=\scriptsize] {$g$} (lookup.north);
\end{tikzpicture}
\caption{\textbf{Data flow of PGNM.} Fitting produces the preset $\{K,g\}$ from
aligned training pairs. The filter branch (top) estimates dependence from selected
pixels; the separately fitted lookup branch (bottom) uses all pixels. Synthesis
builds the latent bank once per preset and image size and renders each output from
one window.}
\label{fig:supp-spec-flow}
\end{figure*}

This section translates the model of Sec.~\ref{sec:supp-equations} into an
implementation. Fitting produces two arrays: convolution filters that set the
latent field's spatial and color dependence, and a lookup that maps each latent
value and clean intensity to a residual. Synthesis first constructs Gaussian
fields with the fitted filters, then applies the lookup pixel by pixel.
Fig.~\ref{fig:supp-spec-flow} summarizes these stages,
Tab.~\ref{tab:supp-spec-constants} lists their fixed settings, and
Algs.~\ref{alg:supp-fit}--\ref{alg:supp-render} specify the numerical operations.
Sec.~\ref{sec:supp-spec-experiments} records the data splits and random seeds used
in our experiments; Tab.~\ref{tab:supp-spec-checkpoints} provides fitted values for
checking a reimplementation.

We retain the notation of the main paper (Sec.~\mainref{sec:method}) and
Sec.~\ref{sec:supp-equations}. In array form, $K$ stores the filters $A_{ci}$,
$g$ stores the values of $f_c$ in Eq.~\mainref{eq:fit_lookup}, and the bank stores
the main paper's latent templates.

\subsection{Interfaces and conventions}

\paragraph{Images.}
Each training pair contains aligned clean and noisy RGB arrays of the same
shape, with values in $[0,1]$; different pairs may have different image sizes.
An 8-bit image is divided by 255 before use. Fitting uses float64. Synthesis
converts the clean image to float32, evaluates the lookup in float64, and returns
a float32 image. Delivery to 8 bits uses round-half-up conversion,
$\lfloor255\operatorname{clip}(v,0,1)+0.5\rfloor$.

\paragraph{Preset.}
The learned part of a preset consists of two arrays. The convolution tensor $K[y,x,i,o]$, of shape
$25\times25\times3\times3$, holds the filter from input channel $i$ to output
channel $o$ at offset $(y-12,x-12)$: $K[y,x,i,o]=A_{oi}(y-12,x-12)$ in Eq.~\mainref{eq:filter}
of the main paper, with $\sum_{y,x,i}K[y,x,i,o]^2=1$ for every $o$. The lookup
$g[c,k,m]=f_c(x_k,t_m)$, of shape $3\times256\times25$, stores the residual quantiles of
Eq.~\mainref{eq:fit_lookup} at the clean-intensity grid points $x_k=k/255$ and the
Gaussian grid points $t_m=-3+0.25m$. Its values are signed residuals in units of the
image range ($|g|\le1$), nondecreasing in $m$. These are the two fitted arrays;
the remaining settings are fixed as listed in
Tab.~\ref{tab:supp-spec-constants}. The lookup carries residual amplitude, so it
is used directly at synthesis without an additional strength multiplier.

\paragraph{Functions.}
\textsc{Fit}(pairs) returns a preset from aligned (clean, noisy) training pairs
(Alg.~\ref{alg:supp-fit}). \textsc{Warmup}$(H,W,s_0,\dots,s_3)$ builds a bank of
four latent fields from four integer seeds, and \textsc{Render}$(x,\mathit{rng})$
synthesizes one noisy image (Alg.~\ref{alg:supp-render}).

\paragraph{Random numbers.}
$\mathit{rng}$ is a NumPy \texttt{Generator} created by
\texttt{numpy.random.}\allowbreak\texttt{default\_rng(seed)}, with the PCG64 bit generator.
Latent values come from \texttt{standard\_normal(}\allowbreak\texttt{shape, dtype=float32)},
and $\mathit{rng}.\mathrm{int}(n)$
denotes \texttt{integers(n)}, uniform on $\{0,\dots,n-1\}$. Draws are made in the
order written.

\paragraph{Operators.}
$U_7$ is the $7\times7$ box mean of each channel with symmetric boundary
extension that repeats the edge pixel ($\dots cba|abc\dots$;
\texttt{scipy.ndimage} mode \texttt{reflect}). $\operatorname{rint}$ rounds to the
nearest integer, ties to even. $\Phi$ and $\phi$ are the standard normal CDF and
density. Percentiles interpolate linearly between order statistics. The DFT is
$\widetilde G(\boldsymbol\omega)=\sum_{\mathbf d}G(\mathbf d)\,e^{-2\pi\mathrm i\langle\boldsymbol\omega,\mathbf d\rangle/64}$ on a
$64\times64$ grid, applied to every matrix entry: the transform of Sec.~\ref{sec:supp-equations}
at the frequencies $2\pi\boldsymbol\omega/64$.

\begin{table*}[tp]
\centering
\small
\setlength{\tabcolsep}{2.5pt}
\begin{tabular}{@{}llll@{}}
\toprule
Stage & Quantity & Value & Used in \\
\midrule
Flatness & local window, contrast & $7\times7$ box ($U_7$), $\max_c$ of local std & Alg.~\ref{alg:supp-fit} \\
 & threshold $\tau$ & $\operatorname{clip}(2\times\text{2nd percentile},\,0.003,\,0.030)$ & \\
 & percentile sample & every 4th row and column, all training images & \\
 & flat mask & contrast $\le\tau$ (inclusive) & \\
Pilot strength & knots & 17, at $k/16$; nearest-knot bins & Alg.~\ref{alg:supp-pilot} \\
 & subbins per bin; copies of subbin means & 32; about 512 & \\
 & minimum pixels per knot; predicted clipping limit & 500; 0.35 & \\
 & bisection interval, iterations; rejection & $[10^{-6},2]$, 60; $\hat\sigma\ge1.999$ & \\
 & smoothing; floor at use & one pass of $(\frac14,\frac12,\frac14)$, ends held; $10^{-6}$ & \\
Correlation & clipping guard & $2.5\,\sigma_c(x_c)$, all channels & Alg.~\ref{alg:supp-corr} \\
 & lags & $\{-16,\dots,16\}^2$, all $3\times3$ channel pairs & \\
 & taper & 1 for $\|\mathbf d\|\le8$, $\cos^2$ to 0 at $\|\mathbf d\|=16$ & \\
 & spectral grid & $64\times64$ & \\
 & kernel support & $25\times25$, energy 1 per output & \\
Lookup & histogram & $3\times256$ clean codes $\times$ 511 residual codes & Alg.~\ref{alg:supp-lookup} \\
 & pooling & $\pm\rho$ codes, $\rho\ge2$; 2,000 samples or all codes & \\
 & Gaussian grid points & 25, $t_m=-3+0.25m$ & \\
Synthesis & bank & 4 fields of $2H\times2W\times3$, float32 & Alg.~\ref{alg:supp-render} \\
 & window & bank index $\ne$ previous; cyclic offsets in $[0,2H)\times[0,2W)$ & \\
 & lookup evaluation & bilinear; linear extrapolation in $t$ & \\
\bottomrule
\end{tabular}
\caption{\textbf{Constants of PGNM.} All values are fixed for every dataset.}
\label{tab:supp-spec-constants}
\end{table*}

\subsection{Fitting}

Fitting follows two branches that use the same aligned training pairs but
estimate different properties (Fig.~\ref{fig:supp-spec-flow}). The filter branch
measures dependence where residuals are least affected by image structure and
clipping. The lookup branch estimates the observed residual distribution at each
clean intensity using all training pixels. The pilot and pixel masks belong only
to the filter branch. The implementation makes repeated passes over the training
pairs, loading one pair at a time.

\paragraph{Selecting and normalizing residuals.}
Alg.~\ref{alg:supp-fit} first selects flat pixels using the largest RGB local
standard deviation in a $7\times7$ window. Alg.~\ref{alg:supp-pilot} then estimates
how residual strength changes with clean intensity using the clipped-Gaussian
model of Eqs.~\ref{eq:supp-pilot-moment}--\ref{eq:supp-pilot}. This pilot supplies
the normalization scale and clipping guard; the lookup supplies the residual
distribution at synthesis. The subbin means and replication counts in
Alg.~\ref{alg:supp-pilot} approximate each bin's clean-intensity distribution
without retaining its individual pixels.

\paragraph{From correlations to filters.}
Alg.~\ref{alg:supp-corr} combines flatness with a guard defined by the clean
intensity and pilot in every RGB channel. If this combined mask selects zero
pixels over the entire fitting set, correlation fitting uses the flat mask alone
for every image, retaining the same pilot normalization. Otherwise the combined
mask is retained throughout the fit. Normalized residuals are centered within
each image; products and valid-pair counts are then pooled across images for each
displacement. Both endpoints must pass the shared mask and remain in the same
image. Zero-lag variances normalize the covariances to correlations. The spectral
construction of Eqs.~\ref{eq:supp-2d-spectrum}--\ref{eq:supp-filters} converts these
measurements into $K$ by tapering, spectral projection, inverse transformation,
cropping, and output-energy normalization.

\paragraph{From histograms to a lookup.}
Alg.~\ref{alg:supp-lookup} counts every residual, including textured and clipped
observations. For each clean code, neighboring intensity histograms are pooled
until they contain at least 2,000 samples or cover the full intensity range.
The lookup stores the residual quantiles at the 25 probabilities $\Phi(t_m)$.
Treating mass as uniform within each residual-code bin lets these quantiles vary
continuously between integer codes.

A fit requires at least two accepted pilot knots per channel, positive selected
residual variances, and nonzero retained filter energy. The numerical checks are
given in the algorithms. Insufficient data or a degenerate estimate produces an
explicit fitting failure; the mask changes only in the globally empty-guard case
described above.

\begin{algorithm*}[tp]
\caption{\textsc{Fit}: one preset from $N$ aligned training pairs.}
\label{alg:supp-fit}
\small
\begin{algorithmic}[1]
\Require pairs $(x^{(n)},y^{(n)})$, float64 arrays $H_n\times W_n\times3$ in $[0,1]$, $n=1,\dots,N$
\For{$n=1,\dots,N$}
    \State $\kappa^{(n)}(\mathbf p)\gets\max_c\sqrt{\max\bigl(U_7(x_c^2)(\mathbf p)-U_7(x_c)(\mathbf p)^2,\,0\bigr)}$
           \Comment{clean-image contrast}
\EndFor
\State $\tau\gets\operatorname{clip}\bigl(2\cdot\operatorname{percentile}_{2}\{\kappa^{(n)}(4i,4j)\}_{n,i,j},\;0.003,\;0.030\bigr)$
\State $M_{\mathrm{flat}}^{(n)}\gets[\kappa^{(n)}\le\tau]$ for every $n$
\State $\sigma\gets\Call{Pilot}{\{x^{(n)},\,y^{(n)}-x^{(n)},\,M_{\mathrm{flat}}^{(n)}\}}$
       \Comment{Alg.~\ref{alg:supp-pilot}; curves $\sigma_c$, $c\in\{R,G,B\}$}
\If{the guarded mask of Alg.~\ref{alg:supp-corr} selects at least one pixel in the whole set}
    \State $R\gets\Call{CrossCorrelation}{\text{pairs},M_{\mathrm{flat}},\sigma,\text{guard}=\text{true}}$
\Else
    \State $R\gets\Call{CrossCorrelation}{\text{pairs},M_{\mathrm{flat}},\sigma,\text{guard}=\text{false}}$
           \Comment{same pilot normalization}
\EndIf
\State $K\gets\Call{Factor}{R}$ \Comment{Alg.~\ref{alg:supp-corr}}
\State $g\gets\Call{Lookup}{\text{pairs}}$ \Comment{Alg.~\ref{alg:supp-lookup}; all pixels, no mask}
\State \Return $\{K,g\}$
\end{algorithmic}
\end{algorithm*}

\begin{algorithm*}[tp]
\caption{\textsc{Pilot}: clipping-aware strength curves from flat pixels.}
\label{alg:supp-pilot}
\small
\begin{algorithmic}[1]
\Require clean images $x^{(n)}$, residuals $r^{(n)}=y^{(n)}-x^{(n)}$, masks $M_{\mathrm{flat}}^{(n)}$
\State knots $\xi_k=k/16$, $k=0,\dots,16$; arrays $\mathit{Sq},\mathit{Num}\in\mathbb R^{3\times17}$ and
       $\mathit{Cnt},\mathit{Sum}\in\mathbb R^{3\times17\times32}$, all zero
\For{each $n$, channel $c$ and pixel $\mathbf p$ with $M_{\mathrm{flat}}^{(n)}(\mathbf p)$}
    \State $v\gets x_c^{(n)}(\mathbf p)$; \quad $k\gets\min(\lfloor16v+\tfrac12\rfloor,16)$ \Comment{nearest knot}
    \State $\mathit{Sq}[c,k]\mathrel{+}=r_c^{(n)}(\mathbf p)^2$; \quad $\mathit{Num}[c,k]\mathrel{+}=1$
    \State $f\gets\operatorname{clip}\bigl(16v-k+\tfrac12,\,0,\,1-10^{-9}\bigr)$; \quad $j\gets\lfloor32f\rfloor$
           \Comment{position inside the bin}
    \State $\mathit{Cnt}[c,k,j]\mathrel{+}=1$; \quad $\mathit{Sum}[c,k,j]\mathrel{+}=v$
\EndFor
\For{$c=1,2,3$}
    \For{$k=0,\dots,16$ with $\mathit{Num}[c,k]\ge500$}
        \State $\bar m\gets\mathit{Sq}[c,k]/\mathit{Num}[c,k]$; \quad
               \textbf{if} $\bar m\le0$ \textbf{then} skip this knot
        \State $\mathit{Rep}[j]\gets\max\bigl(\operatorname{rint}(512\,\mathit{Cnt}[c,k,j]/\textstyle\sum_{j'}\mathit{Cnt}[c,k,j']),\,1\bigr)$
               for every occupied subbin $j$ ($\mathit{Cnt}[c,k,j]>0$)
        \State $\mathcal I\gets$ multiset holding each subbin mean $\mathit{Sum}[c,k,j]/\mathit{Cnt}[c,k,j]$ $\mathit{Rep}[j]$ times
               \Comment{subbin means repeated by weight}
        \If{$\operatorname{mean}_{I\in\mathcal I}m(2,I)<\bar m$}
            \State $\hat\sigma\gets2$ \Comment{$m$ of Eq.~\ref{eq:supp-pilot-moment}; rejected below}
        \Else
            \State $\sigma_-\gets10^{-6}$, $\sigma_+\gets2$
            \For{60 bisection steps}
                \State $\sigma_0\gets(\sigma_-+\sigma_+)/2$
                \If{$\operatorname{mean}_{I\in\mathcal I}m(\sigma_0,I)<\bar m$}
                    \State $\sigma_-\gets\sigma_0$
                \Else
                    \State $\sigma_+\gets\sigma_0$
                \EndIf
            \EndFor
            \State $\hat\sigma\gets(\sigma_-+\sigma_+)/2$
        \EndIf
        \State $\gamma\gets\operatorname{mean}_{I\in\mathcal I}\bigl[\Phi(-I/\hat\sigma)+\Phi(-(1-I)/\hat\sigma)\bigr]$
               \Comment{predicted clipped fraction}
        \State \textbf{if} $\hat\sigma<1.999$ and $\gamma\le0.35$ \textbf{then} accept $\sigma_c[k]\gets\hat\sigma$
    \EndFor
    \State require at least two accepted knots; fill the others by linear interpolation in $\xi$,
           constant beyond the outermost accepted knots
    \State $\sigma_c[k]\gets\tfrac14\sigma_c[k-1]+\tfrac12\sigma_c[k]+\tfrac14\sigma_c[k+1]$ for $k=1,\dots,15$
           (from the unsmoothed values; ends kept); $\sigma_c\gets\max(\sigma_c,0)$
\EndFor
\State \Return $\sigma_c(v)=\max\bigl(\operatorname{interp}(\operatorname{clip}(v,0,1);\,\xi,\,\sigma_c),\,10^{-6}\bigr)$
       \Comment{piecewise linear in $v$}
\end{algorithmic}
\end{algorithm*}

\begin{algorithm*}[tp]
\caption{\textsc{CrossCorrelation} and \textsc{Factor}: from residual correlations
to the convolution tensor.}
\label{alg:supp-corr}
\small
\begin{algorithmic}[1]
\Procedure{CrossCorrelation}{pairs, $M_{\mathrm{flat}}$, $\sigma$, guard}
    \State $P[a,b,\mathbf d]\gets0$, $Q[\mathbf d]\gets0$ for $a,b\in\{1,2,3\}$, $\mathbf d\in\{-16,\dots,16\}^2$
    \For{$n=1,\dots,N$}
        \State $s_c(\mathbf p)\gets\sigma_c(x_c^{(n)}(\mathbf p))$ for every pixel and channel
        \State $M\gets M_{\mathrm{flat}}^{(n)}$; \textbf{if} guard \textbf{then}
               $M(\mathbf p)\gets M(\mathbf p)\wedge\bigwedge_c\bigl[2.5\,s_c(\mathbf p)<x_c^{(n)}(\mathbf p)<1-2.5\,s_c(\mathbf p)\bigr]$
        \State \textbf{if} $M$ has fewer than 2 pixels \textbf{then} skip this image
        \State $u_c\gets(y_c^{(n)}-x_c^{(n)})/s_c$; \quad
               $w_c\gets\bigl(u_c-\operatorname{mean}_{M}u_c\bigr)\cdot M$ \Comment{zero outside $M$}
        \State $P[a,b,\mathbf d]\mathrel{+}=\sum_{\mathbf p} w_a(\mathbf p)\,w_b(\mathbf p+\mathbf d)$; \quad
               $Q[\mathbf d]\mathrel{+}=\sum_{\mathbf p} M(\mathbf p)\,M(\mathbf p+\mathbf d)$
               \Comment{pairs inside image $n$ only}
    \EndFor
    \State $C\gets P/\max(Q,1)$; \quad $v_a\gets C[a,a,(0,0)]$; \quad require $v_a>10^{-12}$
    \State \Return $R[a,b,\mathbf d]=C[a,b,\mathbf d]/\sqrt{v_av_b}$
\EndProcedure
\Statex
\Procedure{Factor}{$R$}
    \State $\eta(\mathbf d)\gets\cos^2\!\bigl(\tfrac\pi2\operatorname{clip}((\|\mathbf d\|_2-8)/8,\,0,\,1)\bigr)$
    \State $R^{\mathrm s}[a,b,\mathbf d]\gets\tfrac12\bigl(R[a,b,\mathbf d]+R[b,a,-\mathbf d]\bigr)\,\eta(\mathbf d)$
           \Comment{symmetrize and taper, Eq.~\ref{eq:supp-taper}}
    \State $G\gets0\in\mathbb R^{64\times64\times3\times3}$; \quad
           $G[d_y\bmod64,\,d_x\bmod64]\gets R^{\mathrm s}[\cdot,\cdot,\mathbf d]$ for every $\mathbf d$
    \State $\widetilde G\gets\operatorname{DFT}(G)$ \Comment{transform each channel-pair lag array}
    \For{each frequency $\boldsymbol\omega$ of the $64\times64$ DFT}
        \State $S_{\boldsymbol\omega}\gets\tfrac12\bigl(\widetilde G(\boldsymbol\omega)+\widetilde G(\boldsymbol\omega)^{\mathsf H}\bigr)
               =V\operatorname{diag}(\lambda_1,\lambda_2,\lambda_3)V^{\mathsf H}$
        \State $W(\boldsymbol\omega)\gets V\operatorname{diag}\bigl(\sqrt{\max(\lambda_l,0)}\bigr)V^{\mathsf H}$
               \Comment{PSD projection and square root}
    \EndFor
    \State $\Gamma\gets\operatorname{Re}\operatorname{IDFT}\bigl(\overline{W}\bigr)$
           \Comment{$\Gamma[\mathbf d,o,i]$: filter from input $i$ to output $o$}
    \State $K[y,x,i,o]\gets\Gamma[((y-12)\bmod64,\,(x-12)\bmod64),o,i]$ for $y,x\in\{0,\dots,24\}$
    \State $e_o\gets\sum_{y,x,i}K[y,x,i,o]^2$ for each $o$; \quad require $e_o>10^{-12}$
    \State $K[\cdot,\cdot,\cdot,o]\gets K[\cdot,\cdot,\cdot,o]/\sqrt{e_o}$ for each $o$
    \State \Return $K$
\EndProcedure
\end{algorithmic}
\end{algorithm*}

\begin{algorithm*}[tp]
\caption{\textsc{Lookup}: conditional quantiles from all training pixels.}
\label{alg:supp-lookup}
\small
\begin{algorithmic}[1]
\State $\mathcal H\gets0\in\mathbb N^{3\times256\times511}$
\For{every pair $n$, pixel $\mathbf p$ and channel $c$}
    \State $k\gets\operatorname{clip}(\operatorname{rint}(255\,x_c(\mathbf p)),0,255)$
    \State $e\gets\operatorname{clip}(\operatorname{rint}(255\,(y_c(\mathbf p)-x_c(\mathbf p))),-255,255)+255$;
           \quad $\mathcal H[c,k,e]\mathrel{+}=1$ \Comment{residual code $e-255$}
\EndFor
\For{$c=1,2,3$ and $k=0,\dots,255$}
    \State $\rho\gets2$
    \Repeat
        \State $k_-\gets\max(0,k-\rho)$, $k_+\gets\min(255,k+\rho)$
        \State $n_{\mathrm{pool}}\gets\sum_{k'=k_-}^{k_+}\sum_e\mathcal H[c,k',e]$
        \State \textbf{if} $n_{\mathrm{pool}}<2000$ and $(k_-,k_+)\ne(0,255)$ \textbf{then} $\rho\gets\rho+1$
    \Until{$n_{\mathrm{pool}}\ge2000$ or $(k_-,k_+)=(0,255)$}
    \State $P_e\gets\sum_{k'=k_-}^{k_+}\mathcal H[c,k',e]/n_{\mathrm{pool}}$;
           \quad $F_e\gets\sum_{e'\le e}P_{e'}$, \quad $e=0,\dots,510$
    \For{$m=0,\dots,24$}
        \State $q\gets\Phi(t_m)$; \quad $i\gets\min(\min\{e:F_e\ge q\},\,510)$; \quad
               $F_-\gets F_{i-1}$ ($0$ if $i=0$)
        \State $\nu_m\gets(i-255-\tfrac12)+(q-F_-)/\max(P_i,10^{-300})$
               \Comment{linear inside the unit bin of code $i-255$}
    \EndFor
    \State $g[c,k,m]\gets\operatorname{clip}\bigl(\max_{m'\le m}\nu_{m'}/255,\,-1,\,1\bigr)$
           \Comment{running maximum keeps $g$ monotone}
\EndFor
\State \Return $g$
\end{algorithmic}
\end{algorithm*}

\begin{algorithm*}[tp]
\caption{\textsc{Warmup} and \textsc{Render}: the latent bank and one synthesized image.}
\label{alg:supp-render}
\small
\begin{algorithmic}[1]
\Procedure{BuildField}{$h,w,\mathit{rng}$}
    \State $z\gets\mathit{rng}.\texttt{standard\_normal}((h+24,\,w+24,\,3),\ \text{float32})$
    \State $t_o\gets\sum_{i}z_i\ast K[\cdot,\cdot,i,o]$ at the $h\times w$ positions where the kernel lies inside $z$
           \Comment{true convolution}
    \Statex \hspace{\algorithmicindent}\hspace{\algorithmicindent}(\texttt{scipy.signal.fftconvolve}, mode \texttt{valid})
    \State \Return $t$ \Comment{$h\times w\times3$; unit variance per channel}
\EndProcedure
\Statex
\Procedure{Warmup}{$H,W,s_0,\dots,s_3$}
    \State $F_j\gets\text{float32}\bigl(\Call{BuildField}{2H,2W,\texttt{default\_rng}(s_j)}\bigr)$ for $j=0,\dots,3$;
           \quad $\mathit{prev}\gets-1$
\EndProcedure
\Statex
\Procedure{Render}{$x\in[0,1]^{H\times W\times3}$ float32, $\mathit{rng}$}
    \State $\mathcal J\gets[\,j\in(0,1,2,3):j\ne\mathit{prev}\,]$; \quad
           $j\gets\mathcal J[\mathit{rng}.\mathrm{int}(|\mathcal J|)]$; \quad
           $a\gets\mathit{rng}.\mathrm{int}(2H)$; \quad $b\gets\mathit{rng}.\mathrm{int}(2W)$
    \For{every pixel $\mathbf p=(p_y,p_x)$ and channel $c$}
        \State $t\gets F_j[(p_y+a)\bmod2H,\,(p_x+b)\bmod2W,\,c]$
        \State $\hat y_c(\mathbf p)\gets\operatorname{clip}\bigl(x_c(\mathbf p)+\Call{Eval}{g_c,x_c(\mathbf p),t},\,0,\,1\bigr)$
    \EndFor
    \State $\mathit{prev}\gets j$; \quad \Return $\hat y$
\EndProcedure
\Statex
\Function{Eval}{$g_c,x,t$}
    \State $\alpha\gets255\operatorname{clip}(x,0,1)$; \quad $k\gets\min(\lfloor\alpha\rfloor,254)$; \quad $\theta_x\gets\alpha-k$
    \State $\beta\gets(t+3)/0.25$; \quad $m\gets\operatorname{clip}(\lfloor\beta\rfloor,0,23)$; \quad $\theta_t\gets\beta-m$
           \Comment{$\theta_t\notin[0,1]$ extrapolates linearly}
    \State $g_-\gets g_c[k,m]+\theta_t\,(g_c[k,m+1]-g_c[k,m])$; \quad
           $g_+\gets g_c[k+1,m]+\theta_t\,(g_c[k+1,m+1]-g_c[k+1,m])$
    \State \Return $g_-+\theta_x\,(g_+-g_-)$
\EndFunction
\end{algorithmic}
\end{algorithm*}

\subsection{Synthesis}

Synthesis separates field construction from repeated rendering
(Alg.~\ref{alg:supp-render}). \textsc{Warmup} filters independent Gaussian samples
with the nine channel-pair kernels using FFT-based convolution. It stores four
float32 fields of size $2H\times2W\times3$ and reuses them for the same preset and
output image size. The filter-energy constraint of Eq.~\ref{eq:supp-energy}
gives each latent value population variance one.

Each call to \textsc{Render} draws a field index, a row offset, and a column
offset, in that order. It reads an $H\times W$ cyclic window, applies the lookup,
adds the signed residual to the clean intensity, and clips to $[0,1]$.
\textsc{Eval} interpolates bilinearly in clean intensity and Gaussian value;
beyond $t=\pm3$ it extends the first or last line segment.

Cyclic indexing permits any offset without constructing another field. The stored
fields themselves are not periodic, so their usual short-range correlation does
not continue across a wrap seam. For large $H,W$, approximately three quarters of
random windows cross at least one field boundary. This bank-based procedure is
used for every reported SIDD and film score. Denoising training instead constructs
a fresh field for each patch, as specified in
Sec.~\ref{sec:supp-spec-experiments}.

\subsection{Data and random seeds}
\label{sec:supp-spec-experiments}

\begin{table}[tp]
\centering
\footnotesize
\setlength{\tabcolsep}{3pt}
\renewcommand{\arraystretch}{0.9}
\begin{tabular}{@{}ll@{}}
\toprule
Class & Held-out captures (scene instance / capture) \\
\midrule
G4 400 & \texttt{0123\_006\_G4\_00400\_00160\_3200\_N} /010, /011 \\
G4 800 & \texttt{0047\_002\_G4\_00800\_00640\_3200\_L} /010 \\
       & \texttt{0149\_007\_G4\_00800\_00800\_4400\_L} /010 \\
GP 1600 & \texttt{0157\_007\_GP\_01600\_01600\_5500\_N} /010 \\
        & \texttt{0200\_010\_GP\_01600\_03200\_5500\_N} /011 \\
GP 3200 & \texttt{0182\_008\_GP\_03200\_03200\_5500\_N} /011 \\
GP 6400 & \texttt{0036\_002\_GP\_06400\_03200\_3200\_N} /010 \\
GP 10000 & \texttt{0065\_003\_GP\_10000\_08460\_4400\_N} /010 \\
IP 1000 & \texttt{0069\_003\_IP\_01000\_02000\_3200\_N} /011 \\
IP 1600 & \texttt{0117\_005\_IP\_01600\_04160\_5500\_N} /010 \\
        & \texttt{0137\_006\_IP\_01600\_01600\_5500\_N} /011 \\
        & \texttt{0194\_009\_IP\_01600\_04000\_3200\_N} /010 \\
        & \texttt{0195\_009\_IP\_01600\_04000\_5500\_L} /010 \\
IP 2000 & \texttt{0070\_003\_IP\_02000\_04000\_3200\_N} /011 \\
N6 400 & \texttt{0146\_007\_N6\_00400\_00400\_4400\_N} /011 \\
N6 800 & \texttt{0055\_003\_N6\_00800\_01000\_5500\_N} /011 \\
       & \texttt{0075\_004\_N6\_00800\_00080\_3200\_L} /010 \\
N6 3200 & \texttt{0050\_002\_N6\_03200\_03200\_5500\_L} /010 \\
        & \texttt{0145\_007\_N6\_03200\_03200\_4400\_N} /011 \\
S6 400 & \texttt{0126\_006\_S6\_00400\_00200\_4400\_L} /010 \\
       & \texttt{0154\_007\_S6\_00400\_00400\_5500\_L} /011 \\
S6 800 & \texttt{0010\_001\_S6\_00800\_00350\_3200\_N} /011 \\
       & \texttt{0081\_004\_S6\_00800\_00160\_4400\_L} /010 \\
       & \texttt{0177\_008\_S6\_00800\_00200\_5500\_L} /010 \\
S6 1600 & \texttt{0152\_007\_S6\_01600\_01600\_5500\_L} /010 \\
        & \texttt{0178\_008\_S6\_01600\_00400\_5500\_L} /010 \\
S6 3200 & \texttt{0013\_001\_S6\_03200\_01250\_3200\_L} /010 \\
        & \texttt{0015\_001\_S6\_03200\_01600\_5500\_L} /011 \\
        & \texttt{0016\_001\_S6\_03200\_01600\_5500\_N} /011 \\
\bottomrule
\end{tabular}
\caption{\textbf{SIDD-Medium captures held out from fitting.} PGNM is
fitted on every crop of all other captures of the class.}
\label{tab:supp-spec-heldout}
\end{table}

\begin{table*}[tp]
\sidecaptioned{\small
\setlength{\tabcolsep}{5pt}
\begin{tabular}{@{}lrrrrrlr@{}}
\toprule
Preset & Pairs & Captures & $\tau$ & Flat (\%) & Corr.\ (\%) & Corr.\ mask & Energy \\
\midrule
\multicolumn{8}{@{}l}{\textit{SIDD, one preset per camera--ISO class ($256\times256$ crops)}} \\
G4 400 & 1{,}008 & 4 & 0.0087 & 20.7 & 16.7 & flat + guard & 1.0000 \\
G4 800 & 1{,}512 & 6 & 0.0076 & 38.4 & 26.0 & flat + guard & 1.0000 \\
GP 1600 & 1{,}152 & 6 & 0.0088 & 38.1 & 32.8 & flat + guard & 1.0000 \\
GP 3200 & 192 & 1 & 0.0280 & 29.6 & 3.6 & flat + guard & 0.9999 \\
GP 6400 & 576 & 3 & 0.0193 & 50.0 & 21.7 & flat + guard & 0.9999 \\
GP 10000 & 192 & 1 & 0.0209 & 58.8 & 4.6 & flat + guard & 0.9999 \\
IP 1000 & 576 & 3 & 0.0061 & 58.2 & 12.7 & flat + guard & 1.0000 \\
IP 1600 & 3{,}072 & 16 & 0.0070 & 42.5 & 15.3 & flat + guard & 0.9989 \\
IP 2000 & 192 & 1 & 0.0062 & 60.0 & 21.9 & flat + guard & 1.0000 \\
N6 400 & 663 & 3 & 0.0080 & 18.4 & 12.7 & flat + guard & 1.0000 \\
N6 800 & 1{,}326 & 6 & 0.0069 & 24.0 & 20.8 & flat + guard & 1.0000 \\
N6 3200 & 1{,}768 & 8 & 0.0230 & 45.4 & 5.9 & flat + guard & 1.0000 \\
S6 400 & 1{,}008 & 4 & 0.0074 & 34.6 & 26.1 & flat + guard & 1.0000 \\
S6 800 & 2{,}268 & 9 & 0.0070 & 33.7 & 21.3 & flat + guard & 1.0000 \\
S6 1600 & 1{,}512 & 6 & 0.0104 & 10.7 & 3.3 & flat + guard & 1.0000 \\
S6 3200 & 2{,}772 & 11 & 0.0133 & 23.9 & 2.4 & flat + guard & 0.9999 \\
\midrule
\multicolumn{8}{@{}l}{\textit{Film grain, one preset per radius (gray, replicated to RGB, 3-pixel border removed)}} \\
$r=0.025$ & 106 & -- & 0.0030 & 37.2 & 36.9 & flat + guard & 1.0000 \\
$r=0.05$ & 84 & -- & 0.0030 & 46.5 & 45.1 & flat + guard & 1.0000 \\
$r=0.075$ & 104 & -- & 0.0030 & 44.1 & 43.0 & flat + guard & 1.0000 \\
$r=0.1$ & 104 & -- & 0.0030 & 41.6 & 40.5 & flat + guard & 1.0000 \\
$r=0.2$ & 113 & -- & 0.0030 & 45.1 & 40.3 & flat + guard & 1.0000 \\
$r=0.4$ & 114 & -- & 0.0030 & 42.0 & 25.4 & flat + guard & 1.0000 \\
$r=0.6$ & 101 & -- & 0.0030 & 41.4 & 9.4 & flat + guard & 1.0000 \\
$r=0.8$ & 86 & -- & 0.0030 & 47.2 & 47.2 & flat only & 1.0000 \\
\bottomrule
\end{tabular}}
{\caption{\textbf{Checkpoints of the fitted presets.} Pairs: training pairs; Captures: SIDD-Medium captures they come from. $\tau$: flatness threshold (Alg.~\ref{alg:supp-fit}). Flat and Corr.: shares of all training pixels in the flat mask and in the mask used for correlation. Corr.\ mask: flat and clipping guard, or the flat-only fallback. Energy: smallest share, over output channels, of the factor's energy retained by the $25\times25$ crop before normalization.}
\label{tab:supp-spec-checkpoints}}
\end{table*}

The same PGNM estimator is used for both datasets: one preset per SIDD
camera--ISO class and one per film-grain radius. This subsection specifies the
training observations and random streams for each preset;
Tab.~\ref{tab:supp-spec-checkpoints} gives values for checking the resulting fits.

\paragraph{SIDD.}
One preset is fitted per camera--ISO class on SIDD-Medium sRGB~\cite{abdelhamed2018sidd}.
Each full image pair is cut into $256\times256$ crops on a 256-pixel grid in raster
order, with a final row and column of crops flush with the bottom and right
borders (\eg 252 crops for a $3000\times5328$ image). A \emph{capture} is one of
the two image pairs (\texttt{010}, \texttt{011}) of a scene instance. Some captures of
each class are held out (Tab.~\ref{tab:supp-spec-heldout}): one generator,
\texttt{default\_rng(1234)}, visits the classes in sorted order and permutes each
class's sorted list of $n$ captures, and the last $\max(1,\lceil0.2n\rceil)$ captures of
the permutation are held out. Every crop from a held-out capture stays outside the PGNM fit. Fitting uses
every crop of the remaining captures (Tab.~\ref{tab:supp-spec-checkpoints}).

For evaluation, the class preset renders each $256\times256$ clean block of the SIDD Validation set once per
generator run, with a bank built for $256\times256$ and the seeds below, and the
output is delivered to 8 bits.

\paragraph{Film grain.}
One preset is fitted per grain radius using GrainNet's released training
pairs~\cite{lesne2025grain}. Source images identical up to rotation or reflection
form one group, so transformed copies remain in the same partition. The groups
are sorted by their first file name and permuted by \texttt{default\_rng(1234)}.
Of the $n$ groups, the last $\max(1,\lfloor0.2n+0.5\rfloor)$ are held out and
excluded from fitting (Sec.~\ref{sec:supp-baselines},
Tab.~\ref{tab:supp-spec-checkpoints}).

The grain renderer also blurs the clean image. We therefore first estimate the
expected image $\mu$ and fit the residual model around it
(Sec.~\mainref{sec:datasets} of the main paper). For each radius, a $7\times7$
image filter $B$ predicts grainy codes from windows of clean codes. Estimation
first removes a 3-pixel border from both images, then uses all complete
$7\times7$ windows within that crop; the regression target centers consequently
lie at least six pixels inside the original image. $B$ is fitted without an
intercept by float64 least-squares normal equations.

Write the original clean and grainy images in code units as
$x_{\mathrm{code}}=255x$ and $y_{\mathrm{code}}=255y$. The expected image is
$\mu=\operatorname{clip}(B\star x_{\mathrm{code}},0,255)/255$, computed on the
whole image by correlation with mode \texttt{reflect}; the main paper writes
this stage as $B*x$. Before fitting, $\mu$ is cast to float32, and
$y_{\mathrm{code}}$ is converted to float32 and divided by 255. These normalized
arrays are cropped by three pixels on each side, replicated to three identical
channels, and passed to Alg.~\ref{alg:supp-fit}, which uses float64.
This 3-pixel exclusion for residual fitting is distinct from the regression-window
requirement above. Rendering uses float32 $\mu$ replicated to three channels,
and the first output channel is delivered.

\paragraph{Seeds.}
Setup and window selection use separate random streams. Each experiment uses
five generator runs, each with its own four-field bank. On both datasets, the bank of run
$\mathit{run}\in\{0,\dots,4\}$ is built from the seeds $6100+4\,\mathit{run},\dots,6103+4\,\mathit{run}$. The window generator $\mathit{rng}$ of each output image is seeded
from the SHA-256 digest $D$ of a UTF-8 string:
{\raggedright
\begin{itemize}
\item SIDD: $D$ of \texttt{pgnm-protocol3|cvpr-sidd|}\allowbreak\texttt{$\langle$class$\rangle$|$\langle$image$\rangle$|}\allowbreak\texttt{$\langle2026+\mathit{run}\rangle$|$\langle$block$\rangle$},
      where $\langle$class$\rangle$ is written like \texttt{S6\_800}, $\langle$image$\rangle$ is the Validation image
      index (0--39) and $\langle$block$\rangle$ the block index (0--31); seed $=$ (first 8 bytes of $D$,
      big-endian) $\bmod (2^{31}-1)$.
\item Film: $D$ of \texttt{cvpr-film-pgnm:}\allowbreak\texttt{$\langle$radius$\rangle$:$\langle$image$\rangle$:$\langle\mathit{run}\rangle$},
      with the radius written to three decimals (\texttt{0.025}) and the test image's
      two-digit identifier (\texttt{01}--\texttt{60}); seed $=$ first 4 bytes of $D$, little-endian.
\end{itemize}
\par}
\noindent Before each output image, $\mathit{prev}$ is reset to $-1$, so in our experiments
the field is drawn uniformly from all four; the rule $j\ne\mathit{prev}$ matters only
for consecutive renders without a reset, such as video frames. With these seeds and
NumPy's generator, the procedure determines every output up to floating-point
rounding in the FFT convolution.

\paragraph{Denoising.}
The denoising presets are fitted in the same way, on every crop of each class's
denoising training images. Training does not use the bank: for every training patch
and epoch, a fresh field \textsc{BuildField}$(96,96,\mathit{rng})$ is built, and
\textsc{Render}'s per-pixel loop reads it with offsets $a=b=0$. Each random stream is
seeded with the first 8 bytes, read big-endian and shifted right by one bit, of the
SHA-256 digest of a string, for training seed $\langle$seed$\rangle$, epoch $\langle$epoch$\rangle=1,\dots,30$ and
position $\langle$position$\rangle$ of the patch in the epoch's order:
{\raggedright
\begin{itemize}
\item field: \texttt{CVPR-DENOISE/1|pgnm-field|}\allowbreak\texttt{$\langle$seed$\rangle$|$\langle$epoch$\rangle$|$\langle$position$\rangle$};
\item patch order: \texttt{CVPR-DENOISE/1|permutation|}\allowbreak\texttt{$\langle$seed$\rangle$|$\langle$epoch$\rangle$};
\item augmentation: \texttt{CVPR-DENOISE/1|augment|}\allowbreak\texttt{$\langle$seed$\rangle$|$\langle$epoch$\rangle$|$\langle$position$\rangle$}.
\end{itemize}
\par}

\section{Metric details}
\label{sec:supp-metrics}

The scores in Sec.~\mainref{sec:metrics} of the main paper examine complementary
properties. Conditional $W_1$ compares residual distributions at similar clean
intensities; spatial and color correlation errors compare relationships across
locations and channels; KL compares residual histograms within small tiles.
Film-grain evaluation also compares the appearance of delivered images, while
denoising evaluates reconstruction of clean images after training on each noise
source. Lower values are better for the synthesis errors and appearance
distances; higher PSNR and SSIM are better for denoising.

We first specify the common observations and intensity groups, then each
estimator and its aggregation. Alg.~\ref{alg:supp-sidd-scores} summarizes one
SIDD source image and one generator run. The film-grain adaptations follow below.

\begin{algorithm*}[tp]
\caption{SIDD scores of one source image and one generator run. All statistics
except KL, which is averaged over tiles, are pooled over the 32 blocks; pixel pairs
never cross block boundaries. Zero-variance and support checks follow the text.}
\label{alg:supp-sidd-scores}
\small
\begin{algorithmic}[1]
\Require delivered code arrays: clean $X_8$, real $Y_8$, generated $\widehat Y_8\in\{0,\dots,255\}^{32\times256\times256\times3}$
\State $r\gets (Y_8-X_8)/255$, \quad $\widehat r\gets(\widehat Y_8-X_8)/255$
\For{$c=1,2,3$}
    \State $j_c(p)\gets\min(\lfloor 16X_{8,c}(p)/255\rfloor,15)$ for every pixel $p$
    \State $\{b\}_c\gets\Call{Merge}{\text{counts of }j_c,\,512}$ \Comment{groups of adjacent bins}
\EndFor
\State $E_{W_1}\gets 255\cdot\frac13\sum_c\sum_b\frac{n_{cb}}{n_c}\,
       \operatorname{mean}\bigl|\operatorname{sort}(r_c[b])-\operatorname{sort}(\widehat r_c[b])\bigr|$
\State $z\gets\Call{Standardize}{r}$, \quad $\widehat z\gets\Call{Standardize}{\widehat r}$
\For{$c=1,2,3$ and $\delta\in\{-8,\dots,8\}^2$}
    \State $A_c(\delta)\gets\operatorname{mean}_{(p,p+\delta)\in P_\delta} z_c(p)z_c(p+\delta)$;
           \quad $\rho_c(\delta)\gets A_c(\delta)/A_c(0)$ \Comment{same for $\widehat z$}
\EndFor
\State $E_{\mathrm{spatial}}\gets\operatorname{RMS}_{c,\;\delta\ne0}\bigl(\rho_c(\delta)-\widehat\rho_c(\delta)\bigr)$
       \Comment{$3\times288$ values}
\For{$\delta\in\{(0,1),(1,0),(1,1),(1,-1)\}$ and ordered $(a,b)$, $a\ne b$}
    \State $d_{ab}(\delta)\gets\sqrt{\operatorname{mean}_{P_\delta}[z_a(p)^2]}\,
           \sqrt{\operatorname{mean}_{P_\delta}[z_b(p+\delta)^2]}$
    \State $\kappa_{ab}(\delta)\gets\operatorname{mean}_{P_\delta}[z_a(p)z_b(p+\delta)]/\max(d_{ab}(\delta),10^{-12})$
           \Comment{same for $\widehat z$}
\EndFor
\State $E_{\mathrm{color}}\gets\operatorname{RMS}_{\delta,(a,b)}\bigl(\kappa_{ab}(\delta)-\widehat\kappa_{ab}(\delta)\bigr)$
       \Comment{24 values}
\For{each block $k$ and each of its 64 nonoverlapping $32\times32$ tiles $t$}
    \State $h\gets$ counts of $255\,r[k,t]$ (all channels) in bins $[-260,-256),\dots,[256,260]$;
           \quad $P\gets(h+1)/\textstyle\sum(h+1)$; \quad $Q$ likewise from $\widehat r$
    \State $D_{kt}\gets\sum_j P_j\log(P_j/Q_j)$ \Comment{$p_j$, $q_j$ of Eq.~\mainref{eq:metric_kl}}
\EndFor
\State $D_{\mathrm{KL}}\gets\operatorname{mean}_k\operatorname{mean}_t D_{kt}$
\State \Return $E_{W_1},\;E_{\mathrm{spatial}},\;E_{\mathrm{color}},\;D_{\mathrm{KL}}$
\Statex
\Procedure{Standardize}{$r$}
    \For{each channel $c$ and group $b$ of $c$}
        \State $m\gets\operatorname{mean}(r_c[b])$, \quad $s\gets\operatorname{std}(r_c[b])$ \Comment{population std}
        \State $z_c[b]\gets(r_c[b]-m)/\max(s,1/255)$
    \EndFor
    \State \Return $z$
\EndProcedure
\Statex
\Procedure{Merge}{counts $n_0,\dots,n_{15}$, minimum $m$}
    \State $G\gets[\{0\},\{1\},\dots,\{15\}]$ \Comment{ordered list of groups of bin indices}
    \While{$|G|>1$ and some group has fewer than $m$ pixels}
        \State $g\gets$ the undersized group with the fewest pixels (ties: lowest first bin)
        \State $h\gets$ the neighbor of $g$ in $G$ whose mean bin index is closest to that of $g$
               (ties: lower first bin)
        \State replace $g$ and $h$ by their union
    \EndWhile
    \State \Return $G$
\EndProcedure
\end{algorithmic}
\end{algorithm*}

\paragraph{Residuals after 8-bit delivery.}
Synthesis metrics compare the images delivered by each method. A continuous
output $v$ in $[0,1]$ units is converted to 8-bit codes by
$\lfloor255\operatorname{clip}(v,0,1)+0.5\rfloor$.
On SIDD, let $X_8$, $Y_8$, and $\widehat Y_8$ denote the clean, real noisy, and
generated code arrays used in Alg.~\ref{alg:supp-sidd-scores}. Their normalized
values are $x=X_8/255$, $y=Y_8/255$, and $\widehat y=\widehat Y_8/255$, as in the
main paper. The signed residuals are $r=y-x$ and $\widehat r=\widehat y-x$;
thus $255r$ and $255\widehat r$ are integer code residuals in $[-255,255]$.

\paragraph{Pixels included in evaluation.}
Every SIDD block pixel contributes, including textured and clipped observations.
The 32 blocks of one source image form one scoring unit per generator run:
intensity statistics pool their pixels, and correlations pool products and
pair counts without forming pairs across block boundaries. Evaluation uses
neither the flatness mask nor the clipping guard used for fitting. Film scores
use the image interior after removing three pixels from each edge.

\paragraph{Grouping similar clean intensities.}
Conditional statistics compare residuals where the clean intensity is similar.
For each channel $c$, assign pixel $p$ to
$j_c(p)=\min(\lfloor16x_c(p)\rfloor,15)$. The groups depend only on the clean
image, so real and generated residuals, and all methods, share exactly the same
pixels in each group.
To avoid estimating a distribution from too few pixels, adjacent bins are merged
until each group has at least 512 pixels or only one group remains
(\textsc{Merge} in Alg.~\ref{alg:supp-sidd-scores}). At each step, the
undersized group with the fewest pixels is merged with the adjacent group whose
unweighted mean bin index is closest. Ties select the group with the lower first
bin. Write $b$ for a merged group, $n_{cb}$ for its pixel count, and
$n_c=\sum_b n_{cb}$.

\paragraph{Conditional distribution mismatch: $W_1$.}
This error responds to differences in residual bias, strength, and distribution
shape at a given clean intensity. In channel $c$ and group $b$, the real and
generated lists contain residuals from the same $n_{cb}$ pixels. Sorting both lists pairs
matching empirical percentiles; the mean absolute difference is their exact
1-Wasserstein distance. The score of
Eq.~\mainref{eq:metric_conditional_w1} weights groups by $n_{cb}/n_c$, averages
channels equally, and multiplies by 255 to report code units.
On SIDD's integer residual grid, the same distance before this final factor is
$\frac{1}{255}\sum_{v=-256}^{256}|F_{cb}(v/255)-G_{cb}(v/255)|$, where $F_{cb}$
and $G_{cb}$ are the empirical CDFs. This equivalent histogram calculation avoids
sorting all block pixels.

\paragraph{Centering and scaling before correlation.}
To keep conditional bias and strength differences from dominating the
correlation comparison, each residual is centered and scaled within its
clean-intensity group:
\begin{equation}
    z_c(p)=\frac{r_c(p)-m_{cb}}{\max(s_{cb},\,1/255)},\qquad j_c(p)\in b.
\end{equation}
Here $m_{cb}$ and $s_{cb}$ are the group's mean and population standard deviation.
They are computed separately from the real and generated residuals, giving
$z$ and $\widehat z$ on the shared groups. This is an evaluation normalization;
it does not use the pilot strength fitted in Sec.~\ref{sec:supp-spec}.

\paragraph{Spatial dependence within each channel.}
Spatial correlation measures how residuals at two locations vary together as
their separation changes. For lag $\delta=(\delta_y,\delta_x)\in\{-8,\dots,8\}^2$,
let $P_\delta$ contain pixel pairs $(p,p+\delta)$ whose endpoints are both in the
scoring region and the same block. We compute
\begin{equation}
    A_c(\delta)=\frac{1}{|P_\delta|}\sum_{(p,p+\delta)\in P_\delta}z_c(p)z_c(p+\delta),
    \qquad \rho_c(\delta)=\frac{A_c(\delta)}{A_c(0)}.
\end{equation}
The spatial error is the root mean square of
$\rho_c(\delta)-\widehat\rho_c(\delta)$ over all channels and the 288 nonzero
lags. Each lag and its opposite enter separately and have equal values.
Zero-padded block FFTs, with at least eight pixels of padding in each dimension,
compute these sums without wrap-around.

\paragraph{Color dependence at neighboring locations.}
Color correlation measures whether a residual in one channel varies with a
neighboring residual in another channel. We use the four row--column lags
$\delta\in\{(0,1),(1,0),(1,1),(1,-1)\}$ and all six ordered pairs $(a,b)$,
$a\ne b$. For each, normalize the mean product by the endpoint root mean squares:
\begin{equation}
\begin{aligned}
    d_{ab}(\delta)&=
    \sqrt{\frac{1}{|P_\delta|}\sum z_a(p)^2}\,
    \sqrt{\frac{1}{|P_\delta|}\sum z_b(p+\delta)^2},\\
    \kappa_{ab}(\delta)&=
    \frac{\frac{1}{|P_\delta|}\sum z_a(p)z_b(p+\delta)}
         {\max(d_{ab}(\delta),10^{-12})},
\end{aligned}
\end{equation}
with all sums over $P_\delta$. The color error is the root mean square of
$\kappa_{ab}(\delta)-\widehat\kappa_{ab}(\delta)$ over these 24 values.
It excludes same-pixel channel correlation (zero lag).

\paragraph{Constant residuals and insufficient support.}
If a generated channel has $\widehat A_c(0)\le10^{-12}$, its spatial correlation
is set to zero; the denominator floor above also defines its color products.
If any real channel has $A_c(0)\le10^{-12}$, both SIDD correlation errors are
undefined. They are also undefined when the region has fewer than 4,096 pixels
or any lag has fewer than 512 pairs. All complete SIDD scoring units satisfy
these size requirements.

\paragraph{Local histogram mismatch: KL divergence.}
KL compares the frequency of residual values within small regions. It pools
channels and clean intensities in each tile; it does not encode the arrangement
of residuals inside that tile. Each $256\times256$ block is divided into 64
nonoverlapping $32\times32$ tiles. The 3,072 integer code residuals in a tile
are counted in 130 width-four bins with edges $-260,-256,\dots,260$ (the last
bin closed). Adding one count per bin gives $p_j=(n_j+1)/(3072+130)$ for the
real residuals and $q_j$ for the generated residuals.
Eq.~\mainref{eq:metric_kl} then computes $D_{\mathrm{KL}}(p\|q)$ with natural
logarithms. The source score averages tiles within each block, then the 32
blocks. The $256\times256$ block KL, reported only in this supplement, uses the
same histogram estimator on the whole block. NM-FlowGAN's released code applies
the estimator before rounding its output; here every method is scored after
8-bit delivery.

\paragraph{Equal camera weighting on SIDD.}
Scores are first computed for each source image and generator run, then averaged
over the five runs. The main table (Tab.~\mainref{tab:sidd}) gives each camera
equal weight, each camera--ISO class equal weight within its camera, and each
source image equal weight within its class. This prevents cameras with more
sources or ISO settings from dominating the result. The 19-source block applies
the same rule to its own sources.

\paragraph{Film residuals around the expected image.}
The film renderer changes the image mean through blur as well as adding grain.
Following Sec.~\ref{sec:supp-spec-experiments}, residuals are therefore measured
around the continuous expected image
$\mu=\operatorname{clip}(B\star X_8,0,255)/255$, where $X_8$ is the clean gray
code image. The fitted $7\times7$ filter $B$ is applied by whole-image
correlation with reflected boundaries, and $\mu$ is not rounded.
The real and generated residuals are $r=Y_8/255-\mu$ and
$\widehat r=\widehat Y_8/255-\mu$.

\paragraph{Film scoring and repeated renders.}
Conditional $W_1$ and spatial correlation use one gray channel and the
three-pixel interior, with $\mu$ replacing $x$ in the intensity groups.
Because $\mu$ is continuous, $W_1$ compares sorted residual lists directly.
Spatial correlation requires at least one pair at every nonzero lag and
nondegenerate reference variance, with the same generated-variance rule as
above. Color correlation and KL are not reported for film grain.
For each test image and radius, five outputs from a method are compared with
five Newson reference renders: every output is paired with every reference,
giving 25 scores. We average these scores, then images equally within each
radius, then radii equally. The independent Newson render is included as a
method with five seeds separate from the reference seeds.

\paragraph{Overall image appearance: DISTS and LPIPS.}
These distances compare delivered images rather than residuals, complementing
the distribution and correlation errors. DISTS~\cite{ding2022dists} and
LPIPS~\cite{zhang2018lpips} (AlexNet) are evaluated for each film output--reference
pair after cropping three pixels from each image edge, with the implementations
and inputs of GrainNet's released evaluation script~\cite{lesne2025grain}.
Gray images are represented as three identical channels, explicitly for DISTS
and by channel broadcasting for LPIPS, with values in $[0,1]$. LPIPS inputs are
not rescaled to $[-1,1]$. Aggregation follows the same 25-pair, image, and radius
averages as the residual scores.

\paragraph{Clean-image reconstruction: PSNR and SSIM.}
Denoising scores compare a denoiser's output with the clean block: PSNR measures
pixel reconstruction error, while SSIM compares local luminance, contrast, and
structure. Under NM-FlowGAN's protocol, noisy 8-bit inputs are divided by 256;
outputs are clipped to $[0,1]$, multiplied by 256, rounded half up, and clipped
to 255 before scoring. PSNR uses peak value 255. SSIM follows scikit-image's
defaults for 8-bit RGB: a $7\times7$ uniform window, $K_1=0.01$, $K_2=0.03$,
and sample covariance. Each channel's SSIM map is averaged after discarding
its three-pixel border, and the channel scores are averaged. Both PSNR and
SSIM are computed per block and averaged equally over the 1,280 Validation
blocks.

\FloatBarrier
\section{Baselines}
\label{sec:supp-baselines}

\begin{table*}[tp]
\sidecaptioned{\small
\setlength{\tabcolsep}{8pt}
\begin{tabular}{@{}lll@{}}
\toprule
Method & Released code & Commit \\
\midrule
NAFlow & \texttt{github.com/dongjinkim9/NAFlow} & \texttt{533e1eac581a} \\
DANet & \texttt{github.com/zsyOAOA/DANet} & \texttt{ea4d71416815} \\
sRGB-Flow & \texttt{github.com/SamsungLabs/Noise2NoiseFlow} & \texttt{54cc3c11e4fc} \\
NM-FlowGAN & \texttt{github.com/YoungJooHan/NM-FlowGAN} & \texttt{63983e2c2c5e} \\
GrainNet & \texttt{github.com/Gwilherm-LESNE/Neural\_Film\_Grain\_Rendering} & \texttt{865c49400405} \\
Newson & \texttt{github.com/alasdairnewson/film\_grain\_rendering\_gpu} & \texttt{5695ff2abea7} \\
\bottomrule
\end{tabular}}
{\caption{\textbf{Released code and checkpoints used for the baselines}, at the
listed commits.}
\label{tab:supp-baseline-code}}
\end{table*}

\paragraph{Common protocol on SIDD.}
Every baseline uses its released checkpoint and sampling code at its native
settings (Tab.~\ref{tab:supp-baseline-code}), except GMDCN, which we implemented. Each
$256\times256$ Validation block is generated separately at native size and batch
1, five times with derived seeds, and every output is clipped to $[0,1]$ and rounded
half up to 8 bits like PGNM's. Baselines receive the camera and ISO of the block's
class where their model accepts them; none reads the real noisy block.

\paragraph{Methods.}
\begin{itemize}
\item \textbf{NAFlow}~\cite{kim2024naflow} samples from its class-conditional
      latent distribution for the block's camera--ISO label at the released
      sampling temperature; its noise-aware mode, which reads a real noisy image,
      is not used.
\item \textbf{GMDCN}~\cite{song2023camera} is our implementation of the published
      model. It predicts the residual from the clean image and a camera--ISO
      condition map and adds it to the clean image.
\item \textbf{DANet}~\cite{yue2020danet} uses the generator of its released model.
      It receives no camera or ISO input.
\item \textbf{sRGB-Flow}~\cite{kousha2022srgb} takes the clean image, camera and
      ISO. Its checkpoint is trained on 15 camera--ISO classes, 7 of our 16 (G4
      ISO\,800, IP ISO\,1600, N6 ISO\,800 and the four S6 classes), which form the
      19-source block.
\item \textbf{NM-FlowGAN}~\cite{han2023nmflowgan} takes the clean image, camera and
      ISO, and its checkpoint is trained on the same 15 classes. It accepts the
      ISO values 100, 400, 800, 1600 and 3200 for every camera, but we report it
      only on the 19-source block of its training classes.
\item \textbf{NeCA-W}~\cite{fu2023neca} is not included: its released generator
      requires an encoding of a real noisy image at sampling time, and
      we found no reference-free inference rule consistent with our protocol.
\end{itemize}

\paragraph{Film grain.}
Fitting uses GrainNet's released training pairs~\cite{lesne2025grain}. Their 199
source images, with flipped and rotated copies merged, were split with seed 1234
into 159 for fitting and 40 held out and unused (evaluation uses GrainNet's 60
separate test images), which leaves 84--114 fitting pairs per
radius (Tab.~\ref{tab:supp-spec-checkpoints}). PGNM and Zhang23 add grain to the
same expected image $\mu=B*x$ (Sec.~\ref{sec:supp-spec-experiments}); GrainNet and
the independent Newson render receive the clean image in their native pipelines.
\begin{itemize}
\item \textbf{Zhang23}~\cite{zhang2023grain} is our implementation of the published
      equations. Its grain-size parameter is estimated from the ratio of lag-1 and
      lag-$\sqrt2$ correlations of flat-pixel residuals (clipped to $[0.3,4]$), and
      its radius by least squares on Zhang23's variance expression across clean
      intensities, after subtracting 8-bit quantization noise; both are fitted on
      the $B$-filtered fitting pairs inside PGNM's flat masks. Rendering adds
      Gaussian noise with the predicted intensity-dependent variance, filtered by a
      separable 33-tap kernel, to $\mu$. At $r=0.8$ the measured correlations vanish
      and the grain-size parameter reaches its lower bound, consistent with the
      approximation no longer holding for resolved grain.
\item \textbf{GrainNet}~\cite{lesne2025grain} uses the released checkpoint with the
      clean gray image, radius and a seed as input.
\item \textbf{Newson}~\cite{newson2017stochastic}: references and the independent
      run use the released GPU renderer, changed only to accept an explicit random
      seed, with grain standard deviation 0, filter standard deviation 0.8 pixels,
      zoom 1 and $10^5$ Monte Carlo samples. Each image and radius has five
      reference renders and five independent renders from separate seed sets.
\end{itemize}

\FloatBarrier
\section{Cost measurement setup}
\label{sec:supp-cost}

This section describes how the parameter counts and times of
Tab.~\mainref{tab:cost} of the main paper were obtained.

\paragraph{Hardware and environment.}
All rows ran one at a time on one machine with an Intel Core i9-14900K. Every
process was pinned to four efficiency cores and limited to two threads (OpenMP,
MKL, Numba and PyTorch), at the lowest scheduling priority and with a 16\,GB
memory cap. GPU rows used one NVIDIA RTX A6000, capped at 12\,GB, with the same
CPU pinning. Each method ran in its own environment with the PyTorch version its
code requires (1.8.1 to 2.5.1, CUDA 11.1 to 12.1). The arithmetic is
single precision; cuDNN was allowed to use TF32 convolutions in every environment,
and two environments (DANet, sRGB-Flow) also allowed TF32 matrix products.

\paragraph{Input and timed region.}
The input is one synthetic $1920\times1080$ 8-bit RGB frame of color ramps, batch 1.
SIDD rows use the G4 ISO\,800 condition and its PGNM preset; film rows use the
radius 0.1 setting, applied to each color channel. The timed region starts from the
8-bit frame in host memory and ends with the method's noisy frame, in floating
point, in host memory; it includes type conversion and all host--device copies,
and CUDA is synchronized before and after each call. Model loading, the first call
and the final 8-bit quantization are excluded; the quantization, common to all
methods, took 50--100\,ms on the two CPU threads. GPU rows report the median of 50
calls after 10 warm-up calls, CPU rows the median of 10 calls after one.

\paragraph{PGNM rows.}
\emph{New field per frame} draws $(H+24)\times(W+24)\times3$ Gaussian values, filters
them with the $25\times25\times3\times3$ tensor (FFT convolution on CPU, direct
convolution on GPU), and applies the lookup, addition and clipping; this is
comparable to the baselines, which draw new noise for every frame. \emph{Bank
playback} is the renderer of all reported results (Alg.~\ref{alg:supp-render}): it
reads one cyclic window of the four $2H\times2W\times3$ float32 fields (398\,MB at this
resolution). \emph{Bank construction} builds these fields once per preset and
resolution; on CPU it includes compiling the lookup kernel, and on GPU it is the
process's first convolution. The bank pays for itself after
$307/(20.4-10.2)\approx30$ frames on GPU and $23.45/(1.04-0.051)\approx24$ on CPU.
The GPU rows use a PyTorch port of the renderer. Filtering the same white noise,
the port's latent field differs from the CPU field by at most $1.02\times10^{-5}$
(unit variance); given the same field, its noisy image differs by at most
$1.8\times10^{-7}$ before rounding and is identical after 8-bit delivery, on a
$256\times384$ crop (height by width) of the same frame. The film PGNM preset has the same array
sizes as the SIDD one and was not timed separately.

\paragraph{Baselines.}
Neural baselines run their generators (released, or ours for GMDCN) through our adapters
(Sec.~\ref{sec:supp-baselines}) in inference mode, including their per-call random
seeding; NM-FlowGAN's reflection padding to 1088 rows and cropping back are inside
the timed region. GrainNet, a gray model, runs once per color channel. Zhang23 has
no GPU implementation, and its time excludes the shared blur $B$.

\paragraph{Parameter counts.}
A PGNM preset holds $25\cdot25\cdot3\cdot3=5{,}625$ filter values and
$3\cdot256\cdot25=19{,}200$ lookup values; the fixed Gaussian knot grid, the
residual step and the bank seeds are not counted. Film presets add the 49 values of
the shared $7\times7$ blur, and Zhang23 has two fitted values per radius plus the
same blur. Neural baselines count the learned parameters of their noise generators.

\FloatBarrier

\end{document}